\PassOptionsToPackage{numbers,sort,compress}{natbib}
\PassOptionsToPackage{dvipsnames,table}{xcolor}
\documentclass[11pt,letterpaper,twocolumn,teaser]{planstyle}

\newcommand{\modelname}{\textup{\textsc{EditVid}}}
\newcommand{\modelnamebf}{\textup{\textbf{\textsc{EditVid}}}}

\usepackage{bm}
\IfFileExists{placeins.sty}{\usepackage{placeins}}{\newcommand{\FloatBarrier}{}}

\definecolor{best}{RGB}{255,214,170}
\definecolor{second}{RGB}{255,235,210}
\definecolor{rowgray}{RGB}{235,235,235}

\usepackage[T1]{fontenc}
\usepackage{microtype}           
\usepackage{nicefrac}            
\usepackage{xspace}              
\usepackage{amsmath, amssymb, amsfonts, mathtools, mathrsfs}
\usepackage{dsfont}              

\usepackage{booktabs}            
\usepackage{nicematrix}         
\usepackage{multirow}
\usepackage{makecell}
\usepackage{bigstrut}
\usepackage{enumitem}            
\setitemize{label=\textbullet, leftmargin=*, nolistsep}
\usepackage{graphicx}
\usepackage{adjustbox}           
\usepackage{float}               
\usepackage{wrapfig}             
\usepackage{xcolor}                
\usepackage{color}
\IfFileExists{bxcoloremoji.sty}{\usepackage{bxcoloremoji}}{} 
\IfFileExists{fontawesome5.sty}{\usepackage{fontawesome5}}{} 
\IfFileExists{pifont.sty}{\usepackage{pifont}}{}             
\IfFileExists{algorithm.sty}{%
\usepackage{algorithm}
}{%
\newenvironment{algorithm}[1][]{\begin{figure}[#1]}{\end{figure}}
}
\IfFileExists{algpseudocode.sty}{%
\usepackage{algpseudocode}
}{%
\newenvironment{algorithmic}[1][]{\begin{enumerate}}{\end{enumerate}}
\newcommand{\Require}{\item[\textbf{Require:}]}
\newcommand{\State}{\item}
\newcommand{\Statex}{\item[]}
\newcommand{\For}[1]{\item[\textbf{for}] #1 \textbf{do}}
\newcommand{\EndFor}{\item[\textbf{end for}]}
\newcommand{\algorithmicindent}{1.5em}
}
\IfFileExists{ulem.sty}{\usepackage[normalem]{ulem}}{} 
\IfFileExists{lipsum.sty}{\usepackage{lipsum}}{}       
\IfFileExists{rotating.sty}{\usepackage{rotating}}{}   
\IfFileExists{stackengine.sty}{\usepackage{stackengine}}{} 
\IfFileExists{listings.sty}{\usepackage{listings}}{}   
\IfFileExists{soul.sty}{\usepackage{soul}}{}           
\IfFileExists{gradient-text.sty}{\usepackage{gradient-text}}{} 
\IfFileExists{pgfplots.sty}{\usepackage{pgfplots}\pgfplotsset{compat=newest}}{} 
\IfFileExists{pgfplotstable.sty}{\usepackage{pgfplotstable}}{} 
\IfFileExists{tikz.sty}{\usepackage{tikz}}{}           
\IfFileExists{svg.sty}{\usepackage{svg}}{}             

\providecommand{\ie}{\textit{i.e.},\xspace}  
\providecommand{\eg}{\textit{e.g.},\xspace}  
\providecommand{\etc}{\textit{etc}.\xspace}  

\IfFileExists{pifont.sty}{%
\newcommand{\cmark}{\textcolor{ForestGreen}{\ding{51}}}
\newcommand{\xmark}{\textcolor{red}{\ding{55}}}
}{%
\newcommand{\cmark}{\textcolor{ForestGreen}{\(\checkmark\)}}
\newcommand{\xmark}{\textcolor{red}{\(\times\)}}
}

\usepackage{natbib}
\usepackage[capitalize,noabbrev]{cleveref}
\crefname{equation}{Eq.}{Eqs.}
\crefformat{section}{\S#2#1#3}
\crefformat{subsection}{\S#2#1#3}
\crefformat{subsubsection}{\S#2#1#3}
\crefrangeformat{section}{\S\S#3#1#4 to~#5#2#6}
\crefmultiformat{section}{\S\S#2#1#3}{ and~#2#1#3}{, #2#1#3}{ and~#2#1#3}

\title{One Editor, Many Edits: A Unified Training-Free Framework for Diverse Video Editing}

\author{%
  \textbf{Adheesh Sunil Juvekar, Onkar Kishor Susladkar, Kiet A. Nguyen,}\\
  \textbf{Muntasir Wahed, Nabeel Bashir, Xiaona Zhou, Tianjiao Yu,}\\
  \textbf{Vedant Shah, and Ismini Lourentzou}
}
\affil{University of Illinois Urbana-Champaign}

\renewcommand{\insertteaserfigure}{%
  \begin{center}
    \vspace{-0.4cm}
    \includegraphics[width=0.9\linewidth]{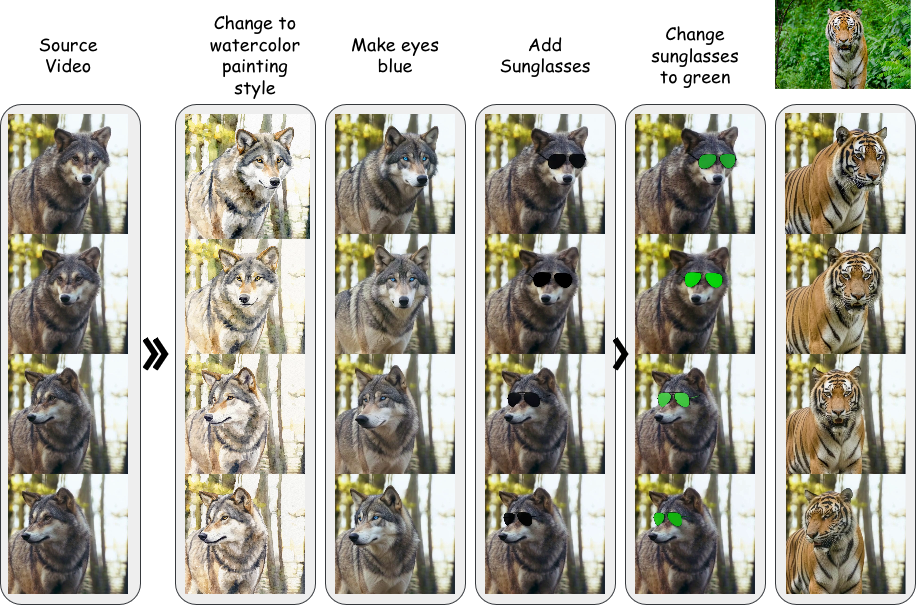}
    \captionsetup{hypcap=false}
    \captionsetup{font=small}
    \captionof{figure}{\textbf{\modelname{} enables training-free, temporally consistent video editing across diverse edit types.} Our method supports style transfer, attribute change, object insertion, attribute refinement, part-level modification, and subject-guided replacement while preserving temporal consistency.}
    \label{fig:teaser}
  \end{center}
}

\correspondingauthor{$^*$ Preprint.}

\begin{document}

\setlength{\abovedisplayskip}{4pt}
\setlength{\belowdisplayskip}{4pt}
\setlength{\abovedisplayshortskip}{2pt}
\setlength{\belowdisplayshortskip}{2pt}

\setabstractlogo[7.5mm]{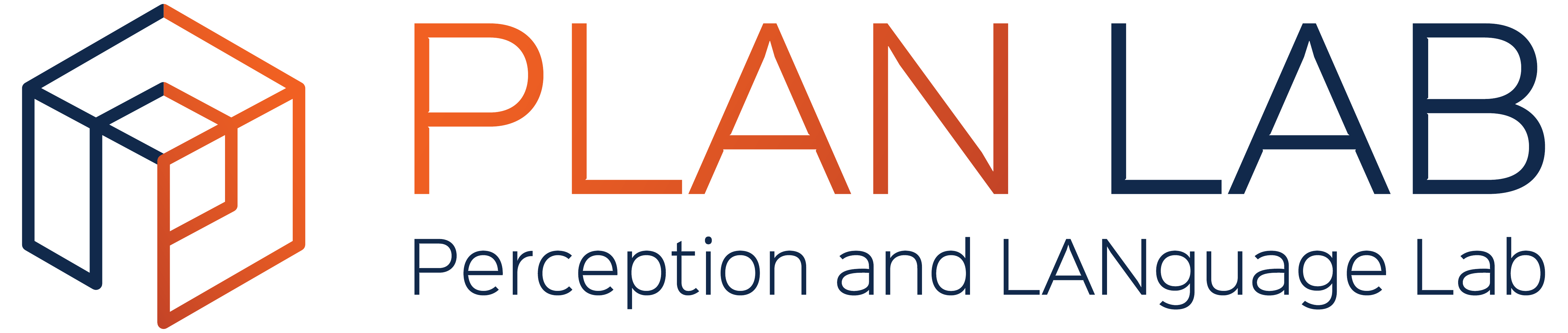}

\begin{abstract}
Video editing spans diverse editing paradigms, yet achieving high-quality instruction-guided and subject-guided editing within a single unified framework remains challenging.
We introduce \modelnamebf{}, a training-free framework combining sparse causal memory for local coherence, correspondence-based post-attention token injection for long-range identity preservation, and soft latent blending for edit locality. 
The same framework supports instruction-guided and reference-guided edits, including style transfer, attribute modification, object insertion, part-level editing, and subject replacement.  
On FiVE, \modelname{} achieves 78.16 FiVE-Acc, compared with 58.95 for the strongest evaluated training-free baseline, while obtaining competitive results on IVEBench. 
A user study further shows a 51.8\% overall preference for \modelname{} over 7 competing methods.\looseness-1 \\
\href{https://plan-lab.github.io/editvid}{https://plan-lab.github.io/editvid}
\end{abstract}
\maketitle

\section{Introduction}
\label{sec:intro}

Image editing has advanced rapidly with modern multimodal diffusion transformers (MM-DiTs)~\citep{esser2024scaling,flux2024,labs2025flux,avrahami2025stable,kulikov2025flowedit,susladkar2026rewardflow}, which provide strong instruction following, reference-image conditioning, and high-fidelity manipulation across diverse editing tasks. Extending these capabilities to video, however, requires handling temporal coherence without sacrificing the semantic control and edit diversity of the image model. Naively applying an image editor to each frame independently lacks temporal coupling, producing flicker, identity drift, inconsistent edits, and unintended changes to the background.
Dedicated video editing models address these requirements through video-specific training~\citep{jiang2025vace,zi2025senorita,bai2026scaling,yang2025unified,lin2026kiwiedit}, while training-free approaches~\cite{li2024vidtome,geyer2024tokenflow,yang2025videograin,jiao2026streamedit} instead reuse pretrained image or video generative priors~\citep{li2025five,cai2025dfvedit,li2026flowdirector}. This raises a central question \emph{whether the rich editing capabilities of a modern image MM-DiT be extended to videos without video-specific training?}

Addressing this question requires understanding \emph{how temporal information should be introduced into an MM-DiT}. Existing image-model-based video editing methods establish useful mechanisms such as cross-frame attention, feature propagation, and correspondence-based reuse~\citep{qi2023fatezero,geyer2024tokenflow,cong2024flatten,li2024vidtome}. Modern MM-DiTs, however, expose a different representation structure where visual and conditioning tokens interact through multimodal transformer blocks, while spatial relationships in attention are encoded through rotary positional embeddings (RoPE)~\citep{su2024roformer,heo2024rotary}. Consequently, effective temporal coupling in MM-DiTs depends on the choice of representation space and temporal scope of cross-frame interaction, both of which interact with the model’s positional and semantic conditioning.\looseness-1

We find that these choices are fundamentally different for short- and long-range temporal consistency. Consecutive frames retain strong spatial continuity, making adjacent-frame attention states useful for stabilizing local appearance and motion. As temporal distance and spatial displacement increase, however, directly reusing attention states becomes sensitive to the relative geometry encoded by RoPE. 
This observation motivates a \emph{local--global decomposition} that uses attention-level memory where geometric continuity is reliable, and correspondence-guided visual feature transfer where long-range identity matters.

Motivated by this distinction, we introduce \modelnamebf{}, a training-free framework that extends frozen image MM-DiT editors to temporally consistent video editing. At the \emph{local} level, sparse causal memory exposes each frame to the immediately preceding frame's key--value states, providing short-range temporal coherence while keeping temporal context bounded. At the \emph{global} level, \modelname{} establishes high-confidence, cycle-consistent correspondences between an anchor and subsequent frames and injects matched visual representations \emph{after attention}. This enables long-range appearance and identity transfer without requiring distant tokens to interact through cross-frame RoPE-modulated attention. Temporal consistency alone is insufficient for reliable editing, as even coherent outputs may alter backgrounds or other instruction-irrelevant content. We therefore derive continuous, timestep-dependent preservation weights from the discrepancy between source and edited trajectories, adaptively retaining source content where preservation is needed while allowing the requested edit to dominate elsewhere. Together, these mechanisms decompose video editing into \emph{local temporal coherence, long-range identity preservation, and edit locality}, without video-specific training or a separate image-to-video propagation model.

We evaluate \modelname{} on the FiVE~\citep{li2025five} and IVEBench~\cite{chen2026ivebench} benchmarks, complemented by VLM-based evaluation on a curated set spanning subject-guided and general video editing, as well as a comprehensive user study. Among training-free methods, \modelname{} achieves the highest FiVE-Acc and competitive performance on IVEBench while maintaining strong video fidelity. Controlled ablations further show that using only the immediately preceding frame as memory outperforms retaining a longer temporal history, while correspondence filtering improves robustness to challenging temporal changes.
Our contributions are summarized as follows:

\begin{itemize}[itemsep=0.4ex, parsep=0pt, topsep=-2pt, leftmargin=0.55cm]
    \item We introduce \textbf{\modelname{}}, a training-free framework that supports both instruction-guided and subject-guided video editing, leveraging MM-DiT-based image editors as strong priors across diverse editing settings.    
    
    \item We identify temporal context as a key factor in MM-DiT representation reuse and develop a RoPE-aware local--global design that uses adjacent-frame key--value memory for short-range coherence and confidence- and cycle-consistent token transfer for long-range preservation.

    \item Comprehensive quantitative, human, robustness, and cross-backbone evaluations demonstrate \modelname{}'s strong performance in temporally consistent video editing without video-specific training.

\end{itemize}
\section{Related Work}
\label{sec:related}

\subsection{Training-free Video Editing}
Training-free video editing often repurposes text-to-image diffusion models~\cite{rombach2022high,podell2024sdxl}. These methods invert the source video into noisy latents~\cite{song2021denoising} and edit the resulting trajectory under temporal consistency constraints. Existing approaches can be categorized by the stage at which they intervene. Attention-level methods~\cite{qi2023fatezero,geyer2024tokenflow,cong2024flatten,yang2025videograin,wang2025videodirector,jeong2024ground} extend temporal context by appending KV states from neighboring frames using correspondence information~\cite{geyer2024tokenflow}, optical flow~\cite{cong2024flatten}, masks~\cite{yang2025videograin}, \etc These methods typically reuse inversion features during generation through attention fusion~\cite{qi2023fatezero} or spatio-temporal guidance~\cite{wang2025videodirector}. At the token level, VidToMe~\cite{li2024vidtome} performs local and global token merging to improve consistency. Noise- or latent-level approaches~\cite{lu2024fuse,kara2024rave,teodoro2026primedit,cohen2024slicedit} alter denoising dynamics through latent fusion~\cite{lu2024fuse}, stochastic rearrangements~\cite{kara2024rave}, instance-aware scheduling~\cite{teodoro2026primedit}, or spatio-temporal slicing~\cite{cohen2024slicedit}. More recent works~\cite{cai2025dfvedit,chen2026contextflow,li2025five,jiao2026streamedit,li2026flowdirector} operate on video-native generators~\cite{wan2025wan,kong2024hunyuanvideo,hacohen2024ltx,jin2025pyramidal}, editing directly in latent or flow spaces through trajectory manipulation~\cite{li2025five}, context augmentation~\cite{chen2026contextflow}, or source-conditioned streaming generation~\cite{jiao2026streamedit}. Orthogonal pipelines~\cite{ku2024anyv2v} combine image editors~\cite{brooks2023instructpix2pix,wang2024instantid,chen2024anydoor} with image-to-video models~\cite{ren2024consisti2v,chen2024seine,zhang2023i2vgen} to obtain temporally consistent edits without per-video optimization.\looseness-1

These works expose a central trade-off: image-prior methods offer editability but require explicit temporal coupling, while video-prior methods inherit temporal coherence but remain constrained by their underlying generators. As a result, strong image-editing priors remain underexplored for video editing. \modelname{} addresses this gap through sparse causal memory, correspondence-based token injection, and soft latent blending.\looseness-1

\subsection{Subject-Guided Generation \& Editing}
Subject-guided generation has advanced through reference-conditioned generators~\cite{sang2026lynx,jiang2025vace,pan2026id}, personalization methods based on optimization or staged tuning~\cite{abdal2025dynamic,huang2025conceptmaster}, and scalable multi-subject conditioning~\cite{chen2025multi}. Training-based subject-guided editing~\cite{huang2025dive,gu2024videoswap,lin2026kiwiedit,jiang2025vace} learns subject-aware modules or adapters for identity-preserving replacement and personalized edits, often using learned correspondences~\cite{huang2025dive,gu2024videoswap} or unified instruction--reference architectures~\cite{lin2026kiwiedit,jiang2025vace}. In contrast, training-free approaches avoid training or test-time tuning by transferring subject cues in diffusion feature space~\cite{chen2025zero} or propagating edits from key frames with image-to-video models~\cite{ku2024anyv2v}.\looseness-1

Although these works show that subject-guided editing benefits from reference cues, existing approaches require learned subject modules, test-time tuning, diffusion-feature transfer, or external image-to-video propagation. \modelname{} instead preserves subject appearance within a training-free framework through correspondence-based token injection from an anchor frame, leveraging the position-disentangled behavior of MM-DiT visual tokens without optimization.\looseness-1

\section{Preliminaries}
\label{sec:prelim}

\noindent\textbf{Conditional Flow Matching.}
\begin{figure*}[t]
\centering
\includegraphics[width=0.97\linewidth]{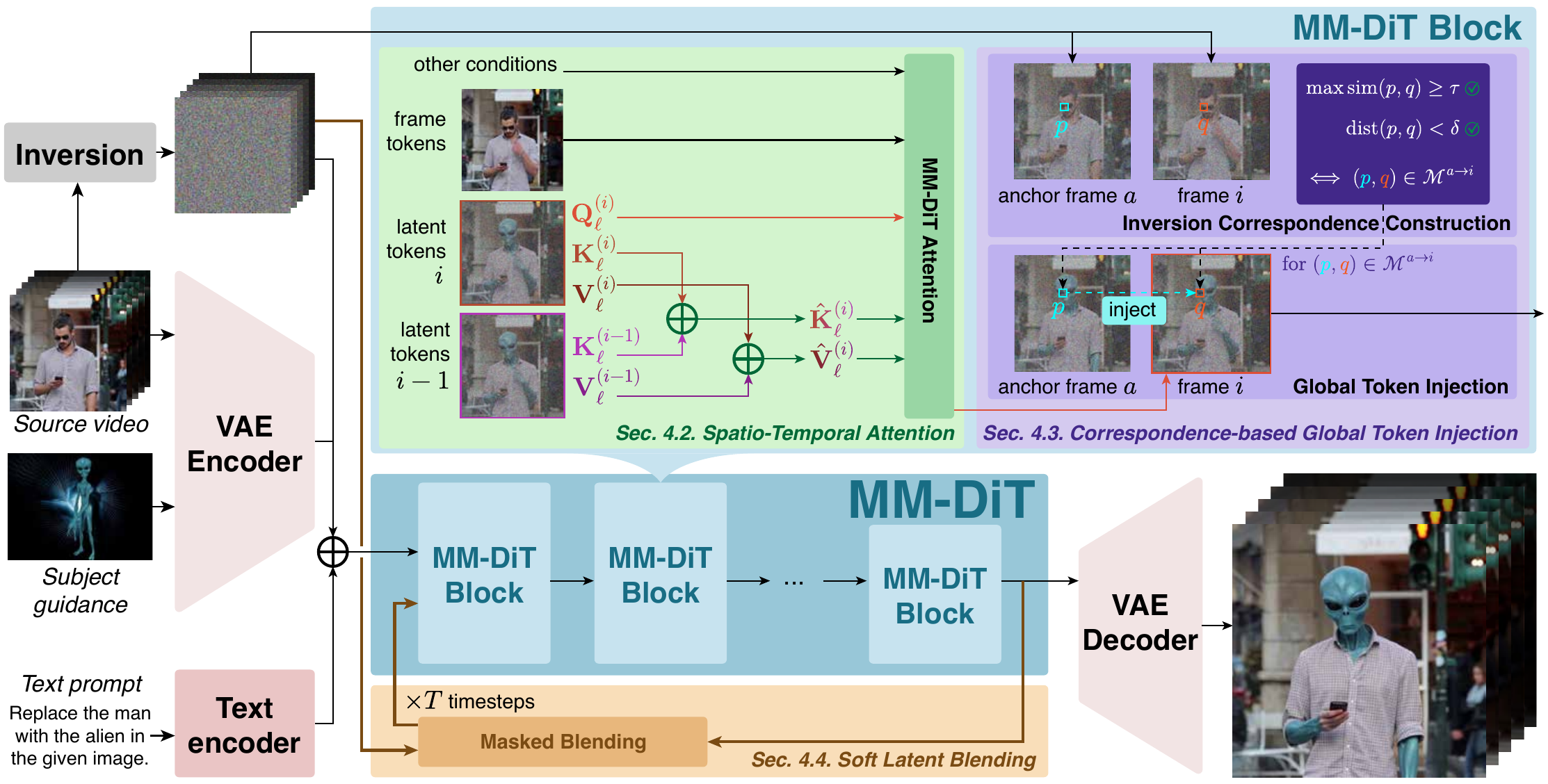}
\caption{\textbf{\modelname{} overview.} Given a source video and optional subject guidance, our training-free framework injects sparse spatio-temporal cues into a frozen MM-DiT. During attention, sparse causal memory reuses selected key--value states from the previous frame, while correspondence-based global token injection preserves subject identity and spatial alignment from an anchor frame. Soft latent blending preserves unedited regions while keeping the edit driven by the target prompt or reference.}
\label{fig:method_overview}
\end{figure*}

Let $\pi_1=\mathcal{N}(0,I)$ and $\pi_0(\cdot\mid \mathbf{c})=p_{\text{data}}(\cdot\mid \mathbf{c})$.
We define a conditional probability path $\{\pi_t(\cdot\mid \mathbf{c})\}_{t\in[0,1]}$ and learn a conditional vector field
\begin{equation}
    \dot{\mathbf{z}}(t)=v_\theta(\mathbf{z}(t),t,\mathbf{c}),\qquad \mathbf{z}(1)\sim\pi_1,
\end{equation}
whose induced flow $\phi^{\mathbf{c}}_{0\leftarrow 1}$ maps $\mathbf{z}(1)$ to $\mathbf{z}(0)$, \ie $\mathbf{z}(0)=\phi^{\mathbf{c}}_{0\leftarrow 1}(\mathbf{z}(1))$, with pushforward $(\phi^{\mathbf{c}}_{0\leftarrow 1})_\#\pi_1$.
Given a target velocity $\mathbf{u}(\mathbf{z},t,\mathbf{c})$ consistent with $\{\pi_t(\cdot\mid\mathbf{c})\}$ (\ie satisfying the conditional continuity equation), conditional flow matching minimizes
\begin{equation}
    \begin{aligned}
        \mathcal{L}_{\text{CFM}}(\theta)
        &=\mathbb{E}_{\mathbf{c}}\,
        \mathbb{E}_{t\sim\mathcal{U}[0,1]}\,
        \mathbb{E}_{\mathbf{z}\sim \pi_t(\cdot\mid\mathbf{c})} \\
        &\quad \Big[
        \big\|v_\theta(\mathbf{z},t,\mathbf{c})
        -\mathbf{u}(\mathbf{z},t,\mathbf{c})\big\|_2^2
        \Big].
    \end{aligned}
\end{equation}
In practice, $\mathbf{z}\sim\pi_t(\cdot\mid\mathbf{c})$ is obtained by sampling $(\mathbf{z}_0,\mathbf{z}_1)\sim q(\mathbf{z}_0,\mathbf{z}_1\mid\mathbf{c})$ with $\mathbf{z}_0\sim p_{\text{data}}(\cdot\mid\mathbf{c})$, $\mathbf{z}_1\sim\mathcal{N}(0,I)$, and setting $\mathbf{z}=\psi_t(\mathbf{z}_0,\mathbf{z}_1)$, $\mathbf{u}=\partial_t \psi_t(\mathbf{z}_0,\mathbf{z}_1)$.
Here, \(\mathbf{z}\) denotes a flow-matching latent state. In the MM-DiT editor, \(\mathbf{Z}^{(i)}\) denotes the VAE latent grid of frame \(i\), while \(\mathbf{x}^{(i)}(t)\) denotes visual tokens obtained by packing/projecting the noised VAE latent grid at diffusion time \(t\).

\noindent\textbf{Multimodal Diffusion Transformers (MM-DiT).}
We parameterize the conditional velocity field with MM-DiT~\cite{peebles2023scalable, esser2024scaling} operating on VAE latent tokens.
Given a frame $\mathbf{X}^{(i)}\in\mathbb{R}^{H\times W\times 3}$, a frozen autoencoder $\mathcal{E}$ with downsampling factor $s$ produces a VAE latent grid
$
\mathbf{Z}^{(i)}=\mathcal{E}(\mathbf{X}^{(i)})
\in
\mathbb{R}^{\frac{H}{s}\times\frac{W}{s}\times D'}
$
which is noised at diffusion time \(t\), packed, and projected into MM-DiT visual tokens $\mathbf{x}^{(i)}(t)\in\mathbb{R}^{N_v\times D}$, where $N_v=\frac{H}{s}\frac{W}{s}$.
Conditioning inputs (\eg text, reference images) are encoded as tokens $\mathbf{c}^{(i)}\in\mathbb{R}^{N_c\times D}$.

MM-DiT architectures typically process these tokens in two stages.
Early \emph{double-stream} blocks maintain separate visual and conditioning streams with cross-modal interaction, while later \emph{single-stream} blocks apply self-attention to the concatenated sequence $\mathbf{h}^{(i)}_{\ell-1}(t) = \big[\mathbf{x}^{(i)}_{\ell-1}(t)\,;\,\mathbf{c}^{(i)}_{\ell-1}\big]$.
At each layer, diffusion time $t$ modulates token features, and positional information is injected through rotary positional embeddings (RoPE)~\cite{su2024roformer}.
Given a token at a multi-dimensional coordinate $\mathbf{p}$, RoPE applies a coordinate-dependent rotation matrix $\mathcal{R}(\mathbf{p})$ to its query $\mathbf{q}$ and key $\mathbf{k}$:
\begin{equation}
\widetilde{\mathbf{q}} = \mathcal{R}(\mathbf{p})\mathbf{q},
\qquad
\widetilde{\mathbf{k}} = \mathcal{R}(\mathbf{p})\mathbf{k}.
\end{equation}
This phase rotation explicitly encodes relative positional geometry into the attention similarity, providing a unified coordinate system across the token streams.
Although attention is performed jointly over both streams, the output head predicts velocity updates exclusively for the visual tokens.
Thus, the conditioning tokens act purely as semantic context, allowing the transformer to realize the conditional field
$
v_\theta\!\big(\mathbf{x}^{(i)}(t), t, \mathbf{c}^{(i)}\big)
$
through repeated attention between editable visual tokens and fixed conditioning.

\section{Methodology}
\label{sec:method}

\subsection{Overview}

Let $\mathcal{V}=\{\mathbf{X}^{(i)}\}_{i=1}^N$ denote an input video of $N$ frames, and let
$\mathbf{x}^{(i)}(t)\in\mathbb{R}^{N_v\times D}$ be the visual latent tokens of frame $i$ at diffusion time $t$.
As illustrated in \cref{fig:method_overview}, our method performs video editing directly in the \emph{visual latent manifold}. 
Temporal interactions are introduced only through the visual latent tokens $\mathbf{x}^{(i)}(t)$, while the conditioning tokens $\mathbf{c}^{(i)}$ (\eg text or reference-image embeddings) remain unchanged and serve purely as semantic context.

This separation between editable visual tokens and fixed conditioning tokens is also reflected in the 4D-RoPE parameterization. 
Each token is assigned a coordinate $\mathbf{p}=(t_p,h,w,\ell_p)$, allowing visual latent tokens and conditioning tokens to occupy distinct coordinate slots in the shared attention space.
Our temporal module operates exclusively on attention states derived from the visual tokens, leaving the conditioning stream and its positional assignments unchanged.
Thus, semantic guidance from text or reference images is preserved, while temporal information is propagated entirely through the editable visual latent tokens.

\subsection{Spatio-Temporal Attention}

To propagate temporal information through the latent stream, we introduce a sparse causal context in visual latent space.
For each frame \(i\), we construct a compact latent context
\begin{equation}
\mathcal{C}^{(i)}_x(t)=
\{\mathbf{x}^{(i-1)}(t)\}, \quad i>1,
\end{equation}
which contains visual latent tokens from the immediately preceding frame.
This previous-frame anchor promotes temporal smoothness by allowing information to propagate causally across frames.
The frame-wise latent dynamics thus follow\looseness-1
\begin{equation}
\frac{d\mathbf{x}^{(i)}(t)}{dt}
=
v_\theta\!\left(\mathbf{x}^{(i)}(t),\, t,\, \mathbf{c}^{(i)},\, \mathcal{C}^{(i)}_x(t)\right),
\end{equation}

Within the MM-DiT backbone, this temporal context is realized by augmenting the attention context of the current frame with visual latent tokens from the previous frame.
Leveraging RoPE, tokens from consecutive frames maintain consistent positional relationships in the attention space, allowing key--value states from the previous frame to serve as temporal context. 
Since RoPE represents positions through relative phase rotations, tokens with small temporal offsets remain geometrically compatible in the attention similarity.

Let $\mathbf{Q}^{(i)}_\ell$, $\mathbf{K}^{(i)}_\ell$, and $\mathbf{V}^{(i)}_\ell$ denote query, key, and value tensors for frame $i$ at layer $\ell$.
For $i>1$, we form
\begin{equation}
\widehat{\mathbf{K}}^{(i)}_\ell
=
\big[
\mathbf{K}^{(i)}_\ell;\,
\mathbf{K}^{(i-1)}_\ell
\big],
\qquad
\widehat{\mathbf{V}}^{(i)}_\ell
\!=\!
\big[
\mathbf{V}^{(i)}_\ell;\,
\mathbf{V}^{(i-1)}_\ell
\big],
\end{equation}
with a no-anchor specialization for \(i=1\).
The updated representation is then
\begin{equation}
\mathbf{x}^{(i)}_\ell(t)
\!=\!
\mathrm{Attn}\!\left(
\mathbf{Q}^{(i)}_\ell,\,
\widehat{\mathbf{K}}^{(i)}_\ell,\,
\widehat{\mathbf{V}}^{(i)}_\ell
\right).
\end{equation}

This design introduces a recurrent temporal inductive bias in which each frame inherits motion and edit continuity from its predecessor.
Because the temporal context is limited to the immediately preceding frame, the conditioning footprint remains constant with respect to video length.
The resulting local temporal dependency can be written as
\begin{equation}
    \begin{aligned}
        p(\mathbf{X}^{(1:N)} \mid \mathbf{c})
        &=
        p(\mathbf{X}^{(1)} \mid \mathbf{c}^{(1)}) \\
        &\quad\times
        \prod_{i=2}^{N}
        p\!\left(
        \mathbf{X}^{(i)}
        \mid
        \mathbf{X}^{(i-1)}, \mathbf{c}^{(i)}
        \right).
    \end{aligned}
\end{equation}
allowing editing to be rolled out to long videos while keeping context and computation limited.

\subsection{Correspondence-based Token Injection}

While causal attention propagates temporal information across frames, it does not explicitly enforce spatial consistency across corresponding regions. To better stabilize object identity and appearance, we introduce correspondence-based global token injection that transfers anchor-frame information to matched spatial locations across the video. The method consists of: (1) inversion-based correspondence construction for reliable anchor-to-frame patch matches, (2) global token injection for temporally aligned feature transfer, and (3) correspondence dropout for robustness to noisy matches.\looseness-1

\noindent\textbf{Inversion correspondence construction.}
Given the input video \(\mathcal{V}=\{\mathbf{X}^{(i)}\}_{i=1}^{N}\), we first perform inversion and extract token sets
\(\{\mathbf{g}^{(i)}_{p}\}_{p=1}^{N_v}\) from the penultimate double-stream block of the inverted trajectory at a fixed correspondence time \(t_{\mathrm{corr}}=0.25\).
Let \(a=1\) denote the anchor frame. For each target frame \(i\), we compute a similarity matrix between the anchor frame and frame \(i\):
\begin{equation}
S^{a\rightarrow i}_{pq}
=
\mathrm{sim}\!\left(\mathbf{g}^{(a)}_{p},\mathbf{g}^{(i)}_{q}\right),
\label{eq:inv_sim_matrix}
\end{equation}
where $\mathrm{sim}(\cdot,\cdot)$ denotes cosine similarity between patch tokens.
To obtain reliable correspondences, we use nearest-neighbor matching with confidence and cycle-consistency constraints. Specifically,
\begin{equation}
q^{*}(p)=\arg\max_q S^{a\rightarrow i}_{pq},
\quad
p^{*}(q)=\arg\max_p S^{a\rightarrow i}_{pq}.
\end{equation}
and we accept a pair \((p,q^{*}(p))\) into \(\mathcal{M}^{a\rightarrow i}\) if 
\begin{equation}
\begin{aligned}
(p,q^{*}(p))\in\mathcal{M}^{a\to i}
\Longleftrightarrow\;&
S^{a\to i}_{p q^{*}(p)}\geq\tau,\\
&\hspace{-2cm}
\|\mathrm{pos}(p)-\mathrm{pos}(p^{*}(q^{*}(p)))\|_2<r.
\end{aligned}
\label{eq:inv_cycle_constraints}
\end{equation}
Here, \(\mathrm{pos}(\cdot)\) maps a patch index to its 2D spatial coordinate, while \(\tau\) and \(r\) control the similarity threshold and cycle-consistency radius, respectively.
This procedure yields a reliable correspondence map \(\mathcal{M}^{a\rightarrow i}\) between the anchor frame and each target frame. The same map is reused across all layers where global token injection is applied.

\noindent\textbf{Global token injection.}
During editing, we use the anchor-to-frame correspondence maps to propagate token information across the full video. 
Let \(\{\mathbf{h}^{(i)}_{\ell,t,q}\}_{q=1}^{N_v}\) denote the denoising tokens of frame \(i\) at layer \(\ell\) and diffusion time \(t\).
For each retained correspondence \((p,q)\in\mathcal{M}^{a\rightarrow i}\), we replace the target-frame token with the matched anchor token:
\begin{equation}
\tilde{\mathbf{h}}^{(i)}_{\ell,t,q}
\!=\!
\mathbf{h}^{(a)}_{\ell,t,p},
\qquad
\forall (p,q)\in\mathcal{M}^{a\rightarrow i}, \quad \ell\in\mathcal{L}_{\mathrm{inj}}.
\label{eq:global_token_replace}
\end{equation}
Unmatched locations retain their original representations, \ie
\(\tilde{\mathbf{h}}^{(i)}_{\ell,t,q}=\mathbf{h}^{(i)}_{\ell,t,q}\) when no valid anchor correspondence maps to location \(q\).
This correspondence-based global token injection allows each frame to receive aligned appearance information from the anchor frame, improving temporal consistency under large motion and occlusion.

\noindent\textbf{Correspondence dropout.}
To improve robustness to imperfect correspondences and prevent over-reliance on a fixed match set, we randomly subsample the valid correspondences following~\cite{yao2025freegraftor}.
Specifically, we retain a fraction \(1-p_{\mathrm{drop}}\) of \(\mathcal{M}^{a\rightarrow i}\) and apply Eq.~\eqref{eq:global_token_replace} only to this subset.

Together with causal attention, global token injection stabilizes object identity and appearance by enforcing sparse, spatially aligned feature propagation across the video.

\subsection{Soft Latent Blending}
In addition to token-level temporal propagation, we apply soft latent blending to preserve unedited regions.
Following prior work~\citep{avrahami2023blended}, we blend the edited latent trajectory $\mathbf{x}^{(i)}_k$ with the source trajectory $\mathbf{x}^{(i),\mathrm{src}}_k$ (obtained from inversion) at each discrete denoising step $k$.
To avoid rigidly masking out moving subjects, we compute a soft spatial mask derived from the continuous latent changes.

Specifically, we compute a per-token \(L_1\) difference map between the edited and source latents. This map is normalized using low and high quantiles and optionally smoothed with a Gaussian filter to yield a soft mask \(\hat{m}^{(i)}_k \in [0, 1]\) (more details are provided in Appendix~\ref{sec:app_soft_blending}).
The effective preservation weight is defined as \(w^{(i)}_{k,p} = \alpha\left(1-\hat{m}^{(i)}_{k,p}\right)\), where \(\alpha\in[0,1]\) controls the blending strength.
Finally, we blend the edited and source visual tokens:
\begin{equation}
\tilde{\mathbf{x}}^{(i)}_{k,p}
=
\left(1-w^{(i)}_{k,p}\right)\mathbf{x}^{(i)}_{k,p}
+
w^{(i)}_{k,p}\mathbf{x}^{(i),\mathrm{src}}_{k,p}.
\label{eq:latent_blending}
\end{equation}
This softly anchors regions that remain close to the source trajectory, while allowing regions with large latent changes to follow the target edit, effectively preserving background structure in unedited areas.\looseness-1

\begin{table*}[t]

\centering

\caption{\textbf{Quantitative comparison on FiVE~\cite{li2025five} benchmark and metrics}. Highlighted values are \colorbox{best}{\textbf{best}} within each method section. $^{*}$requires additional depth maps, and $^{\dagger}$requires additional masks.}

\setlength{\tabcolsep}{9pt}

\resizebox{0.99\linewidth}{!}{%

\begin{tabular}{lcc ccccc}

\toprule

\multicolumn{1}{c}{\textbf{Method}} &

\multicolumn{1}{c}{\textbf{Foundation}} &

\multicolumn{1}{c}{\textbf{Subject-guided}} &

\multicolumn{5}{c}{\textbf{FiVE-Acc Metrics}~\cite{li2025five}} \\

\cmidrule(lr){4-8}

& \multicolumn{1}{c}{\textbf{Model}} &

\multicolumn{1}{c}{\textbf{Edits}} &

FiVE-YN \(\uparrow\) & FiVE-MC \(\uparrow\) & FiVE-U \(\uparrow\) & FiVE-\(\cap\) \(\uparrow\) & FiVE-Acc \(\uparrow\) \\

\midrule

Pyramid-Edit~\cite{li2025five} 

& Video & \xmark 

& 33.67 

& 54.01 

& 56.36

& 31.31 

& 43.84 \\

Wan-Edit~\cite{li2025five} 

& Video & \xmark 

& 41.41 

& 52.53 

& 55.72 

& 38.22 

& 46.97 \\

FlowDirector~\cite{li2026flowdirector}

& Video & \xmark

& \cellcolor{best}\textbf{49.16}

& \cellcolor{best}\textbf{68.74}

& \cellcolor{best}\textbf{70.41}

& \cellcolor{best}\textbf{47.49}

& \cellcolor{best}\textbf{58.95} \\

StreamEdit (SF)~\cite{jiao2026streamedit}

& Video & \xmark

& 43.09

& 61.09

& 64.11

& 40.07

& 51.94 \\

\midrule

DMT*~\cite{yatim2024space} 

& Image & \xmark 

& \cellcolor{best}\textbf{34.78} 

& \cellcolor{best}\textbf{62.06} 

& \cellcolor{best}\textbf{62.98} 

& \cellcolor{best}\textbf{33.86} 

& \cellcolor{best}\textbf{48.42} \\

VideoGrain\(^\dagger\)~\cite{yang2025videograin} 

& Image & \xmark 

& 30.50 

& 43.97 

& 44.30 

& 30.17 

& 37.23 \\

\midrule

TokenFlow~\cite{geyer2024tokenflow} 

& Image & \xmark 

& 19.36 

& 35.51 

& 36.68 

& 18.18 

& 27.43 \\

VidToMe~\cite{li2024vidtome} 

& Image & \xmark 

& 20.03 

& 33.50 

& 36.20 

& 17.34 

& 26.77 \\

AnyV2V~\cite{ku2024anyv2v} 

& Hybrid & \cmark 

& 30.62 

& 45.42 

& 48.96 

& 27.09 

& 38.02 \\

\modelnamebf\ \textbf{(Ours)} 

& Image & \cmark

& \cellcolor{best}\textbf{69.45} 

& \cellcolor{best}\textbf{86.87} 

& \cellcolor{best}\textbf{88.78} 

& \cellcolor{best}\textbf{67.54} 

& \cellcolor{best}\textbf{78.16} \\

\bottomrule

\end{tabular}%

}
\vspace{-0.3cm}

\label{tab:five_results}

\end{table*}

\section{Experiments}
\label{sec:experiments}

\subsection{Experimental Setup}

\label{sec:exp_setup}
We evaluate \modelname{} on FiVE~\cite{li2025five} and IVEBench~\cite{chen2026ivebench} following the official benchmark protocols, and a curated 50-video set comprising 24 general video-editing examples and 26 subject-guided editing examples. The curated set spans diverse edit types and motion patterns beyond the benchmark-specific examples and is used for our mask-free VLM evaluation and human study. On FiVE, we compare against training-free image-~\cite{yang2025videograin,geyer2024tokenflow,li2024vidtome,yatim2024space}, video-~\cite{li2025five,li2026flowdirector}, and hybrid-prior~\cite{ku2024anyv2v} methods. On IVEBench and the mask-free VLM evaluation, we additionally compare against training-based video editors~\cite{lin2026kiwiedit,jiang2025vace,liao2025context,bai2026qwen}. Unless stated otherwise, \modelname{} uses FLUX.2-Klein-9B with four denoising steps. Additional implementation details are reported in Appendix~\ref{sec:app_implementation_details}.

\subsection{Comparison with State-of-the-Art Methods}
\label{sec:quant}

\noindent\textbf{Edit correctness on FiVE.}
\cref{tab:five_results} shows that \modelnamebf{} achieves the highest overall FiVE-Acc among all training-free methods, \textbf{improving the strongest prior result from 58.95 to 78.16}. The gains are particularly pronounced for the multi-choice, open-ended, and intersection variants, indicating more reliable execution of fine-grained object-level edits rather than merely stronger low-level source similarity.\looseness-1

\begin{table*}[t]
\centering
\setlength{\tabcolsep}{5pt}
\caption{\textbf{Quantitative comparison on FiVE~\cite{li2025five}}. Highlighted values are \colorbox{best}{\textbf{best}} within each method section. $^{*}$requires additional depth maps and $^{\dagger}$requires additional masks. Source Videos row shown for reference.}
\resizebox{0.99\linewidth}{!}{%
\begin{tabular}{lcc*{9}{c}}
\toprule
\multicolumn{1}{c}{\textbf{Method}} &
\multicolumn{1}{c}{\textbf{Found.}} &
\multicolumn{1}{c}{\textbf{Subj.}} &
\multicolumn{9}{c}{\textbf{FiVE Traditional Metrics}~\cite{li2025five}} \\
\cmidrule(lr){4-12}
& \multicolumn{1}{c}{\textbf{Model}}
& \multicolumn{1}{c}{\textbf{Edits}}
& Str. Dist. $\downarrow$
& PSNR $\uparrow$
& LPIPS $\downarrow$
& MSE $\downarrow$
& SSIM $\uparrow$
& CLIP$_S$ $\uparrow$
& CLIP$_{S}^{\text{edit}}$ $\uparrow$
& NIQE $\downarrow$
& MFS $\uparrow$ \\
\midrule
\rowcolor{rowgray}
\textcolor{gray}{Source Videos} & \textcolor{gray}{--} & \textcolor{gray}{--} & 0 & inf & 0 & 0 & 100 & 24.59 & 19.87 & 6.33 & 93.76 \\
\midrule

Pyramid-Edit~\cite{li2025five} & Video & \xmark
& 28.65 & 20.84 & 276.59 & 95.63 & 71.72
& 26.82 & 20.20 & 5.48 & 80.59 \\

Wan-Edit~\cite{li2025five} & Video & \xmark
& 12.53 & 25.57 & 94.61 & 41.84 & 82.55
& 26.39 & 21.23 & 6.54 & 89.43 \\

FlowDirector~\cite{li2026flowdirector} & Video & \xmark

& 21.56

& 22.74

& 107.09

& 65.20

& 81.68

& \cellcolor{best}\textbf{27.93}

& \cellcolor{best}\textbf{22.00}

& 4.98

& 81.22 \\

StreamEdit (SF) & Video & \xmark

& \cellcolor{best}\textbf{10.27}

& \cellcolor{best}\textbf{28.47}

& \cellcolor{best}\textbf{49.84}

& \cellcolor{best}\textbf{21.88}

& \cellcolor{best}\textbf{87.52}

& 26.95

& 21.73

& \cellcolor{best}\textbf{4.38}

& \cellcolor{best}\textbf{90.93} \\

\midrule

DMT*~\cite{yatim2024space} & Image & \xmark
& 85.95 & 14.71 & 404.60 & 372.78 & 51.64
& \cellcolor{best}\textbf{26.66}
& \cellcolor{best}\textbf{21.44}
& 5.24
& 82.30 \\

VideoGrain\(^{\dagger}\)~\cite{yang2025videograin} & Image & \xmark
& \cellcolor{best}\textbf{12.40}
& \cellcolor{best}\textbf{27.05}
& \cellcolor{best}\textbf{185.21}
& \cellcolor{best}\textbf{25.10}
& \cellcolor{best}\textbf{79.13}
& 25.69
& 20.31
& \cellcolor{best}\textbf{4.08}
& \cellcolor{best}\textbf{88.57} \\

\midrule

TokenFlow~\cite{geyer2024tokenflow} & Image & \xmark
& 35.62 & 19.06 & 263.61 & 138.65 & 72.51
& 26.46 & 21.15
& \cellcolor{best}\textbf{4.01}
& 89.00 \\

VidToMe~\cite{li2024vidtome} & Image & \xmark
& 22.37 & 21.15 & 263.91 & 88.75 & 70.69
& 26.84 & 21.05 & 4.68
& \cellcolor{best}\textbf{90.06} \\

AnyV2V~\cite{ku2024anyv2v} & Hybrid & \cmark
& 71.36 & 15.90 & 348.59 & 342.97 & 50.77
& 24.89 & 19.72 & 5.04 & 60.36 \\

\modelnamebf\ \textbf{(Ours)} & Image & \cmark
& \cellcolor{best}\textbf{17.10}
& \cellcolor{best}\textbf{25.97}
& \cellcolor{best}\textbf{218.71}
& \cellcolor{best}\textbf{36.37}
& \cellcolor{best}\textbf{83.31}
& \cellcolor{best}\textbf{28.83}
& \cellcolor{best}\textbf{22.40}
& 4.17
& 85.31 \\

\bottomrule
\end{tabular}%
}
\label{tab:five_traditional}
\end{table*}
                   
\noindent\textbf{Conventional Metrics on FiVE.}
\cref{tab:five_traditional} provides complementary diagnostics of source preservation, semantic alignment, perceptual quality, and motion fidelity. Among image/hybrid methods without auxiliary inputs, \textbf{\modelname{} achieves the strongest performance across reconstruction, perceptual quality, and semantic-alignment metrics, while remaining competitive in motion fidelity.} These results show that \modelname{} preserves strong low-level fidelity and semantic alignment despite relying only on an image MM-DiT, complementing its superior FiVE-Acc in \cref{tab:five_results}, which serves as the primary measure of edit correctness.\looseness-1

\begin{table*}[t]
\centering
\caption{\textbf{Quantitative evaluation on IVEBench and our curated video-editing set.}
TF indicates training-free methods. IVEBench columns report Total Score, Video Quality, Instruction Compliance, and Video Fidelity. PF, EQ, and BC denote Prompt Following, Edit Quality, and Background Consistency. Highlighted values are \colorbox{best}{\textbf{best}} within each method section.}
\label{tab:ivebench_vlm}
\small
\setlength{\tabcolsep}{4.0pt}
\renewcommand{\arraystretch}{0.92}
\resizebox{\textwidth}{!}{
\begin{tabular}{lccccccccccc}
\toprule

\textbf{Method}
& \textbf{TF}
& \multicolumn{4}{c}{\textbf{IVEBench}}
& \multicolumn{3}{c}{\textbf{Subject-guided}}
& \multicolumn{3}{c}{\textbf{General Video Editing}} \\
\cmidrule(lr){3-6}
\cmidrule(lr){7-9}
\cmidrule(lr){10-12}
&
& \textbf{Total}
& \textbf{Quality}
& \textbf{Compliance}
& \textbf{Fidelity}
& \textbf{PF} $\uparrow$
& \textbf{EQ} $\uparrow$
& \textbf{BC} $\uparrow$
& \textbf{PF} $\uparrow$
& \textbf{EQ} $\uparrow$
& \textbf{BC} $\uparrow$ \\
\midrule

LucyEditDev~\cite{liao2025context}
& \xmark
& 0.6220
& \cellcolor{best}\textbf{0.8164}
& 0.3679
& 0.6817
& 2.019
& 3.731
& 6.596
& 4.667
& 6.292
& 8.542 \\

Kiwi-Edit-5B~\cite{lin2026kiwiedit}
& \xmark
& \cellcolor{best}\textbf{0.6936}
& 0.8101
& 0.5182
& \cellcolor{best}\textbf{0.7524}
& \cellcolor{best}\textbf{3.115}
& \cellcolor{best}\textbf{4.327}
& 6.596
& \cellcolor{best}\textbf{8.000}
& \cellcolor{best}\textbf{8.229}
& \cellcolor{best}\textbf{9.458} \\

Qwen-Video-Edit~\cite{bai2026qwen}
& \xmark
& 0.6765
& 0.7752
& \cellcolor{best}\textbf{0.5256}
& 0.7286
& 1.865
& 2.654
& \cellcolor{best}\textbf{7.654}
& 7.812
& 7.979
& 8.958 \\

\midrule

VidToMe~\cite{li2024vidtome}
& \cmark
& 0.6297
& 0.7826
& 0.4010
& 0.7056
& 4.308
& 5.942
& 5.923
& 3.875
& 5.938
& 5.250 \\

WANEdit~\cite{li2025five}
& \cmark
& 0.6072
& 0.7973
& 0.2801
& 0.7441
& 5.962
& 6.654
& 8.942
& 5.125
& 6.625
& 8.375 \\

FlowDirector~\cite{li2026flowdirector}
& \cmark
& 0.6264
& 0.7771
& 0.3765
& 0.7255
& 4.019
& 5.865
& 8.058
& 4.021
& 5.521
& 6.875 \\

StreamEdit~\cite{jiao2026streamedit}
& \cmark
& 0.6229
& \cellcolor{best}\textbf{0.8238}
& 0.2921
& 0.7530
& 4.077
& 6.673
& 8.288
& 5.250
& 7.000
& \cellcolor{best}\textbf{9.792} \\

\modelnamebf{} \textbf{(Ours)}
& \cmark
& \cellcolor{best}\textbf{0.6847}
& 0.8114
& \cellcolor{best}\textbf{0.4587}
& \cellcolor{best}\textbf{0.7839}
& \cellcolor{best}\textbf{7.769}
& \cellcolor{best}\textbf{7.846}
& \cellcolor{best}\textbf{9.558}
& \cellcolor{best}\textbf{9.458}
& \cellcolor{best}\textbf{9.000}
& 9.500 \\

\bottomrule
\end{tabular}
}
\vspace{-0.4cm}
\end{table*}
\noindent\textbf{Broader video-editing evaluation.}
We further evaluate \modelname{} on IVEBench and our curated subject-guided and general video-editing set. As shown in \cref{tab:ivebench_vlm}, \modelname{} achieves the strongest training-free IVEBench results in Total score, Instruction Compliance, and Video Fidelity, while ranking second overall in Total score and obtaining the best Video Fidelity across all methods. On the curated VLM evaluation, \modelname{} leads five of six criteria, including all subject-guided metrics and general-video Prompt Following and Edit Quality, while attaining the second-highest Background Consistency. Together, these results show that \modelname{} performs strongly across both instruction-guided and subject-guided editing while preserving video fidelity.

\definecolor{usEditVid}{HTML}{6A3D9A}
\definecolor{usStreamEdit}{HTML}{1F78B4}
\definecolor{usWANEdit}{HTML}{33A02C}
\definecolor{usTokenFlow}{HTML}{FF7F00}
\definecolor{usFlowDirector}{HTML}{E31A1C}
\definecolor{usVidToMe}{HTML}{A6CEE3}
\definecolor{usPyramidEdit}{HTML}{B2DF8A}
\definecolor{usAnyV2V}{HTML}{FDBF6F}

\begin{figure}[t]
    \centering
    \small
    \newcommand{\uslegendentry}[2]{%
        \tikz[baseline=-0.5ex]\fill[#1] (0,0) rectangle (0.20,0.14);%
        \hspace{0.35em}{\scriptsize #2}%
    }
    \begin{tabular*}{\linewidth}{@{\extracolsep{\fill}}llll@{}}
        \uslegendentry{usEditVid}{\modelnamebf{} \textbf{(Ours)}} &
        \uslegendentry{usStreamEdit}{StreamEdit} &
        \uslegendentry{usWANEdit}{WANEdit} &
        \uslegendentry{usTokenFlow}{TokenFlow} \\
        \uslegendentry{usFlowDirector}{FlowDirector} &
        \uslegendentry{usVidToMe}{VidToMe} &
        \uslegendentry{usPyramidEdit}{PyramidEdit} &
        \uslegendentry{usAnyV2V}{AnyV2V}
    \end{tabular*}

    \vspace{5pt}
    \begin{minipage}{\linewidth}
        \raggedright
        Overall preference\par\nobreak\vspace{1pt}
        \begin{tikzpicture}[x=\linewidth,y=1cm]
            \fill[usEditVid]      (0.0000,0) rectangle (0.5182,0.34);
            \fill[usStreamEdit]   (0.5182,0) rectangle (0.6091,0.34);
            \fill[usWANEdit]      (0.6091,0) rectangle (0.6864,0.34);
            \fill[usTokenFlow]    (0.6864,0) rectangle (0.7636,0.34);
            \fill[usFlowDirector] (0.7636,0) rectangle (0.8318,0.34);
            \fill[usVidToMe]      (0.8318,0) rectangle (0.9045,0.34);
            \fill[usPyramidEdit]  (0.9045,0) rectangle (0.9614,0.34);
            \fill[usAnyV2V]       (0.9614,0) rectangle (1.0000,0.34);
            \node[font=\scriptsize\bfseries,text=white] at (0.2591,0.17) {51.8};
            \node[font=\tiny,text=white] at (0.5637,0.17) {9.1};
            \node[font=\tiny,text=white] at (0.6477,0.17) {7.7};
            \node[font=\tiny] at (0.7250,0.17) {7.7};
            \node[font=\tiny,text=white] at (0.7977,0.17) {6.8};
            \node[font=\tiny] at (0.8682,0.17) {7.3};
            \node[font=\tiny,rotate=90] at (0.9330,0.17) {5.7};
            \node[font=\tiny,rotate=90] at (0.9807,0.17) {3.9};
        \end{tikzpicture}
        Prompt following\par\nobreak\vspace{1pt}
        \begin{tikzpicture}[x=\linewidth,y=1cm]
            \fill[usEditVid]      (0.0000,0) rectangle (0.5318,0.34);
            \fill[usStreamEdit]   (0.5318,0) rectangle (0.6250,0.34);
            \fill[usWANEdit]      (0.6250,0) rectangle (0.7091,0.34);
            \fill[usTokenFlow]    (0.7091,0) rectangle (0.7795,0.34);
            \fill[usFlowDirector] (0.7795,0) rectangle (0.8477,0.34);
            \fill[usVidToMe]      (0.8477,0) rectangle (0.9136,0.34);
            \fill[usPyramidEdit]  (0.9136,0) rectangle (0.9591,0.34);
            \fill[usAnyV2V]       (0.9591,0) rectangle (1.0000,0.34);
            \node[font=\scriptsize\bfseries,text=white] at (0.2659,0.17) {53.2};
            \node[font=\tiny,text=white] at (0.5784,0.17) {9.3};
            \node[font=\tiny,text=white] at (0.6670,0.17) {8.4};
            \node[font=\tiny] at (0.7443,0.17) {7.0};
            \node[font=\tiny,text=white] at (0.8136,0.17) {6.8};
            \node[font=\tiny] at (0.8807,0.17) {6.6};
            \node[font=\tiny,rotate=90] at (0.9364,0.17) {4.5};
            \node[font=\tiny,rotate=90] at (0.9795,0.17) {4.1};
        \end{tikzpicture}
        Visual quality \& temporal consistency\par\nobreak\vspace{1pt}
        \begin{tikzpicture}[x=\linewidth,y=1cm]
            \fill[usEditVid]      (0.0000,0) rectangle (0.4932,0.34);
            \fill[usStreamEdit]   (0.4932,0) rectangle (0.6023,0.34);
            \fill[usWANEdit]      (0.6023,0) rectangle (0.7068,0.34);
            \fill[usTokenFlow]    (0.7068,0) rectangle (0.7909,0.34);
            \fill[usFlowDirector] (0.7909,0) rectangle (0.8545,0.34);
            \fill[usVidToMe]      (0.8545,0) rectangle (0.9182,0.34);
            \fill[usPyramidEdit]  (0.9182,0) rectangle (0.9568,0.34);
            \fill[usAnyV2V]       (0.9568,0) rectangle (1.0000,0.34);
            \node[font=\scriptsize\bfseries,text=white] at (0.2466,0.17) {49.3};
            \node[font=\tiny,text=white] at (0.5477,0.17) {10.9};
            \node[font=\tiny,text=white] at (0.6545,0.17) {10.5};
            \node[font=\tiny] at (0.7489,0.17) {8.4};
            \node[font=\tiny,text=white] at (0.8227,0.17) {6.4};
            \node[font=\tiny] at (0.8864,0.17) {6.4};
            \node[font=\tiny,rotate=90] at (0.9375,0.17) {3.9};
            \node[font=\tiny,rotate=90] at (0.9784,0.17) {4.3};
        \end{tikzpicture}
        Background preservation\par\nobreak\vspace{1pt}
        \begin{tikzpicture}[x=\linewidth,y=1cm]
            \fill[usEditVid]      (0.0000,0) rectangle (0.5000,0.34);
            \fill[usStreamEdit]   (0.5000,0) rectangle (0.6114,0.34);
            \fill[usWANEdit]      (0.6114,0) rectangle (0.7250,0.34);
            \fill[usTokenFlow]    (0.7250,0) rectangle (0.8023,0.34);
            \fill[usFlowDirector] (0.8023,0) rectangle (0.8705,0.34);
            \fill[usVidToMe]      (0.8705,0) rectangle (0.9273,0.34);
            \fill[usPyramidEdit]  (0.9273,0) rectangle (0.9705,0.34);
            \fill[usAnyV2V]       (0.9705,0) rectangle (1.0000,0.34);
            \node[font=\scriptsize\bfseries,text=white] at (0.2500,0.17) {50.0};
            \node[font=\tiny,text=white] at (0.5557,0.17) {11.1};
            \node[font=\tiny,text=white] at (0.6682,0.17) {11.4};
            \node[font=\tiny] at (0.7636,0.17) {7.7};
            \node[font=\tiny,text=white] at (0.8364,0.17) {6.8};
            \node[font=\tiny,rotate=90] at (0.8989,0.17) {5.7};
            \node[font=\tiny,rotate=90] at (0.9489,0.17) {4.3};
            \node[font=\tiny,rotate=90] at (0.9852,0.17) {3.0};
        \end{tikzpicture}
    \end{minipage}
    \caption{\textbf{User Study.}
    Preference share (\%) across 22 participants and 20 video examples.}
    \label{fig:user}
    \vspace{-0.1cm}
\end{figure}
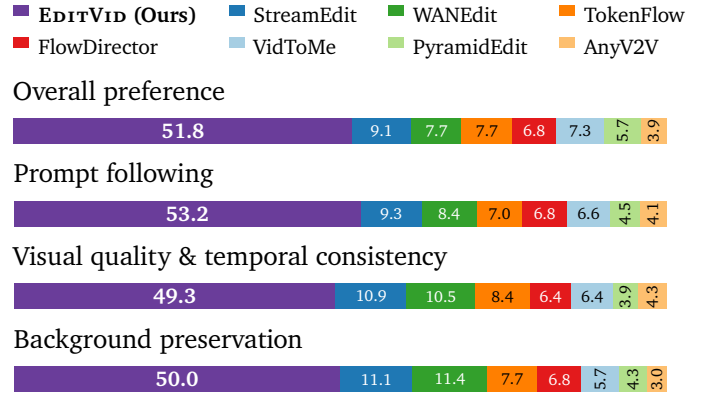
\noindent\textbf{User study.}
We conduct a blind user study with 22 participants over 20 video examples, comparing \modelname{} against seven training-free baselines. For each example, method identities are hidden and the outputs are independently randomized. Participants select the best result according to overall preference, prompt following, visual quality and temporal consistency, and background preservation. This yields 440 participant--video evaluations and 1,760 criterion-level selections. As shown in \cref{fig:user}, \textbf{\modelname{} receives the highest preference share across all four criteria}. These results demonstrate a consistent and substantial human preference for our method across edit accuracy, perceptual quality, temporal coherence, and content preservation.

\subsection{Ablation Studies}
\label{sec:ablation}

\begin{table}[!t]
\centering
\caption{{Module ablation on a FiVE subset} measured by FiVE-Acc and MF-S.}
\label{tab:ablation}
\resizebox{\linewidth}{!}{%
\begin{tabular}{lcccccc}
\toprule

{Method} &
{FiVE-YN}$\uparrow$ &
{FiVE-MC}$\uparrow$ &
{FiVE-U}$\uparrow$ &
{FiVE-$\cap$}$\uparrow$ &
{FiVE-Acc}$\uparrow$ &
{MF-S}$\times 10^{2}\uparrow$ \\
\midrule

\modelname &
{73.33} & {90.00} & {90.00} & {73.33} & {81.67} & {84.88} \\

w/o latent blending &
70.00 & {90.00} & {90.00} & 70.00 & 80.00 & 84.59 \\

w/o spatio-temporal attention &
70.00 & {90.00} & {90.00} & 70.00 & 80.00 & 84.55 \\

w/o global token injection &
{73.33} & {90.00} & {90.00} & {73.33} & {81.67} & 84.72 \\

\bottomrule
\end{tabular}%
\vspace{-0.3cm}
}
\end{table}

\noindent\textbf{Component Analysis.} 
We ablate the core components of our approach on 30 videos sampled across all FiVE edit categories. As shown in \cref{tab:ablation}, the full model achieves the strongest overall combination of edit correctness and motion fidelity. Removing soft latent blending reduces FiVE-Acc from \(81.67\) to \(80.00\), with corresponding drops in FiVE-YN and FiVE-\(\cap\), supporting its role in preserving instruction-irrelevant source content. Removing spatio-temporal attention yields the same FiVE-Acc drop and the lowest MF-S, confirming the importance of local temporal context for motion consistency. Removing global token injection leaves FiVE-Acc unchanged on this subset but lowers MF-S, suggesting that its contribution is more apparent in temporal consistency than in the aggregate edit score. We examine this effect more directly under challenging temporal changes in \cref{tab:correspondence_robustness}.\looseness-1

\begin{figure*}[t!]
\centering
\setlength{\tabcolsep}{0.45pt}
\renewcommand{\arraystretch}{0.62}
\scriptsize
\def\qualtime#1{\raisebox{0.050\linewidth}{\makebox[0.018\linewidth][r]{#1}}}
\resizebox{\textwidth}{!}{%
\begin{tabular}{@{}r@{\hspace{1pt}}cccccccc@{}}
\multicolumn{9}{c}{\small\textbf{Prompt:} Change the elephant's color to blue.} \\[3pt]
 & \textbf{Source} &\textbf{Pyramid-Edit~\cite{li2025five}} & \textbf{TokenFlow~\cite{geyer2024tokenflow}} & \textbf{VidToMe~\cite{li2024vidtome}} & \textbf{Wan-Edit~\cite{li2025five}} & \textbf{StreamEdit~\cite{jiao2026streamedit}} & \textbf{FlowDirector~\cite{li2026flowdirector}} & \makecell{\textbf{\modelname{}}\\\textbf{(Ours)}} \\[-1pt]
&
& \makecell{\xmark\ \scriptsize Pose drift}
& \makecell{\xmark\ \scriptsize Edit failure}
& \makecell{\xmark\ \scriptsize Edit failure}
& \makecell{\xmark\ \scriptsize Edit failure}
& \makecell{\xmark\ \scriptsize Edit failure}
& \makecell{\xmark\ \scriptsize Pose drift}
& \makecell{\cmark\ \scriptsize Consistent edit}
\\[2pt]
 
\qualtime{$t_1$} &
\includegraphics[width=0.120\linewidth,height=0.120\linewidth]{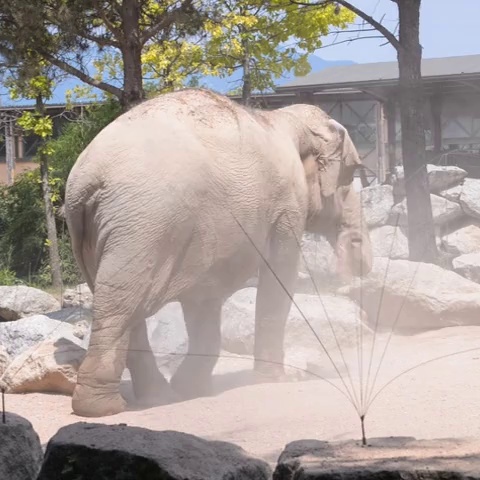} &
\includegraphics[width=0.120\linewidth,height=0.120\linewidth]{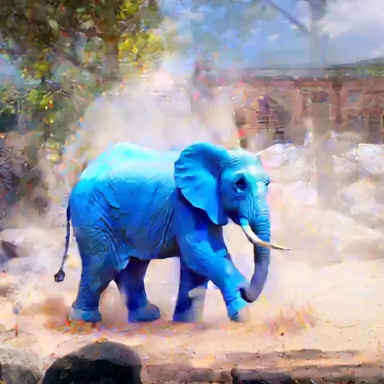} &
\includegraphics[width=0.120\linewidth,height=0.120\linewidth]{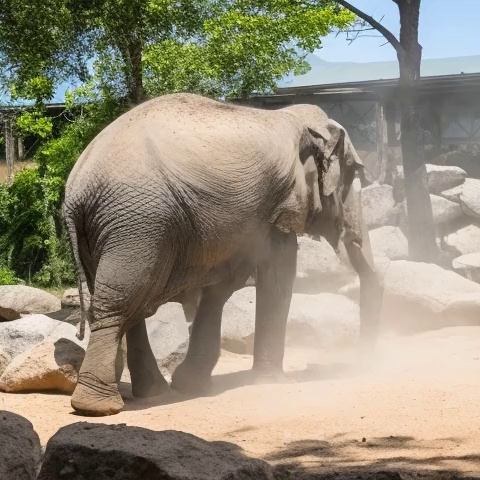} &
\includegraphics[width=0.120\linewidth,height=0.120\linewidth]{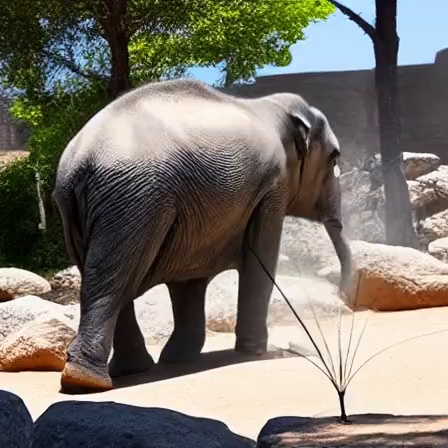} &
\includegraphics[width=0.120\linewidth,height=0.120\linewidth]{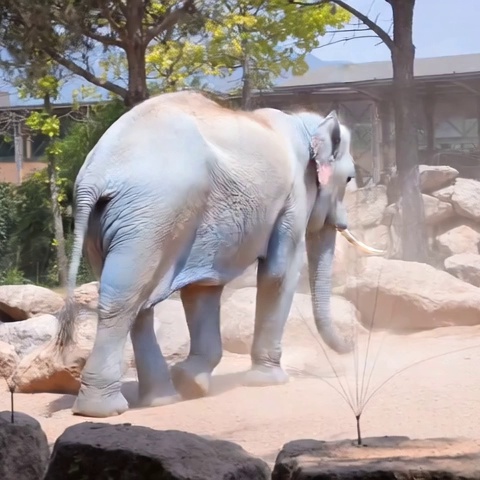} &
\includegraphics[width=0.120\linewidth,height=0.120\linewidth]{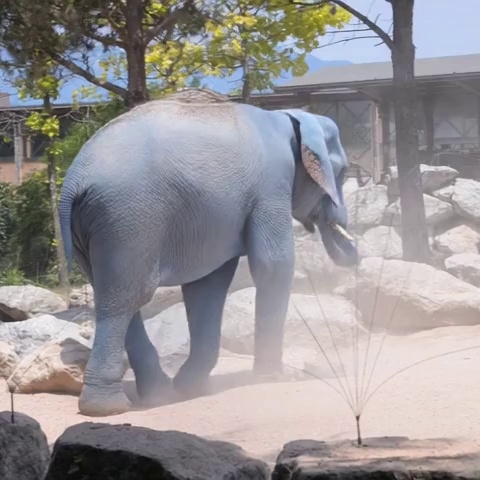} &
\includegraphics[width=0.120\linewidth,height=0.120\linewidth]{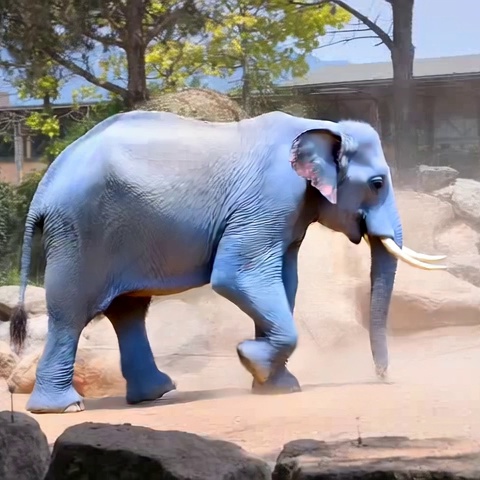} &
\includegraphics[width=0.120\linewidth,height=0.120\linewidth]{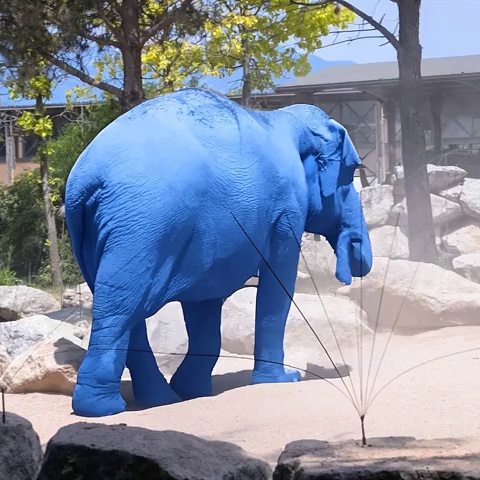} \\[-1.0pt]
\qualtime{$t_2$} &
\includegraphics[width=0.120\linewidth,height=0.120\linewidth]{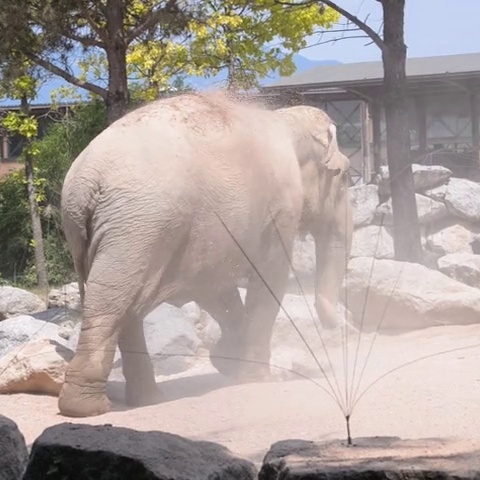} &
\includegraphics[width=0.120\linewidth,height=0.120\linewidth]{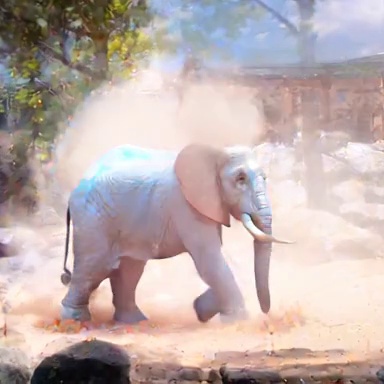} &
\includegraphics[width=0.120\linewidth,height=0.120\linewidth]{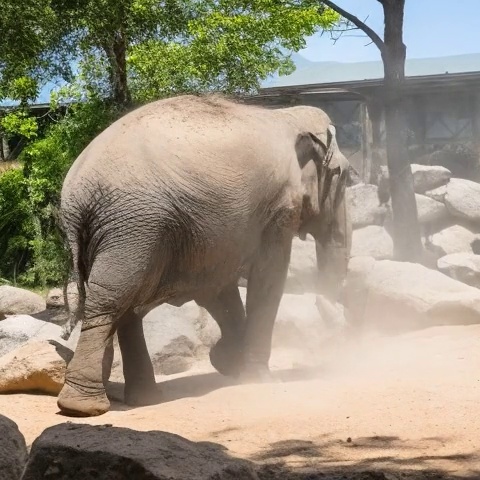} &
\includegraphics[width=0.120\linewidth,height=0.120\linewidth]{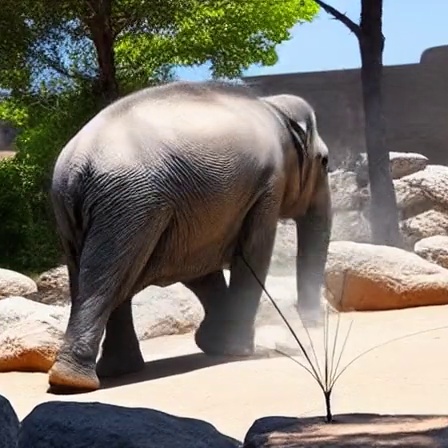} &
\includegraphics[width=0.120\linewidth,height=0.120\linewidth]{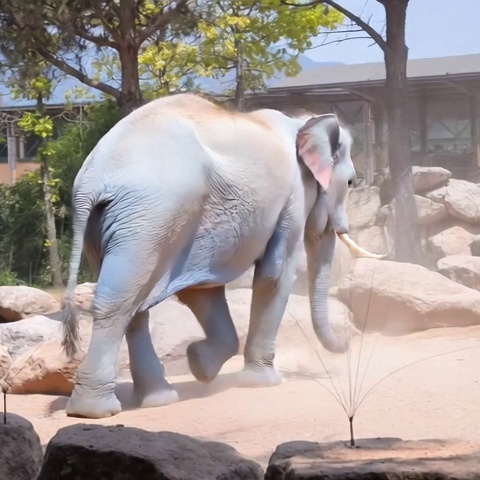} &
\includegraphics[width=0.120\linewidth,height=0.120\linewidth]{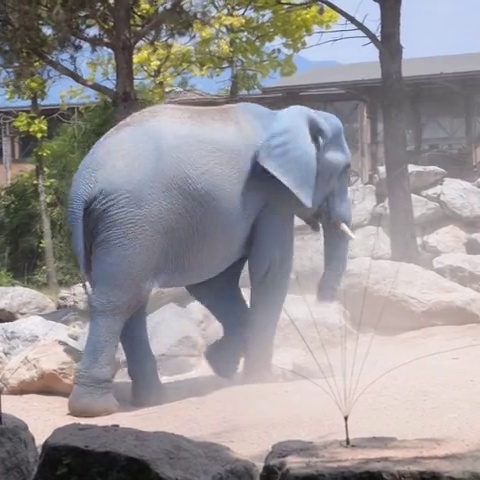} &
\includegraphics[width=0.120\linewidth,height=0.120\linewidth]{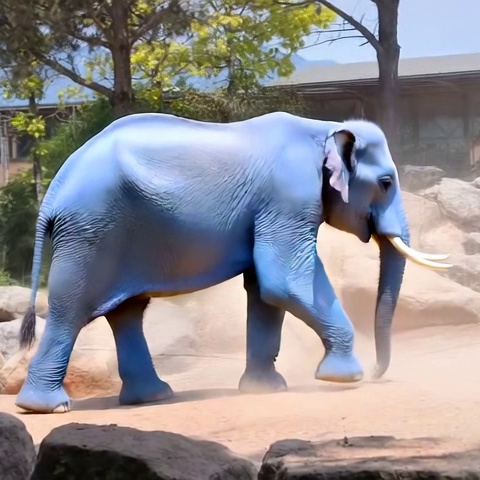} &
\includegraphics[width=0.120\linewidth,height=0.120\linewidth]{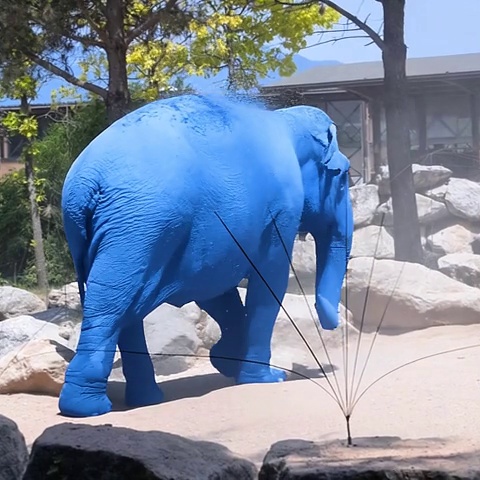} \\[-1.0pt]
\qualtime{$t_3$} &
\includegraphics[width=0.120\linewidth,height=0.120\linewidth]{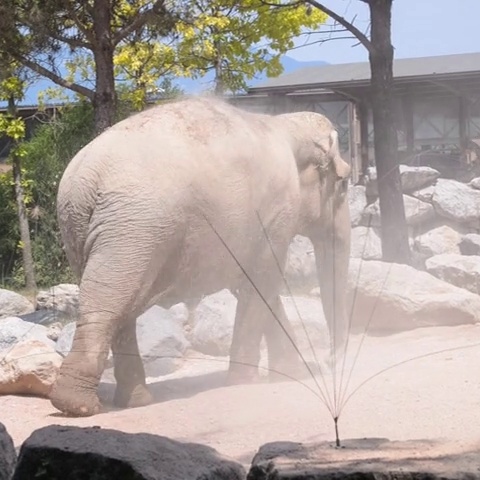} &
\includegraphics[width=0.120\linewidth,height=0.120\linewidth]{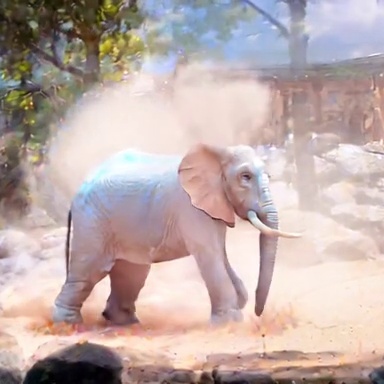} &
\includegraphics[width=0.120\linewidth,height=0.120\linewidth]{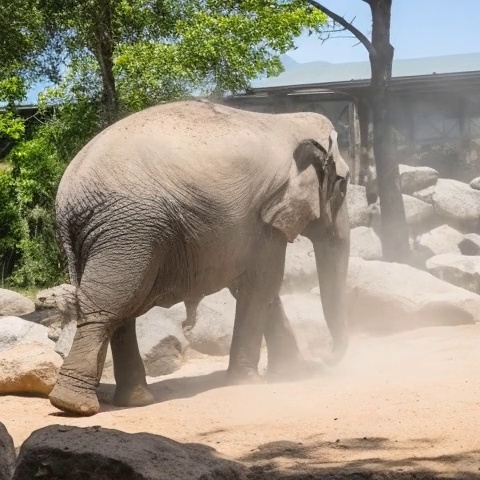} &
\includegraphics[width=0.120\linewidth,height=0.120\linewidth]{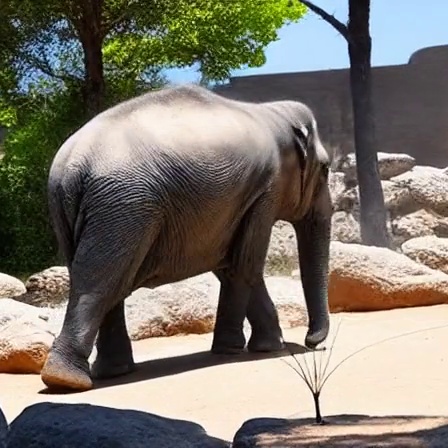} &
\includegraphics[width=0.120\linewidth,height=0.120\linewidth]{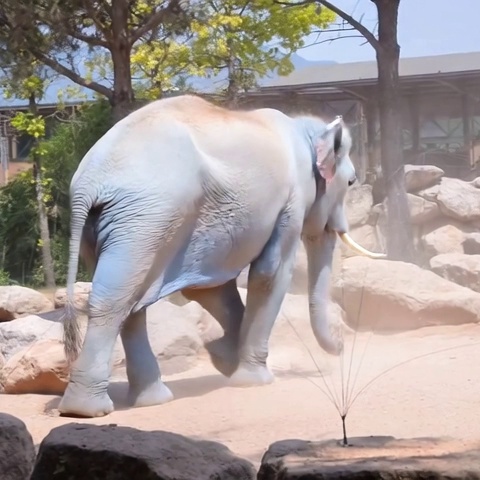} &
\includegraphics[width=0.120\linewidth,height=0.120\linewidth]{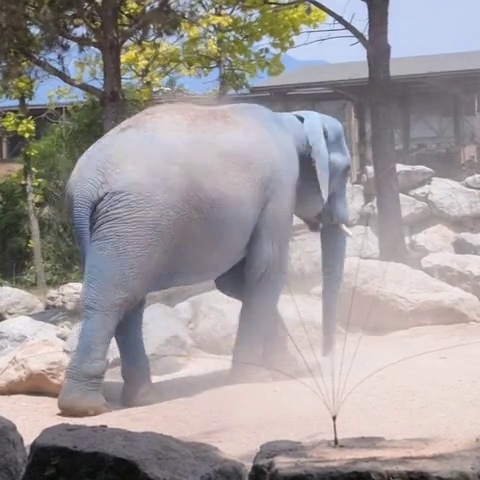} &
\includegraphics[width=0.120\linewidth,height=0.120\linewidth]{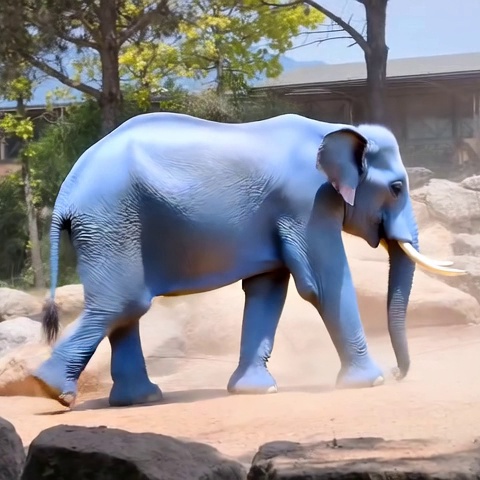} &
\includegraphics[width=0.120\linewidth,height=0.120\linewidth]{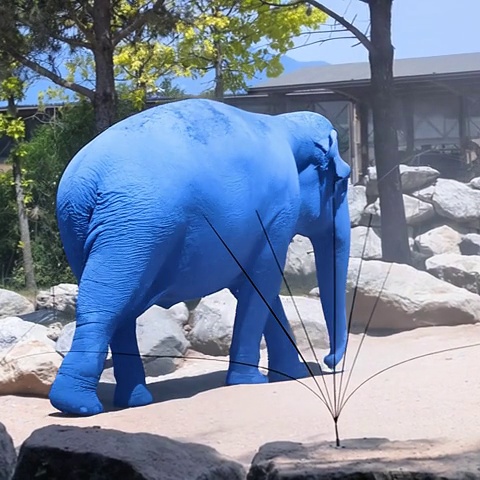} \\[-1.0pt]
\qualtime{$t_4$} &
\includegraphics[width=0.120\linewidth,height=0.120\linewidth]{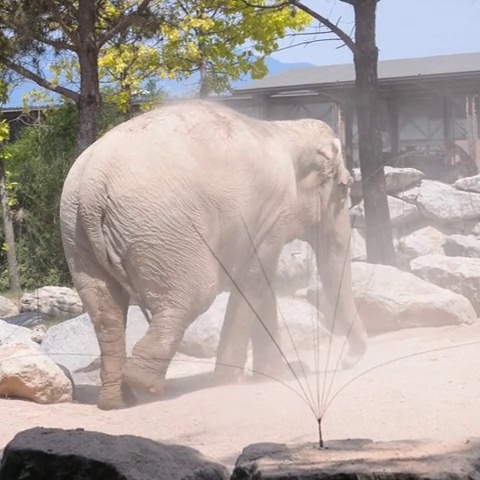} &
\includegraphics[width=0.120\linewidth,height=0.120\linewidth]{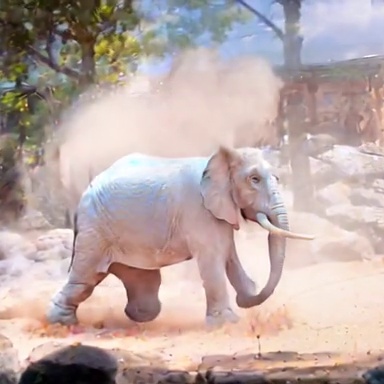} &
\includegraphics[width=0.120\linewidth,height=0.120\linewidth]{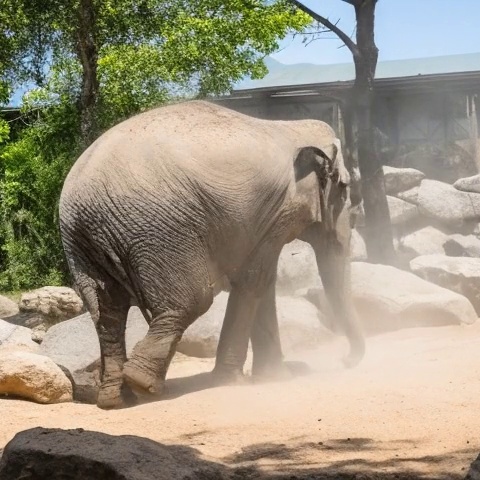} &
\includegraphics[width=0.120\linewidth,height=0.120\linewidth]{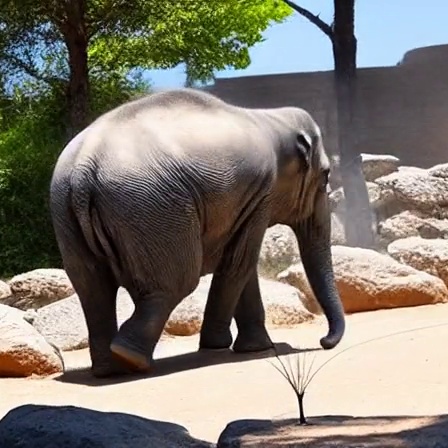} &
\includegraphics[width=0.120\linewidth,height=0.120\linewidth]{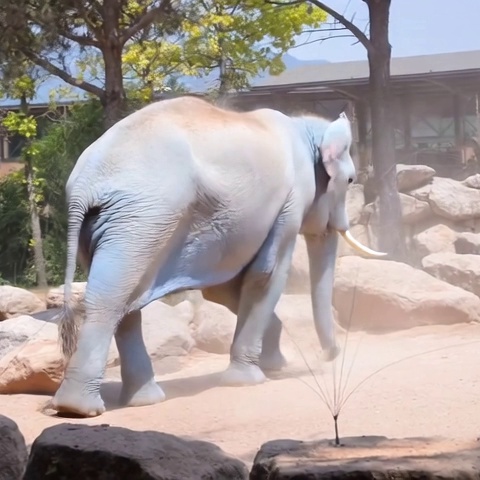} &
\includegraphics[width=0.120\linewidth,height=0.120\linewidth]{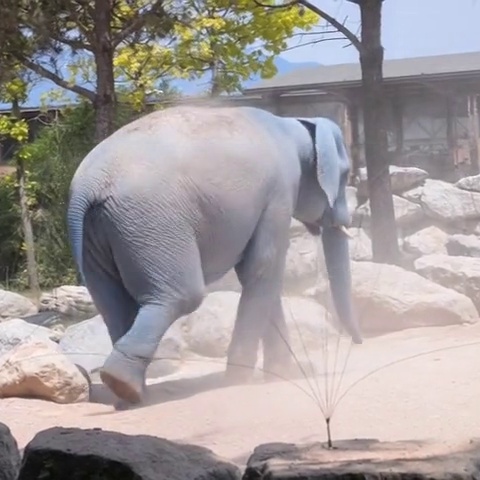} &
\includegraphics[width=0.120\linewidth,height=0.120\linewidth]{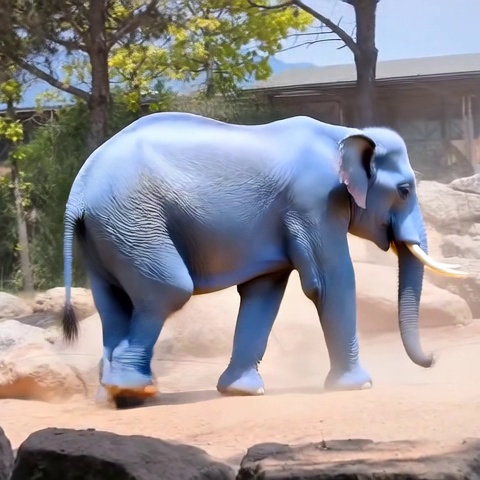} &
\includegraphics[width=0.120\linewidth,height=0.120\linewidth]{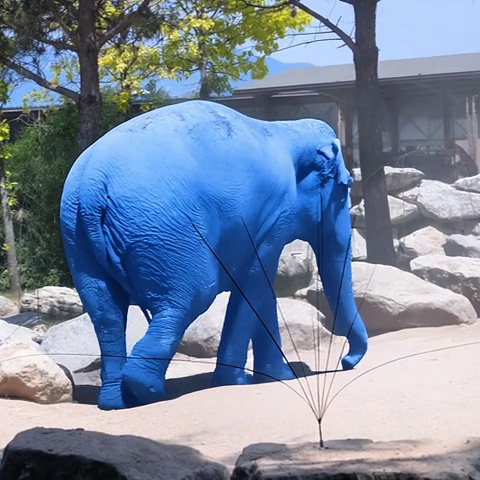} \\[-1.0pt]
\qualtime{$t_5$} &
\includegraphics[width=0.120\linewidth,height=0.120\linewidth]{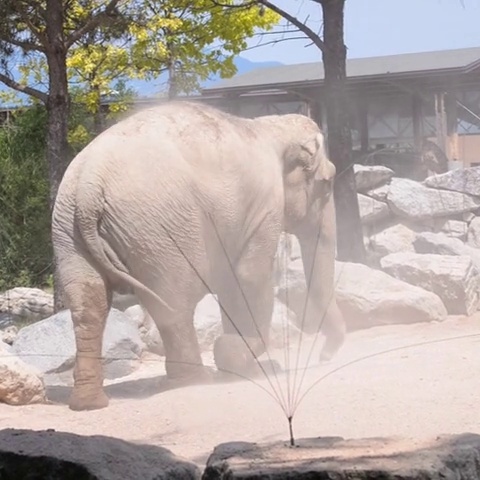} &
\includegraphics[width=0.120\linewidth,height=0.120\linewidth]{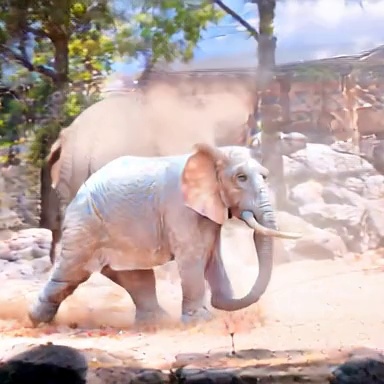} &
\includegraphics[width=0.120\linewidth,height=0.120\linewidth]{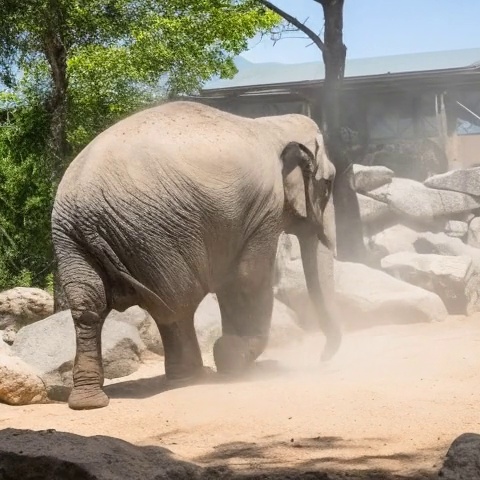} &
\includegraphics[width=0.120\linewidth,height=0.120\linewidth]{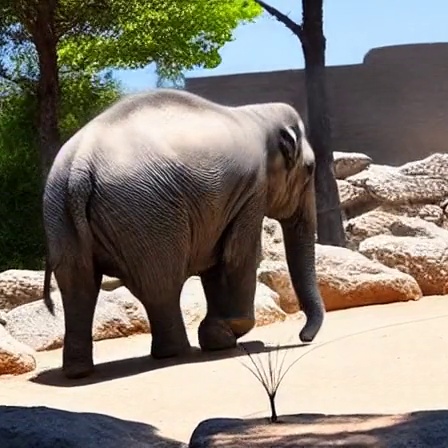} &
\includegraphics[width=0.120\linewidth,height=0.120\linewidth]{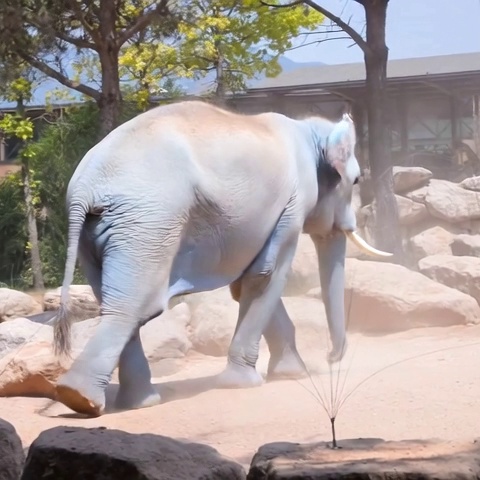} &
\includegraphics[width=0.120\linewidth,height=0.120\linewidth]{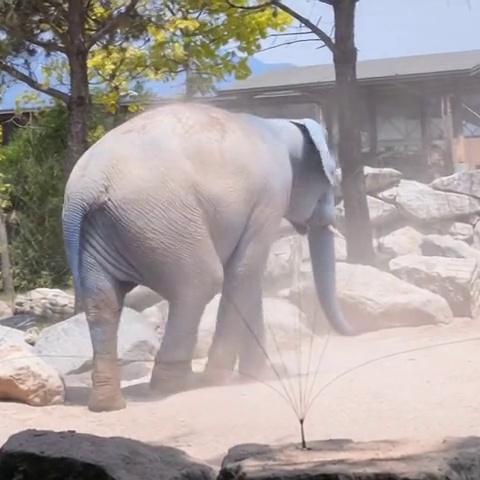} &
\includegraphics[width=0.120\linewidth,height=0.120\linewidth]{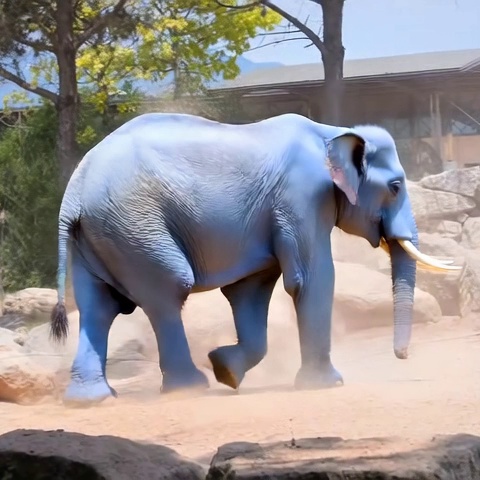} &
\includegraphics[width=0.120\linewidth,height=0.120\linewidth]{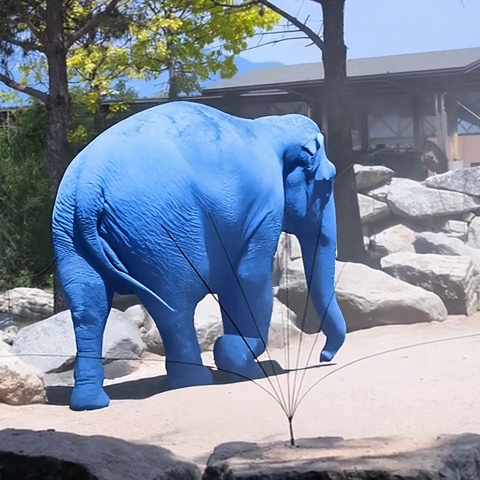} \\[-1.0pt]
\end{tabular}%
}
\caption{\textbf{Training-free video editing comparison.} Given the instruction to change the elephant's color to blue, \modelnamebf{} consistently applies the requested change while preserving the elephant's shape, motion, and background. Frames are sampled at evenly spaced temporal positions throughout each video.}
\label{fig:qual_blue_elephant}
\end{figure*}

\begin{table}[t]
\centering
\caption{\textbf{Effect of temporal scope in spatio-temporal attention} on FiVE-Acc and MF-S.}
\label{tab:temporal_scope}
\resizebox{\linewidth}{!}{
\begin{tabular}{lcccc}
\toprule
Temporal Context & FiVE-Acc $\uparrow$ & MF-S $\uparrow$ & Runtime (s/frame) $\downarrow$ & Memory (GB) $\downarrow$ \\
\midrule
No temporal memory          & 72.50 & 91.39 & \textbf{3.495} & \textbf{38.37} \\
Previous 2 frames           & 72.50 & 89.76 & 4.146 & 54.98 \\
All previous frames         & 62.50 & 49.81 & 5.490 & 90.16 \\
Immediately preceding frame & \textbf{77.50} & \textbf{91.63} & 4.096 & 44.65 \\
\bottomrule
\end{tabular}}
\end{table}

\noindent\textbf{Temporal scope of attention memory.}
\cref{tab:temporal_scope} compares different temporal contexts while keeping the remaining pipeline unchanged. Restricting attention memory to the immediately preceding frame provides the strongest balance of edit correctness and motion preservation, reaching 77.50 FiVE-Acc and 91.63 MF-S. Expanding the context to two previous frames does not improve editing accuracy and increases memory consumption, while attending to all previous frames sharply degrades both FiVE-Acc and MF-S. These results motivate separating temporal coupling by range, as adjacent-frame attention-state reuse is effective under strong local geometric continuity, whereas longer-range interactions become increasingly sensitive to RoPE-modulated geometry and are better handled through token injection.

\begin{table}[!t]
\centering
\caption{\textbf{Correspondence robustness under challenging temporal changes} measured by MF-S.}
\label{tab:correspondence_robustness}
\resizebox{\linewidth}{!}{
\begin{tabular}{lccccc}
\toprule
Method & Viewpoint $\uparrow$ & Occlusion $\uparrow$ & Non-Rigid $\uparrow$ & Structural $\uparrow$ & Overall $\uparrow$ \\
\midrule
w/o Global Injection & 70.97 & 64.90 & 78.51 & \textbf{95.26} & 75.65 \\
Unfiltered Injection & 70.92 & 64.13 & 73.85 & 89.55 & 72.53 \\
\modelnamebf{}        & \textbf{71.96} & \textbf{66.27} & \textbf{78.96} & 94.80 & \textbf{76.23} \\
\bottomrule
\end{tabular}}
\vspace{-0.1cm}
\end{table}

\noindent\textbf{Robustness of long-range correspondence transfer.}
We further isolate the global component under viewpoint changes, occlusions, non-rigid motion, and structural edits, where long-range identity preservation is more demanding. \cref{tab:correspondence_robustness} shows that unfiltered anchor injection is consistently less robust than the complete method and can perform worse than removing global injection altogether. Confidence and cycle-consistency filtering are therefore critical: \modelname{} achieves the strongest overall robustness and the best performance under viewpoint change, occlusion, and non-rigid motion. These results indicate that long-range identity propagation is not obtained by indiscriminate token copying, but depends on selectively transferring correspondences that remain reliable under appearance and geometric change.

\subsection{Qualitative Results}
\label{sec:qual}

\cref{fig:qual_blue_elephant} compares \modelnamebf{} with training-free baselines on an object-level attribute edit. Notably, FlowDirector and StreamEdit, which achieve strong scores on FiVE traditional metrics, preserve the source video well but apply the requested color change only partially or conservatively across the sequence. This reflects a key limitation of traditional preservation metrics, which can reward conservative edits despite incomplete instruction execution, unlike FiVE-Acc which directly measures edit correctness. In contrast, \modelname{} consistently produces the intended blue appearance across frames while preserving the elephant’s structure, motion, and surrounding scene. Other image-based baselines exhibit weaker edit strength, unwanted appearance changes, or greater temporal variation. This example highlights \modelname{}’s ability to balance instruction adherence with strong temporal and source consistency.
\section{Conclusion}
\label{sec:conclusion}

In this work, we introduce \modelnamebf{}, a unified training-free framework for temporally consistent instruction-guided and subject-guided video editing with frozen image editing MM-DiTs. \modelname{} combines sparse causal memory for short-range coherence, correspondence-based global token injection for long-range identity preservation, and adaptive latent blending to preserve instruction-irrelevant content. Across FiVE, IVEBench, mask-free VLM evaluation, and human preference studies, \modelname{} achieves the strongest performance among training-free methods while remaining competitive with recent training-based editors. Controlled temporal-scope and robustness studies further support the proposed local--global decomposition. These results establish modern image-editing priors as a strong foundation for diverse training-free video editing.

{%
  \small
  \bibliographystyle{ieeenat_fullname}
  \bibliography{main}
}

\clearpage
\appendix

\newpage
\setcounter{section}{0}

\section{Additional Inference Details}
\label{sec:additional_details}

\subsection{Algorithm}
\label{sec:algorithm}
Algorithm~\ref{alg:method_inference} summarizes the test-time inference procedure. We partition the input video into contiguous chunks \(\mathcal{V}^{(m)}=\{\mathbf{X}^{(s_m)},\dots,\mathbf{X}^{(e_m)}\}\), where \(s_m\) and \(e_m\) denote the start and end frame indices of chunk \(m\), respectively. For our experiments, we use chunk sizes of 36, 24, or 16, depending on the input video resolution. We use \(a=1\) as the global anchor frame and precompute one correspondence map \(\mathcal{M}^{a\rightarrow i}\) per target frame from inversion features extracted at the penultimate double-stream block and correspondence time \(t_{\mathrm{corr}}=0.25\). For each chunk, we encode the source frames into source conditioning tokens \(\mathbf{y}^{(m)}\) and initialize noisy visual latent tokens \(\mathbf{x}^{(m)}_K\), while the text/reference conditioning tokens are computed once and kept fixed. At each denoising step \(k\), with continuous diffusion time \(t_k\), the model \(v_\theta\) is applied jointly to all frames in the current chunk. Sparse causal memory is injected inside \(v_\theta\) through previous-frame and cached cross-chunk key--value (KV) context, rather than by processing frames sequentially outside the model.

More specifically, KV augmentation is applied in all layers, \ie \(\mathcal{L}_{\mathrm{KV}}\) spans the full set of transformer blocks, so each frame can attend to its immediate predecessor as well as cached temporal memory from the previous chunk. In contrast, correspondence-based global token injection is applied only in a selected subset of vital layers \(\mathcal{L}_{\mathrm{inj}}\). The precomputed map \(\mathcal{M}^{a\rightarrow i}\) is reused at every injection layer to inject anchor-frame denoising tokens into reliable target-frame locations. After each denoising step, the temporal memory is updated for reuse at subsequent chunk boundaries, and after the final step, the edited visual tokens are decoded and reassembled into the output video.\looseness-1

\begin{algorithm}[t!]
\caption{Inference algorithm with spatio-temporal attention and correspondence-based global token injection}
\label{alg:method_inference}
\small
\begin{algorithmic}[1]
\Require Video \(\mathcal{V}=\{\mathbf{X}^{(i)}\}_{i=1}^{N}\), conditioning signal \(\mathbf{c}\), model \(v_\theta\), chunk size \(B\), denoising steps \(K\)
\Require KV-augmentation layers \(\mathcal{L}_{\mathrm{KV}}\), global-injection layers \(\mathcal{L}_{\mathrm{inj}}\), threshold \(\tau\), locality radius \(r\), dropout probability \(p_{\mathrm{drop}}\)
\State Partition \(\mathcal{V}\) into contiguous chunks \(\{\mathcal{V}^{(m)}\}_{m=1}^{M}\), where \(\mathcal{V}^{(m)}=\{\mathbf{X}^{(s_m)},\dots,\mathbf{X}^{(e_m)}\}\)
\State Compute conditioning tokens \(\mathbf{c}_e \leftarrow \mathrm{Enc}_c(\mathbf{c})\)
\State Set anchor index \(a\leftarrow 1\)
\State Set correspondence time \(t_{\mathrm{corr}}\leftarrow0.25\) and correspondence block \(\ell_{\mathrm{corr}}\leftarrow\) penultimate double-stream block
\State Compute inversion token sets \(\mathbf{G}^{(i)}=\{\mathbf{g}^{(i)}_{p}\}_{p=1}^{N_v}\) from \((\ell_{\mathrm{corr}},t_{\mathrm{corr}})\), for all \(i=1,\dots,N\)
\State Construct \(\mathcal{M}^{a\rightarrow i}\leftarrow
\mathrm{Corr}\!\big(\mathbf{G}^{(a)},\,\mathbf{G}^{(i)};\tau,r,p_{\mathrm{drop}}\big)\), for all \(i\)
\State Initialize temporal memory \(\mathcal{S}\leftarrow\varnothing\)
\For{each chunk \(\mathcal{V}^{(m)}\)}
    \State Compute source conditioning tokens \(\mathbf{y}^{(m)} \leftarrow \mathrm{Enc}_x(\mathcal{V}^{(m)})\)
    \State Sample initial noisy visual latents \(\mathbf{x}^{(m)}_K\sim p(\mathbf{x}_K)\)
    \For{\(k=K,\dots,1\)}
        \State Set continuous diffusion time \(t_k\)
        \State \(\mathcal{C}^{(m)}_{x,k}\leftarrow \{\mathbf{x}^{(i-1)}_k\}_{i=s_m+1}^{e_m}\cup \mathcal{S}\)
        \State \(\mathbf{u}^{(m)}_k\leftarrow
        v_\theta\!\big(\mathbf{x}^{(m)}_k,\, t_k,\, \mathbf{c}_e,\, \mathbf{y}^{(m)}\big)\)
        \Statex \hspace{\algorithmicindent}with key--value augmentation in all \(\ell\in\mathcal{L}_{\mathrm{KV}}\) using \(\mathcal{C}^{(m)}_{x,k}\)
        \Statex \hspace{\algorithmicindent}and correspondence-based global token injection in all \(\ell\in\mathcal{L}_{\mathrm{inj}}\) using precomputed \(\mathcal{M}^{a\rightarrow i}\)
        \State \(\mathbf{x}^{(m)}_{k-1}\leftarrow
        \mathrm{SchedulerStep}\!\big(\mathbf{x}^{(m)}_k,\,\mathbf{u}^{(m)}_k,\,t_k\big)\)
        \State \(\mathcal{S}\leftarrow \mathrm{UpdateMemory}\!\big(\mathbf{x}^{(m)}_{k-1}\big)\)
    \EndFor
    \State \(\hat{\mathcal{V}}^{(m)}\leftarrow \mathrm{Dec}_x\!\big(\mathbf{x}^{(m)}_0\big)\)
\EndFor
\State Reassemble \(\hat{\mathcal{V}}=\{\hat{\mathcal{V}}^{(m)}\}_{m=1}^{M}\)
\end{algorithmic}
\end{algorithm}

\subsection{Implementation Details}
\label{sec:app_implementation_details}
We implement \modelname{} on top of FLUX.2-Klein-9B~\cite{flux2024}, using the released model weights, tokenizer/text encoder, and VAE in the Python Diffusers library. Unless noted otherwise, inference uses \texttt{bfloat16}, guidance scale \(1.0\), and \(4\) denoising steps. We apply temporal attention modifications at all denoising steps \(\{0,1,2,3\}\). For correspondence-based global token injection, we compute correspondences from the penultimate double-stream block at \(t_{\mathrm{corr}}=0.25\) and replace tokens in the first 12 MM-DiT blocks identified as vital layers~\cite{avrahami2025stable}. We use confidence threshold \(\tau=0.4\), spatial tolerance radius \(r=1.5\), and dropout probability \(p_{\mathrm{drop}}=0.5\).\looseness-1

\subsection{Soft Latent Blending Details}
\label{sec:app_soft_blending}

As introduced in \cref{sec:method}, we apply soft latent blending to preserve unedited regions of the source video without rigidly masking moving subjects.
Let \(t_k\) denote the continuous diffusion time associated with discrete denoising step \(k\), and write \(\mathbf{x}^{(i)}_k=\mathbf{x}^{(i)}(t_k)\) for the edited visual latent tokens of frame \(i\). Let \(\mathbf{x}^{(i),\mathrm{src}}_k\) denote the corresponding source visual-token trajectory obtained from inversion.
We first compute a per-token difference map
\begin{equation}
d^{(i)}_{k,p}
=
\frac{1}{D}
\left\lVert
\mathbf{x}^{(i)}_{k,p}
-
\mathbf{x}^{(i),\mathrm{src}}_{k,p}
\right\rVert_1,
\label{eq:latent_diff}
\end{equation}
where \(p\) indexes visual latent tokens and \(D\) is the token channel dimension.
We normalize this difference map using low and high quantiles:

\begin{equation}
m^{(i)}_{k,p}
=
\mathrm{clip}
\left(
\frac{
d^{(i)}_{k,p}-Q_{\mathrm{lo}}(d^{(i)}_k)
}{
[
Q_{\mathrm{hi}}(d^{(i)}_k)
-
Q_{\mathrm{lo}}(d^{(i)}_k)
]_{\epsilon}
},
0,1
\right)^{\gamma},
\label{eq:soft_mask}
\end{equation}

where \(Q_{\mathrm{lo}}\) and \(Q_{\mathrm{hi}}\) are the low and high
quantiles, and \([a]_{\epsilon}\coloneqq\max(a,\epsilon)\) ensures
stability. We empirically fix
\(Q_{\mathrm{lo}}=Q_{0.25}\) and \(Q_{\mathrm{hi}}=Q_{0.65}\),
computing them independently over each frame's spatial tokens at every
blending step. We set \(\epsilon=10^{-6}\) and \(\gamma=0.7\).
Gaussian smoothing on the latent grid optionally yields the final mask
\(\hat{m}^{(i)}_k\) used for blending weights \(w^{(i)}_{k,p}\) in the
main text.\looseness-1

\section{Additional Experiments}

\subsection{FiVE Metrics}
\label{sec:app_eval_metrics}

For FiVE~\cite{li2025five}, we report the benchmark's FiVE-Acc suite in \cref{tab:five_results}, which evaluates object-level editing success using VLM-based recognition of the edited object. The suite includes binary yes/no accuracy (FiVE-YN), multi-choice accuracy (FiVE-MC), open-ended accuracy (FiVE-U), their intersection (FiVE-\(\cap\)), and the overall FiVE-Acc score.

We also report FiVE's conventional automatic metrics as complementary diagnostics in \cref{tab:five_traditional}. Following FiVE-Bench~\cite{li2025five}, these metrics are organized into structure preservation (Structure Dist.), background preservation (PSNR, LPIPS, MSE, and SSIM computed outside the editing mask), edit prompt--image consistency (CLIP$_S$ on full images and CLIP$_S^{\mathrm{edit}}$ on masked images), image quality assessment (NIQE), and temporal consistency (MF-S). Since these metrics measure different aspects of preservation, alignment, quality, and temporal stability, we use them alongside FiVE-Acc rather than as the sole measure of edit correctness.\looseness-1

For readability, \cref{tab:five_traditional} follows FiVE's scaled reporting convention. Structure distance and LPIPS are reported after multiplication by \(10^{3}\), MSE after multiplication by \(10^{4}\), and SSIM and MF-S after multiplication by \(10^{2}\). CLIP$_S$ and CLIP$_S^{\mathrm{edit}}$ are also reported on a \(10^{2}\)-scaled score. PSNR is reported in dB, and NIQE is reported in its original scale. Thus, raw values can be recovered by dividing the scaled columns by their corresponding factors.

\begin{table*}[t]
\centering
\small
\setlength{\tabcolsep}{5.5pt}
\caption{\textbf{Comparison on general video editing on our curated subset.}
For trained methods, the best value is highlighted as \colorbox{best}{\textbf{best}}.
For training-free methods, the \colorbox{best}{\textbf{best}} and
\colorbox{second}{second-best} values are highlighted.}
\resizebox{0.99\linewidth}{!}{%
\begin{tabular}{lccccccc}
\toprule
\textbf{Method}
& \makecell{\textbf{Training-}\\\textbf{free}}
& \makecell{\textbf{CLIP-T}\\$\uparrow$}
& \makecell{\textbf{CLIP-Dir}\\$\uparrow$}
& \makecell{\textbf{Warp-Err}\\$\downarrow$}
& \makecell{\textbf{BRISQUE}\\$\downarrow$}
& \makecell{\textbf{PickScore}\\$\uparrow$}
& \makecell{\textbf{PickScore-Dir*}\\$\uparrow$} \\
\midrule

LucyEdit-Dev~\cite{liao2025context}
& \xmark
& \cellcolor{best}\textbf{32.33}
& 0.105
& \cellcolor{best}\textbf{0.56}
& 31.38
& 20.57
& 0.146 \\

Kiwi-Edit~\cite{lin2026kiwiedit}
& \xmark
& 32.10
& 0.138
& 0.61
& \cellcolor{best}\textbf{18.18}
& 20.45
& 0.197 \\

Qwen-Video-Edit~\cite{bai2026qwen}
& \xmark
& 32.01
& \cellcolor{best}\textbf{0.144}
& 0.64
& 26.56
& \cellcolor{best}\textbf{20.85}
& \cellcolor{best}\textbf{0.205} \\

\midrule

Pyramid-Edit~\cite{li2025five}
& \cmark
& 32.59
& 0.071
& 0.61
& 38.48
& 20.07
& 0.120 \\

VidToMe~\cite{li2024vidtome}
& \cmark
& \cellcolor{best}\textbf{33.10}
& 0.089
& 0.62
& 28.21
& 20.60
& 0.134 \\

Wan-Edit~\cite{li2025five}
& \cmark
& 32.40
& 0.061
& \cellcolor{second}0.57
& 26.92
& \cellcolor{best}\textbf{20.99}
& 0.096 \\

StreamEdit~\cite{jiao2026streamedit}
& \cmark
& 31.94
& 0.067
& 0.58
& 32.55
& 20.83
& 0.103 \\

FlowDirector~\cite{li2026flowdirector}
& \cmark
& 32.45
& 0.083
& 0.58
& 25.54
& 20.87
& 0.123 \\

\modelnamebf\ \textbf{(Ours)}
& \cmark
& \cellcolor{second}32.83
& \cellcolor{best}\textbf{0.201}
& \cellcolor{best}\textbf{0.53}
& \cellcolor{best}\textbf{20.38}
& \cellcolor{second}20.92
& \cellcolor{best}\textbf{0.280} \\

\bottomrule
\end{tabular}%
}
\label{tab:vg_metrics}
\end{table*}
\begin{table*}[t]
\centering
\small
\setlength{\tabcolsep}{5pt}
\caption{\textbf{Comparison on subject-guided editing on our curated subset.}
For trained methods, the best value is highlighted as \colorbox{best}{\textbf{best}}.
For training-free methods, the \colorbox{best}{\textbf{best}} and
\colorbox{second}{second-best} values are highlighted.}
\resizebox{0.99\linewidth}{!}{%
\begin{tabular}{lcccccccc}
\toprule
\textbf{Method}
& \makecell{\textbf{Training-}\\\textbf{free}}
& \makecell{\textbf{DINO-Ref}\\$\uparrow$}
& \makecell{\textbf{CLIP-T}\\$\uparrow$}
& \makecell{\textbf{CLIP-Dir}\\$\uparrow$}
& \makecell{\textbf{Warp-Err}\\$\downarrow$}
& \makecell{\textbf{BRISQUE}\\$\downarrow$}
& \makecell{\textbf{PickScore}\\$\uparrow$}
& \makecell{\textbf{PickScore-Dir*}\\$\uparrow$} \\
\midrule

LucyEdit-Dev~\cite{liao2025context}
& \xmark
& 0.196
& \cellcolor{best}\textbf{29.18}
& 0.084
& 0.68
& 35.29
& 19.64
& 0.145 \\

Kiwi-Edit~\cite{lin2026kiwiedit}
& \xmark
& \cellcolor{best}\textbf{0.203}
& 28.83
& \cellcolor{best}\textbf{0.099}
& \cellcolor{best}\textbf{0.65}
& \cellcolor{best}\textbf{21.37}
& 19.61
& \cellcolor{best}\textbf{0.175} \\

Qwen-Video-Edit~\cite{bai2026qwen}
& \xmark
& 0.125
& 27.72
& 0.009
& 0.68
& 26.42
& \cellcolor{best}\textbf{19.93}
& 0.029 \\

\midrule

Pyramid-Edit$^{\dagger}$~\cite{li2025five}
& \cmark
& 0.223
& 30.53
& 0.133
& 0.72
& 38.75
& 19.68
& 0.204 \\

VidToMe~\cite{li2024vidtome}
& \cmark
& 0.246
& \cellcolor{second}30.98
& 0.117
& 0.71
& 33.08
& 19.97
& 0.185 \\

Wan-Edit~\cite{li2025five}
& \cmark
& 0.265
& 30.31
& 0.126
& \cellcolor{second}0.66
& 27.72
& 20.48
& 0.214 \\

StreamEdit~\cite{jiao2026streamedit}
& \cmark
& 0.270
& 30.37
& 0.162
& 0.69
& 32.23
& 20.42
& 0.253 \\

FlowDirector~\cite{li2026flowdirector}
& \cmark
& \cellcolor{second}0.338
& \cellcolor{best}\textbf{31.08}
& \cellcolor{second}0.185
& 0.68
& 25.66
& \cellcolor{second}20.49
& \cellcolor{second}0.281 \\

\modelnamebf\ \textbf{(Ours)}
& \cmark
& \cellcolor{best}\textbf{0.364}
& 30.65
& \cellcolor{best}\textbf{0.190}
& \cellcolor{best}\textbf{0.65}
& \cellcolor{best}\textbf{22.52}
& \cellcolor{best}\textbf{20.59}
& \cellcolor{best}\textbf{0.286} \\

\bottomrule
\end{tabular}%
}
\label{tab:ref_align_metrics}
\end{table*}

\subsection{Evaluation on Diverse Edits}
\label{sec:app_curated_eval}

\begin{figure*}[t]
\centering
\fbox{
\begin{minipage}{0.96\textwidth}
\footnotesize
\ttfamily
\raggedright

You are a strict expert evaluator of video editing systems.

\vspace{0.5em}
You are given:\\
1. SOURCE VIDEO FRAMES sampled in chronological order.\\
2. EDITED VIDEO FRAMES sampled at matching timestamps and in the same order.\\
3. EDIT INSTRUCTION.\\
4. For subject-guided editing only, a REFERENCE IMAGE defining the desired subject identity/appearance.

\vspace{0.5em}
Evaluate the entire edited video, not a single frame. Score each criterion from 1.0 to 10.0:

\vspace{0.5em}
- prompt\_following: How completely and accurately the edited video follows the edit instruction. Penalize missing, partial, incorrect, misplaced, or temporally inconsistent edits. For subject-guided editing, also judge whether the edited subject matches the identity and visual appearance in the reference image across the video.

\vspace{0.5em}
- edit\_quality: Visual quality and realism of the intended edit, including clean boundaries, plausible geometry, lighting, texture, compositing, temporal stability, and absence of flicker, deformation, duplication, or artifacts. Judge the edit itself rather than unrelated source-video quality.

\vspace{0.5em}
- background\_consistency: Preservation of all regions and details that the instruction does not ask to change, compared with the source frames. Penalize changes to background, camera, layout, pose, scale, unrelated objects, colors, lighting, or framing. Do not penalize changes strictly necessary for the requested edit.

\vspace{0.5em}
Use the full 1--10 range. A score of 10 means essentially perfect; 5 means materially flawed but recognizable; 1 means complete failure.

\vspace{0.5em}
Return ONLY one valid JSON object with exactly this schema:

\vspace{0.3em}
\{\\
\quad "prompt\_following": \{"score": 0.0, "reason": "brief evidence"\},\\
\quad "edit\_quality": \{"score": 0.0, "reason": "brief evidence"\},\\
\quad "background\_consistency": \{"score": 0.0, "reason": "brief evidence"\},\\
\quad "overall\_reason": "brief summary"\\
\}

\end{minipage}
}
\caption{\textbf{Mask-free VLM evaluator prompt} used for all VLM-based evaluations reported in \cref{tab:ivebench_vlm}.}
\label{fig:vlm_judge_prompt}
\end{figure*}

\noindent\textbf{Mask-free VLM Evaluation Setup.}
The mask-free VLM results in \cref{tab:ivebench_vlm} demonstrate the strong performance of \modelname{} on our curated evaluation set. The set contains 24 general video-editing examples that cover part-, instance-, and class-level modifications, style and attribute changes, and other diverse edits. In addition, it includes 26 subject-guided examples that require preservation of a reference identity.
For the VLM evaluation, we use Qwen3.6-27B-FP8 as the evaluator. It receives eight source frames and eight edited frames sampled approximately uniformly at corresponding frame indices over their common temporal extent, together with the edit instruction and, when applicable, the reference image. The VLM evaluates prompt following, edit quality, and background consistency of the edits without edit masks. The evaluation prompt is provided in \cref{fig:vlm_judge_prompt}.\looseness-1

\noindent\textbf{Evaluation on Additional Metrics.}
We report evaluation results on additional automatic metrics in \cref{tab:vg_metrics,tab:ref_align_metrics}, providing complementary evidence beyond VLM judgments. CLIP-T measures target-text alignment, whereas CLIP-Dir evaluates whether the source-to-output visual change agrees with the requested textual change. Warp-Err measures temporal consistency, and BRISQUE provides a no-reference assessment sensitive to noise, blur, and other visual distortions, with lower values indicating better perceptual quality. PickScore estimates human-aligned preference for the edited output. We also extend PickScore with a diagnostic directional variant, PickScore-Dir, computed analogously to CLIP-Dir using PickScore’s image and text embeddings to measure alignment with the requested edit direction.  For subject-guided editing, DINO-Ref additionally measures reference-subject fidelity.

Among training-free methods on general video editing, \modelname{} achieves the best CLIP-Dir, Warp-Err, BRISQUE, and PickScore-Dir, while ranking second on CLIP-T and PickScore. The best BRISQUE and Warp-Err results indicate that its stronger edits maintain both perceptual quality and temporal consistency, while the direction-sensitive metrics show improved alignment with the requested transformation. On subject-guided editing, \modelname{} leads training-free methods on six of seven metrics, obtaining the best DINO-Ref, CLIP-Dir, Warp-Err, BRISQUE, PickScore, and PickScore-Dir. In particular, the strong DINO-Ref and BRISQUE scores indicate improved reference-subject fidelity together with higher perceptual quality. Across both settings, \modelname{} also outperforms the best evaluated training-based methods on 10 of the 13 reported metrics and matches the best training-based subject-guided Warp-Err. These results demonstrate that its advantages extend beyond VLM evaluation to edit alignment, temporal consistency, perceptual quality, human-aligned preference, and reference-subject preservation.

\subsection{Performance Profiling}
\label{sec:profiling}
\begin{figure}[t]
    \centering
    \includegraphics[width=0.99\linewidth]{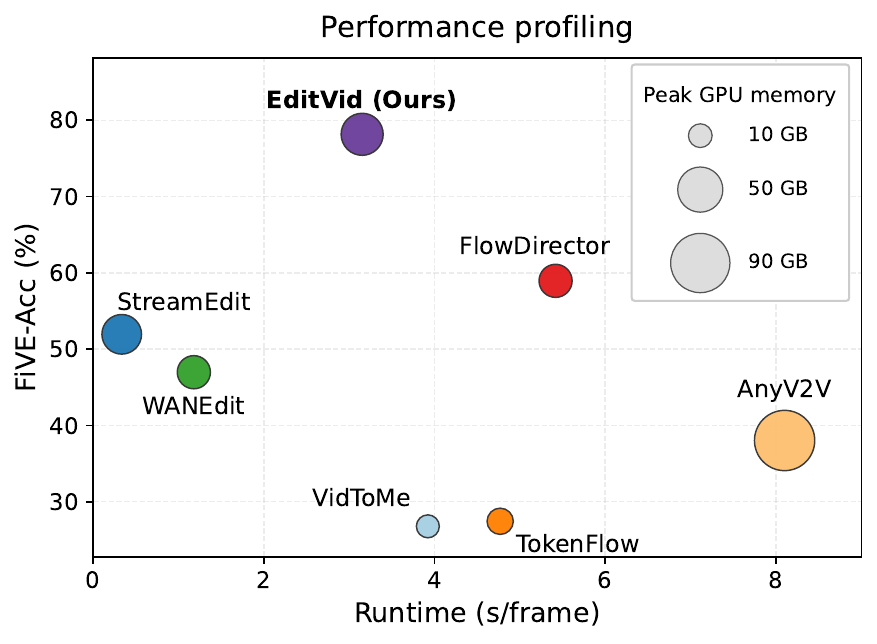}
    \caption{\textbf{Performance profiling.} The x-axis shows runtime in seconds per frame, the y-axis shows FiVE-Accuracy. Bubble size indicates GPU memory usage.}
    \label{fig:performance_profiling}
    \vspace{-0.3cm}
\end{figure}

We further analyze the trade-off between runtime, GPU memory, and editing quality in \cref{fig:performance_profiling}. The x-axis reports inference time in seconds per frame, the y-axis reports FiVE-Acc~\cite{li2025five}, and bubble size indicates peak GPU memory consumption. \modelname{} achieves the highest FiVE-Acc (78.16) at 3.15 seconds per frame, substantially outperforming the baselines, whose FiVE-Acc ranges from 26.77 to 58.95. Notably, \modelname{} is considerably faster than AnyV2V~\cite{ku2024anyv2v}, the only compared baseline supporting both text-guided and subject-guided editing, and FlowDirector, the strongest baseline in terms of FiVE-Acc. Overall, \modelname{} provides a favorable trade-off between editing accuracy, runtime, and memory consumption. All runtime and memory measurements are obtained on a single NVIDIA H200 GPU.

\subsection{Cross-backbone generalization}
To test the backbone generalization, we instantiate the same training-free framework on FLUX.1-Kontext and evaluate it on the 24-video instruction-guided editing subset. As shown in \cref{tab:cross_backbone}, the FLUX.1-Kontext variant achieves strong mask-free VLM scores of 8.125 PF, 8.000 EQ, and 9.042 BC. Although FLUX.2-Klein-9B remains stronger across all three criteria, results show that the proposed temporal design transfers to a different image-editing MM-DiT.\looseness-1

\begin{table}[t]
\centering
\caption{\textbf{Cross-backbone evaluation} on the video editing subset using our mask-free VLM evaluation. Both variants use the same training-free temporal framework. Best values are in \textbf{bold}.}
\label{tab:cross_backbone}
\resizebox{\columnwidth}{!}{
\begin{tabular}{lccc}
\toprule
Backbone & PF $\uparrow$ & EQ $\uparrow$ & BC $\uparrow$ \\
\midrule
\modelname{}-FLUX.1-Kontext & 8.125 & 8.000 & 9.042 \\
\modelname{}-FLUX.2-Klein-9B & \textbf{9.458} & \textbf{9.000} & \textbf{9.500} \\
\bottomrule
\end{tabular}
}
\end{table}

\section{Additional Qualitative Results}
\label{sec:additional_qualitative}

\begin{figure*}[t!]
\centering
\setlength{\tabcolsep}{0.45pt}
\renewcommand{\arraystretch}{0.62}
\scriptsize
\def\qualtime#1{\raisebox{0.050\linewidth}{\makebox[0.018\linewidth][r]{#1}}}
\resizebox{\textwidth}{!}{%
\begin{tabular}{@{}r@{\hspace{1pt}}cccccccc@{}}
\multicolumn{9}{c}{\small\textbf{Prompt:} Change the video to Ghibli art style.} \\[3pt]
& \textbf{Source} & \textbf{AnyV2V~\cite{ku2024anyv2v}} & \textbf{VidToMe~\cite{li2024vidtome}} & \textbf{Pyramid-Edit~\cite{li2025five}} & \textbf{StreamEdit~\cite{jiao2026streamedit}} & \textbf{Wan-Edit~\cite{li2025five}} & \textbf{FlowDirector~\cite{li2026flowdirector}} & \makecell{\textbf{\modelname{}}\\\textbf{(Ours)}} \\[2pt]
\qualtime{$t_1$} &
\includegraphics[width=0.120\linewidth,height=0.120\linewidth]{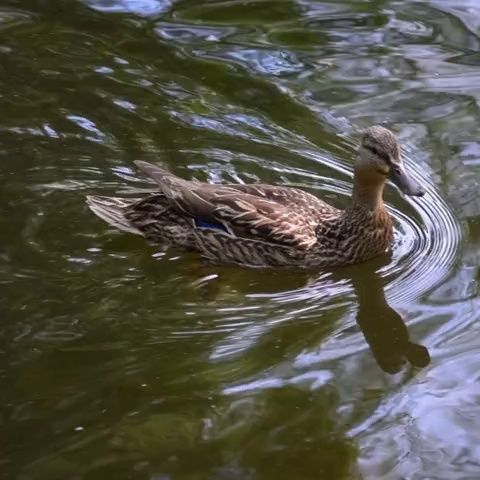} &
\includegraphics[width=0.120\linewidth,height=0.120\linewidth]{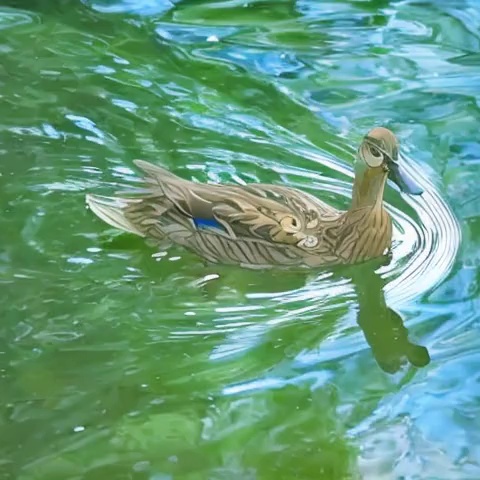} &
\includegraphics[width=0.120\linewidth,height=0.120\linewidth]{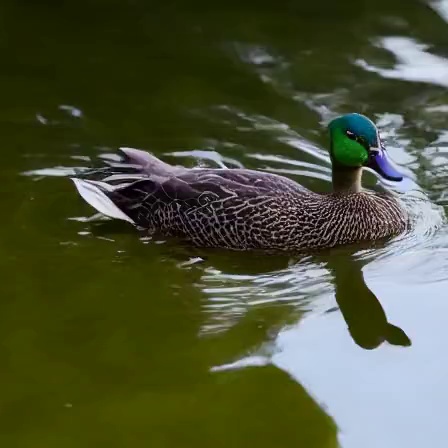} &
\includegraphics[width=0.120\linewidth,height=0.120\linewidth]{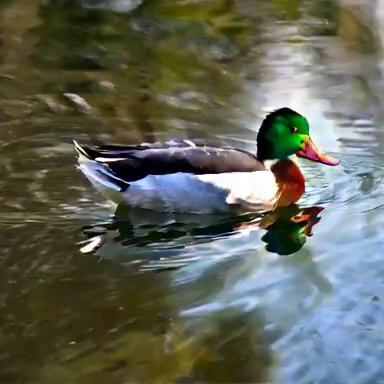} &
\includegraphics[width=0.120\linewidth,height=0.120\linewidth]{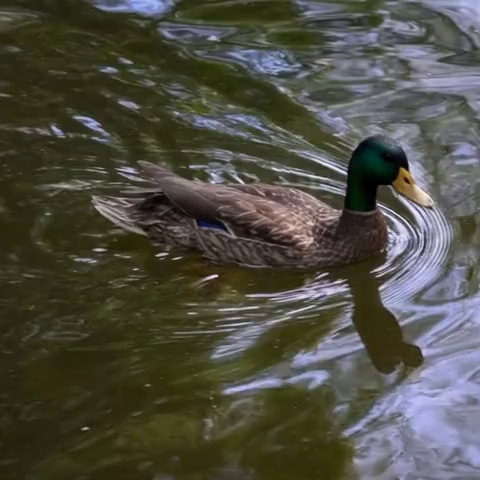} &
\includegraphics[width=0.120\linewidth,height=0.120\linewidth]{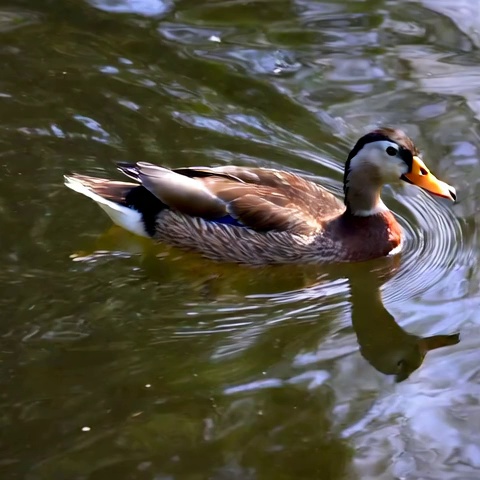} &
\includegraphics[width=0.120\linewidth,height=0.120\linewidth]{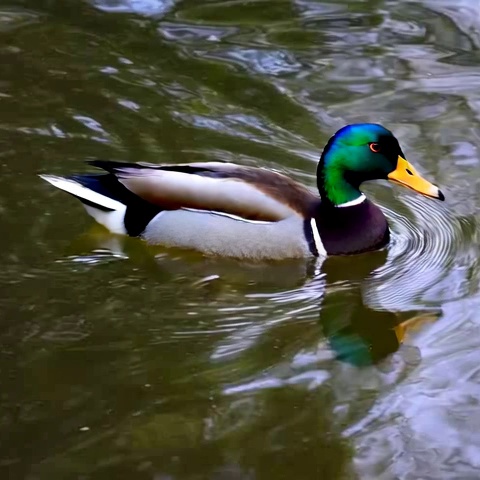} &
\includegraphics[width=0.120\linewidth,height=0.120\linewidth]{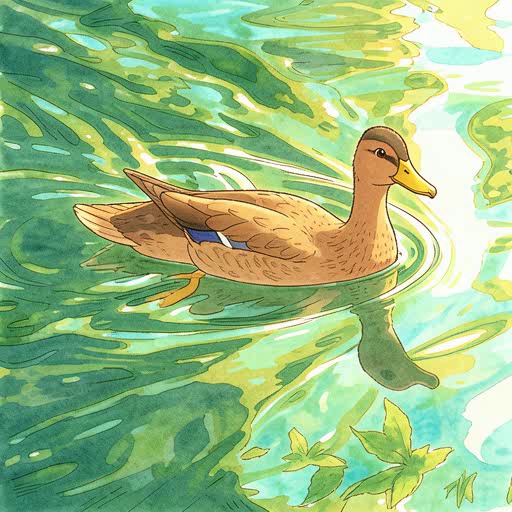} \\[-1.0pt]
\qualtime{$t_2$} &
\includegraphics[width=0.120\linewidth,height=0.120\linewidth]{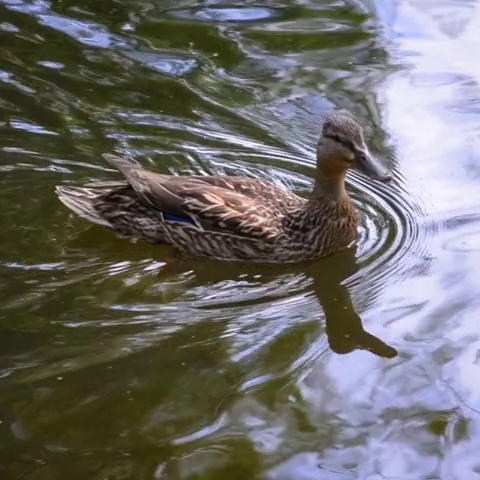} &
\includegraphics[width=0.120\linewidth,height=0.120\linewidth]{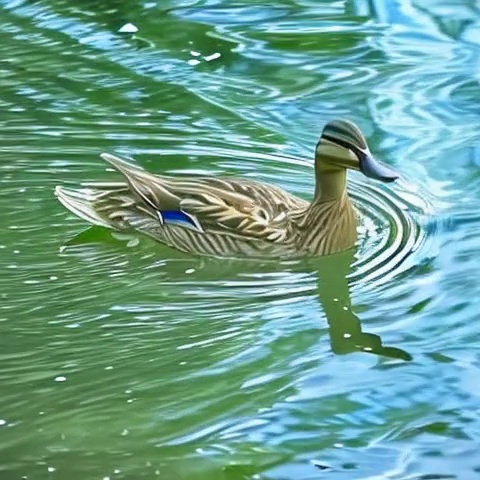} &
\includegraphics[width=0.120\linewidth,height=0.120\linewidth]{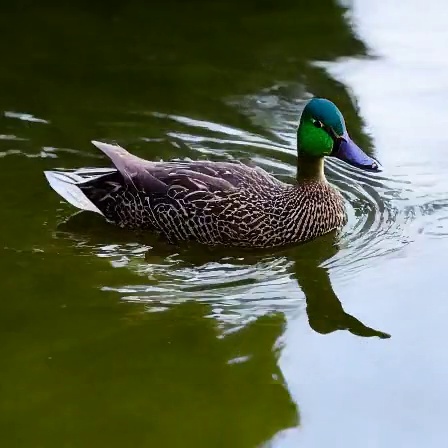} &
\includegraphics[width=0.120\linewidth,height=0.120\linewidth]{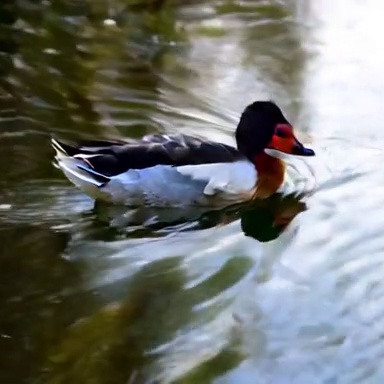} &
\includegraphics[width=0.120\linewidth,height=0.120\linewidth]{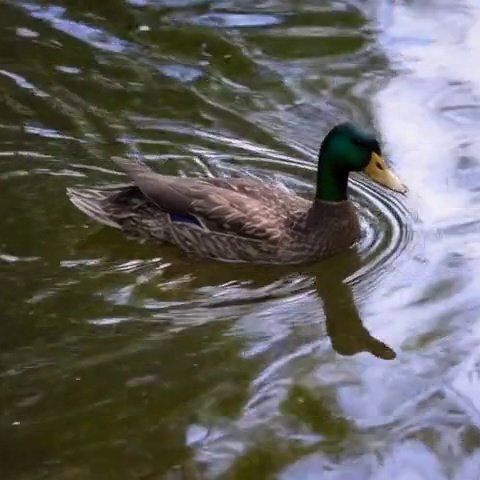} &
\includegraphics[width=0.120\linewidth,height=0.120\linewidth]{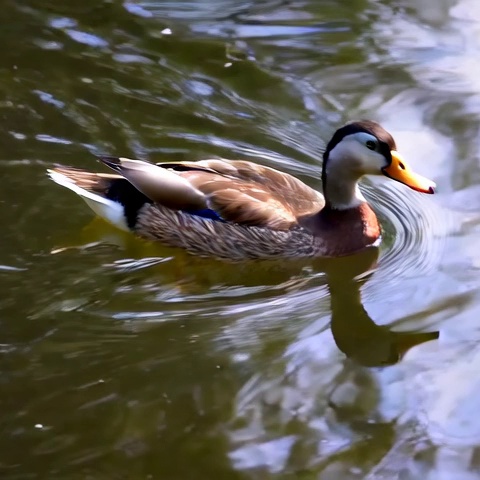} &
\includegraphics[width=0.120\linewidth,height=0.120\linewidth]{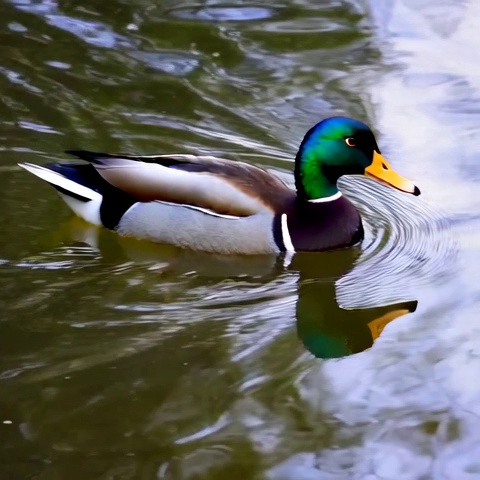} &
\includegraphics[width=0.120\linewidth,height=0.120\linewidth]{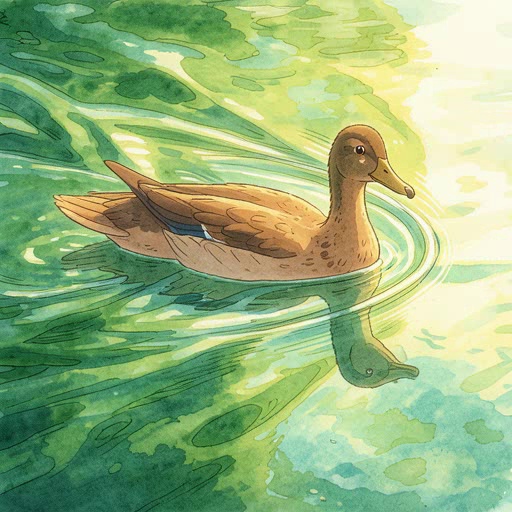} \\[-1.0pt]
\qualtime{$t_3$} &
\includegraphics[width=0.120\linewidth,height=0.120\linewidth]{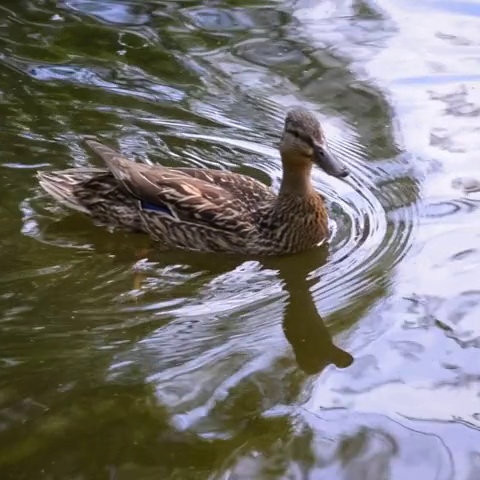} &
\includegraphics[width=0.120\linewidth,height=0.120\linewidth]{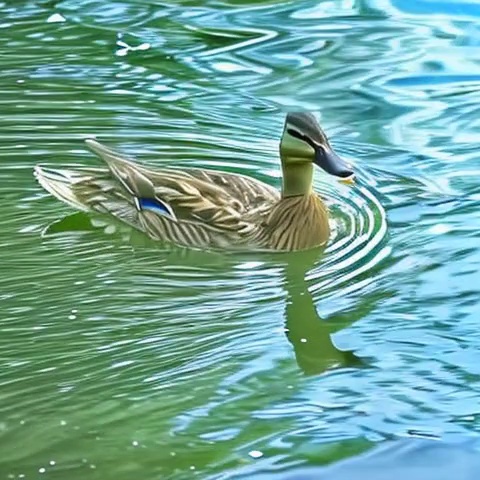} &
\includegraphics[width=0.120\linewidth,height=0.120\linewidth]{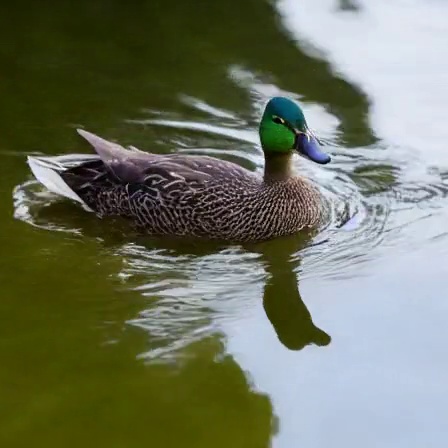} &
\includegraphics[width=0.120\linewidth,height=0.120\linewidth]{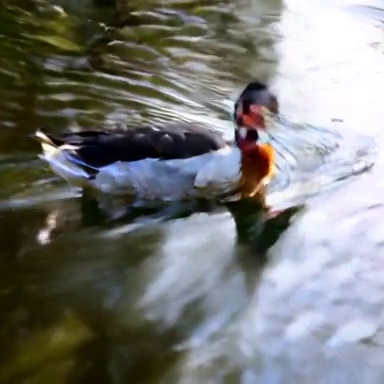} &
\includegraphics[width=0.120\linewidth,height=0.120\linewidth]{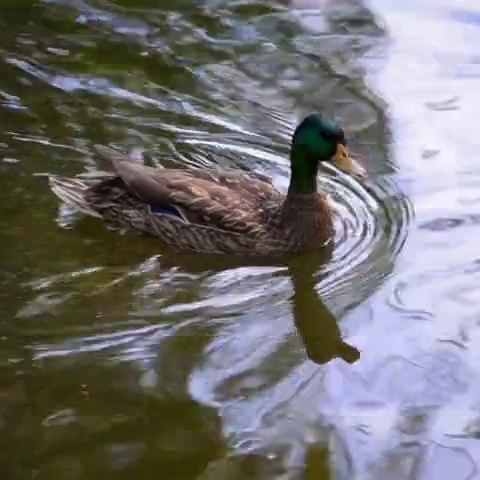} &
\includegraphics[width=0.120\linewidth,height=0.120\linewidth]{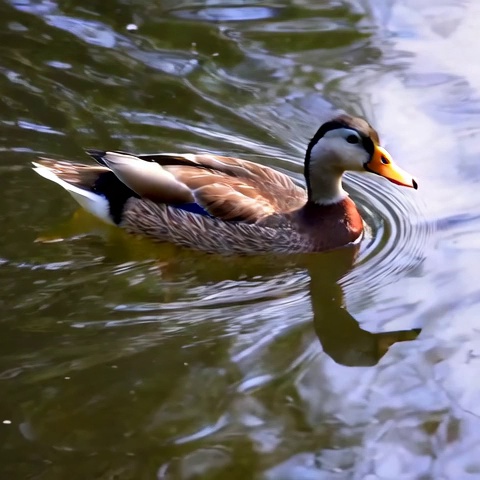} &
\includegraphics[width=0.120\linewidth,height=0.120\linewidth]{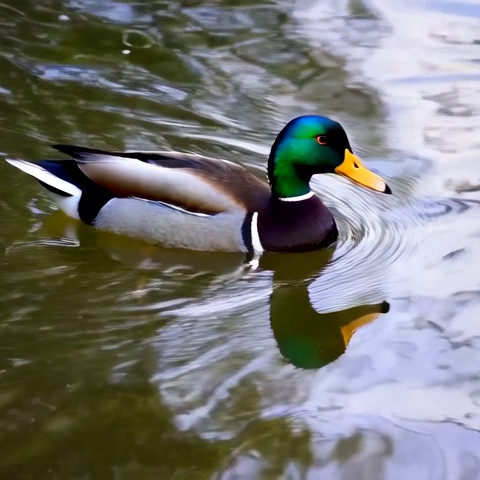} &
\includegraphics[width=0.120\linewidth,height=0.120\linewidth]{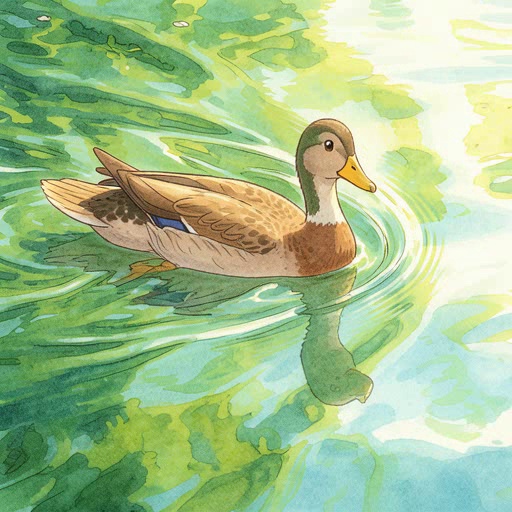} \\[-1.0pt]
\qualtime{$t_4$} &
\includegraphics[width=0.120\linewidth,height=0.120\linewidth]{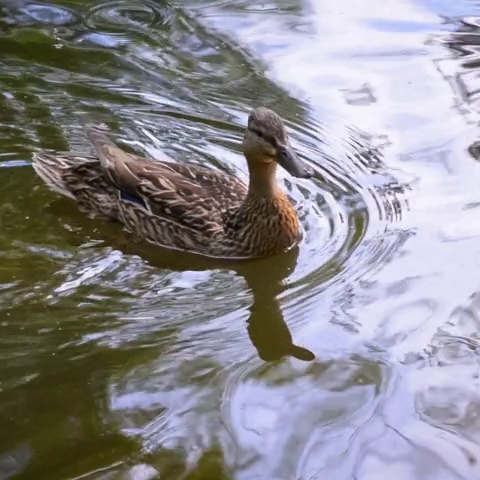} &
\includegraphics[width=0.120\linewidth,height=0.120\linewidth]{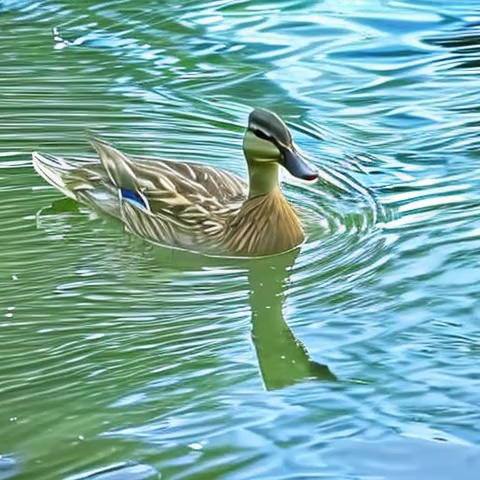} &
\includegraphics[width=0.120\linewidth,height=0.120\linewidth]{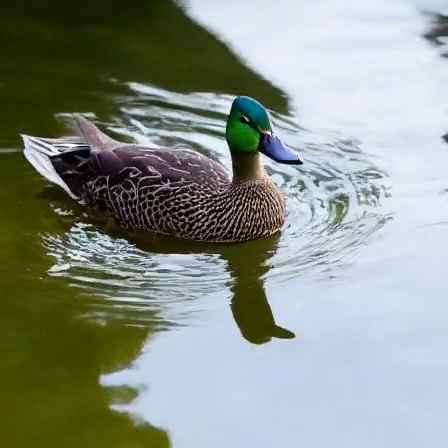} &
\includegraphics[width=0.120\linewidth,height=0.120\linewidth]{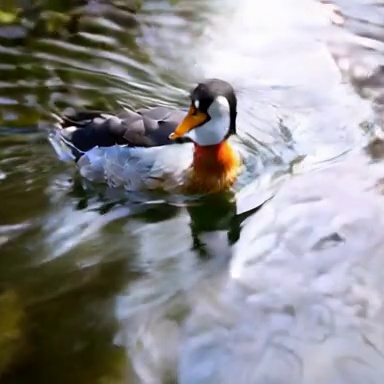} &
\includegraphics[width=0.120\linewidth,height=0.120\linewidth]{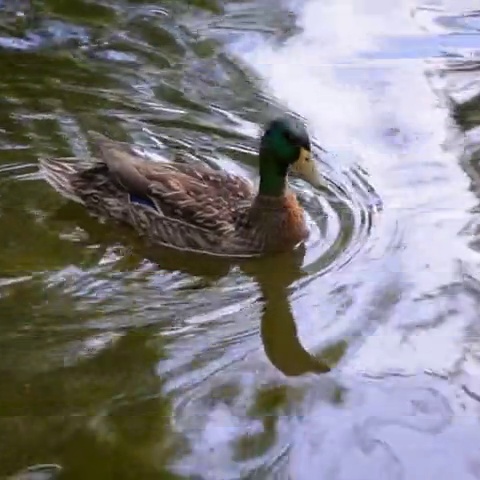} &
\includegraphics[width=0.120\linewidth,height=0.120\linewidth]{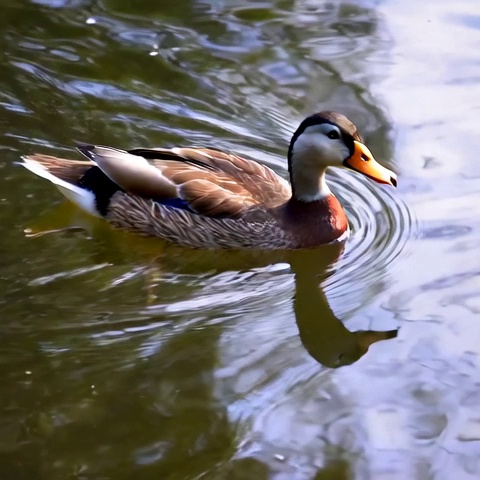} &
\includegraphics[width=0.120\linewidth,height=0.120\linewidth]{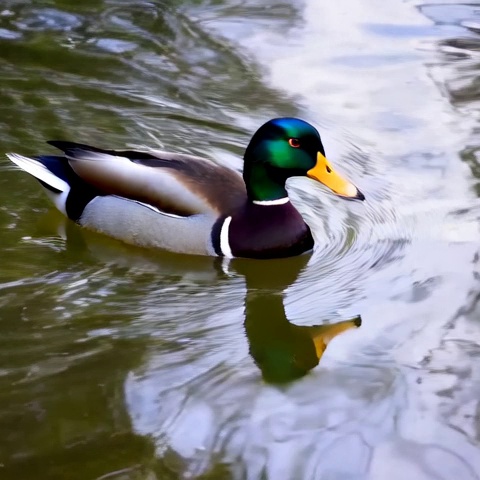} &
\includegraphics[width=0.120\linewidth,height=0.120\linewidth]{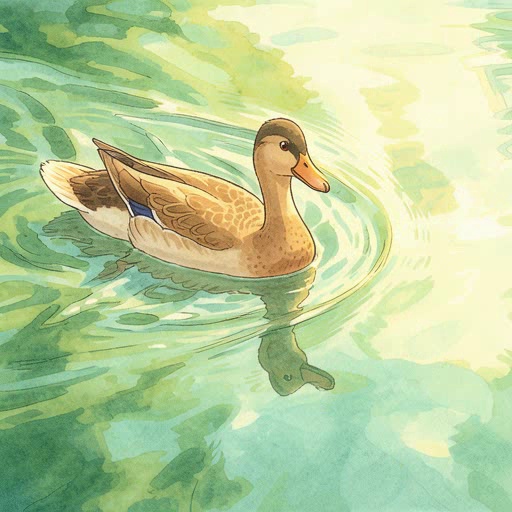} \\[-1.0pt]
\qualtime{$t_5$} &
\includegraphics[width=0.120\linewidth,height=0.120\linewidth]{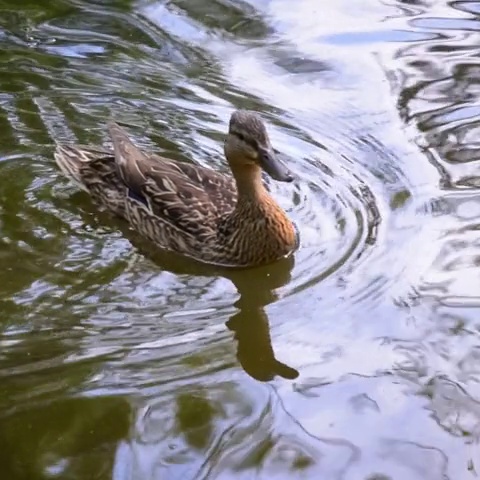} &
\includegraphics[width=0.120\linewidth,height=0.120\linewidth]{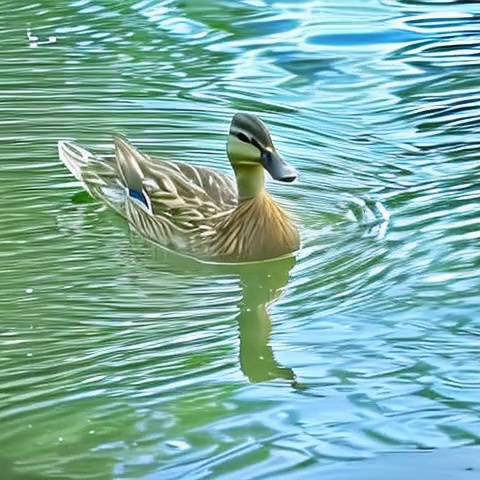} &
\includegraphics[width=0.120\linewidth,height=0.120\linewidth]{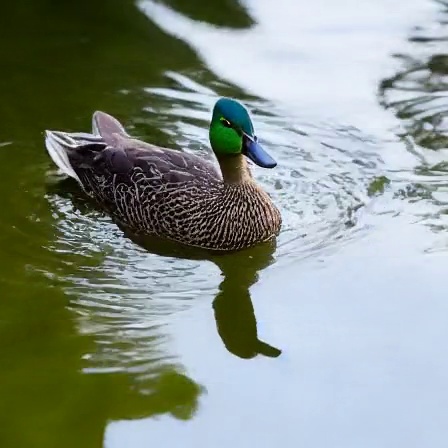} &
\includegraphics[width=0.120\linewidth,height=0.120\linewidth]{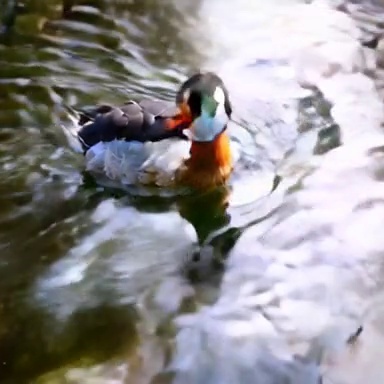} &
\includegraphics[width=0.120\linewidth,height=0.120\linewidth]{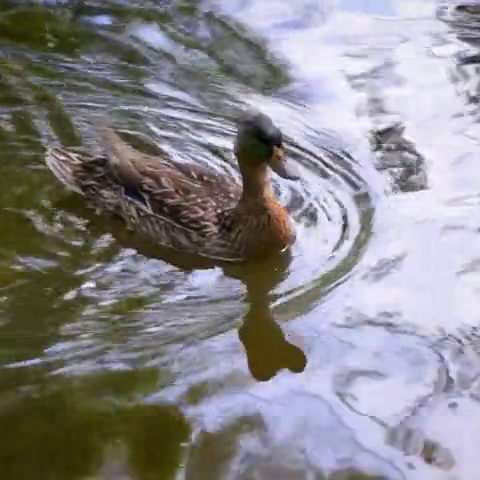} &
\includegraphics[width=0.120\linewidth,height=0.120\linewidth]{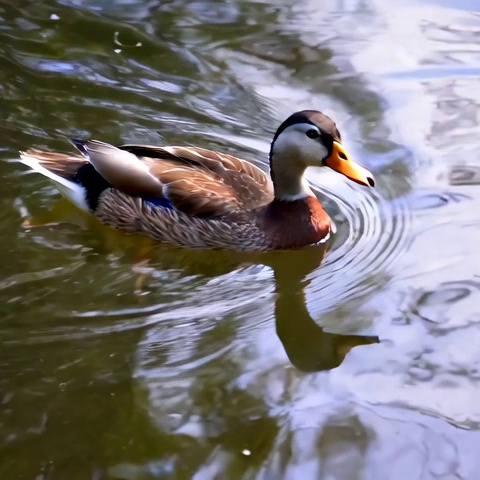} &
\includegraphics[width=0.120\linewidth,height=0.120\linewidth]{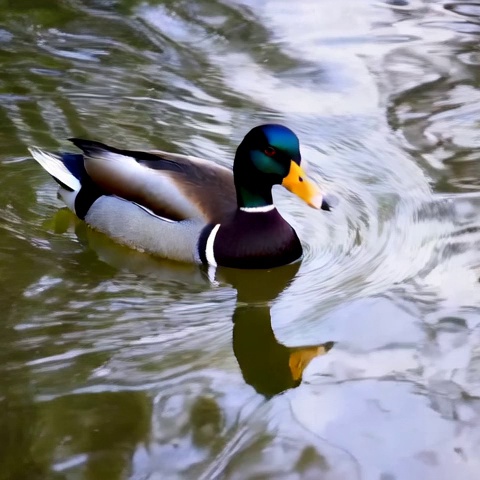} &
\includegraphics[width=0.120\linewidth,height=0.120\linewidth]{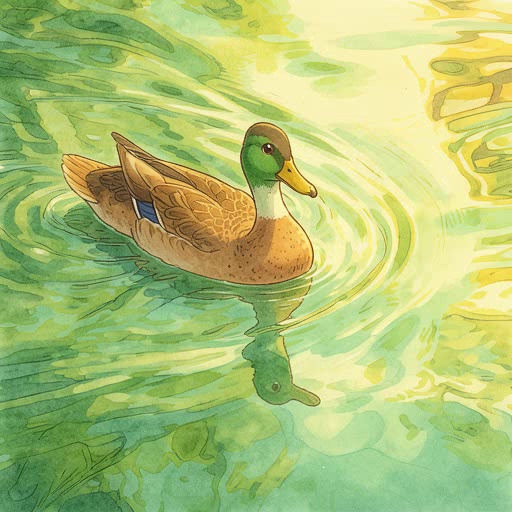} \\[-1.0pt]
\qualtime{$t_6$} &
\includegraphics[width=0.120\linewidth,height=0.120\linewidth]{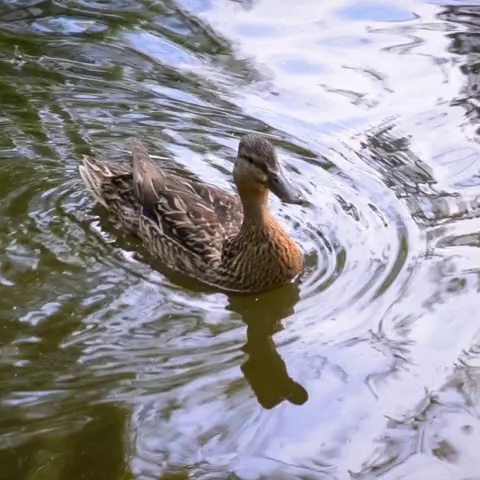} &
\includegraphics[width=0.120\linewidth,height=0.120\linewidth]{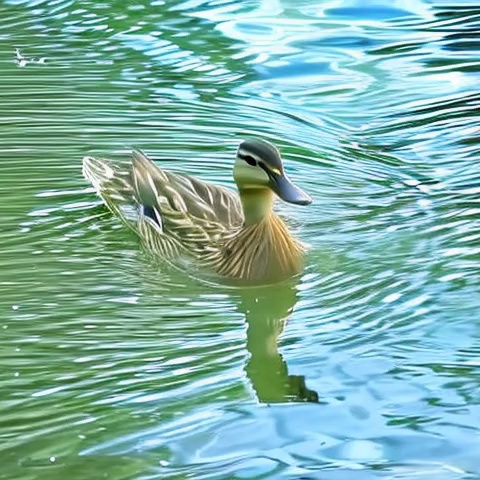} &
\includegraphics[width=0.120\linewidth,height=0.120\linewidth]{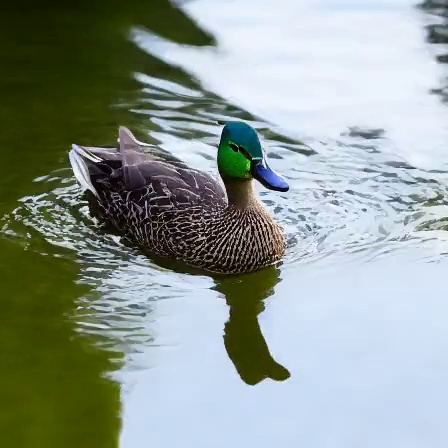} &
\includegraphics[width=0.120\linewidth,height=0.120\linewidth]{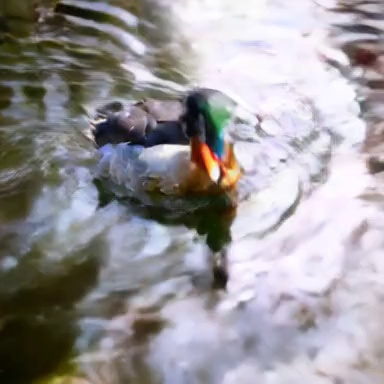} &
\includegraphics[width=0.120\linewidth,height=0.120\linewidth]{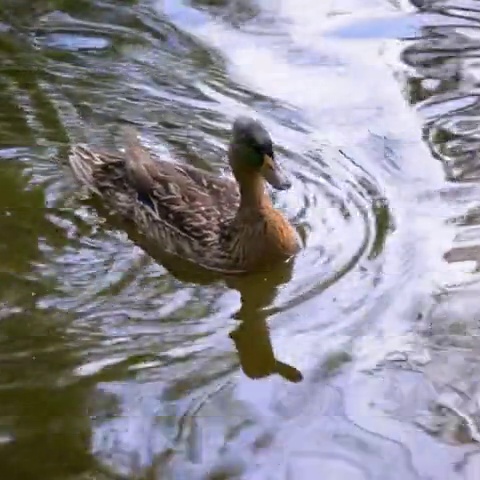} &
\includegraphics[width=0.120\linewidth,height=0.120\linewidth]{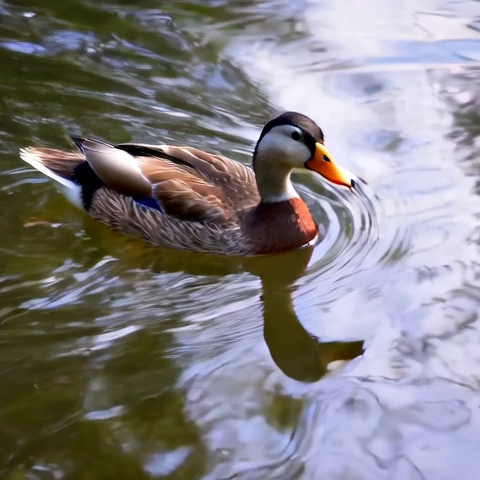} &
\includegraphics[width=0.120\linewidth,height=0.120\linewidth]{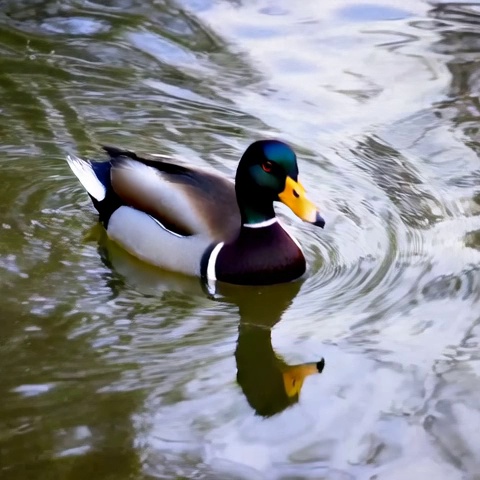} &
\includegraphics[width=0.120\linewidth,height=0.120\linewidth]{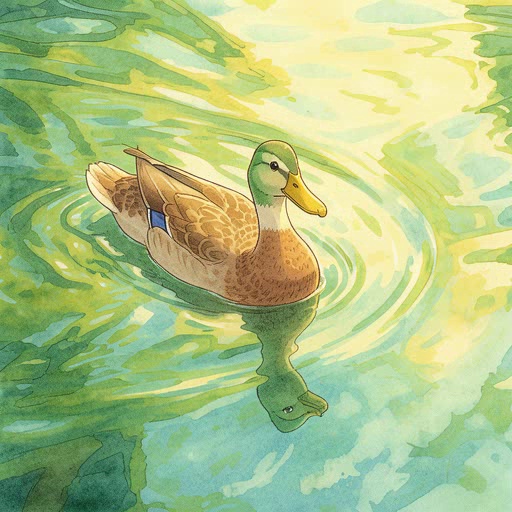} \\[-1.0pt]
\qualtime{$t_7$} &
\includegraphics[width=0.120\linewidth,height=0.120\linewidth]{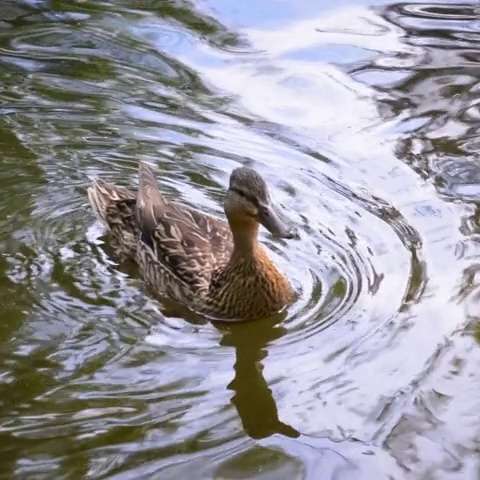} &
\includegraphics[width=0.120\linewidth,height=0.120\linewidth]{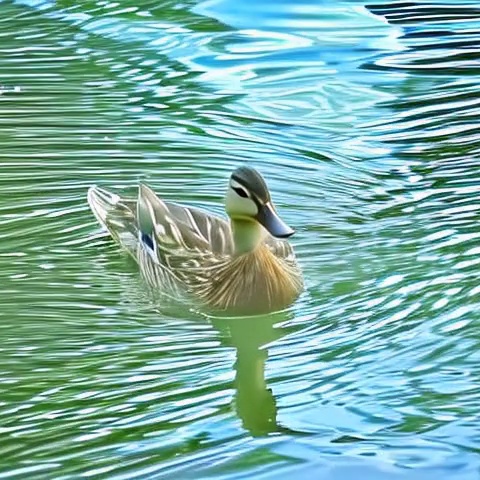} &
\includegraphics[width=0.120\linewidth,height=0.120\linewidth]{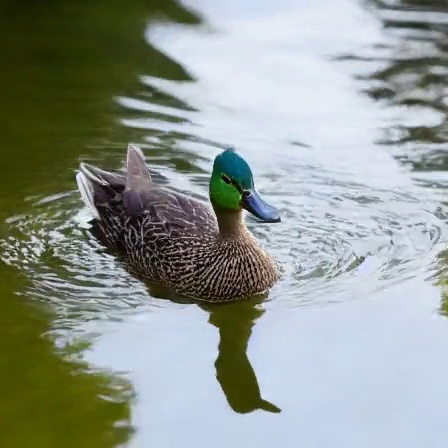} &
\includegraphics[width=0.120\linewidth,height=0.120\linewidth]{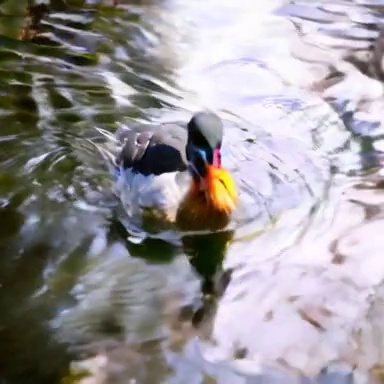} &
\includegraphics[width=0.120\linewidth,height=0.120\linewidth]{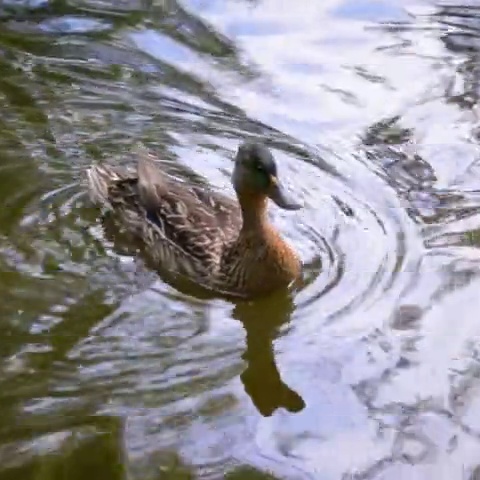} &
\includegraphics[width=0.120\linewidth,height=0.120\linewidth]{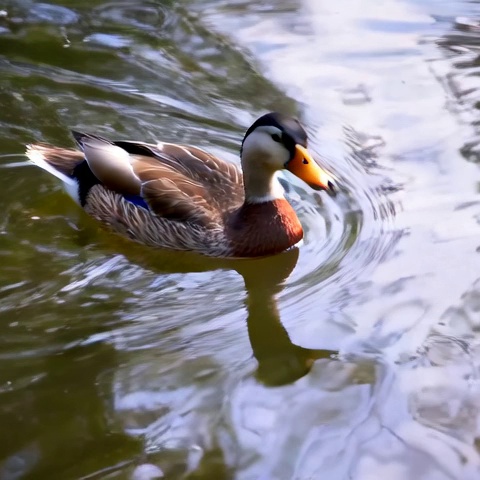} &
\includegraphics[width=0.120\linewidth,height=0.120\linewidth]{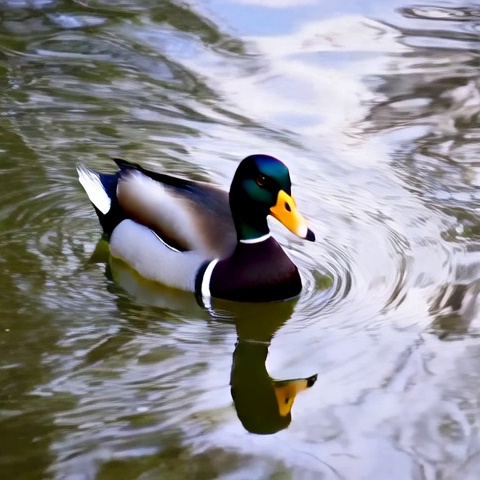} &
\includegraphics[width=0.120\linewidth,height=0.120\linewidth]{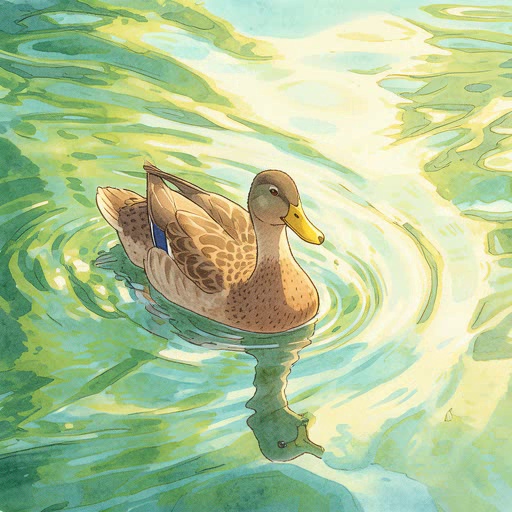} \\[-1.0pt]
\qualtime{$t_8$} &
\includegraphics[width=0.120\linewidth,height=0.120\linewidth]{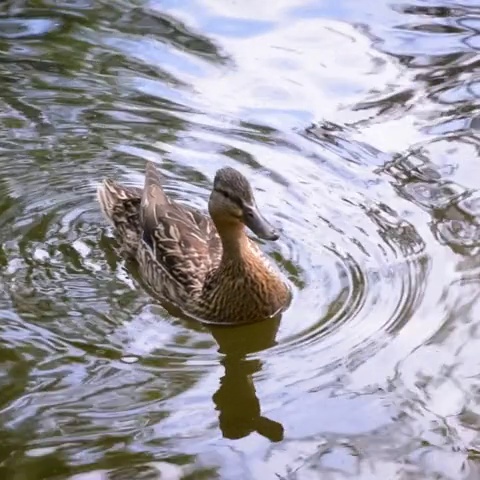} &
\includegraphics[width=0.120\linewidth,height=0.120\linewidth]{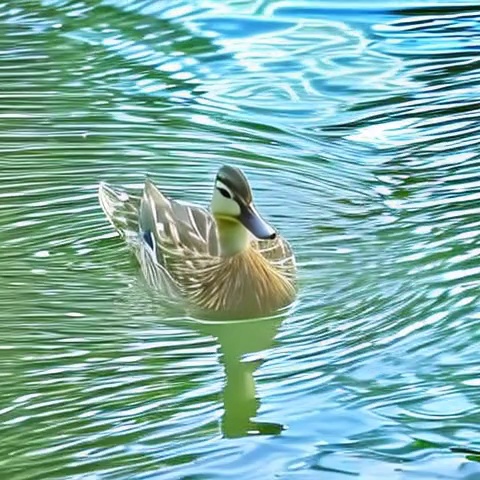} &
\includegraphics[width=0.120\linewidth,height=0.120\linewidth]{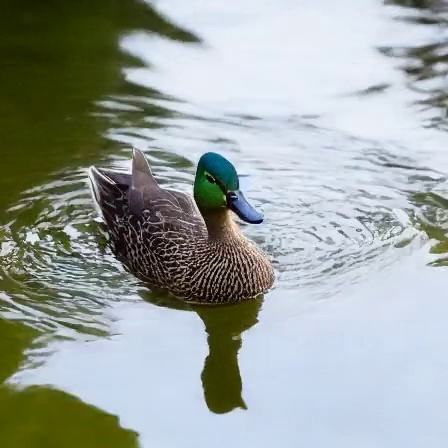} &
\includegraphics[width=0.120\linewidth,height=0.120\linewidth]{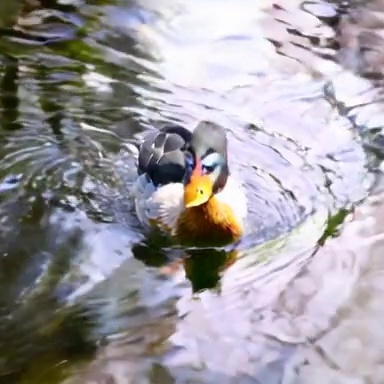} &
\includegraphics[width=0.120\linewidth,height=0.120\linewidth]{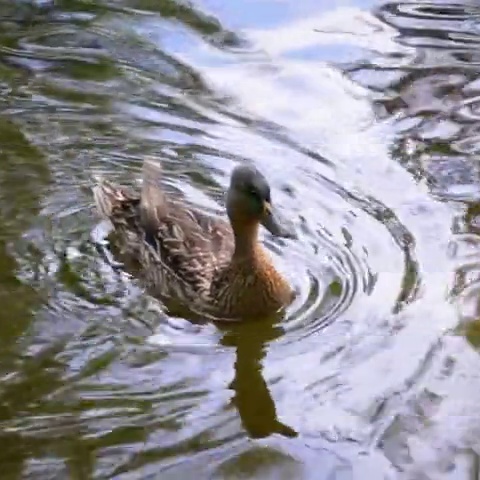} &
\includegraphics[width=0.120\linewidth,height=0.120\linewidth]{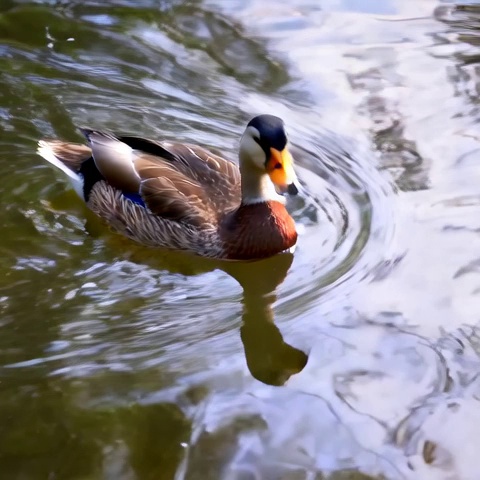} &
\includegraphics[width=0.120\linewidth,height=0.120\linewidth]{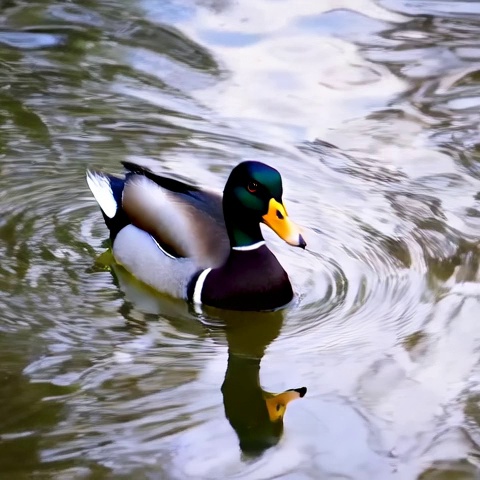} &
\includegraphics[width=0.120\linewidth,height=0.120\linewidth]{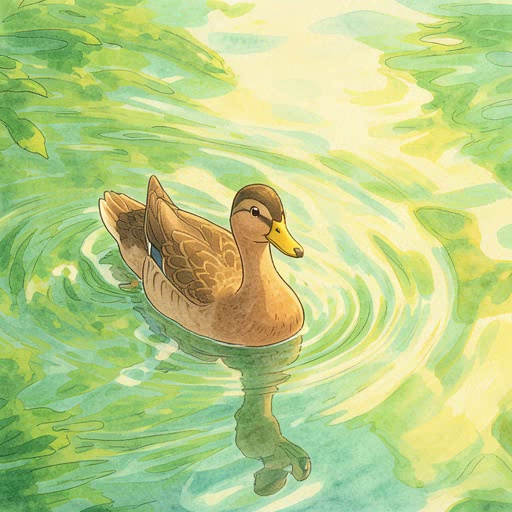} \\[-1.0pt]
\qualtime{$t_9$} &
\includegraphics[width=0.120\linewidth,height=0.120\linewidth]{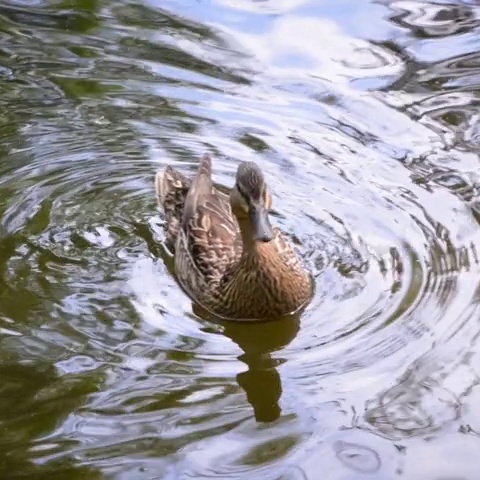} &
\includegraphics[width=0.120\linewidth,height=0.120\linewidth]{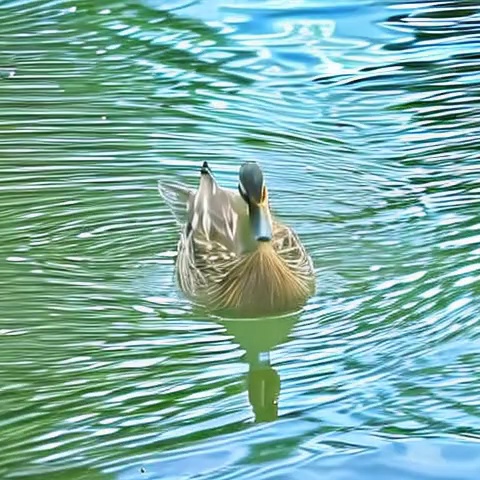} &
\includegraphics[width=0.120\linewidth,height=0.120\linewidth]{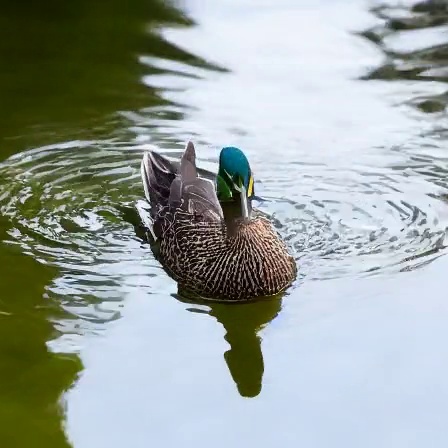} &
\includegraphics[width=0.120\linewidth,height=0.120\linewidth]{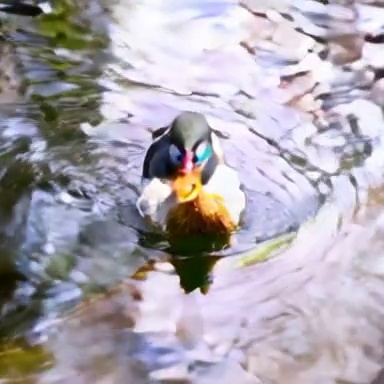} &
\includegraphics[width=0.120\linewidth,height=0.120\linewidth]{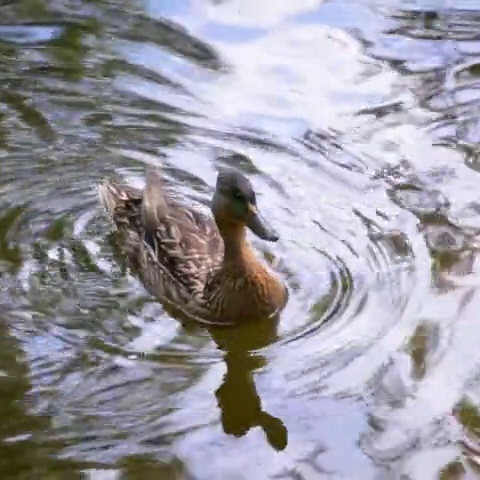} &
\includegraphics[width=0.120\linewidth,height=0.120\linewidth]{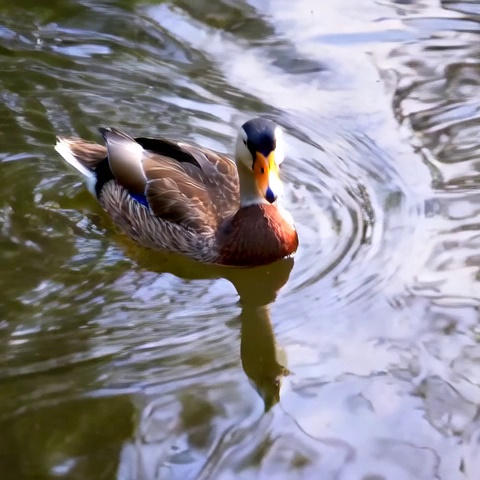} &
\includegraphics[width=0.120\linewidth,height=0.120\linewidth]{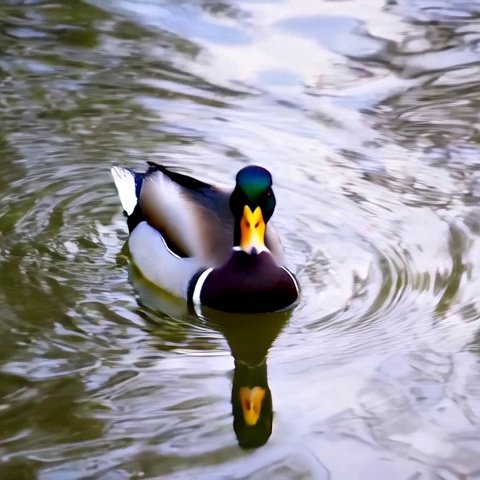} &
\includegraphics[width=0.120\linewidth,height=0.120\linewidth]{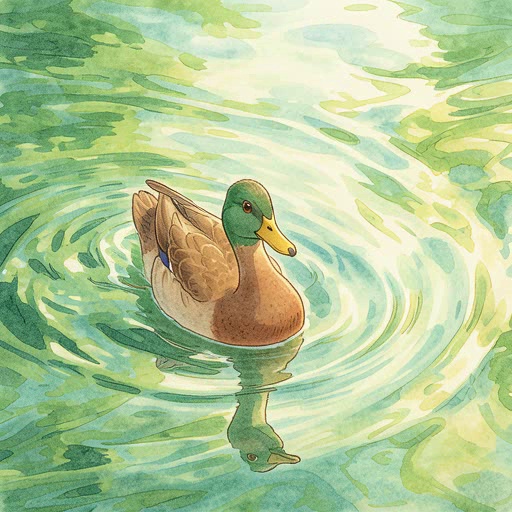} \\[-1.0pt]
\end{tabular}%
}
\caption{\textbf{Global style-transfer comparison.}
Given the instruction to change the mallard duck video to Ghibli art style, \modelnamebf{} applies a consistent stylization while preserving the duck's pose, motion, and surrounding scene. Most baselines either remain close to the source appearance or introduce substantial structural changes; AnyV2V produces a stronger stylization but relies on an external image editor. Frames are sampled at uniformly spaced temporal positions throughout each video.}
\label{fig:qual_ghibli_duck}
\end{figure*}

\begin{figure*}[t!]
\centering
\setlength{\tabcolsep}{0.45pt}
\renewcommand{\arraystretch}{0.62}
\scriptsize
\def\qualtime#1{\raisebox{0.057\linewidth}{\makebox[0.018\linewidth][r]{#1}}}
\resizebox{\textwidth}{!}{%
\begin{tabular}{@{}r@{\hspace{1pt}}cccccccc@{}}
\multicolumn{9}{c}{\small\textbf{Prompt:} Change the material of the butterfly from a living creature to glass.} \\[3pt]
& \textbf{Source} & \textbf{AnyV2V~\cite{ku2024anyv2v}} & \textbf{VidToMe~\cite{li2024vidtome}} & \textbf{Pyramid-Edit~\cite{li2025five}} & \textbf{StreamEdit~\cite{jiao2026streamedit}} & \textbf{Wan-Edit~\cite{li2025five}} & \textbf{FlowDirector~\cite{li2026flowdirector}} & \makecell{\textbf{\modelname{}}\\\textbf{(Ours)}} \\[2pt]
\qualtime{$t_1$} &
\includegraphics[width=0.120\linewidth,height=0.120\linewidth]{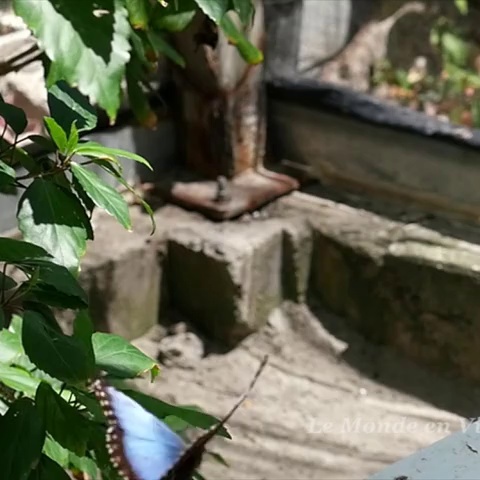} &
\includegraphics[width=0.120\linewidth,height=0.120\linewidth]{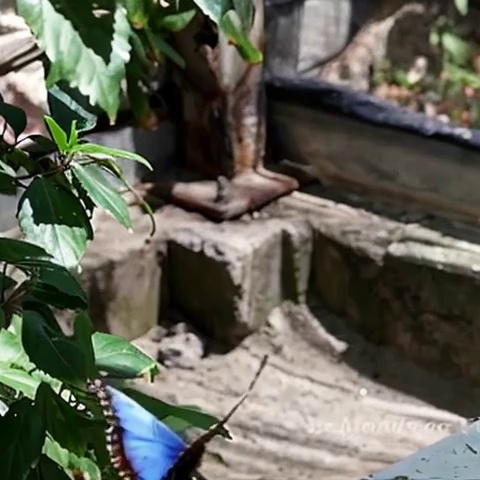} &
\includegraphics[width=0.120\linewidth,height=0.120\linewidth]{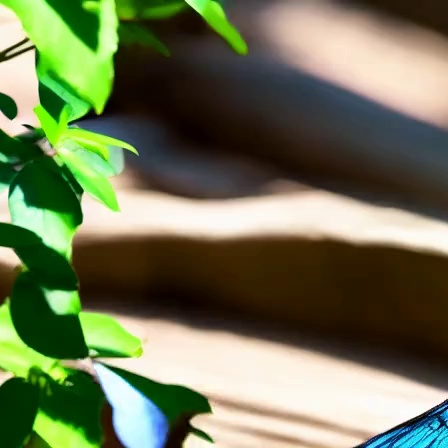} &
\includegraphics[width=0.120\linewidth,height=0.120\linewidth]{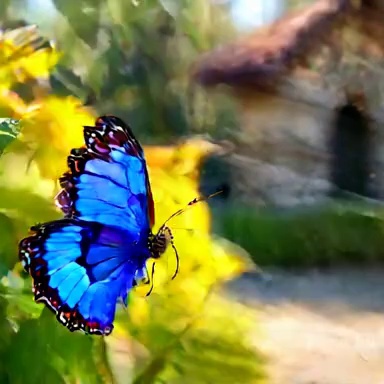} &
\includegraphics[width=0.120\linewidth,height=0.120\linewidth]{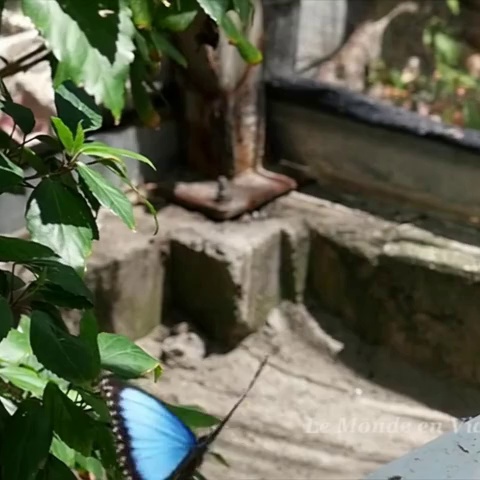} &
\includegraphics[width=0.120\linewidth,height=0.120\linewidth]{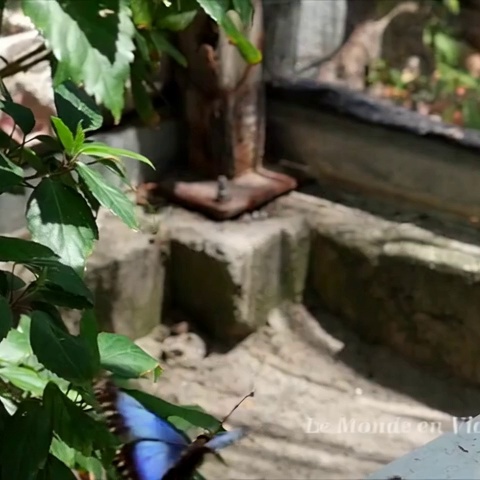} &
\includegraphics[width=0.120\linewidth,height=0.120\linewidth]{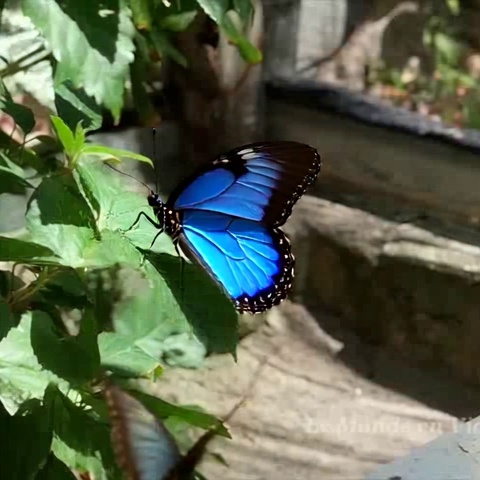} &
\includegraphics[width=0.120\linewidth,height=0.120\linewidth]{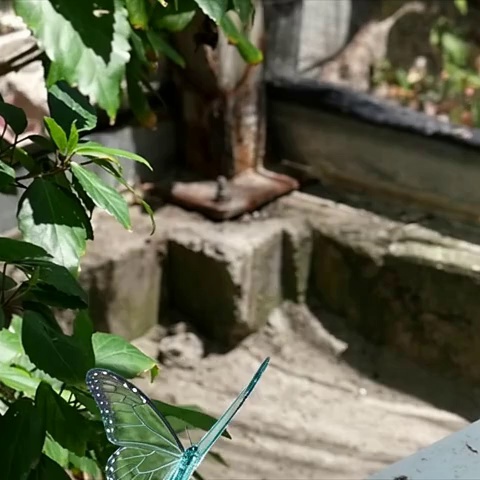} \\[-1.0pt]
\qualtime{$t_2$} &
\includegraphics[width=0.120\linewidth,height=0.120\linewidth]{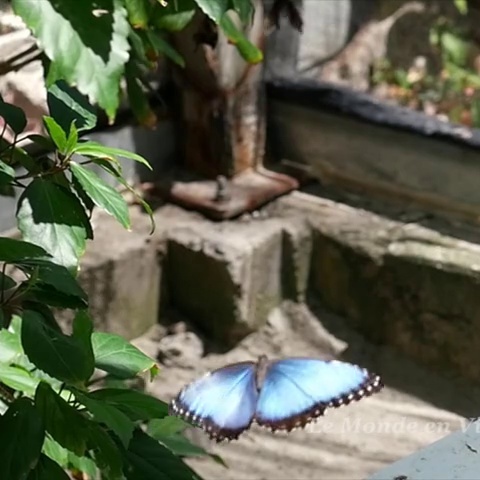} &
\includegraphics[width=0.120\linewidth,height=0.120\linewidth]{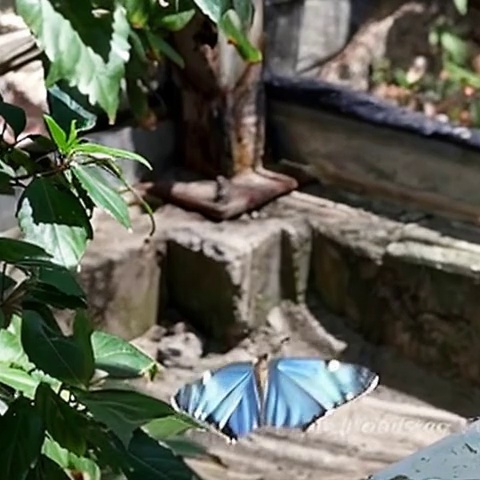} &
\includegraphics[width=0.120\linewidth,height=0.120\linewidth]{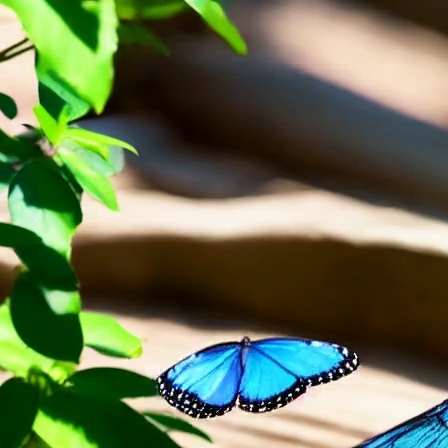} &
\includegraphics[width=0.120\linewidth,height=0.120\linewidth]{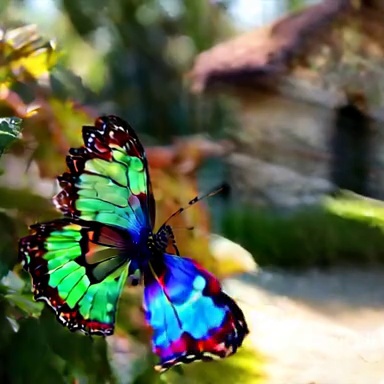} &
\includegraphics[width=0.120\linewidth,height=0.120\linewidth]{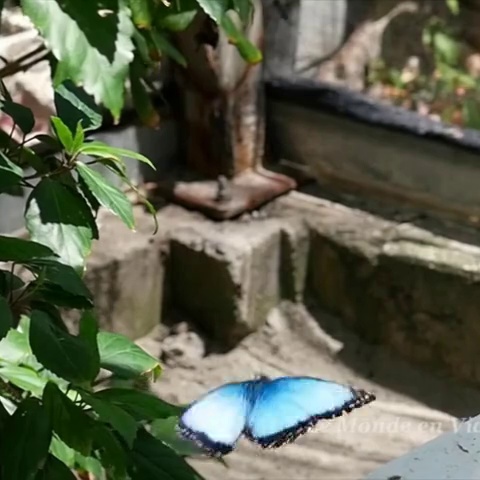} &
\includegraphics[width=0.120\linewidth,height=0.120\linewidth]{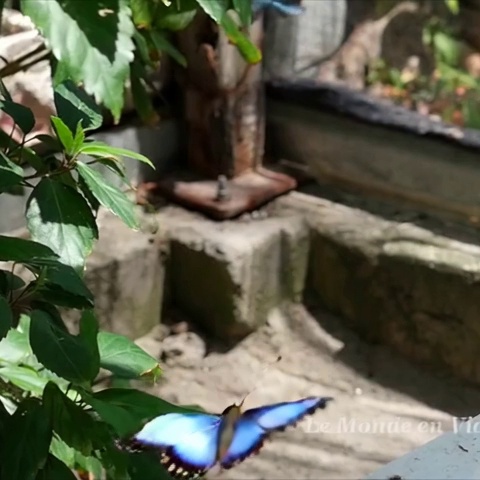} &
\includegraphics[width=0.120\linewidth,height=0.120\linewidth]{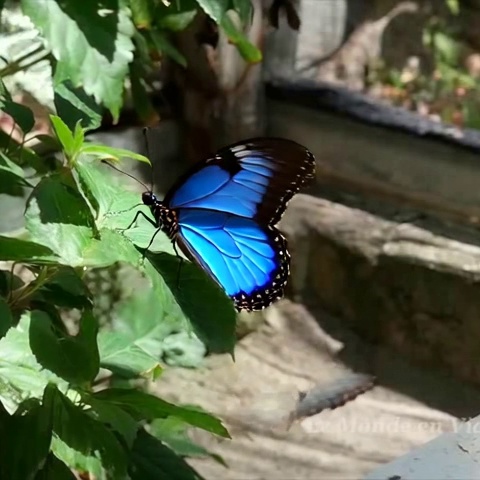} &
\includegraphics[width=0.120\linewidth,height=0.120\linewidth]{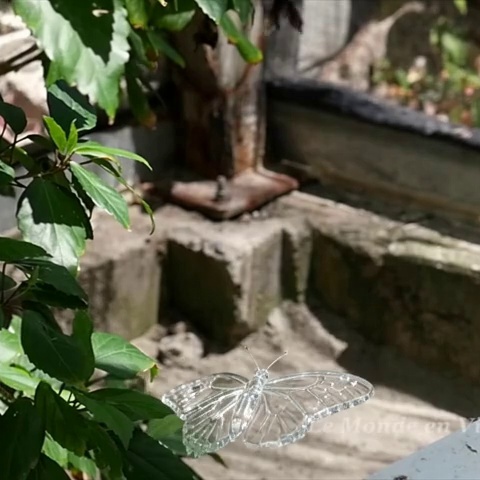} \\[-1.0pt]
\qualtime{$t_3$} &
\includegraphics[width=0.120\linewidth,height=0.120\linewidth]{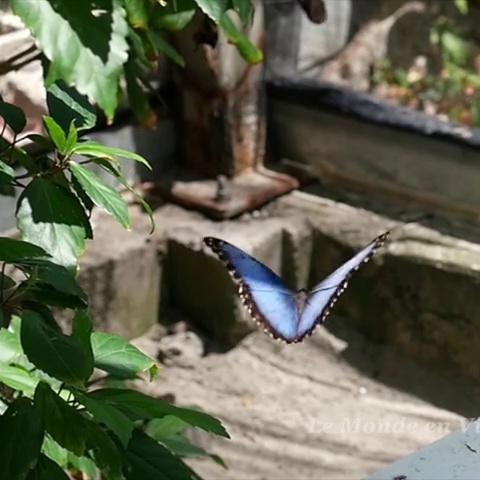} &
\includegraphics[width=0.120\linewidth,height=0.120\linewidth]{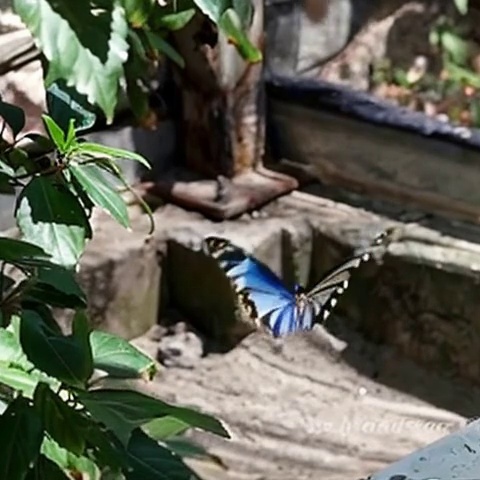} &
\includegraphics[width=0.120\linewidth,height=0.120\linewidth]{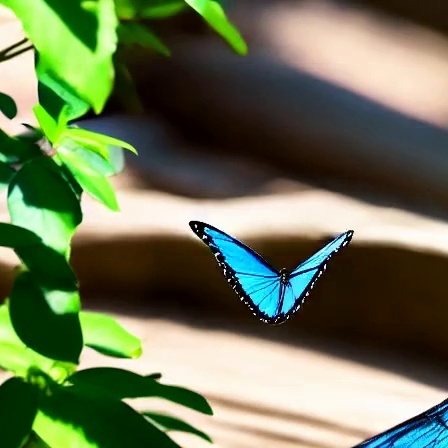} &
\includegraphics[width=0.120\linewidth,height=0.120\linewidth]{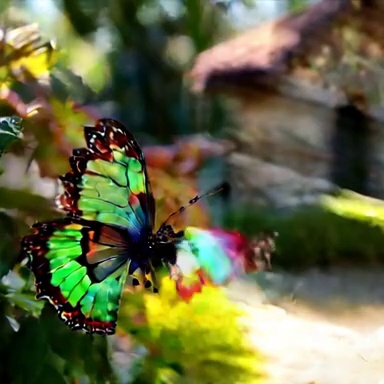} &
\includegraphics[width=0.120\linewidth,height=0.120\linewidth]{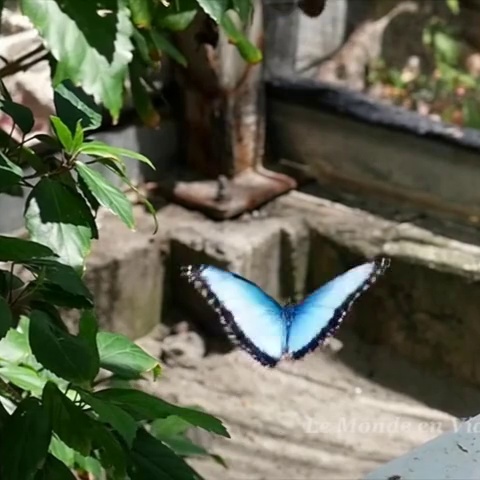} &
\includegraphics[width=0.120\linewidth,height=0.120\linewidth]{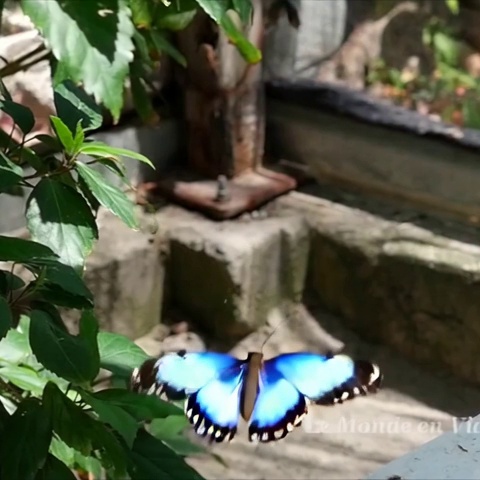} &
\includegraphics[width=0.120\linewidth,height=0.120\linewidth]{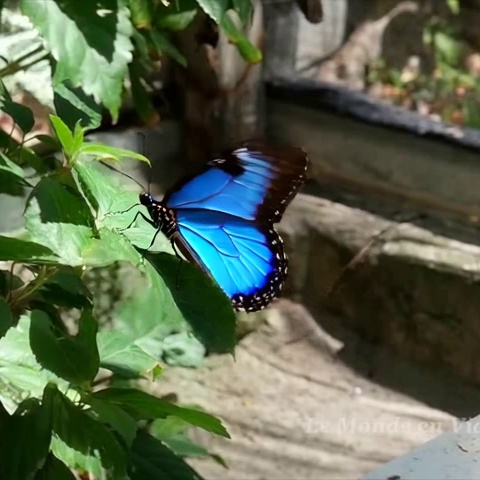} &
\includegraphics[width=0.120\linewidth,height=0.120\linewidth]{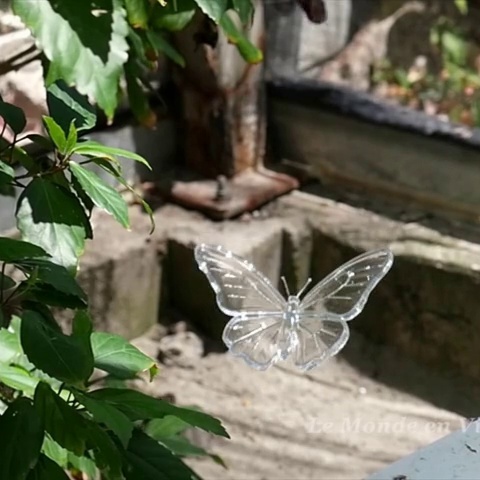} \\[-1.0pt]
\qualtime{$t_4$} &
\includegraphics[width=0.120\linewidth,height=0.120\linewidth]{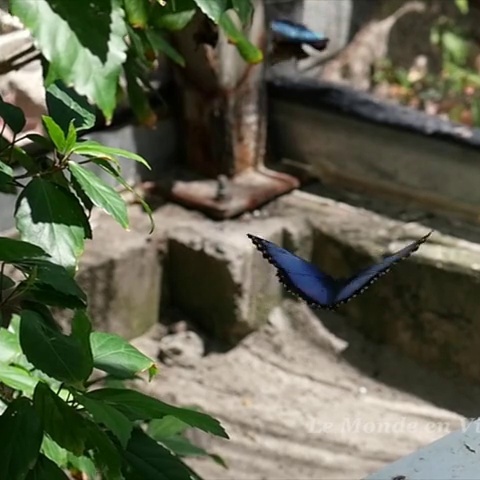} &
\includegraphics[width=0.120\linewidth,height=0.120\linewidth]{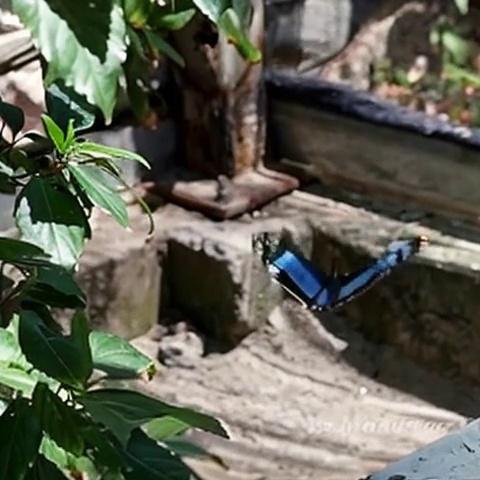} &
\includegraphics[width=0.120\linewidth,height=0.120\linewidth]{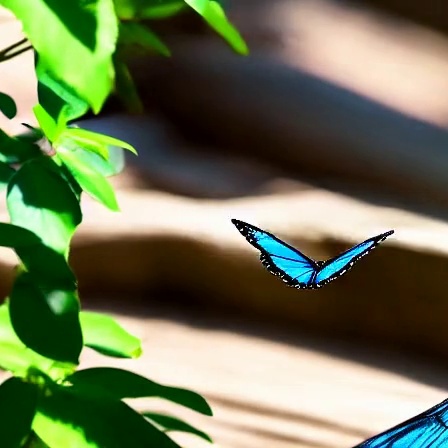} &
\includegraphics[width=0.120\linewidth,height=0.120\linewidth]{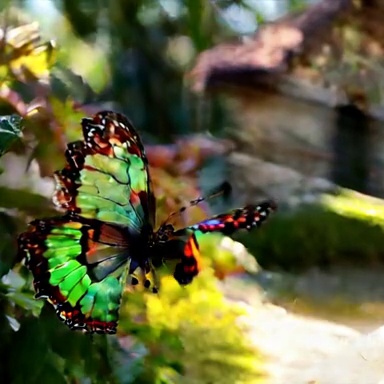} &
\includegraphics[width=0.120\linewidth,height=0.120\linewidth]{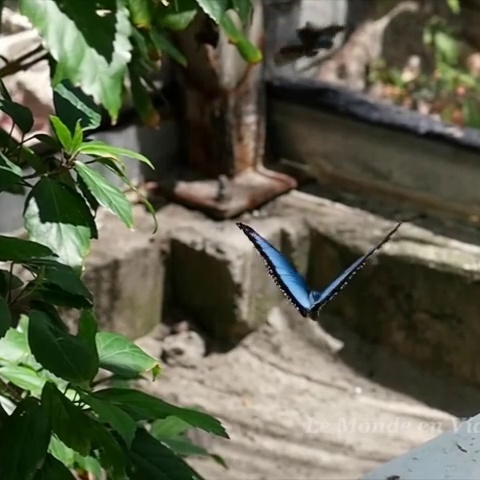} &
\includegraphics[width=0.120\linewidth,height=0.120\linewidth]{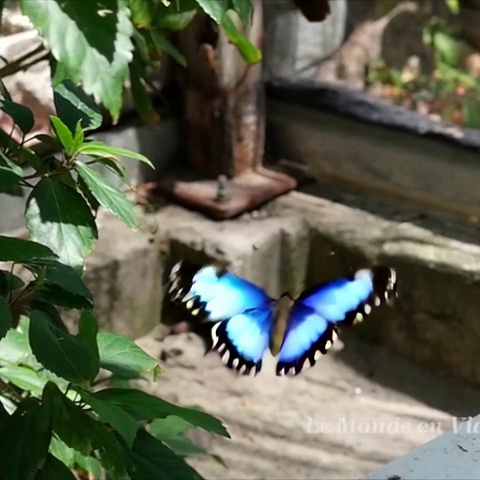} &
\includegraphics[width=0.120\linewidth,height=0.120\linewidth]{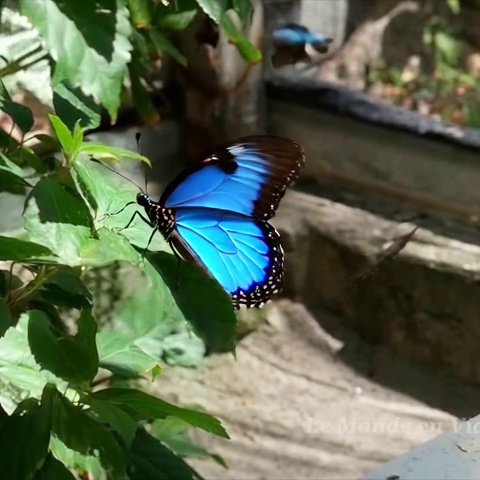} &
\includegraphics[width=0.120\linewidth,height=0.120\linewidth]{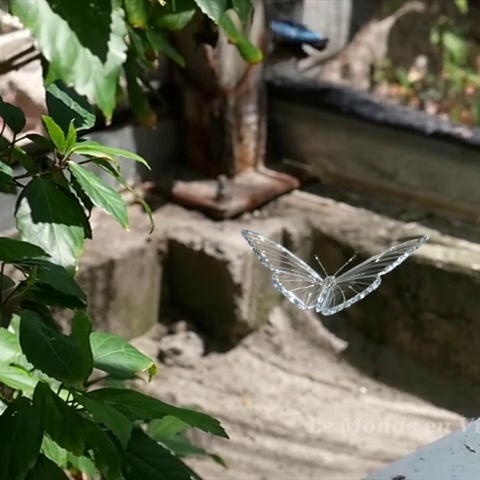} \\[-1.0pt]
\qualtime{$t_5$} &
\includegraphics[width=0.120\linewidth,height=0.120\linewidth]{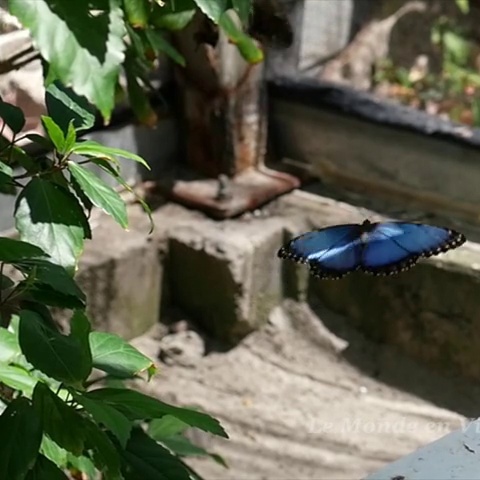} &
\includegraphics[width=0.120\linewidth,height=0.120\linewidth]{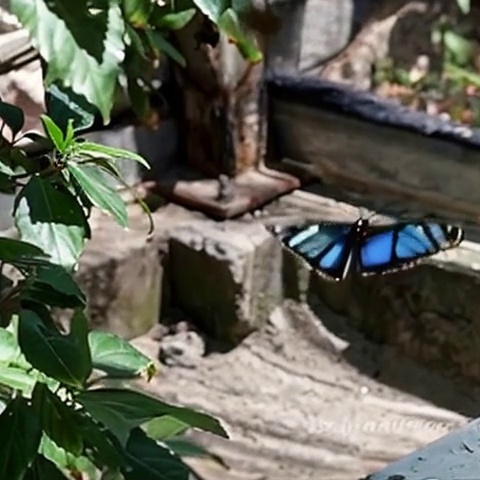} &
\includegraphics[width=0.120\linewidth,height=0.120\linewidth]{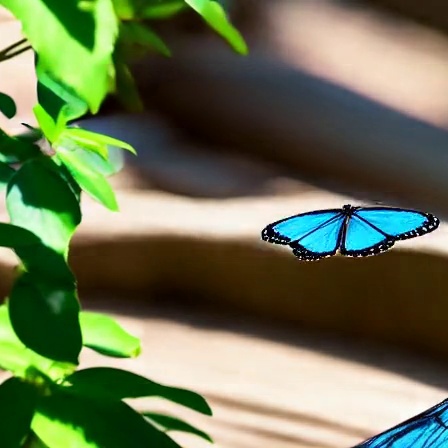} &
\includegraphics[width=0.120\linewidth,height=0.120\linewidth]{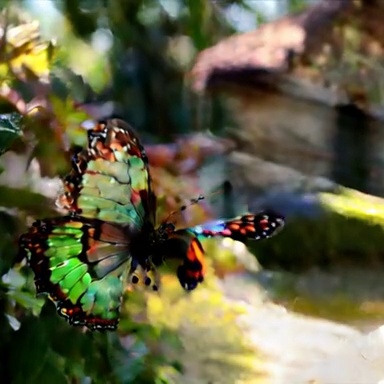} &
\includegraphics[width=0.120\linewidth,height=0.120\linewidth]{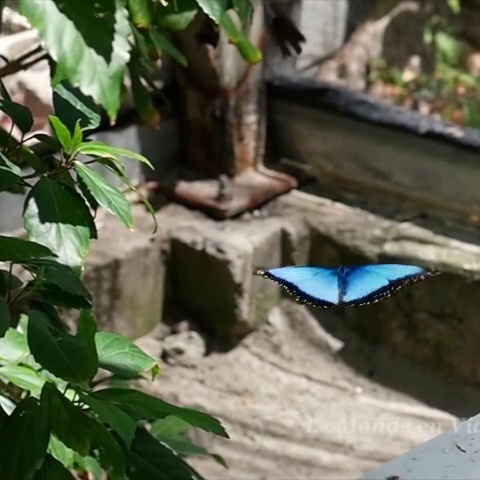} &
\includegraphics[width=0.120\linewidth,height=0.120\linewidth]{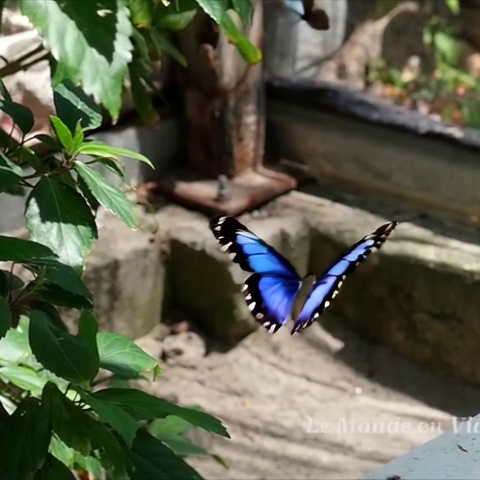} &
\includegraphics[width=0.120\linewidth,height=0.120\linewidth]{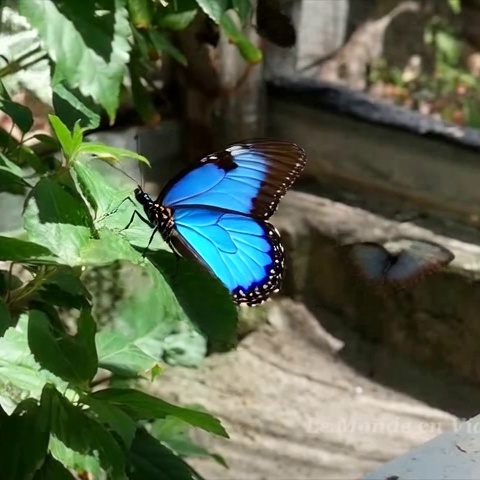} &
\includegraphics[width=0.120\linewidth,height=0.120\linewidth]{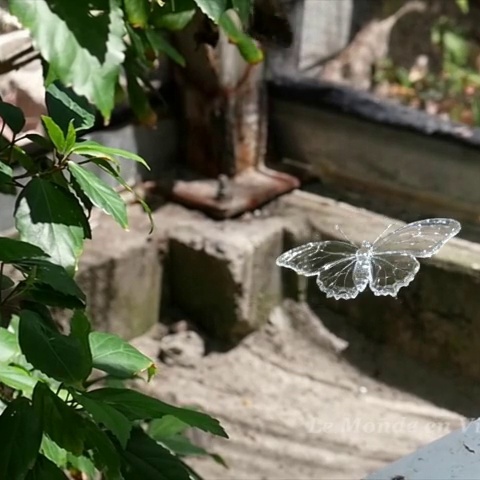} \\[-1.0pt]
\qualtime{$t_6$} &
\includegraphics[width=0.120\linewidth,height=0.120\linewidth]{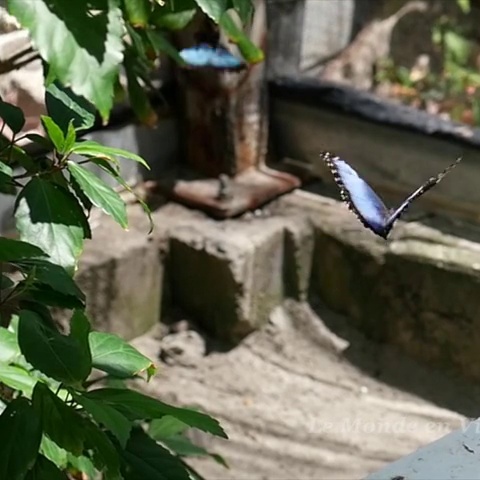} &
\includegraphics[width=0.120\linewidth,height=0.120\linewidth]{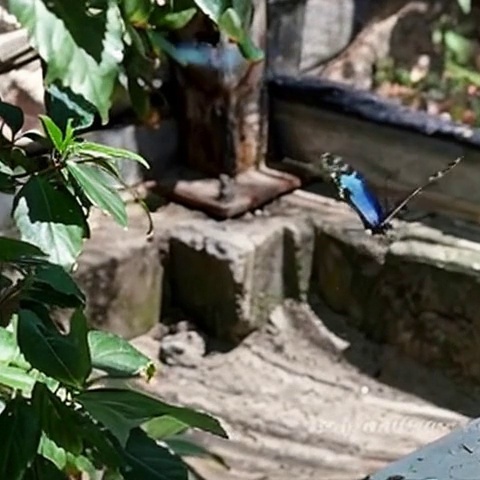} &
\includegraphics[width=0.120\linewidth,height=0.120\linewidth]{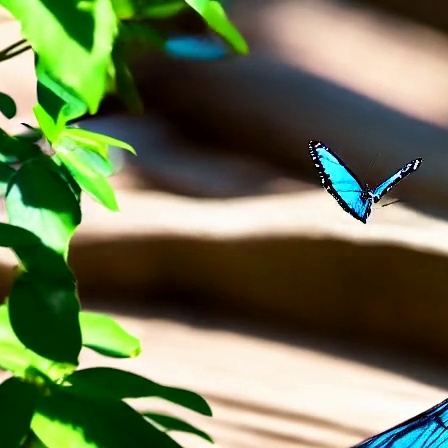} &
\includegraphics[width=0.120\linewidth,height=0.120\linewidth]{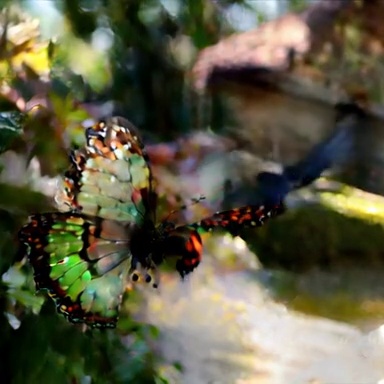} &
\includegraphics[width=0.120\linewidth,height=0.120\linewidth]{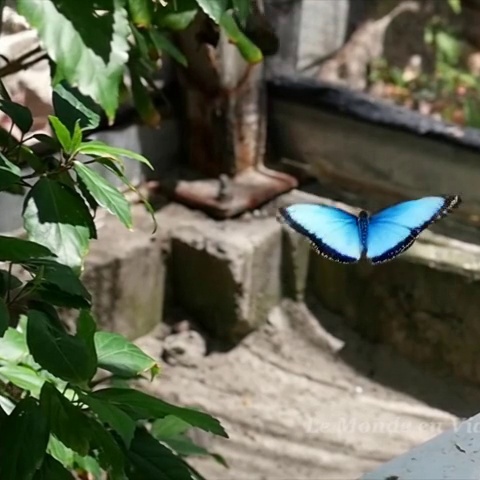} &
\includegraphics[width=0.120\linewidth,height=0.120\linewidth]{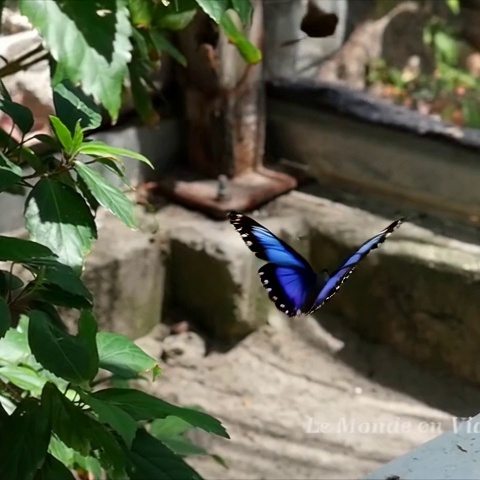} &
\includegraphics[width=0.120\linewidth,height=0.120\linewidth]{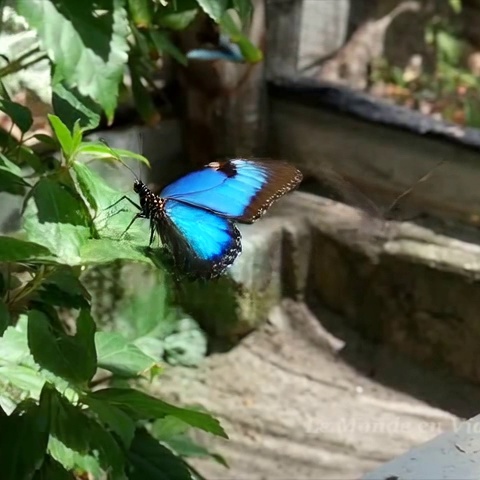} &
\includegraphics[width=0.120\linewidth,height=0.120\linewidth]{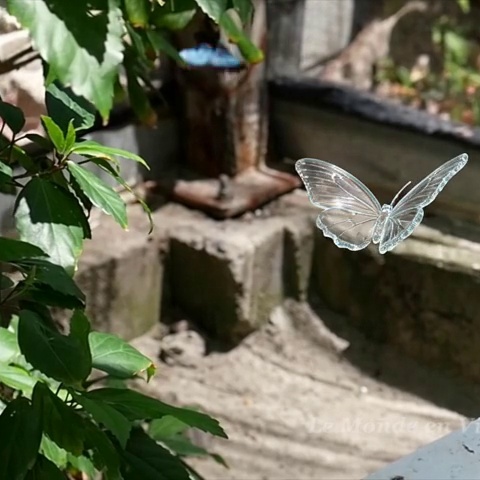} \\[-1.0pt]
\qualtime{$t_7$} &
\includegraphics[width=0.120\linewidth,height=0.120\linewidth]{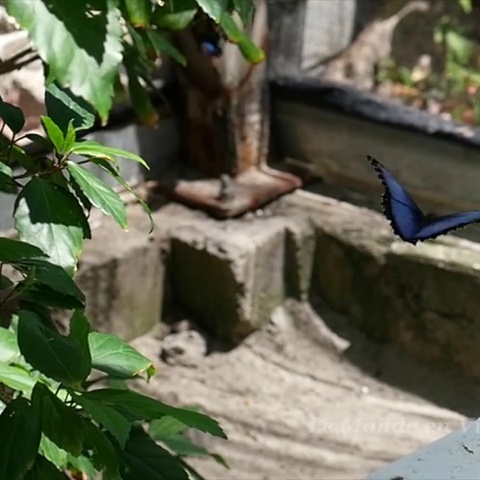} &
\includegraphics[width=0.120\linewidth,height=0.120\linewidth]{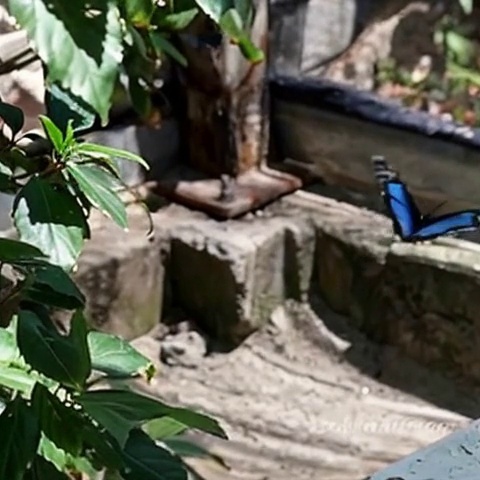} &
\includegraphics[width=0.120\linewidth,height=0.120\linewidth]{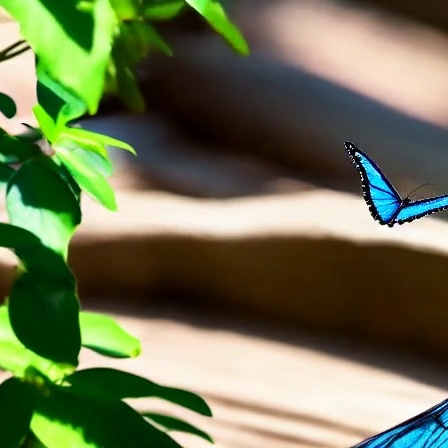} &
\includegraphics[width=0.120\linewidth,height=0.120\linewidth]{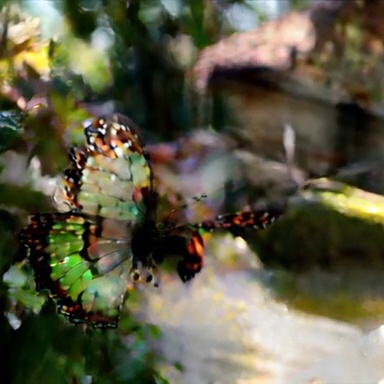} &
\includegraphics[width=0.120\linewidth,height=0.120\linewidth]{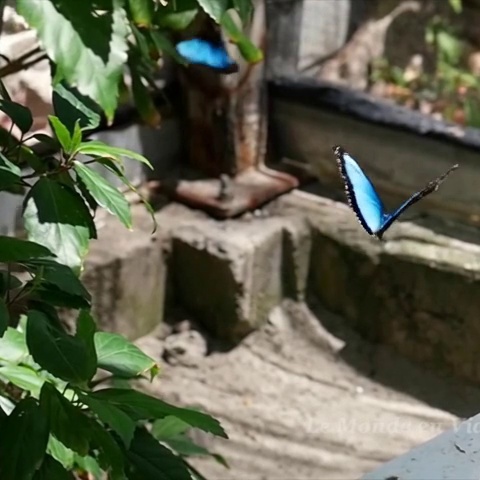} &
\includegraphics[width=0.120\linewidth,height=0.120\linewidth]{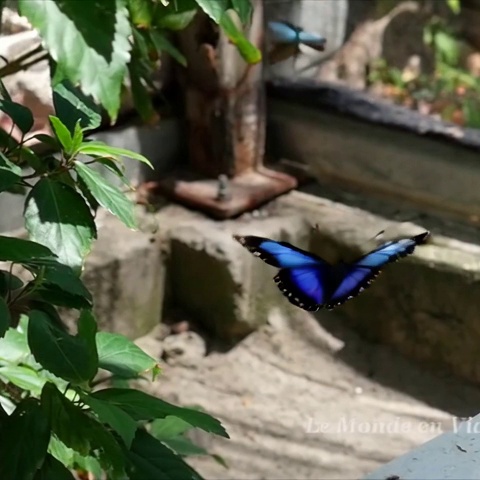} &
\includegraphics[width=0.120\linewidth,height=0.120\linewidth]{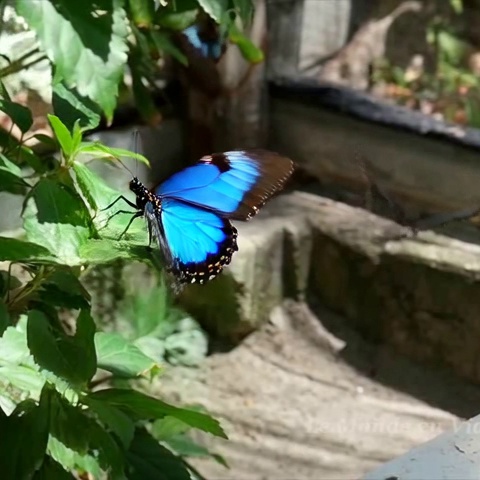} &
\includegraphics[width=0.120\linewidth,height=0.120\linewidth]{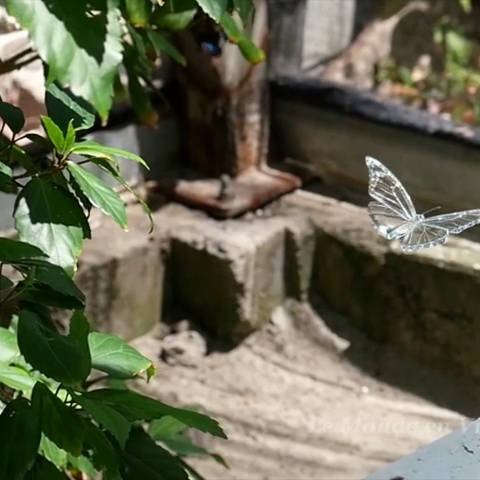} \\[-1.0pt]
\qualtime{$t_8$} &
\includegraphics[width=0.120\linewidth,height=0.120\linewidth]{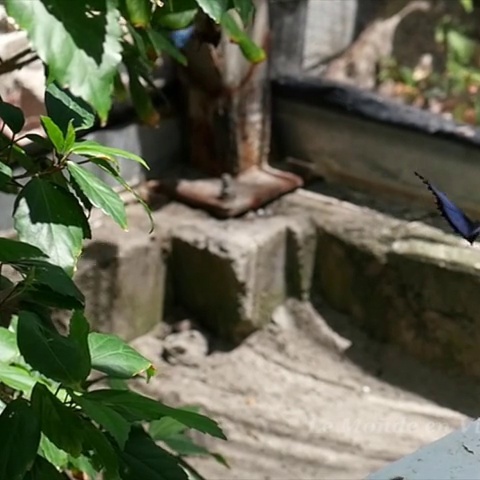} &
\includegraphics[width=0.120\linewidth,height=0.120\linewidth]{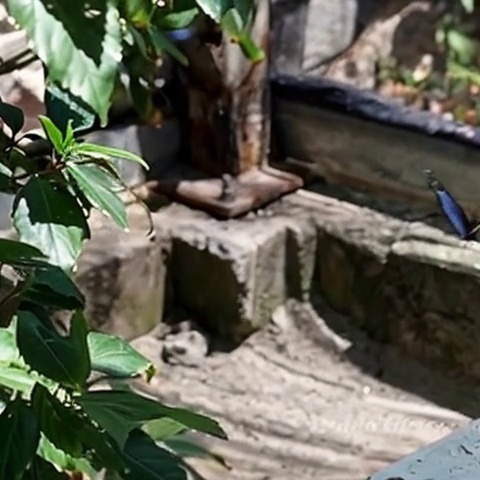} &
\includegraphics[width=0.120\linewidth,height=0.120\linewidth]{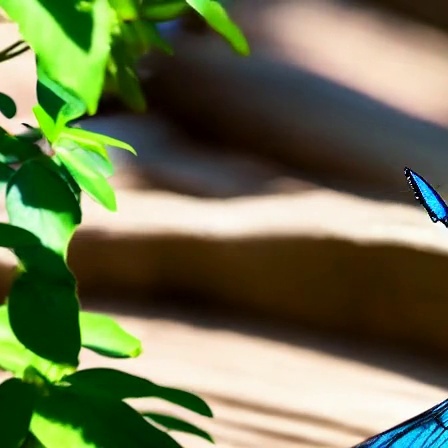} &
\includegraphics[width=0.120\linewidth,height=0.120\linewidth]{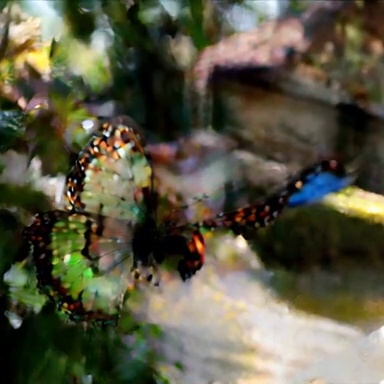} &
\includegraphics[width=0.120\linewidth,height=0.120\linewidth]{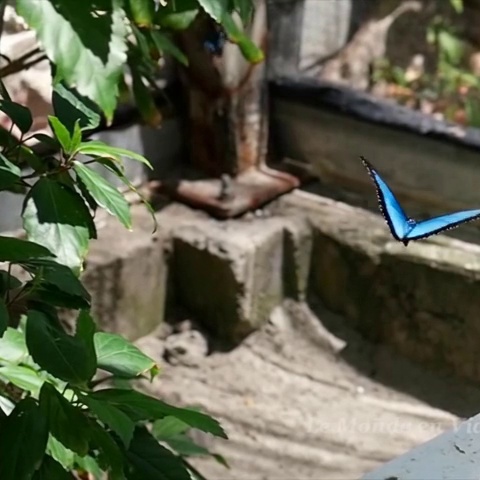} &
\includegraphics[width=0.120\linewidth,height=0.120\linewidth]{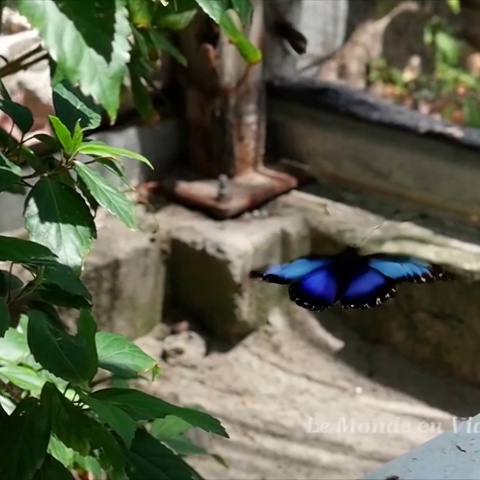} &
\includegraphics[width=0.120\linewidth,height=0.120\linewidth]{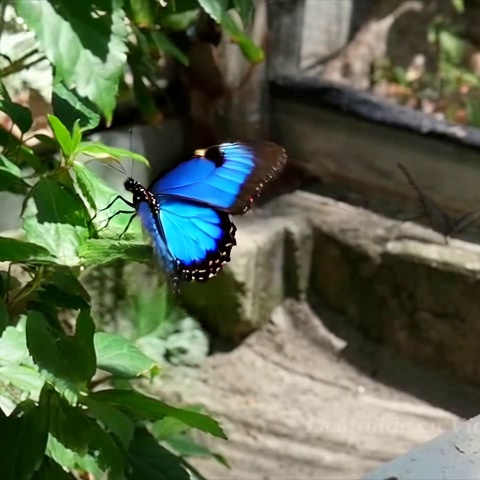} &
\includegraphics[width=0.120\linewidth,height=0.120\linewidth]{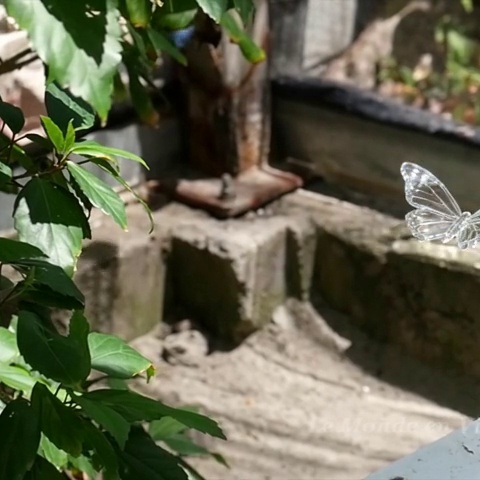} \\[-1.0pt]
\qualtime{$t_9$} &
\includegraphics[width=0.120\linewidth,height=0.120\linewidth]{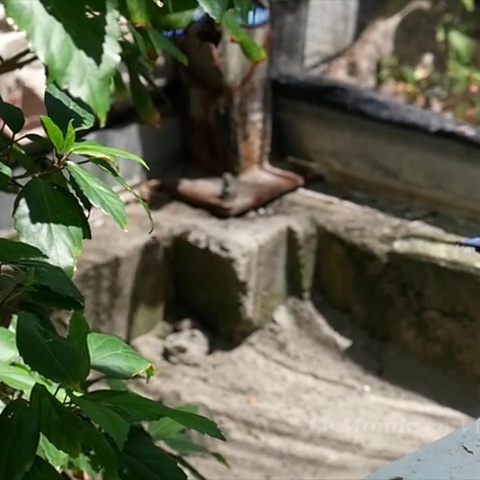} &
\includegraphics[width=0.120\linewidth,height=0.120\linewidth]{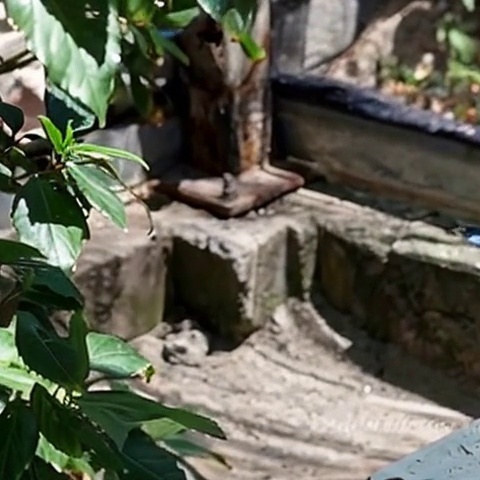} &
\includegraphics[width=0.120\linewidth,height=0.120\linewidth]{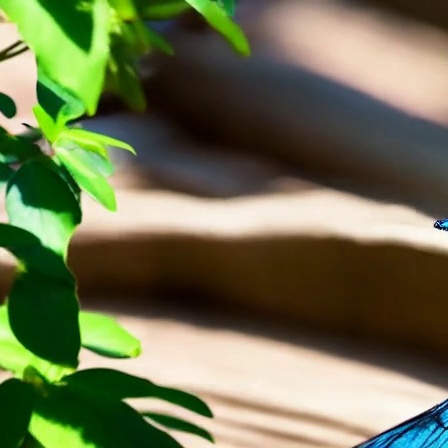} &
\includegraphics[width=0.120\linewidth,height=0.120\linewidth]{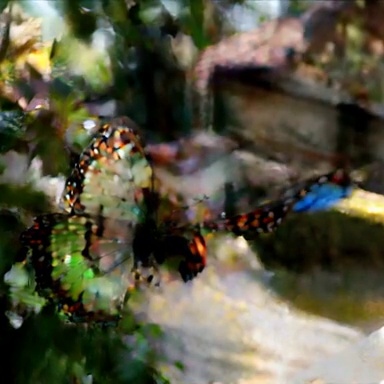} &
\includegraphics[width=0.120\linewidth,height=0.120\linewidth]{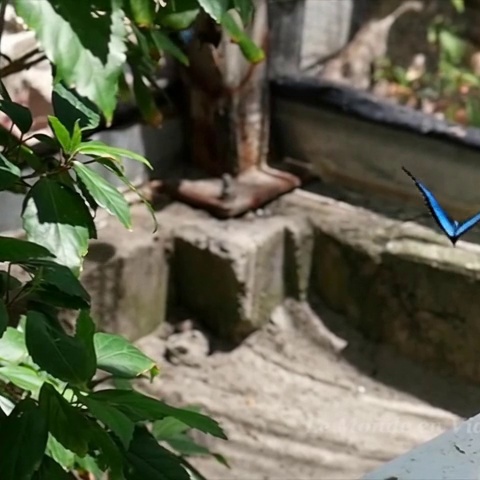} &
\includegraphics[width=0.120\linewidth,height=0.120\linewidth]{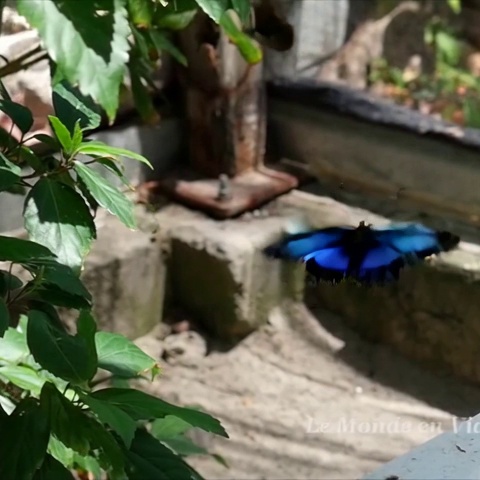} &
\includegraphics[width=0.120\linewidth,height=0.120\linewidth]{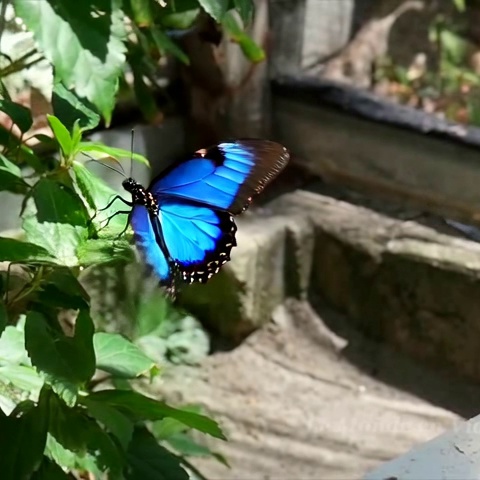} &
\includegraphics[width=0.120\linewidth,height=0.120\linewidth]{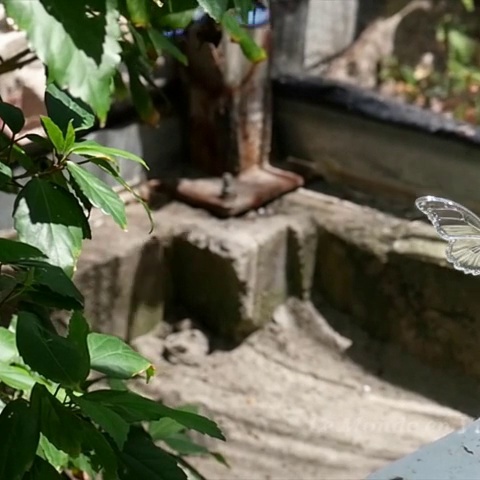} \\[-1.0pt]
\end{tabular}%
}
\caption{\textbf{Object-level material editing.}
Given the instruction to change the butterfly material from a living creature to glass, \modelnamebf{} produces a clear glass appearance while preserving the butterfly's shape, motion, and surrounding scene. Most competing methods largely retain the original appearance or introduce unrelated changes rather than realizing the requested material transformation. Frames are sampled at uniformly spaced temporal positions throughout each video.}
\label{fig:qual_butterfly_glass}
\end{figure*}

\begin{figure*}[t!]
\centering
\setlength{\tabcolsep}{0.45pt}
\renewcommand{\arraystretch}{0.62}
\scriptsize
\def\qualtime#1{\raisebox{0.050\linewidth}{\makebox[0.018\linewidth][r]{#1}}}
\resizebox{\textwidth}{!}{%
\begin{tabular}{@{}r@{\hspace{1pt}}cccccccc@{}}
\multicolumn{9}{c}{\small\textbf{Prompt:} Change the breakdancer to a holographic breakdancer.} \\[3pt]
& \textbf{Source} & \textbf{AnyV2V~\cite{ku2024anyv2v}} & \textbf{VidToMe~\cite{li2024vidtome}} & \textbf{Pyramid-Edit~\cite{li2025five}} & \textbf{StreamEdit~\cite{jiao2026streamedit}} & \textbf{Wan-Edit~\cite{li2025five}} & \textbf{FlowDirector~\cite{li2026flowdirector}} & \makecell{\textbf{\modelname{}}\\\textbf{(Ours)}} \\[2pt]
\qualtime{$t_1$} &
\includegraphics[width=0.120\linewidth,height=0.120\linewidth]{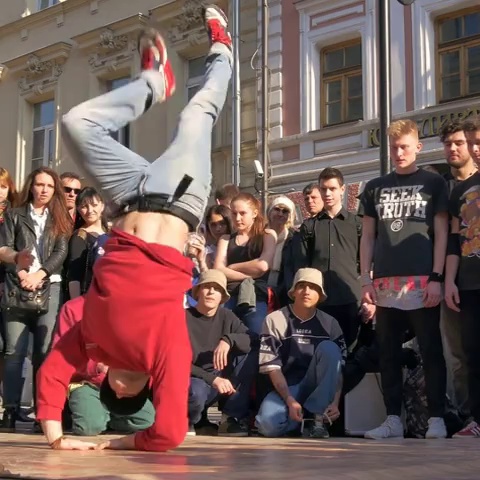} &
\includegraphics[width=0.120\linewidth,height=0.120\linewidth]{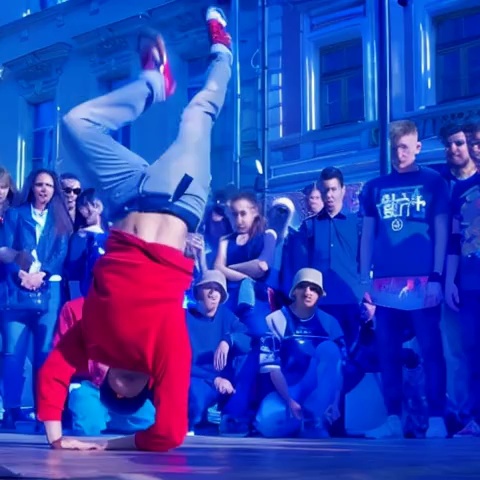} &
\includegraphics[width=0.120\linewidth,height=0.120\linewidth]{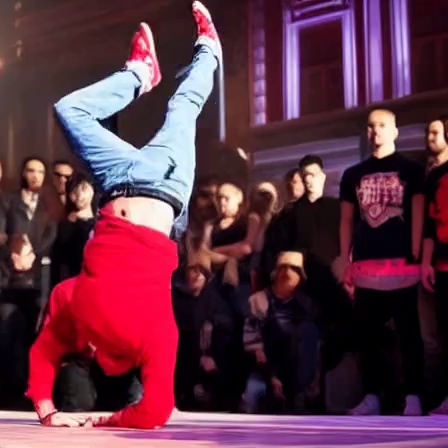} &
\includegraphics[width=0.120\linewidth,height=0.120\linewidth]{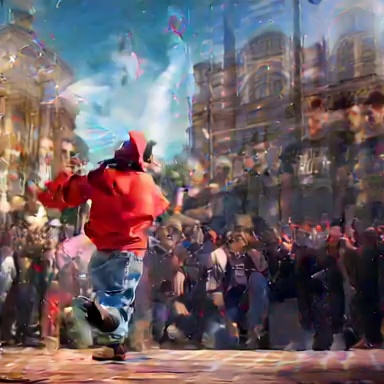} &
\includegraphics[width=0.120\linewidth,height=0.120\linewidth]{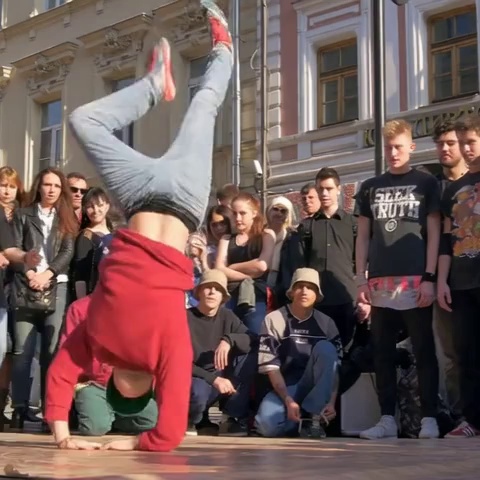} &
\includegraphics[width=0.120\linewidth,height=0.120\linewidth]{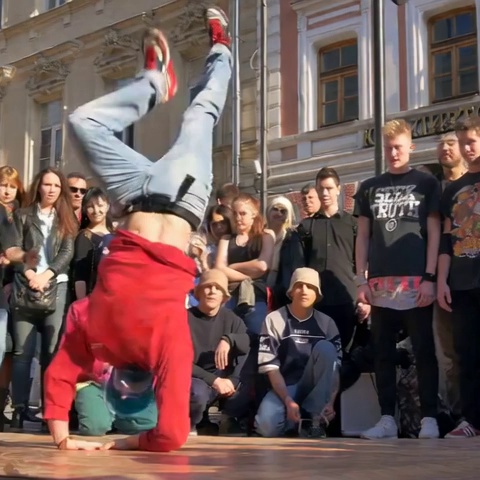} &
\includegraphics[width=0.120\linewidth,height=0.120\linewidth]{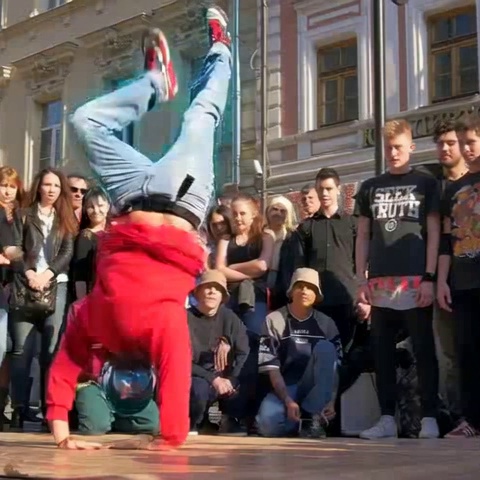} &
\includegraphics[width=0.120\linewidth,height=0.120\linewidth]{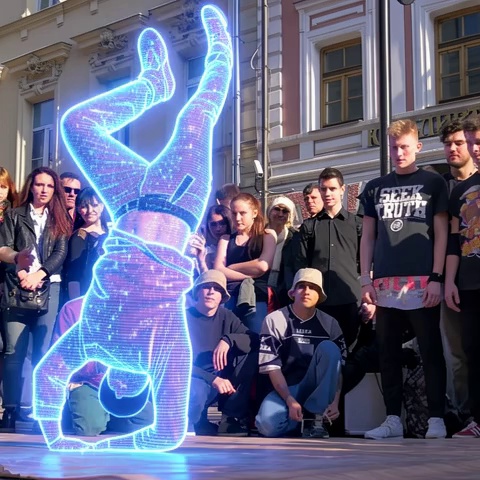} \\[-1.0pt]
\qualtime{$t_2$} &
\includegraphics[width=0.120\linewidth,height=0.120\linewidth]{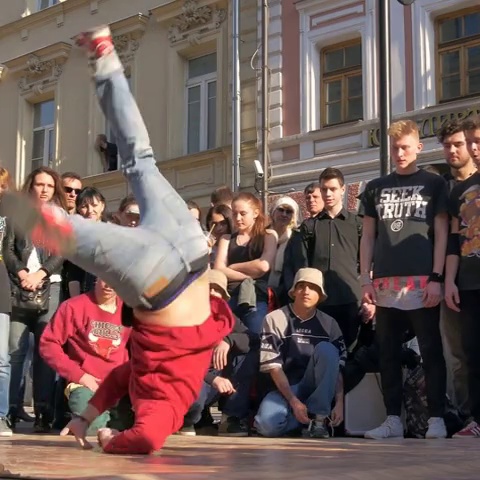} &
\includegraphics[width=0.120\linewidth,height=0.120\linewidth]{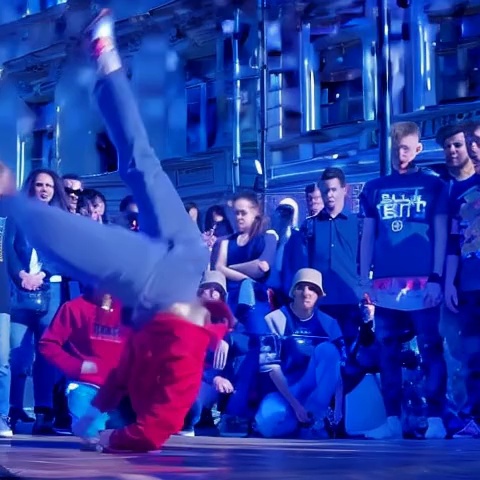} &
\includegraphics[width=0.120\linewidth,height=0.120\linewidth]{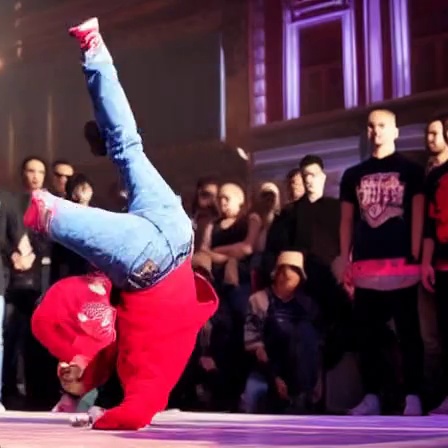} &
\includegraphics[width=0.120\linewidth,height=0.120\linewidth]{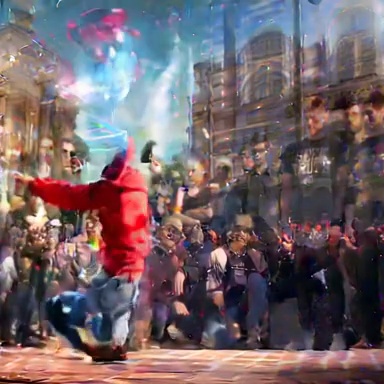} &
\includegraphics[width=0.120\linewidth,height=0.120\linewidth]{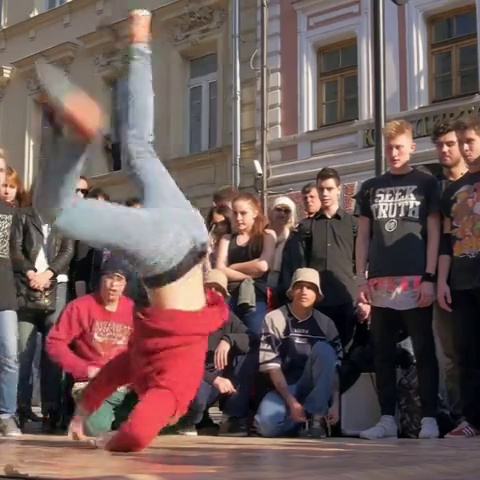} &
\includegraphics[width=0.120\linewidth,height=0.120\linewidth]{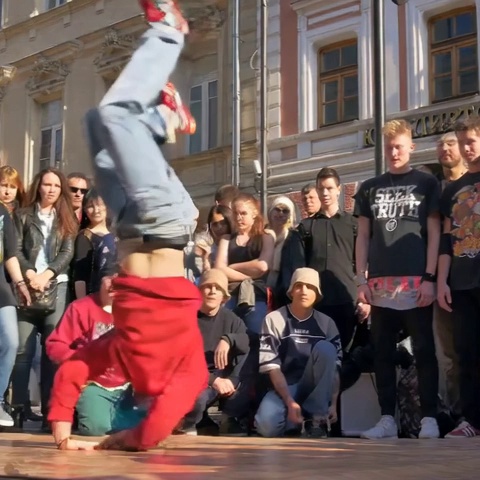} &
\includegraphics[width=0.120\linewidth,height=0.120\linewidth]{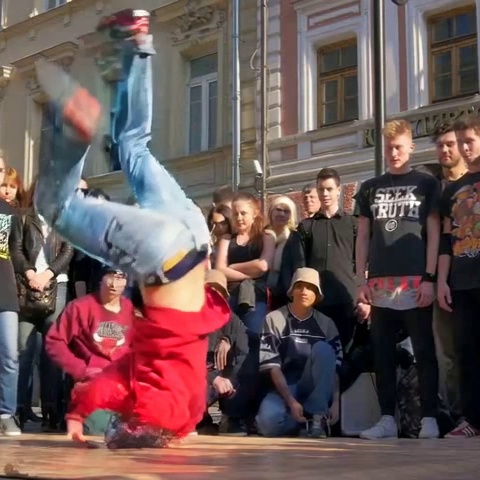} &
\includegraphics[width=0.120\linewidth,height=0.120\linewidth]{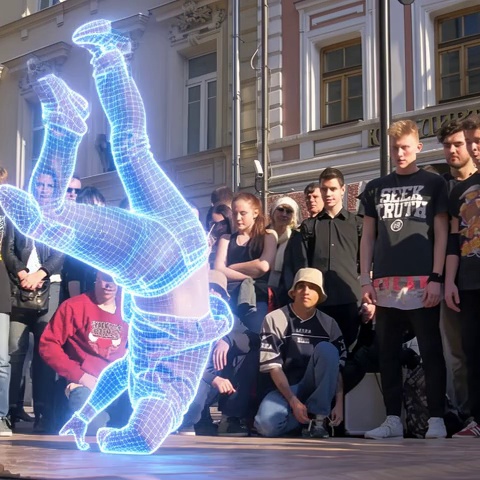} \\[-1.0pt]
\qualtime{$t_3$} &
\includegraphics[width=0.120\linewidth,height=0.120\linewidth]{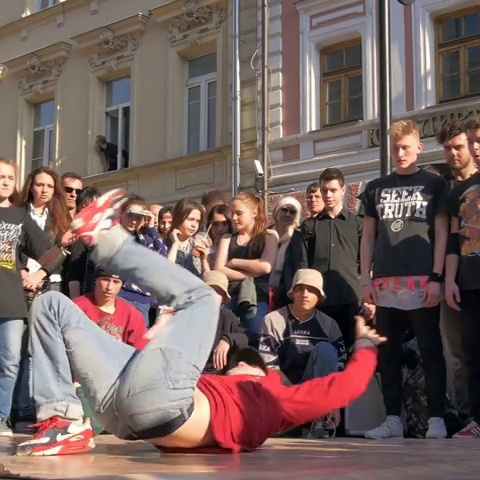} &
\includegraphics[width=0.120\linewidth,height=0.120\linewidth]{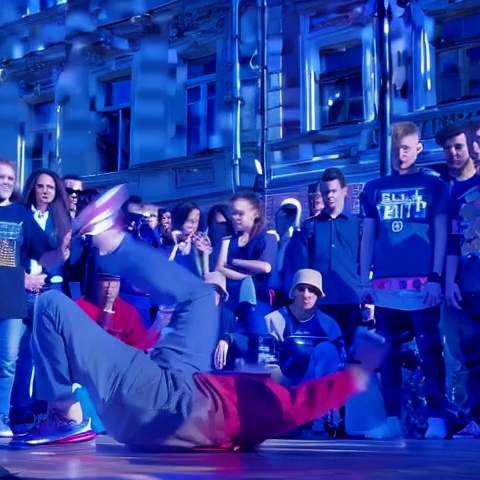} &
\includegraphics[width=0.120\linewidth,height=0.120\linewidth]{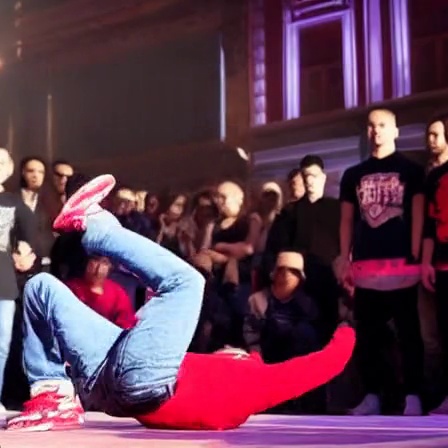} &
\includegraphics[width=0.120\linewidth,height=0.120\linewidth]{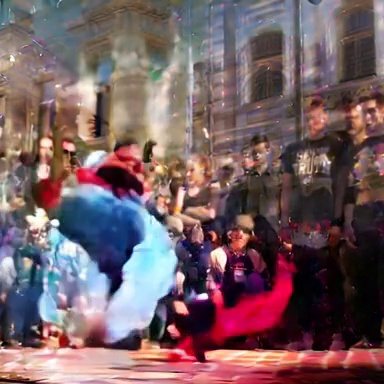} &
\includegraphics[width=0.120\linewidth,height=0.120\linewidth]{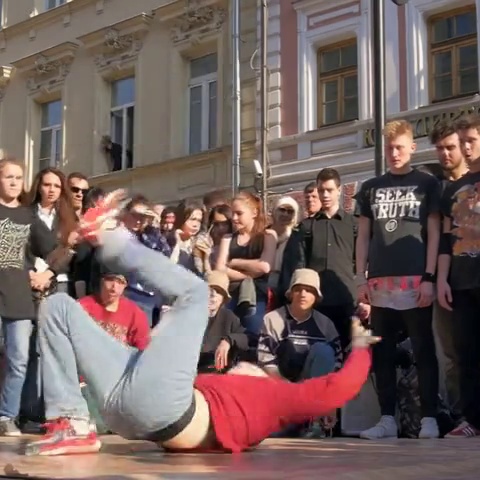} &
\includegraphics[width=0.120\linewidth,height=0.120\linewidth]{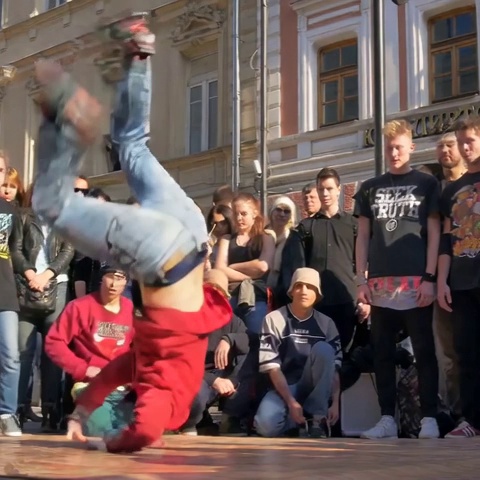} &
\includegraphics[width=0.120\linewidth,height=0.120\linewidth]{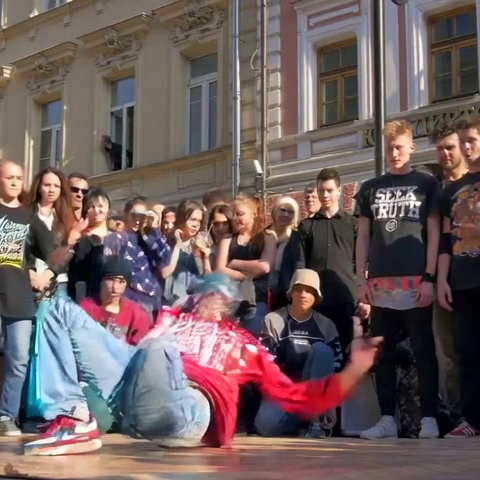} &
\includegraphics[width=0.120\linewidth,height=0.120\linewidth]{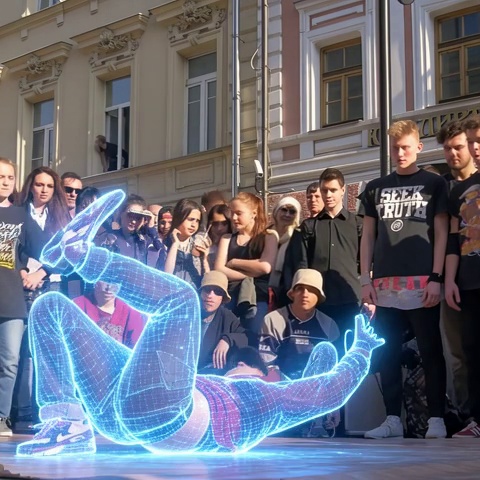} \\[-1.0pt]
\qualtime{$t_4$} &
\includegraphics[width=0.120\linewidth,height=0.120\linewidth]{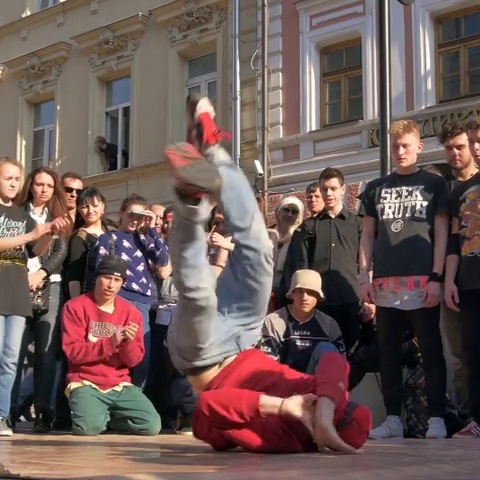} &
\includegraphics[width=0.120\linewidth,height=0.120\linewidth]{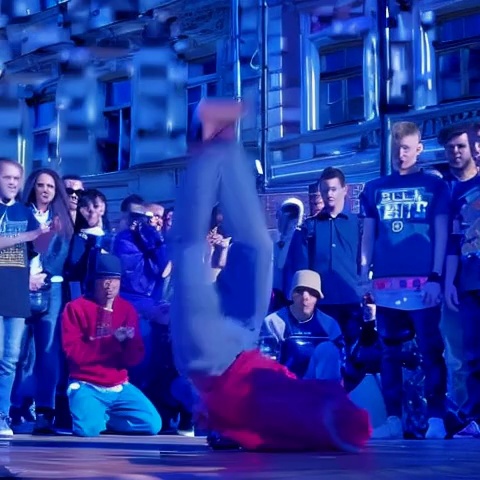} &
\includegraphics[width=0.120\linewidth,height=0.120\linewidth]{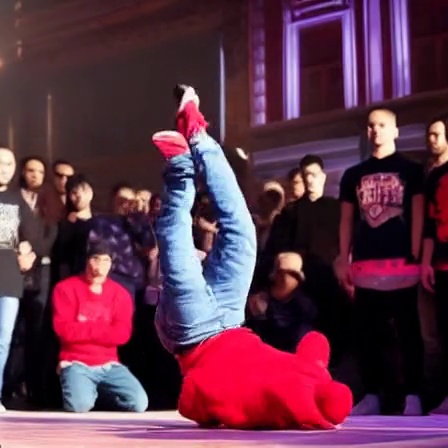} &
\includegraphics[width=0.120\linewidth,height=0.120\linewidth]{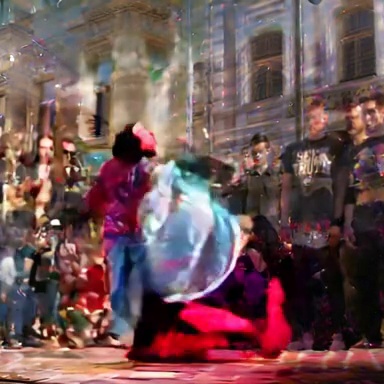} &
\includegraphics[width=0.120\linewidth,height=0.120\linewidth]{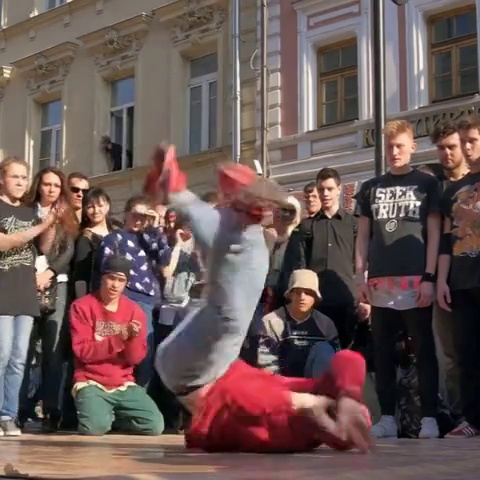} &
\includegraphics[width=0.120\linewidth,height=0.120\linewidth]{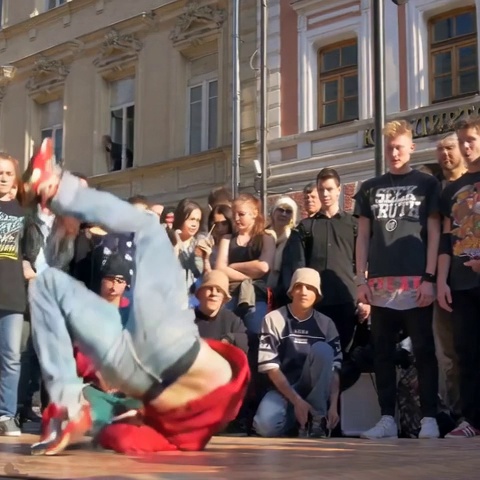} &
\includegraphics[width=0.120\linewidth,height=0.120\linewidth]{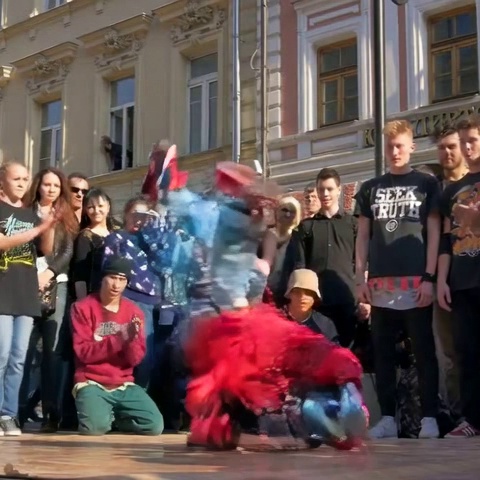} &
\includegraphics[width=0.120\linewidth,height=0.120\linewidth]{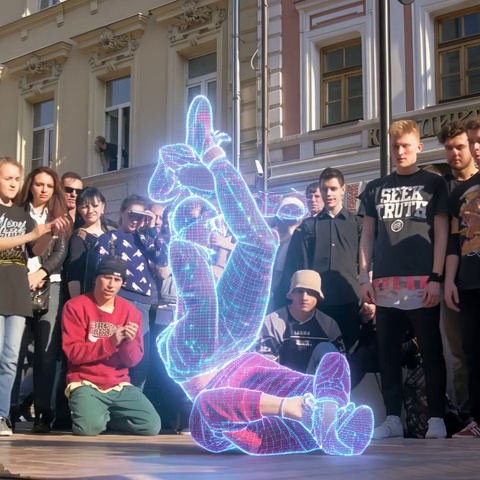} \\[-1.0pt]
\qualtime{$t_5$} &
\includegraphics[width=0.120\linewidth,height=0.120\linewidth]{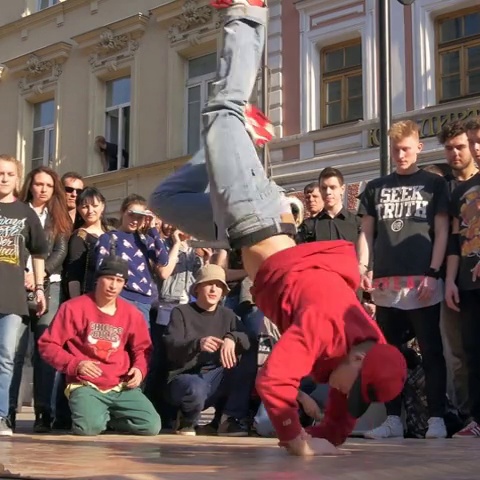} &
\includegraphics[width=0.120\linewidth,height=0.120\linewidth]{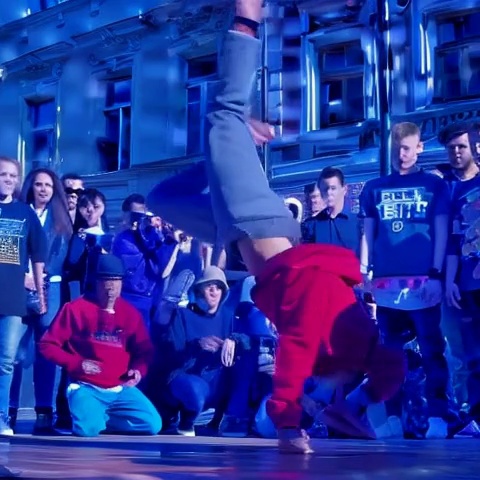} &
\includegraphics[width=0.120\linewidth,height=0.120\linewidth]{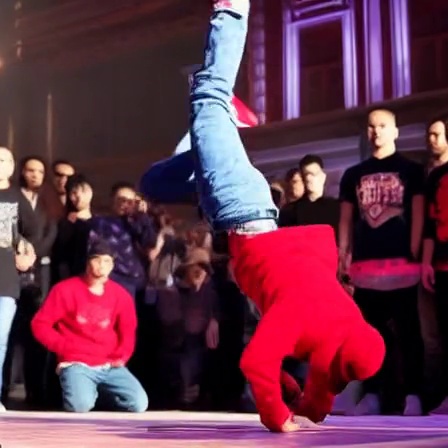} &
\includegraphics[width=0.120\linewidth,height=0.120\linewidth]{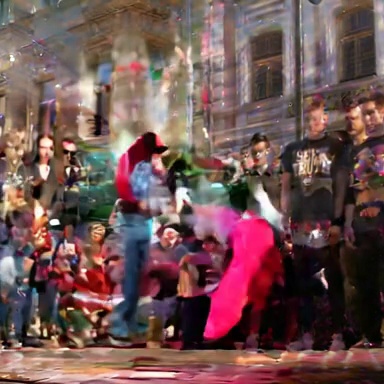} &
\includegraphics[width=0.120\linewidth,height=0.120\linewidth]{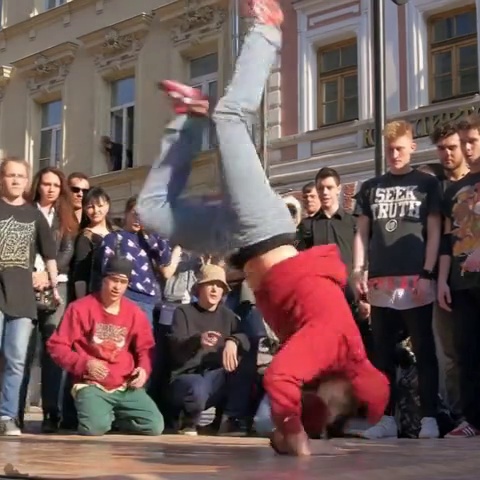} &
\includegraphics[width=0.120\linewidth,height=0.120\linewidth]{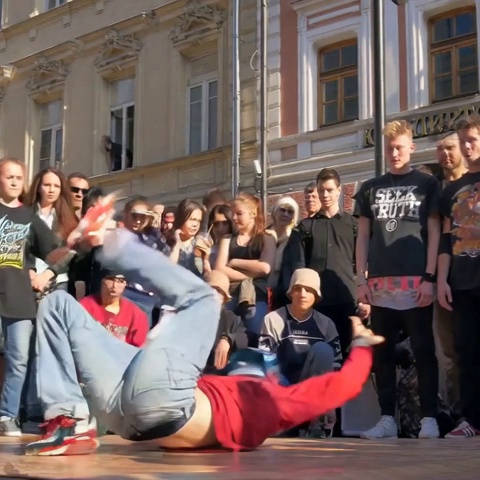} &
\includegraphics[width=0.120\linewidth,height=0.120\linewidth]{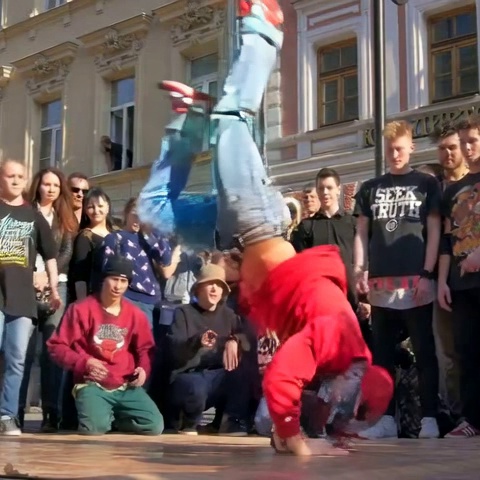} &
\includegraphics[width=0.120\linewidth,height=0.120\linewidth]{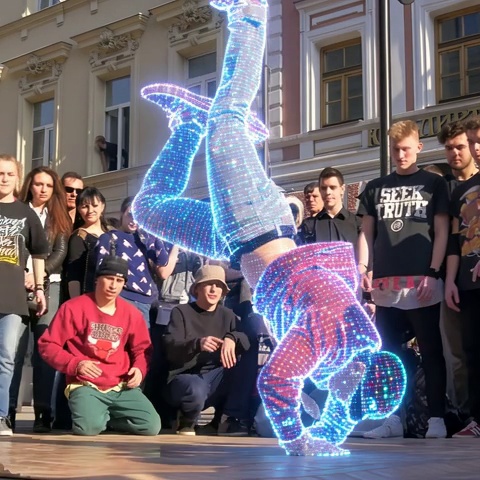} \\[-1.0pt]
\qualtime{$t_6$} &
\includegraphics[width=0.120\linewidth,height=0.120\linewidth]{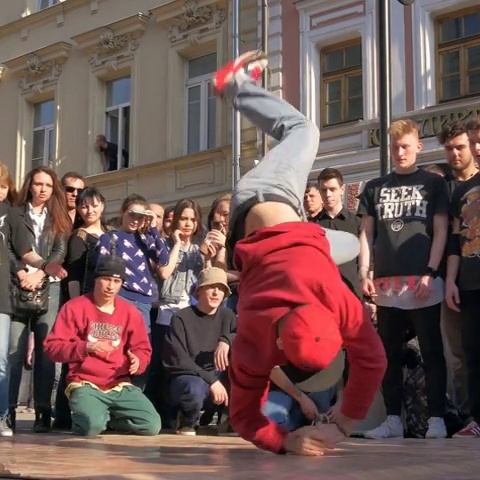} &
\includegraphics[width=0.120\linewidth,height=0.120\linewidth]{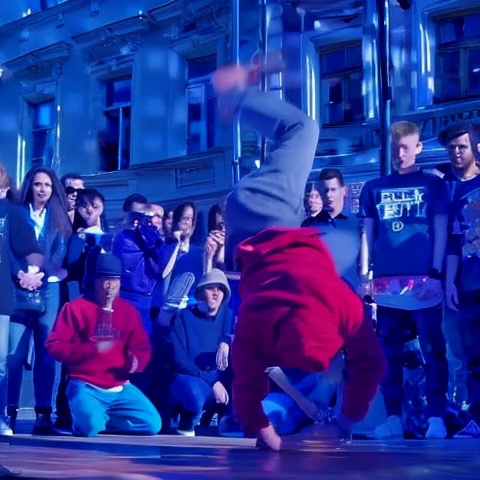} &
\includegraphics[width=0.120\linewidth,height=0.120\linewidth]{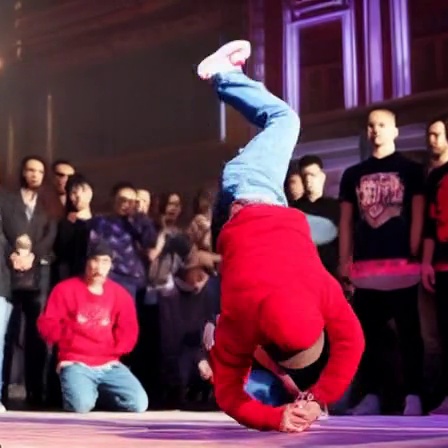} &
\includegraphics[width=0.120\linewidth,height=0.120\linewidth]{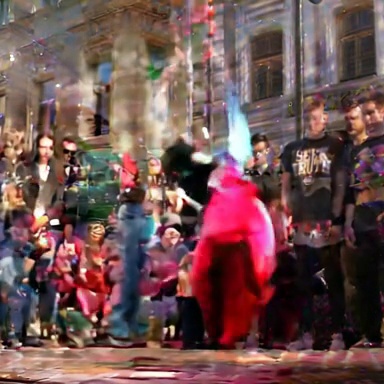} &
\includegraphics[width=0.120\linewidth,height=0.120\linewidth]{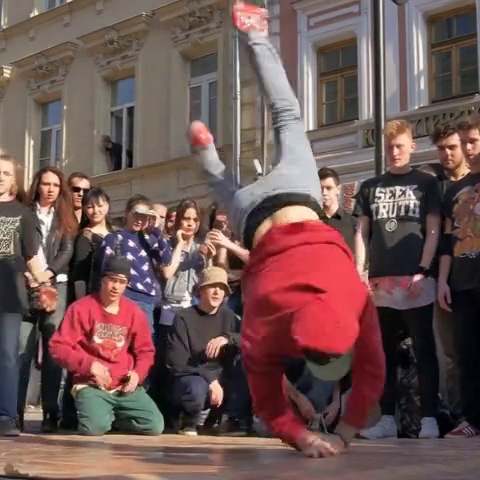} &
\includegraphics[width=0.120\linewidth,height=0.120\linewidth]{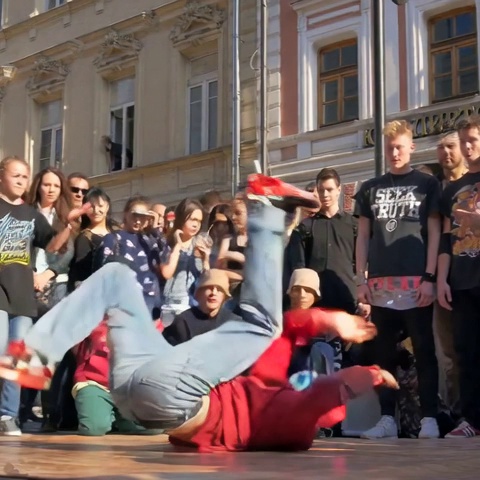} &
\includegraphics[width=0.120\linewidth,height=0.120\linewidth]{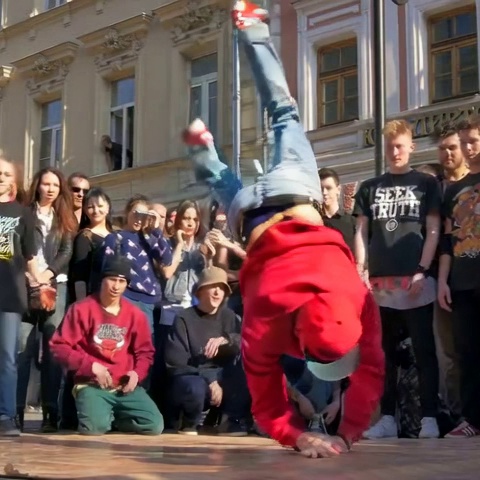} &
\includegraphics[width=0.120\linewidth,height=0.120\linewidth]{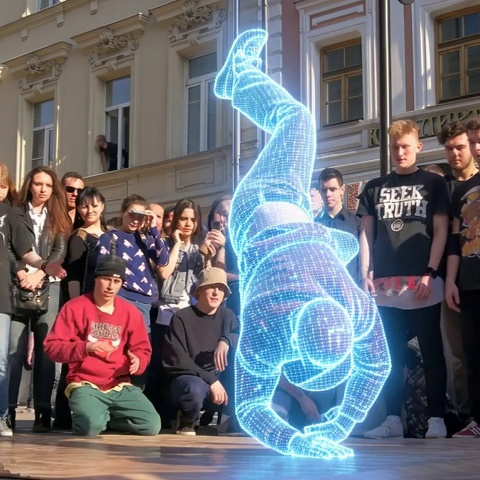} \\[-1.0pt]
\qualtime{$t_7$} &
\includegraphics[width=0.120\linewidth,height=0.120\linewidth]{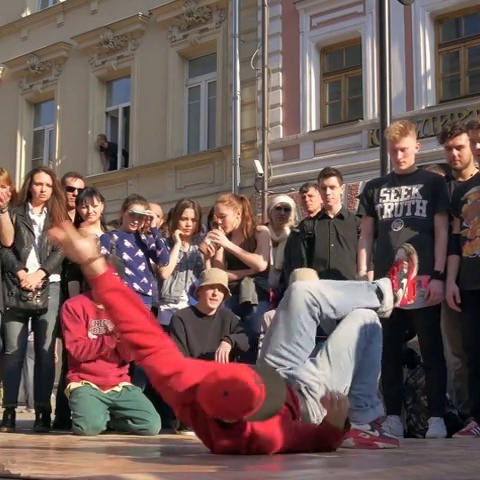} &
\includegraphics[width=0.120\linewidth,height=0.120\linewidth]{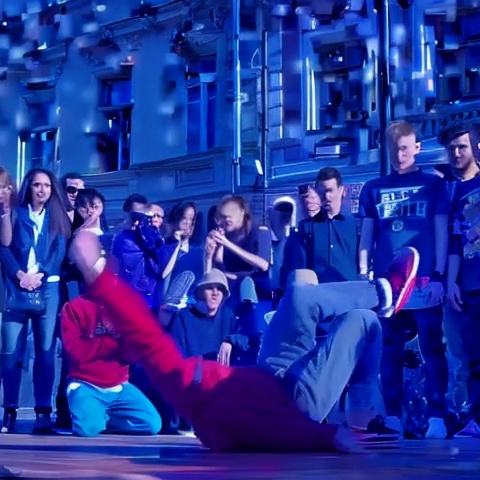} &
\includegraphics[width=0.120\linewidth,height=0.120\linewidth]{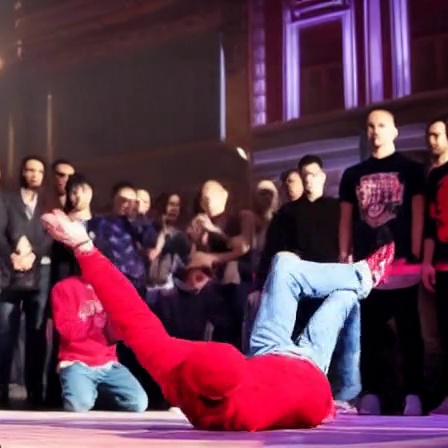} &
\includegraphics[width=0.120\linewidth,height=0.120\linewidth]{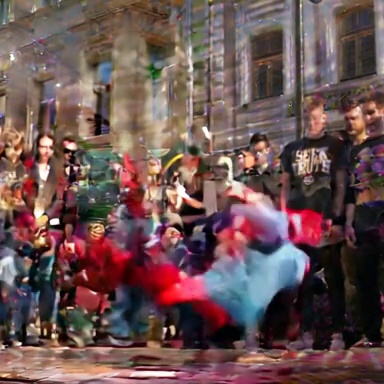} &
\includegraphics[width=0.120\linewidth,height=0.120\linewidth]{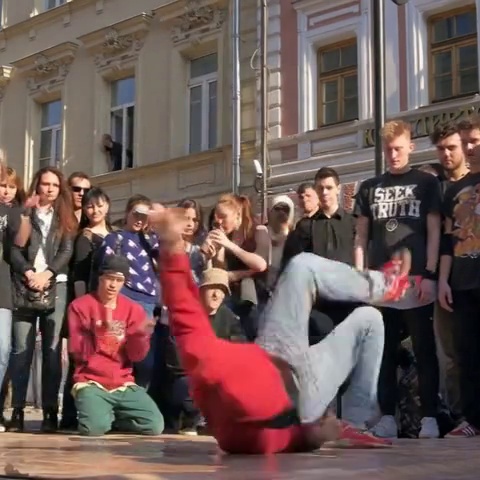} &
\includegraphics[width=0.120\linewidth,height=0.120\linewidth]{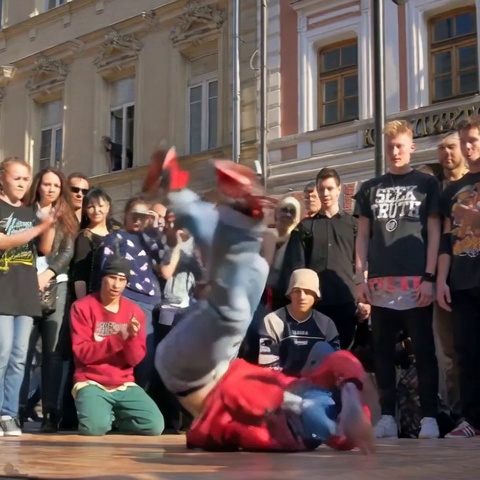} &
\includegraphics[width=0.120\linewidth,height=0.120\linewidth]{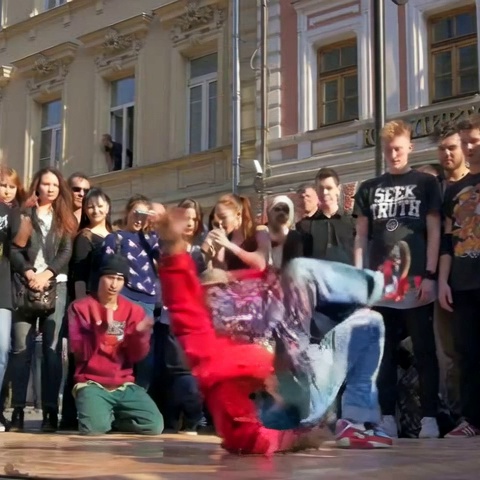} &
\includegraphics[width=0.120\linewidth,height=0.120\linewidth]{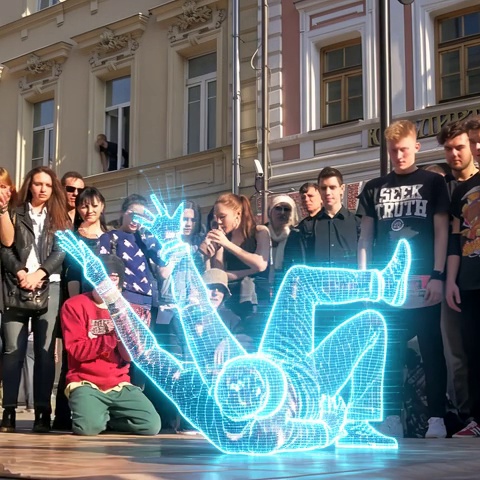} \\[-1.0pt]
\qualtime{$t_8$} &
\includegraphics[width=0.120\linewidth,height=0.120\linewidth]{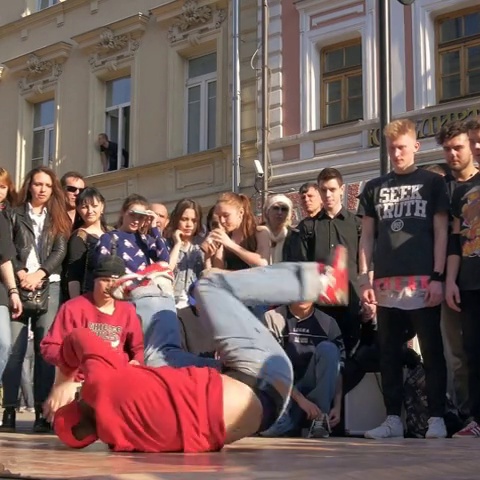} &
\includegraphics[width=0.120\linewidth,height=0.120\linewidth]{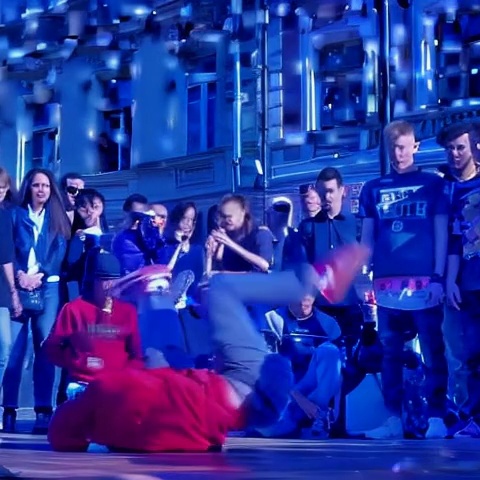} &
\includegraphics[width=0.120\linewidth,height=0.120\linewidth]{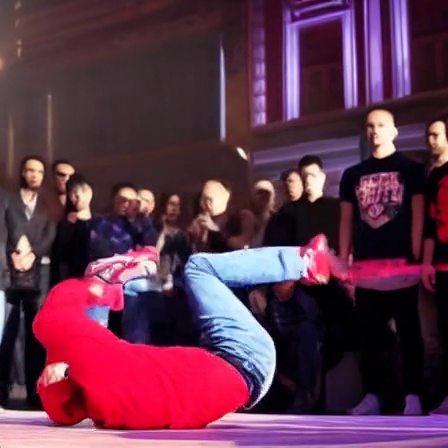} &
\includegraphics[width=0.120\linewidth,height=0.120\linewidth]{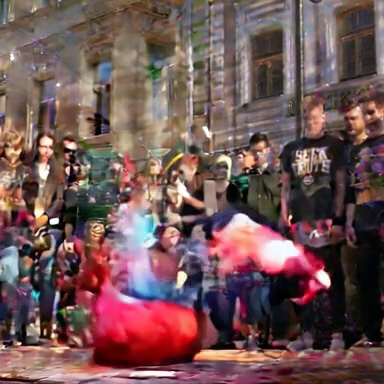} &
\includegraphics[width=0.120\linewidth,height=0.120\linewidth]{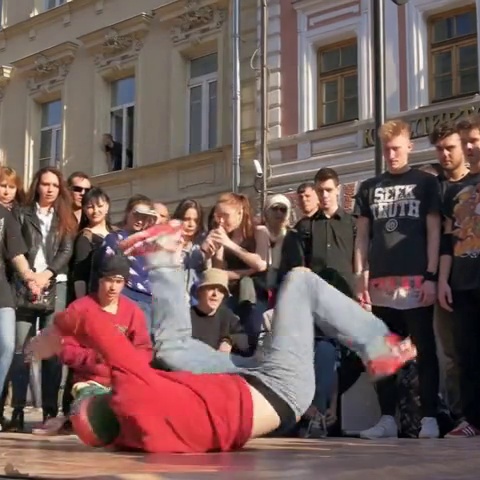} &
\includegraphics[width=0.120\linewidth,height=0.120\linewidth]{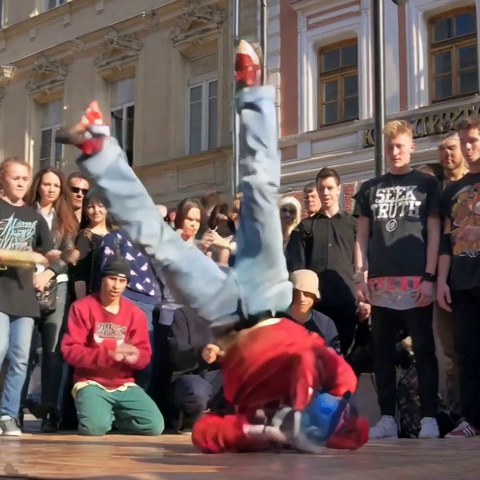} &
\includegraphics[width=0.120\linewidth,height=0.120\linewidth]{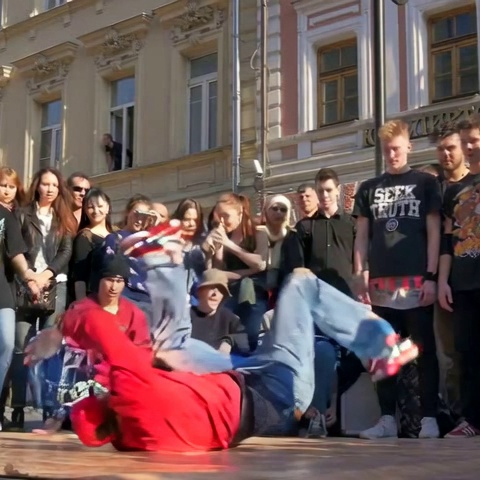} &
\includegraphics[width=0.120\linewidth,height=0.120\linewidth]{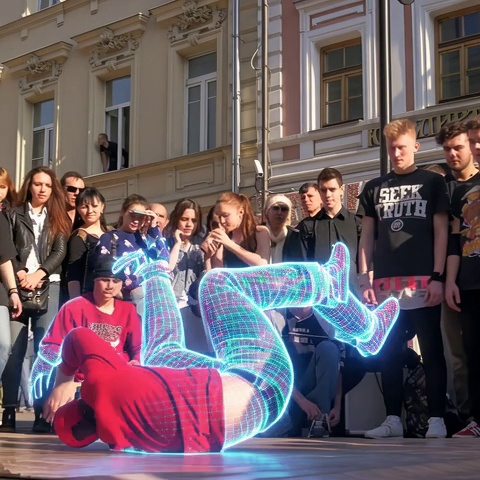} \\[-1.0pt]
\qualtime{$t_9$} &
\includegraphics[width=0.120\linewidth,height=0.120\linewidth]{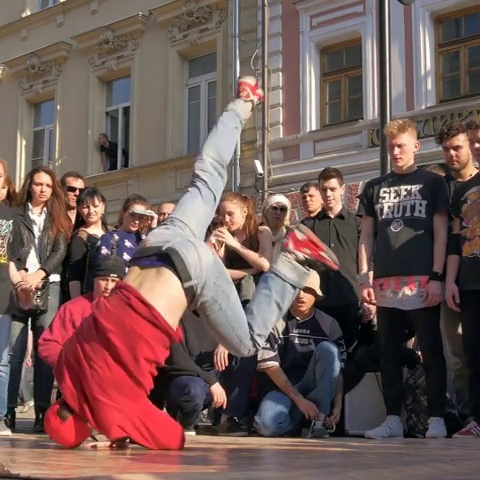} &
\includegraphics[width=0.120\linewidth,height=0.120\linewidth]{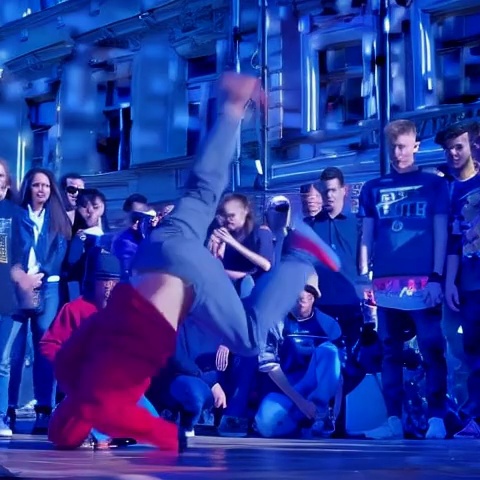} &
\includegraphics[width=0.120\linewidth,height=0.120\linewidth]{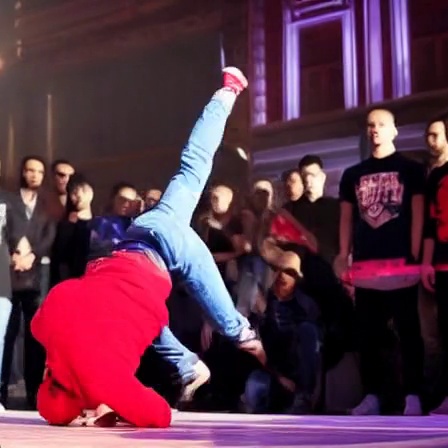} &
\includegraphics[width=0.120\linewidth,height=0.120\linewidth]{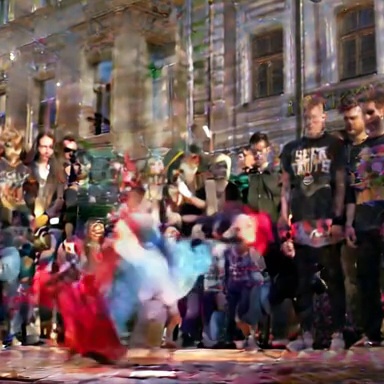} &
\includegraphics[width=0.120\linewidth,height=0.120\linewidth]{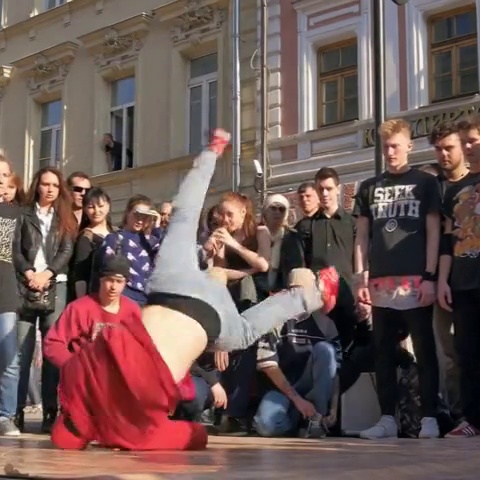} &
\includegraphics[width=0.120\linewidth,height=0.120\linewidth]{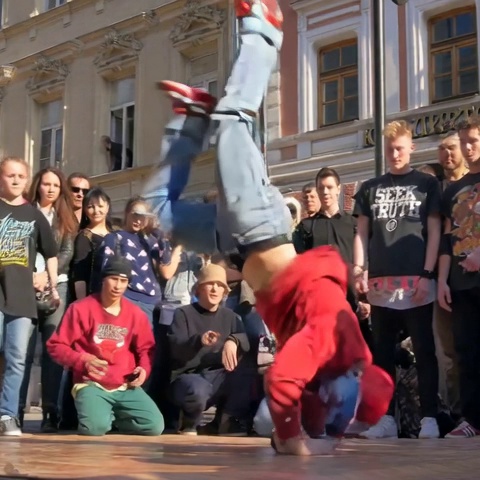} &
\includegraphics[width=0.120\linewidth,height=0.120\linewidth]{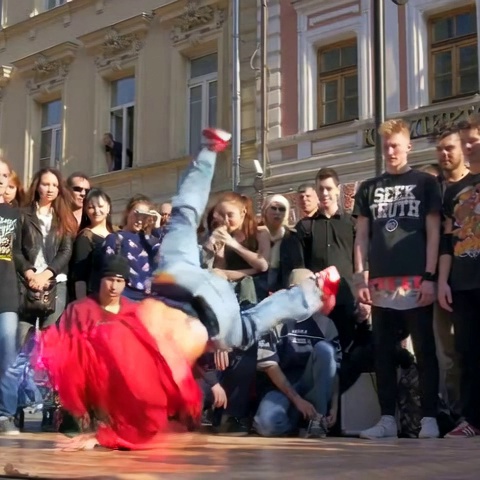} &
\includegraphics[width=0.120\linewidth,height=0.120\linewidth]{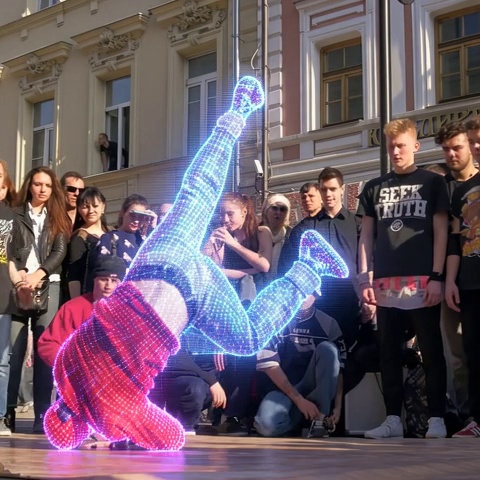} \\[-1.0pt]
\end{tabular}%
}
\caption{\textbf{Object-level appearance editing.}
Given the instruction to change the breakdancer into a holographic breakdancer, \modelnamebf{} applies the requested holographic appearance throughout the sequence while preserving the dancer's highly varying pose and motion. Several baselines either under-apply the edit or introduce broader changes to the subject and surrounding scene. Frames are sampled at uniformly spaced temporal positions throughout each video.}
\label{fig:qual_holographic_breakdancer}
\end{figure*}

\begin{figure*}[t!]
\centering
\setlength{\tabcolsep}{0.45pt}
\renewcommand{\arraystretch}{0.62}
\scriptsize
\def\qualtime#1{\raisebox{0.050\linewidth}{\makebox[0.018\linewidth][r]{#1}}}
\resizebox{\textwidth}{!}{%
\begin{tabular}{@{}r@{\hspace{1pt}}ccccccc@{}}
\multicolumn{8}{c}{\small\textbf{Prompt:} Make the wolf's eyes blue.} \\[3pt]
& \textbf{Source} & \textbf{VidToMe~\cite{li2024vidtome}} & \textbf{Pyramid-Edit~\cite{li2025five}} & \textbf{StreamEdit~\cite{jiao2026streamedit}} & \textbf{Wan-Edit~\cite{li2025five}} & \textbf{FlowDirector~\cite{li2026flowdirector}} & \makecell{\textbf{\modelname{}}\\\textbf{(Ours)}} \\[2pt]
\qualtime{$t_1$} &
\includegraphics[width=0.137\linewidth,height=0.137\linewidth]{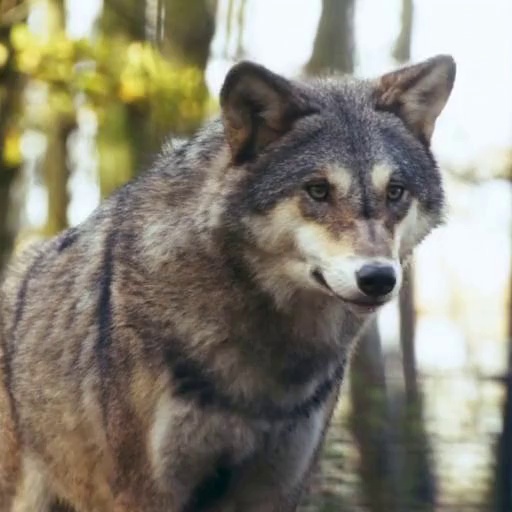} &
\includegraphics[width=0.137\linewidth,height=0.137\linewidth]{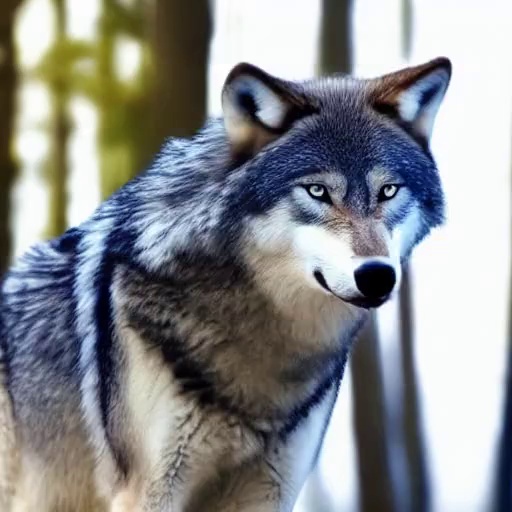} &
\includegraphics[width=0.137\linewidth,height=0.137\linewidth]{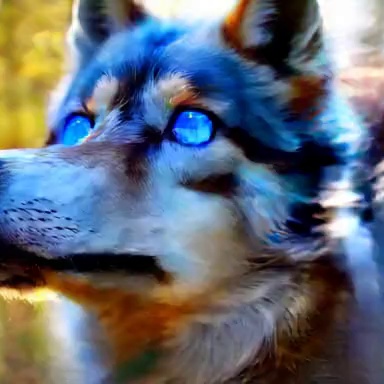} &
\includegraphics[width=0.137\linewidth,height=0.137\linewidth]{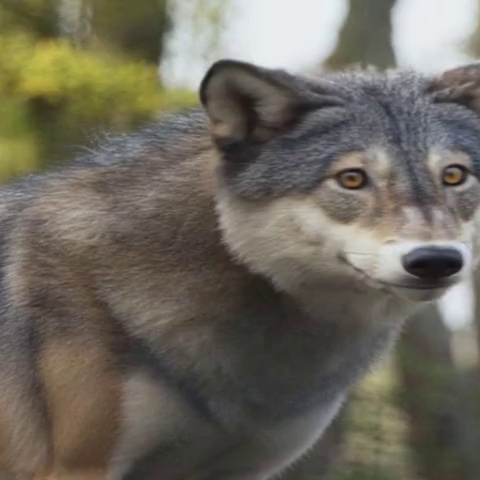} &
\includegraphics[width=0.137\linewidth,height=0.137\linewidth]{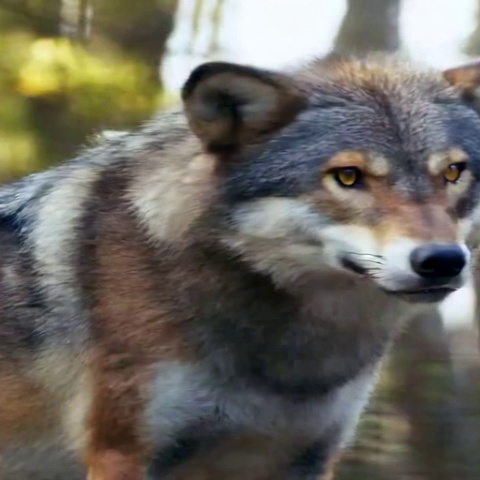} &
\includegraphics[width=0.137\linewidth,height=0.137\linewidth]{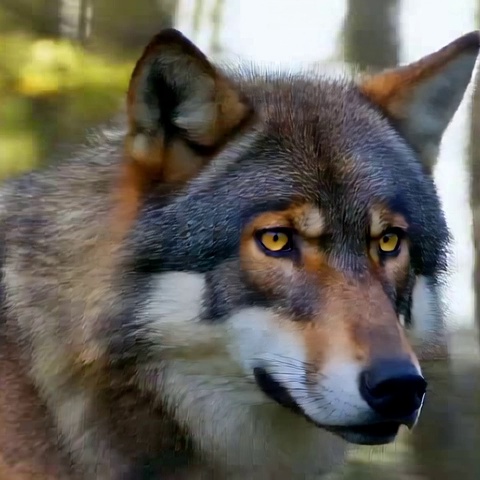} &
\includegraphics[width=0.137\linewidth,height=0.137\linewidth]{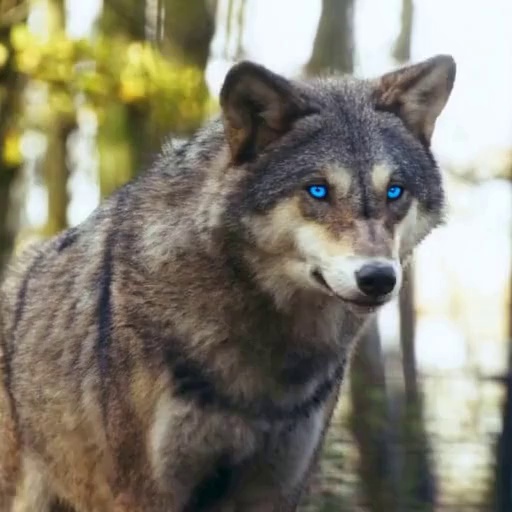} \\[-1.0pt]
\qualtime{$t_2$} &
\includegraphics[width=0.137\linewidth,height=0.137\linewidth]{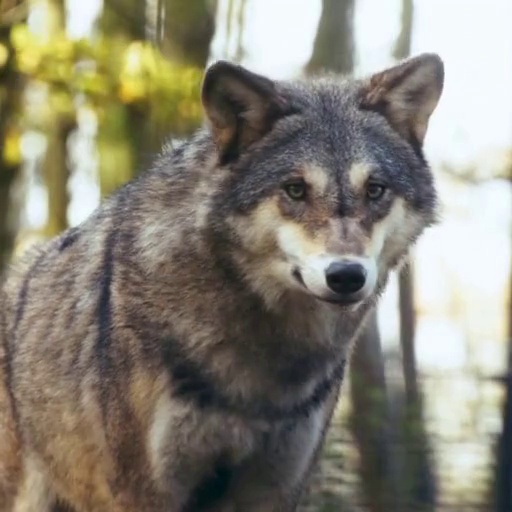} &
\includegraphics[width=0.137\linewidth,height=0.137\linewidth]{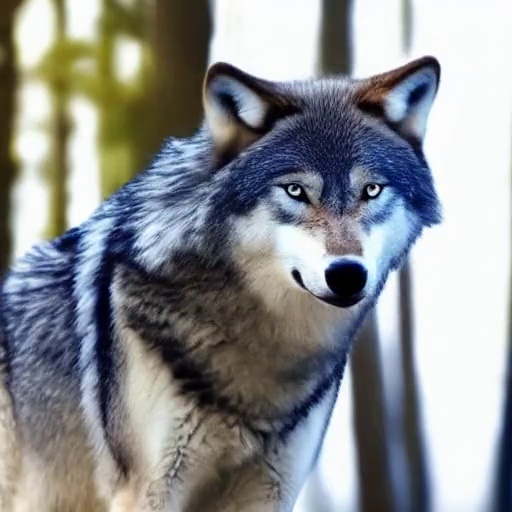} &
\includegraphics[width=0.137\linewidth,height=0.137\linewidth]{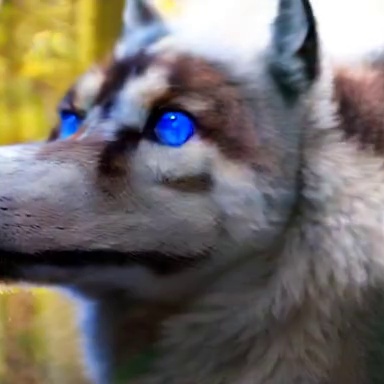} &
\includegraphics[width=0.137\linewidth,height=0.137\linewidth]{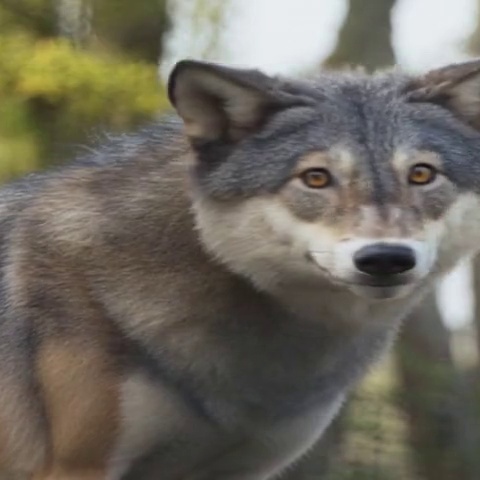} &
\includegraphics[width=0.137\linewidth,height=0.137\linewidth]{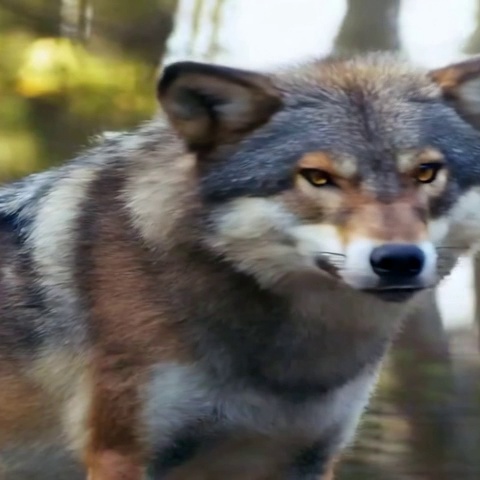} &
\includegraphics[width=0.137\linewidth,height=0.137\linewidth]{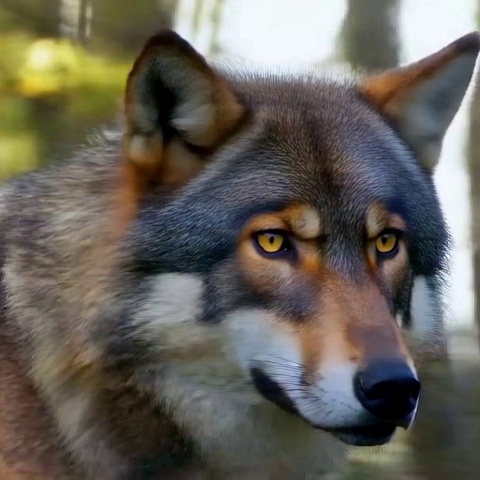} &
\includegraphics[width=0.137\linewidth,height=0.137\linewidth]{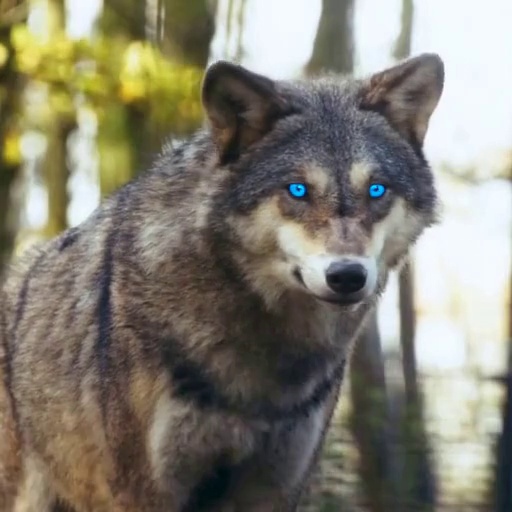} \\[-1.0pt]
\qualtime{$t_3$} &
\includegraphics[width=0.137\linewidth,height=0.137\linewidth]{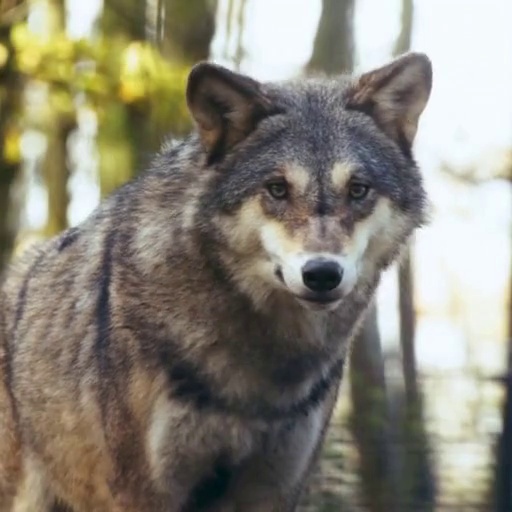} &
\includegraphics[width=0.137\linewidth,height=0.137\linewidth]{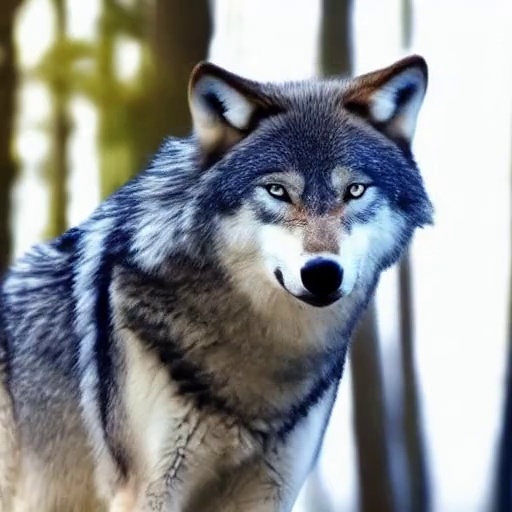} &
\includegraphics[width=0.137\linewidth,height=0.137\linewidth]{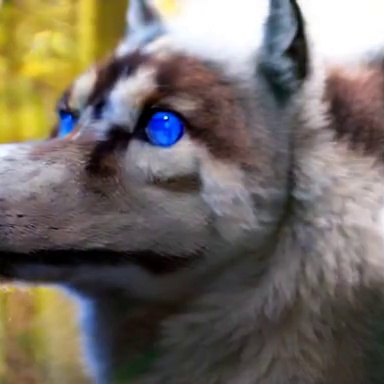} &
\includegraphics[width=0.137\linewidth,height=0.137\linewidth]{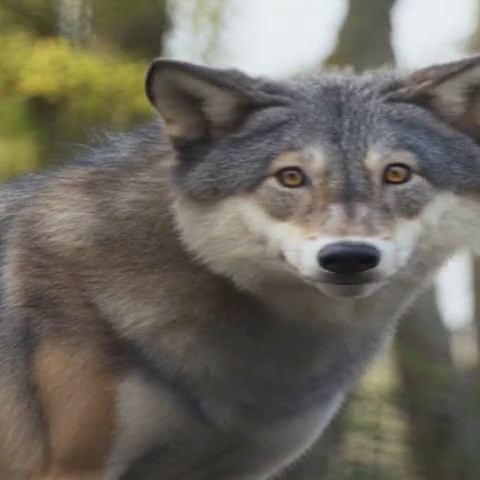} &
\includegraphics[width=0.137\linewidth,height=0.137\linewidth]{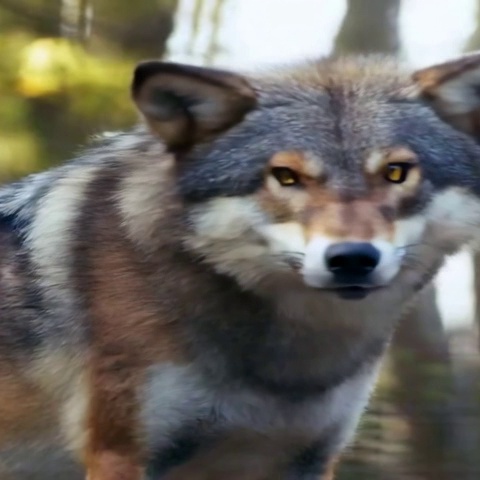} &
\includegraphics[width=0.137\linewidth,height=0.137\linewidth]{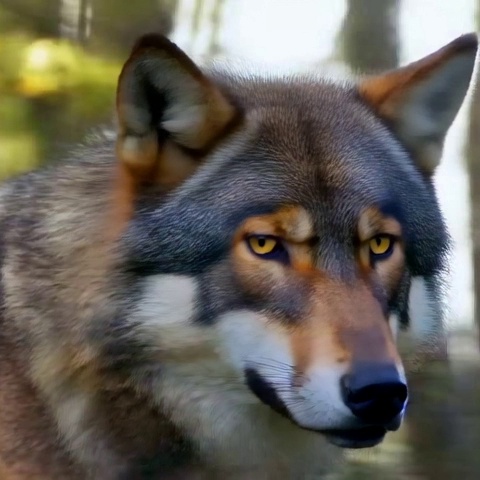} &
\includegraphics[width=0.137\linewidth,height=0.137\linewidth]{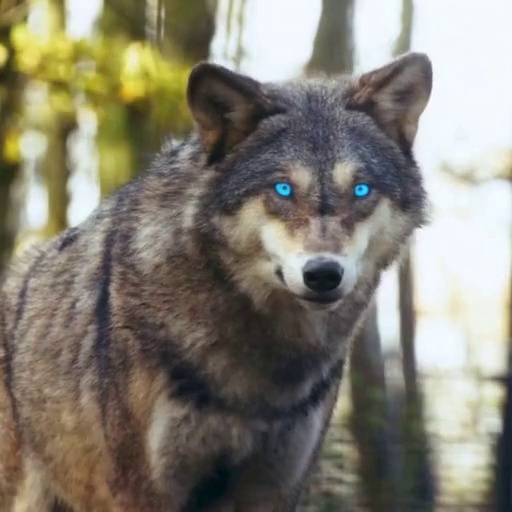} \\[-1.0pt]
\qualtime{$t_4$} &
\includegraphics[width=0.137\linewidth,height=0.137\linewidth]{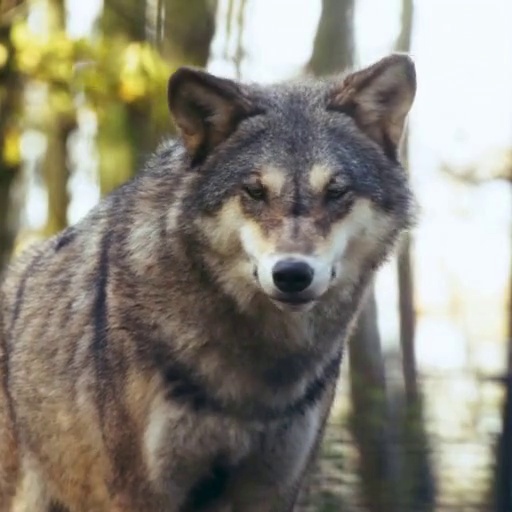} &
\includegraphics[width=0.137\linewidth,height=0.137\linewidth]{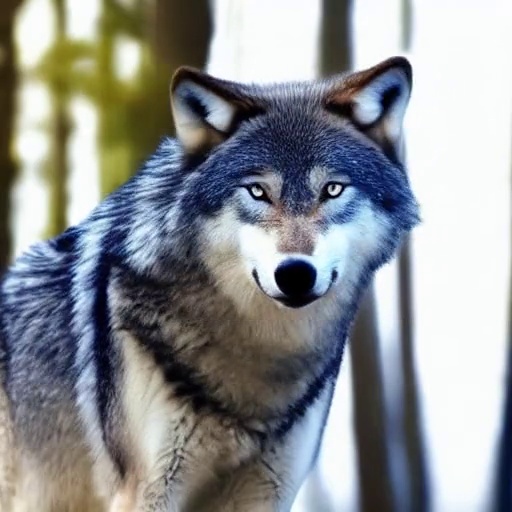} &
\includegraphics[width=0.137\linewidth,height=0.137\linewidth]{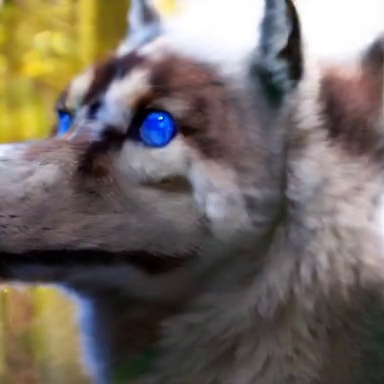} &
\includegraphics[width=0.137\linewidth,height=0.137\linewidth]{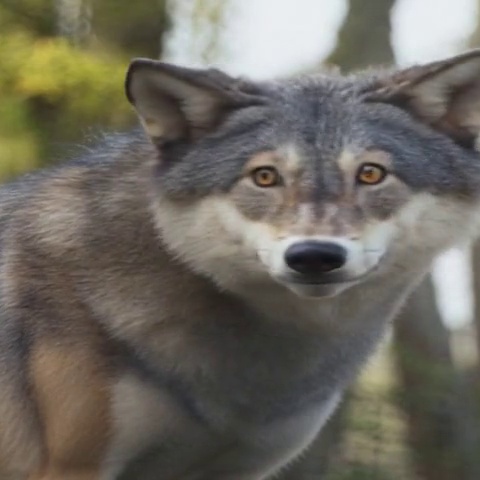} &
\includegraphics[width=0.137\linewidth,height=0.137\linewidth]{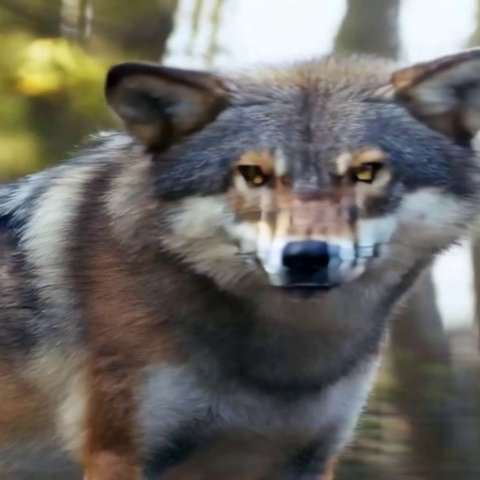} &
\includegraphics[width=0.137\linewidth,height=0.137\linewidth]{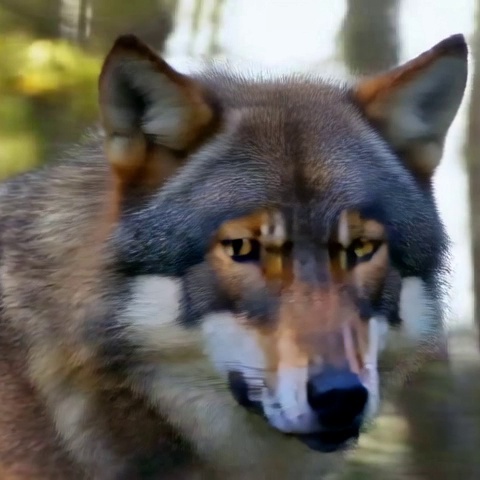} &
\includegraphics[width=0.137\linewidth,height=0.137\linewidth]{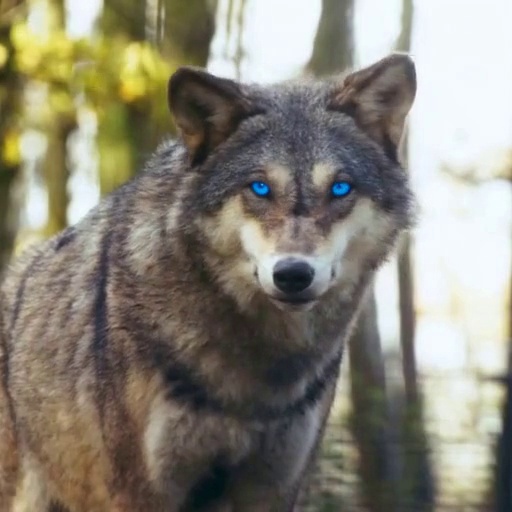} \\[-1.0pt]
\qualtime{$t_5$} &
\includegraphics[width=0.137\linewidth,height=0.137\linewidth]{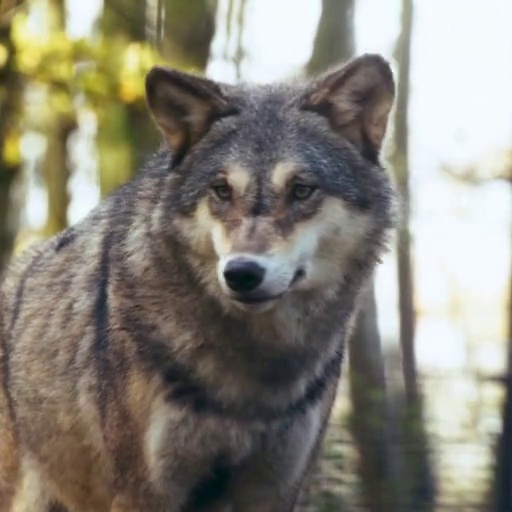} &
\includegraphics[width=0.137\linewidth,height=0.137\linewidth]{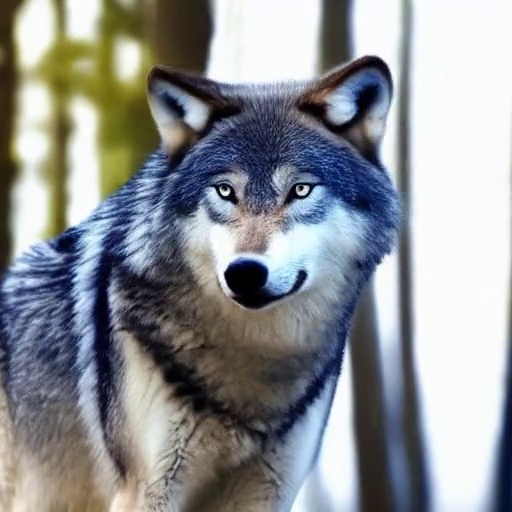} &
\includegraphics[width=0.137\linewidth,height=0.137\linewidth]{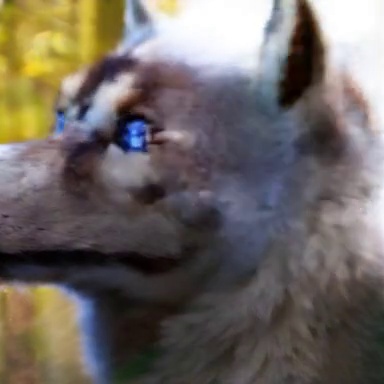} &
\includegraphics[width=0.137\linewidth,height=0.137\linewidth]{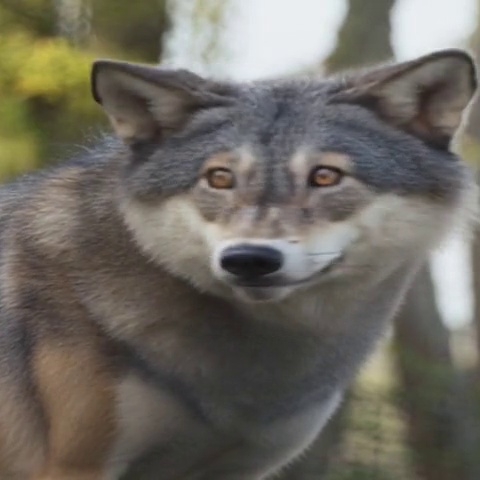} &
\includegraphics[width=0.137\linewidth,height=0.137\linewidth]{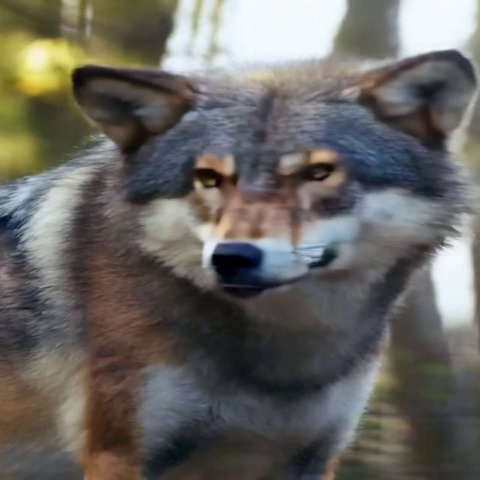} &
\includegraphics[width=0.137\linewidth,height=0.137\linewidth]{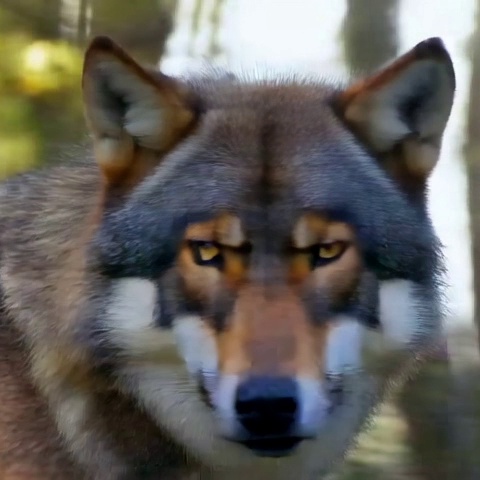} &
\includegraphics[width=0.137\linewidth,height=0.137\linewidth]{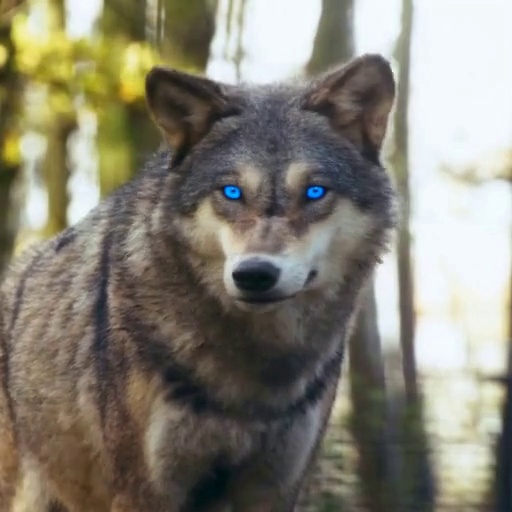} \\[-1.0pt]
\qualtime{$t_6$} &
\includegraphics[width=0.137\linewidth,height=0.137\linewidth]{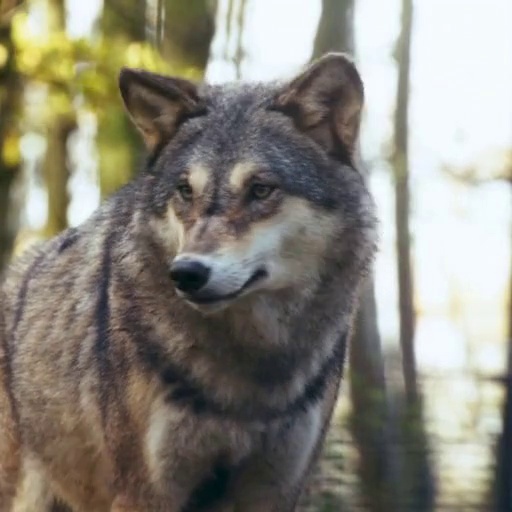} &
\includegraphics[width=0.137\linewidth,height=0.137\linewidth]{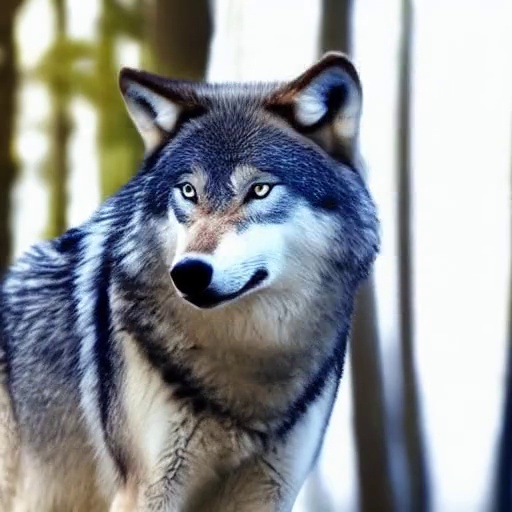} &
\includegraphics[width=0.137\linewidth,height=0.137\linewidth]{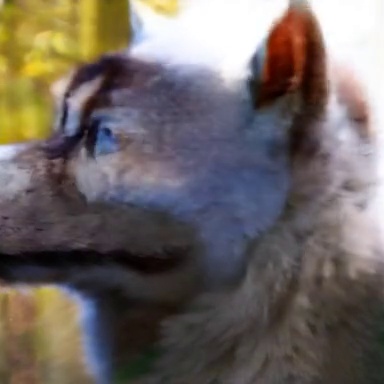} &
\includegraphics[width=0.137\linewidth,height=0.137\linewidth]{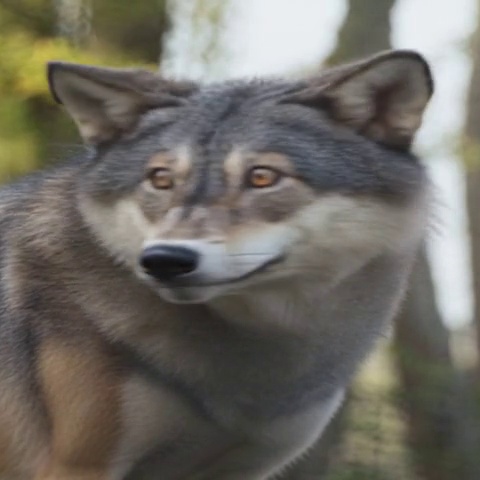} &
\includegraphics[width=0.137\linewidth,height=0.137\linewidth]{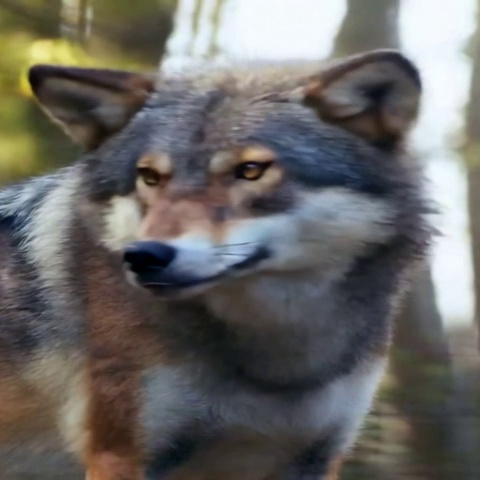} &
\includegraphics[width=0.137\linewidth,height=0.137\linewidth]{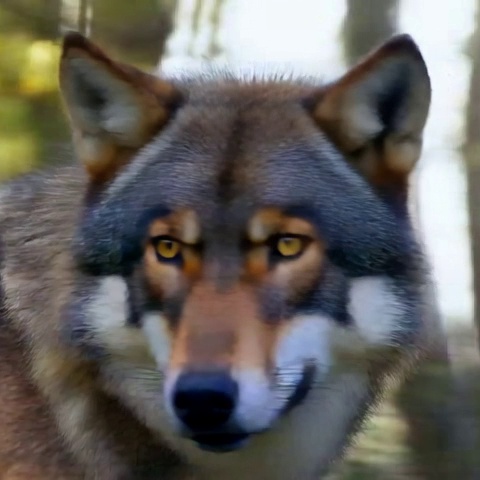} &
\includegraphics[width=0.137\linewidth,height=0.137\linewidth]{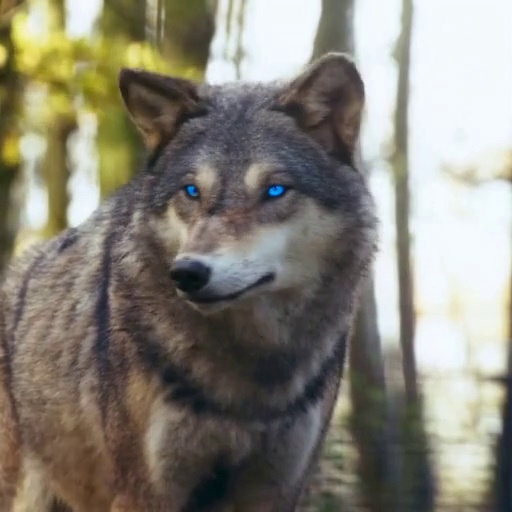} \\[-1.0pt]
\qualtime{$t_7$} &
\includegraphics[width=0.137\linewidth,height=0.137\linewidth]{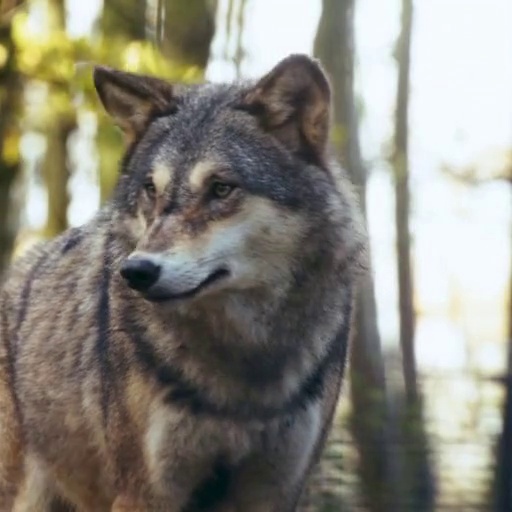} &
\includegraphics[width=0.137\linewidth,height=0.137\linewidth]{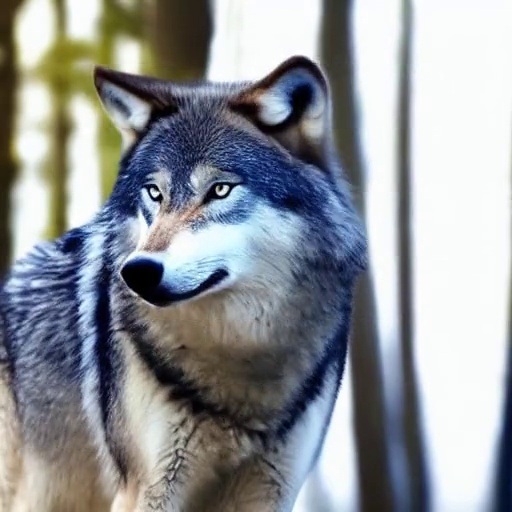} &
\includegraphics[width=0.137\linewidth,height=0.137\linewidth]{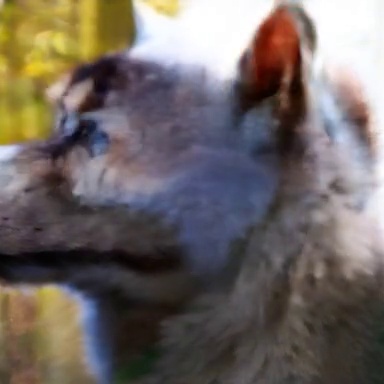} &
\includegraphics[width=0.137\linewidth,height=0.137\linewidth]{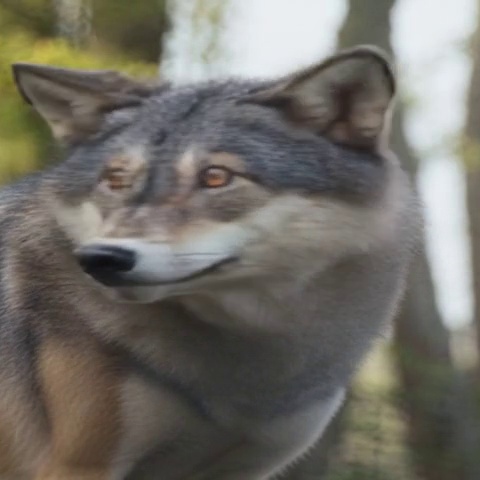} &
\includegraphics[width=0.137\linewidth,height=0.137\linewidth]{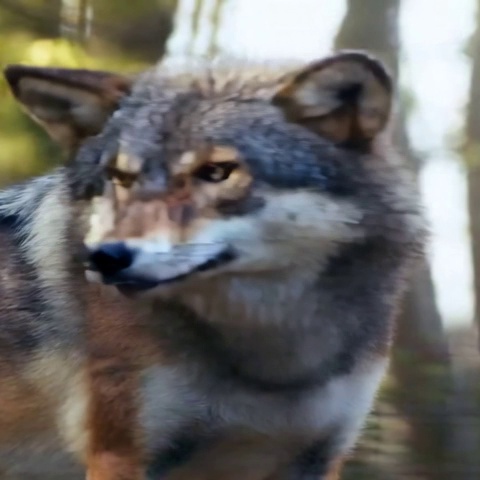} &
\includegraphics[width=0.137\linewidth,height=0.137\linewidth]{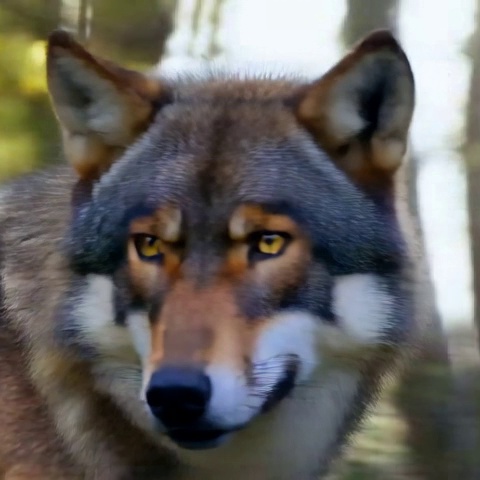} &
\includegraphics[width=0.137\linewidth,height=0.137\linewidth]{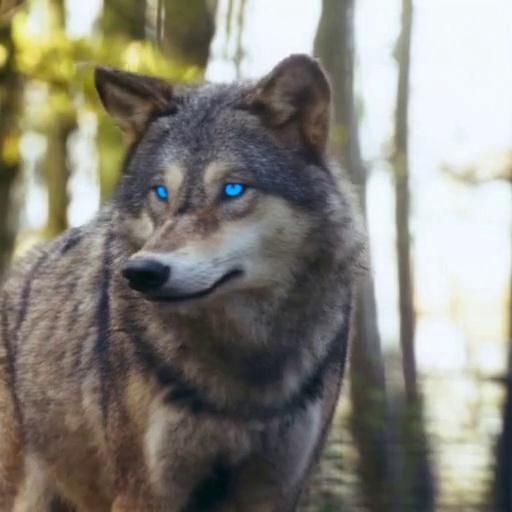} \\[-1.0pt]
\qualtime{$t_8$} &
\includegraphics[width=0.137\linewidth,height=0.137\linewidth]{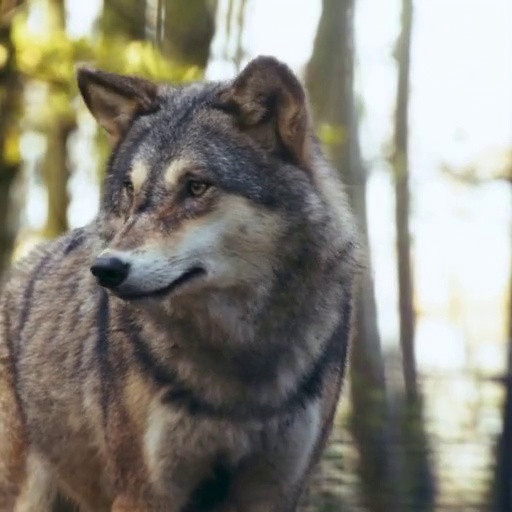} &
\includegraphics[width=0.137\linewidth,height=0.137\linewidth]{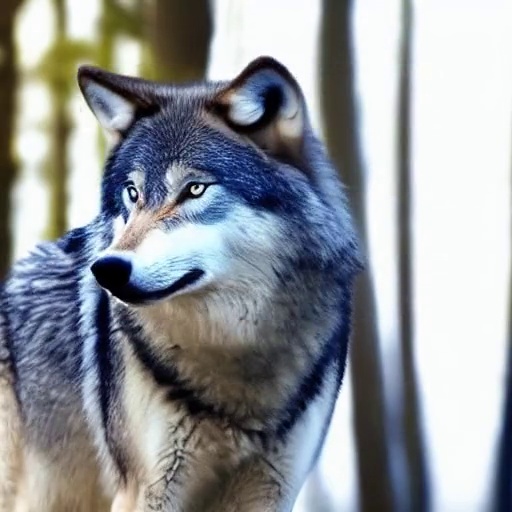} &
\includegraphics[width=0.137\linewidth,height=0.137\linewidth]{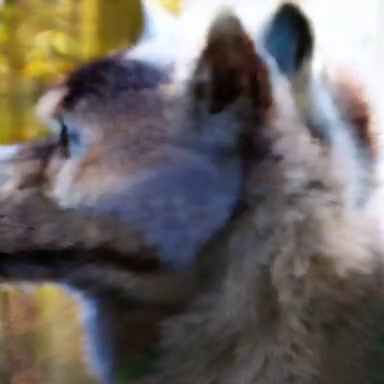} &
\includegraphics[width=0.137\linewidth,height=0.137\linewidth]{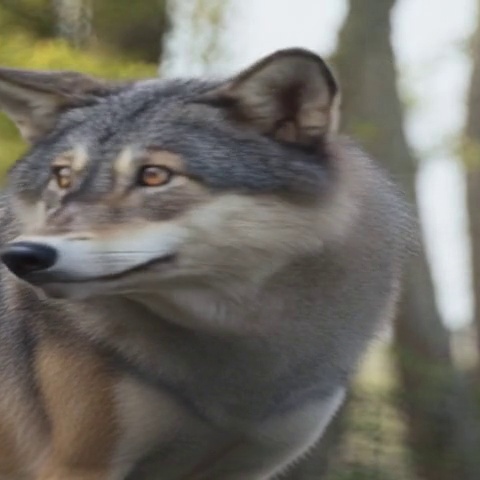} &
\includegraphics[width=0.137\linewidth,height=0.137\linewidth]{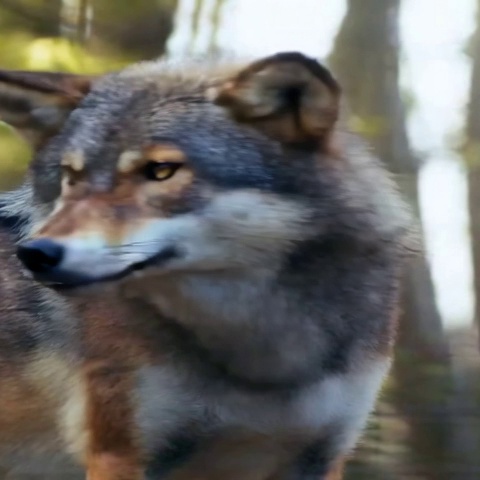} &
\includegraphics[width=0.137\linewidth,height=0.137\linewidth]{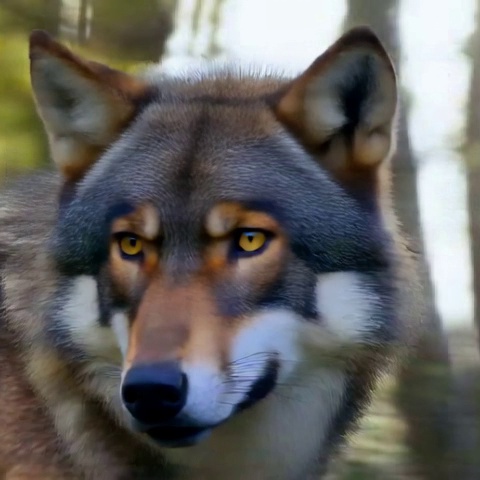} &
\includegraphics[width=0.137\linewidth,height=0.137\linewidth]{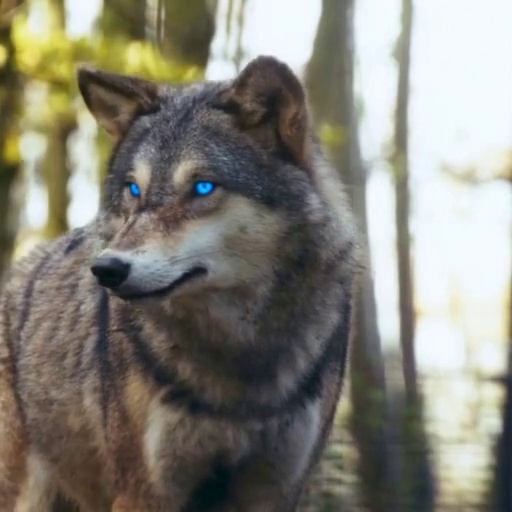} \\[-1.0pt]
\end{tabular}%
}
\caption{\textbf{Localized part-level editing.}
Given the instruction to make the wolf's eyes blue, \modelnamebf{} localizes the edit to the eyes while preserving the wolf's identity, facial structure, and motion across frames. Several baselines either fail to apply the edit consistently or alter broader aspects of the subject's appearance. Frames are sampled at uniformly spaced temporal positions throughout each video.}
\label{fig:qual_wolf_blue_eyes}
\end{figure*}

\begin{figure*}[t!]
\centering
\setlength{\tabcolsep}{0.45pt}
\renewcommand{\arraystretch}{0.62}
\scriptsize
\def\qualtime#1{\raisebox{0.050\linewidth}{\makebox[0.018\linewidth][r]{#1}}}
\resizebox{\textwidth}{!}{%
\begin{tabular}{@{}r@{\hspace{1pt}}cccccccc@{}}
\multicolumn{8}{c}{\small\textbf{Prompt:} Change the person's hair color to white while keeping the hairstyle exactly the same.} \\[3pt]
& \textbf{Source} & \textbf{VidToMe~\cite{li2024vidtome}} & \textbf{Pyramid-Edit~\cite{li2025five}} & \textbf{StreamEdit~\cite{jiao2026streamedit}} & \textbf{Wan-Edit~\cite{li2025five}} & \textbf{FlowDirector~\cite{li2026flowdirector}} & \makecell{\textbf{\modelname{}}\\\textbf{(Ours)}} \\[2pt]
\qualtime{$t_1$} &
\includegraphics[width=0.137\linewidth,height=0.137\linewidth]{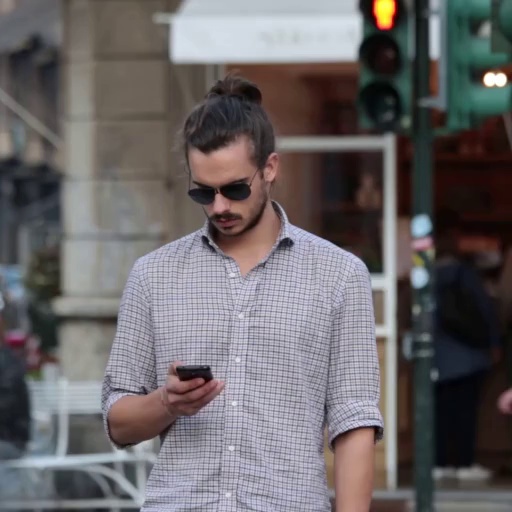} &
\includegraphics[width=0.137\linewidth,height=0.137\linewidth]{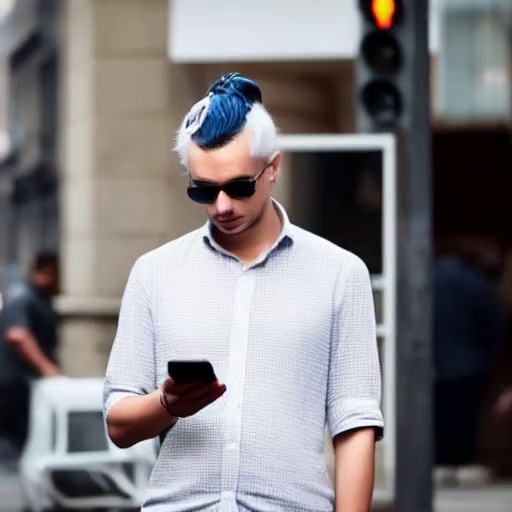} &
\includegraphics[width=0.137\linewidth,height=0.137\linewidth]{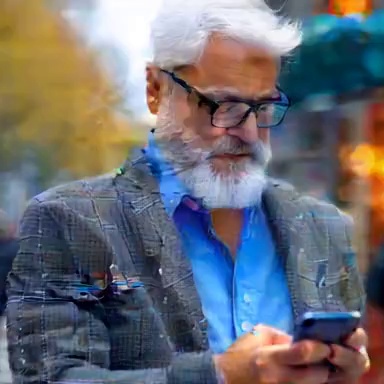} &
\includegraphics[width=0.137\linewidth,height=0.137\linewidth]{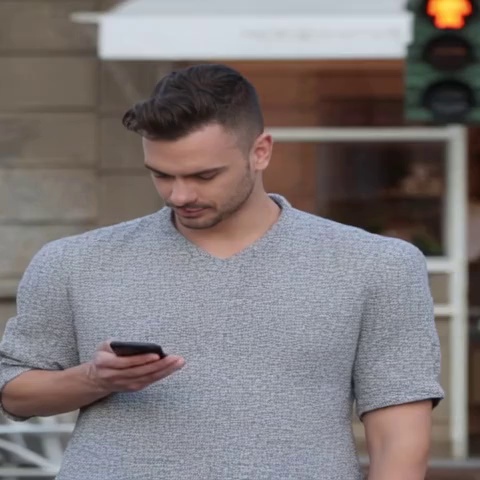} &
\includegraphics[width=0.137\linewidth,height=0.137\linewidth]{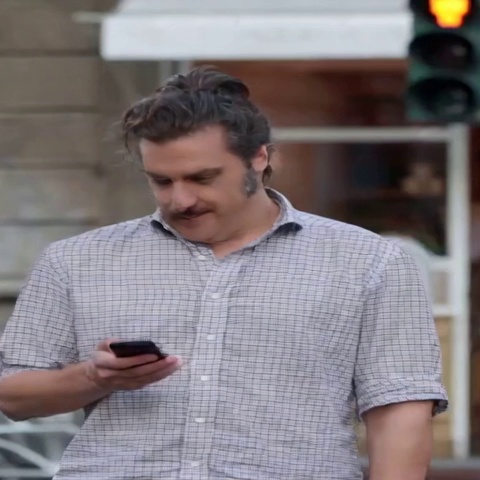} &
\includegraphics[width=0.137\linewidth,height=0.137\linewidth]{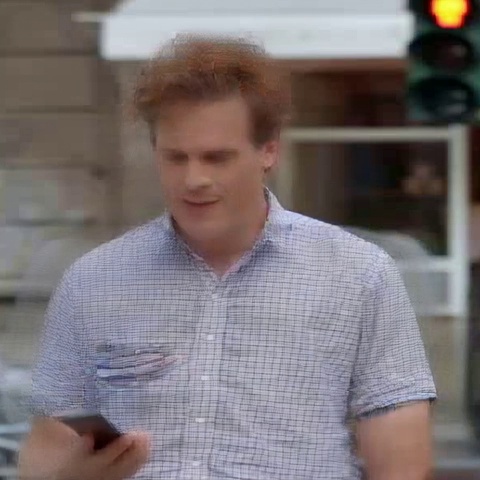} &
\includegraphics[width=0.137\linewidth,height=0.137\linewidth]{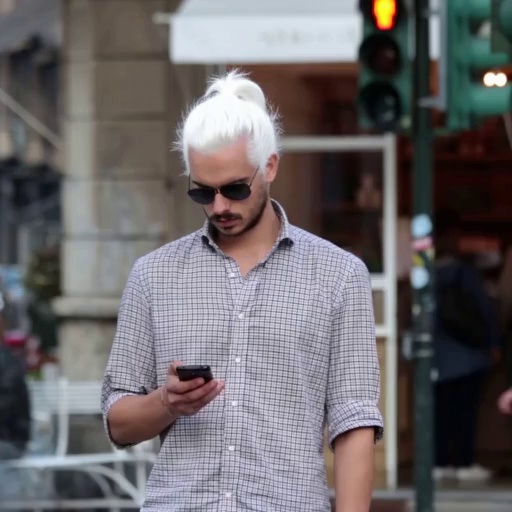} \\[-1.0pt]
\qualtime{$t_2$} &
\includegraphics[width=0.137\linewidth,height=0.137\linewidth]{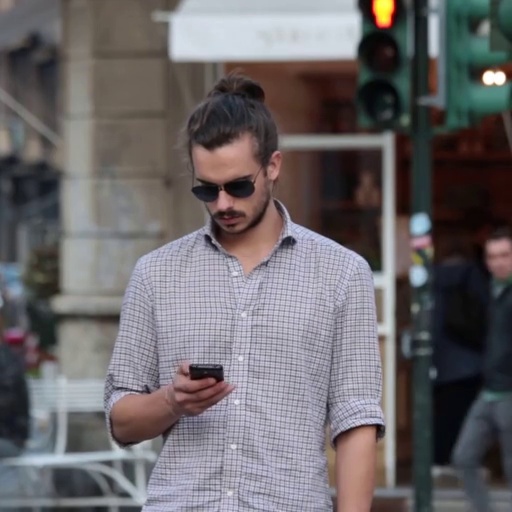} &
\includegraphics[width=0.137\linewidth,height=0.137\linewidth]{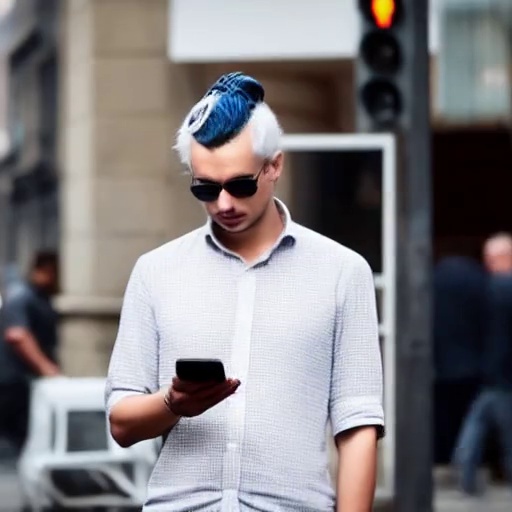} &
\includegraphics[width=0.137\linewidth,height=0.137\linewidth]{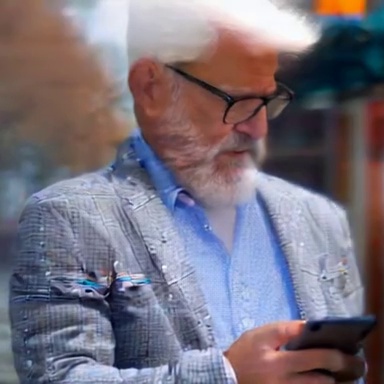} &
\includegraphics[width=0.137\linewidth,height=0.137\linewidth]{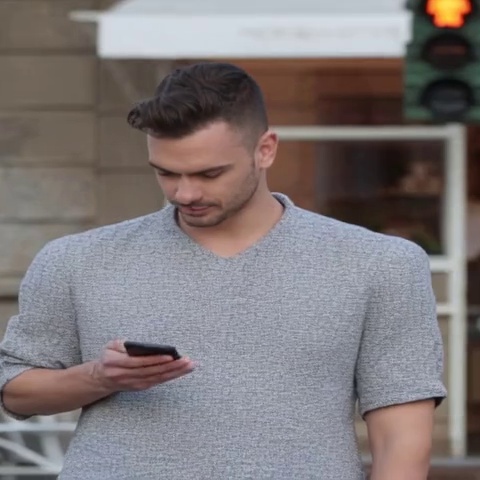} &
\includegraphics[width=0.137\linewidth,height=0.137\linewidth]{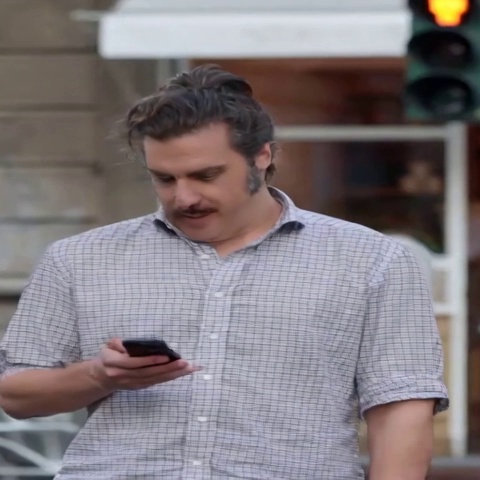} &
\includegraphics[width=0.137\linewidth,height=0.137\linewidth]{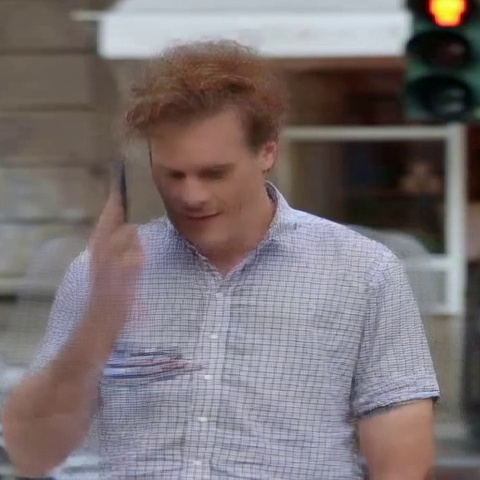} &
\includegraphics[width=0.137\linewidth,height=0.137\linewidth]{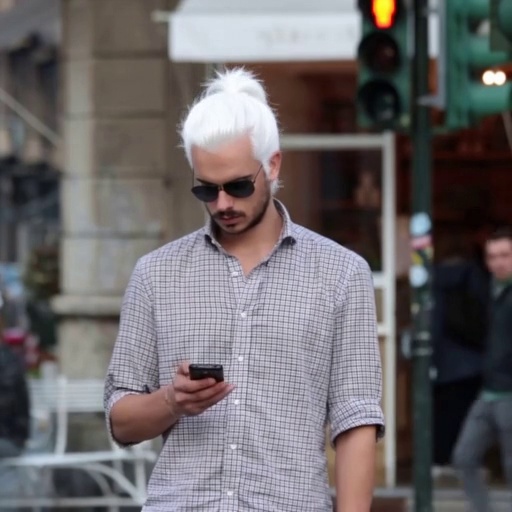} \\[-1.0pt]
\qualtime{$t_3$} &
\includegraphics[width=0.137\linewidth,height=0.137\linewidth]{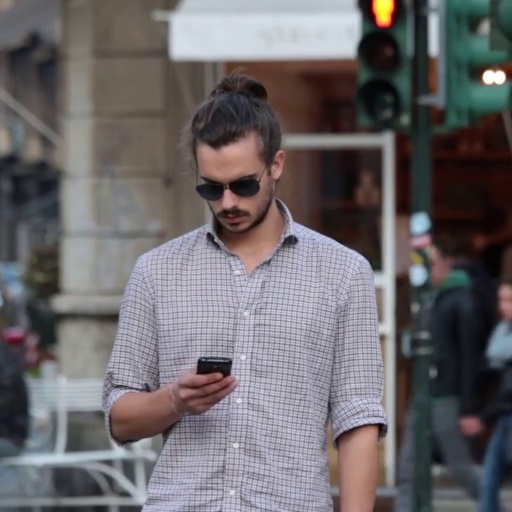} &
\includegraphics[width=0.137\linewidth,height=0.137\linewidth]{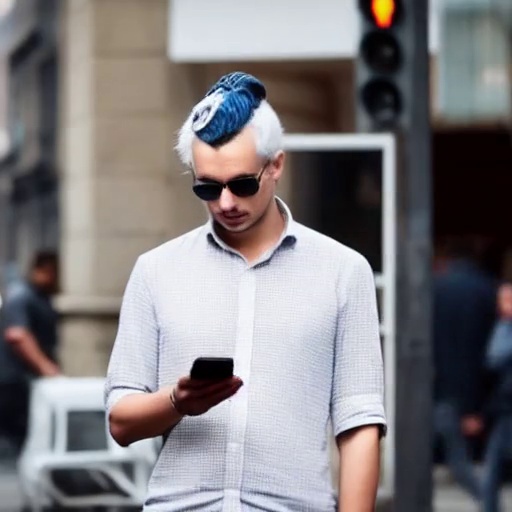} &
\includegraphics[width=0.137\linewidth,height=0.137\linewidth]{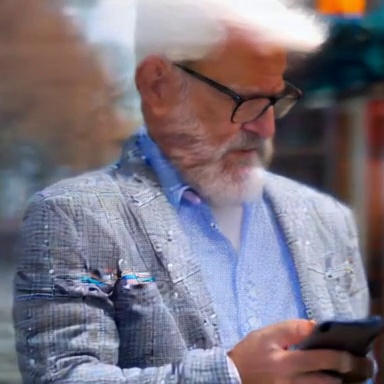} &
\includegraphics[width=0.137\linewidth,height=0.137\linewidth]{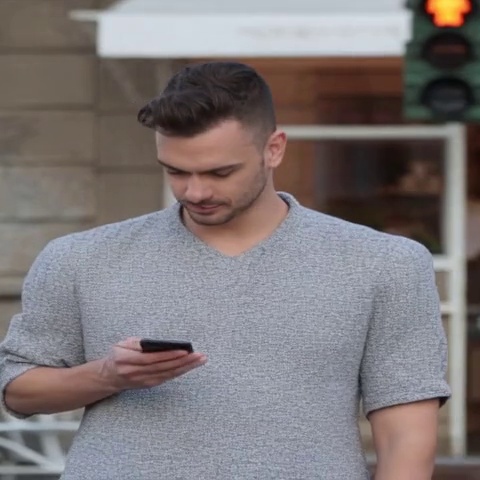} &
\includegraphics[width=0.137\linewidth,height=0.137\linewidth]{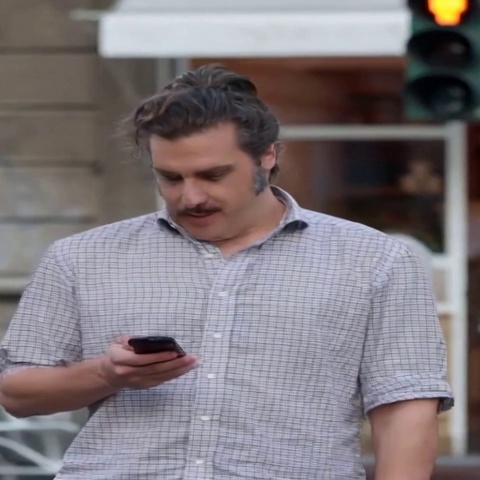} &
\includegraphics[width=0.137\linewidth,height=0.137\linewidth]{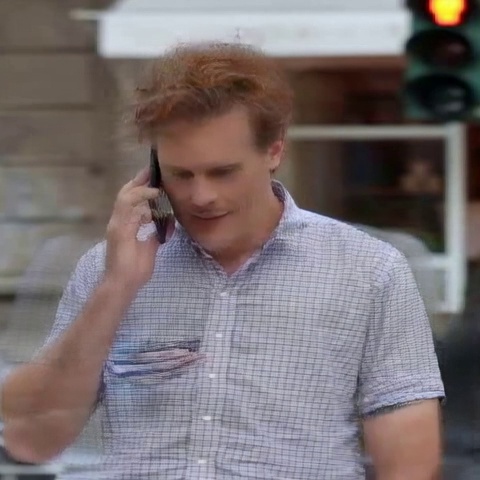} &
\includegraphics[width=0.137\linewidth,height=0.137\linewidth]{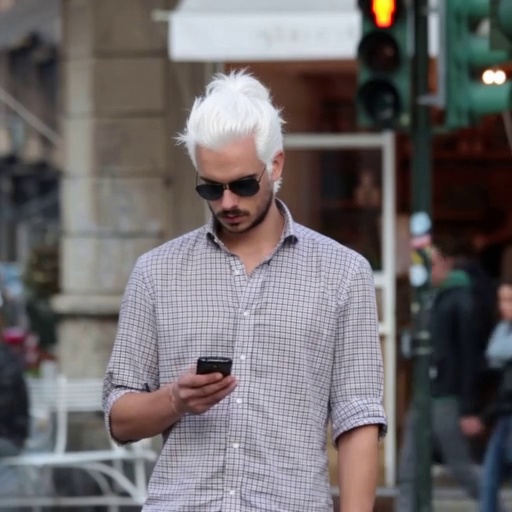} \\[-1.0pt]
\qualtime{$t_4$} &
\includegraphics[width=0.137\linewidth,height=0.137\linewidth]{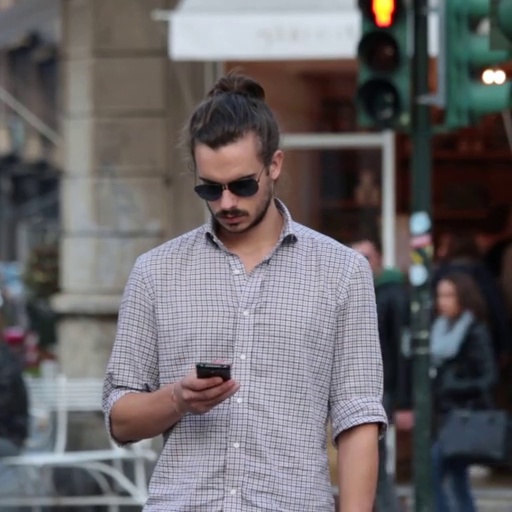} &
\includegraphics[width=0.137\linewidth,height=0.137\linewidth]{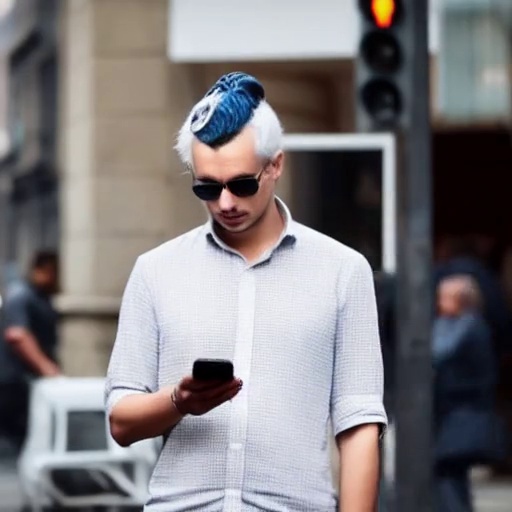} &
\includegraphics[width=0.137\linewidth,height=0.137\linewidth]{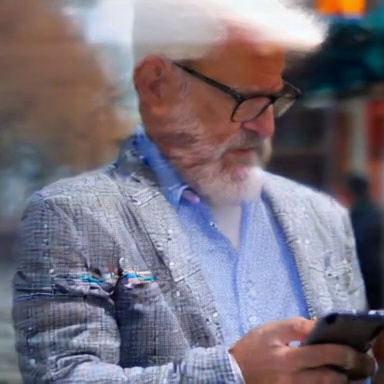} &
\includegraphics[width=0.137\linewidth,height=0.137\linewidth]{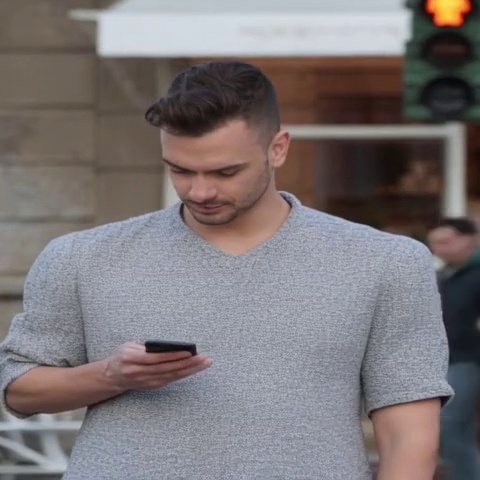} &
\includegraphics[width=0.137\linewidth,height=0.137\linewidth]{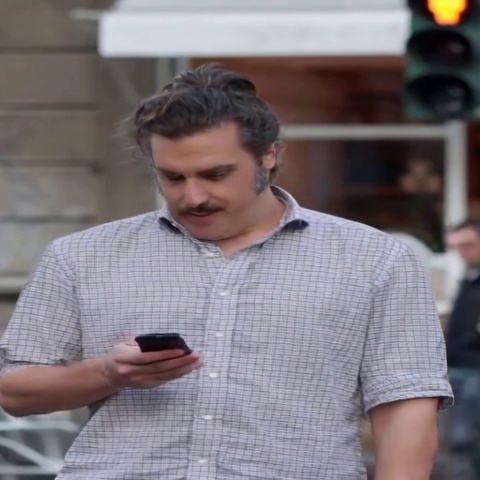} &
\includegraphics[width=0.137\linewidth,height=0.137\linewidth]{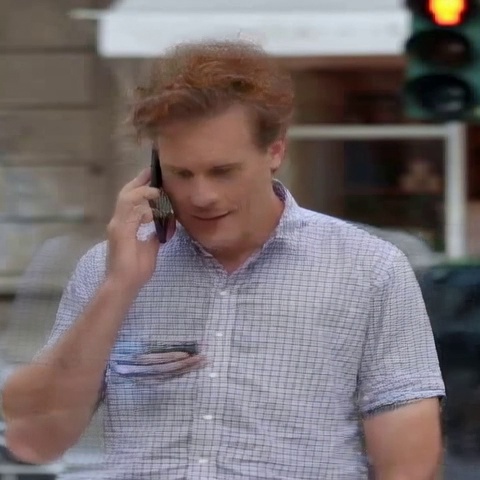} &
\includegraphics[width=0.137\linewidth,height=0.137\linewidth]{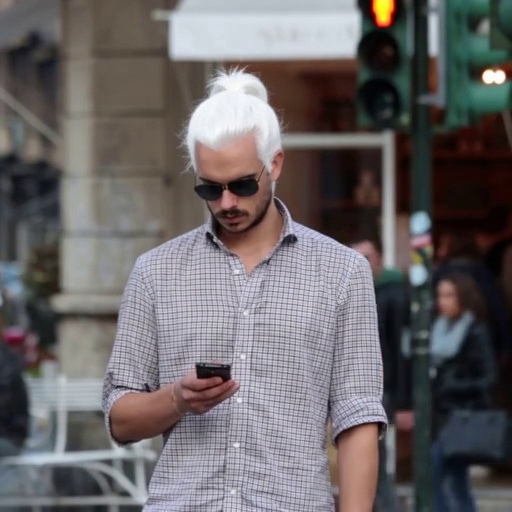} \\[-1.0pt]
\qualtime{$t_5$} &
\includegraphics[width=0.137\linewidth,height=0.137\linewidth]{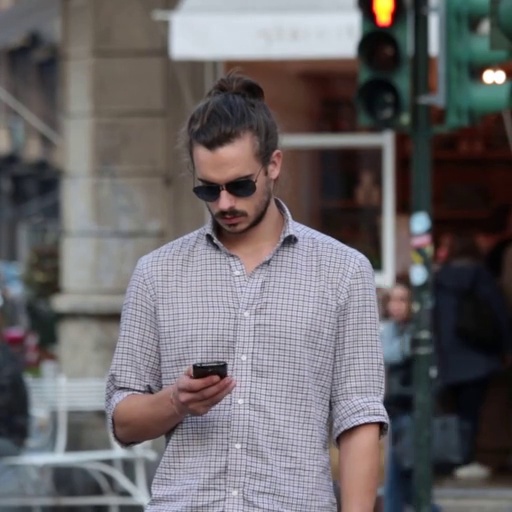} &
\includegraphics[width=0.137\linewidth,height=0.137\linewidth]{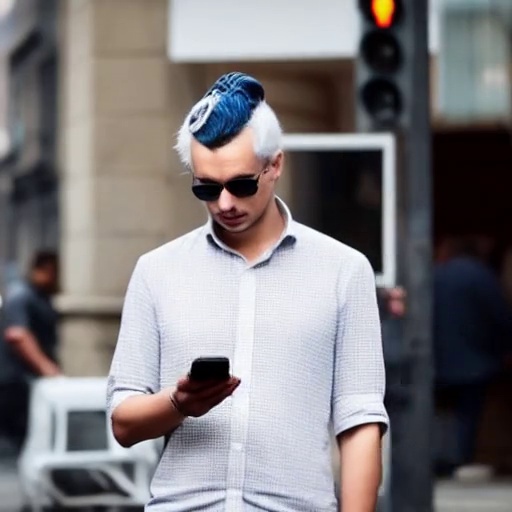} &
\includegraphics[width=0.137\linewidth,height=0.137\linewidth]{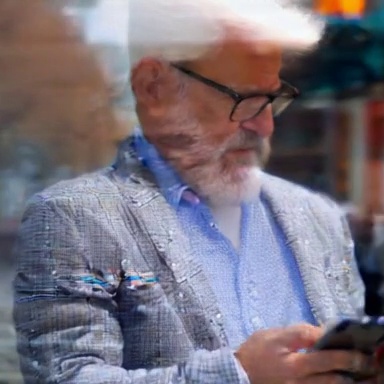} &
\includegraphics[width=0.137\linewidth,height=0.137\linewidth]{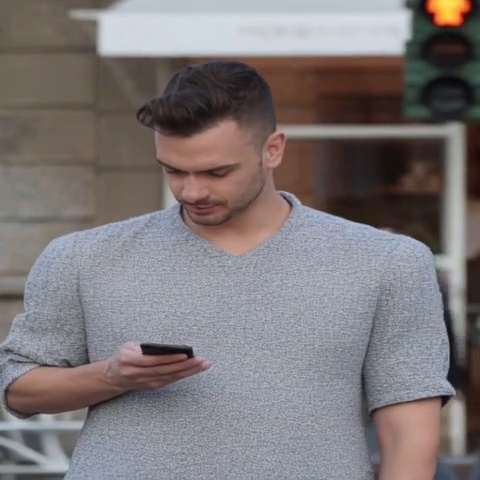} &
\includegraphics[width=0.137\linewidth,height=0.137\linewidth]{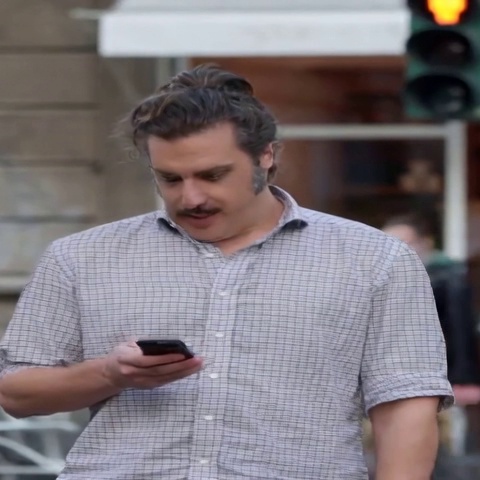} &
\includegraphics[width=0.137\linewidth,height=0.137\linewidth]{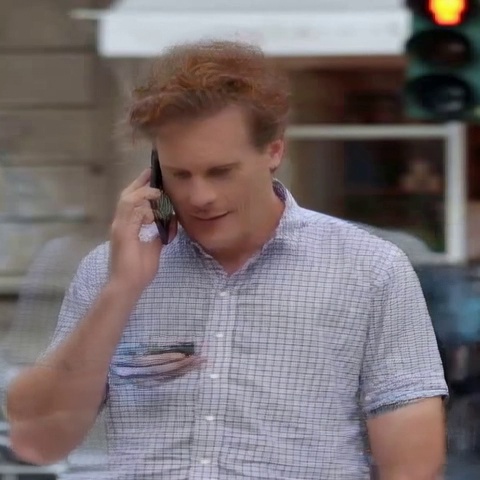} &
\includegraphics[width=0.137\linewidth,height=0.137\linewidth]{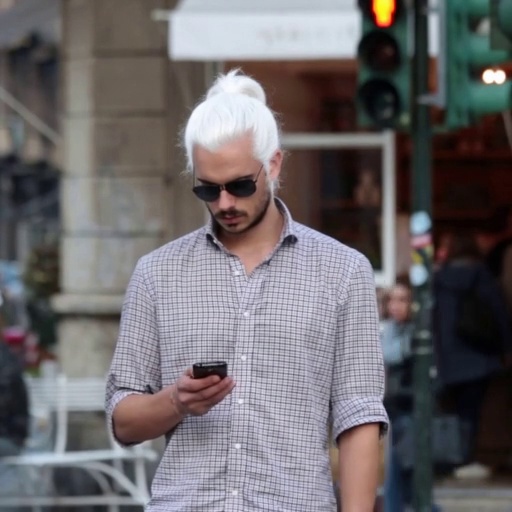} \\[-1.0pt]
\qualtime{$t_6$} &
\includegraphics[width=0.137\linewidth,height=0.137\linewidth]{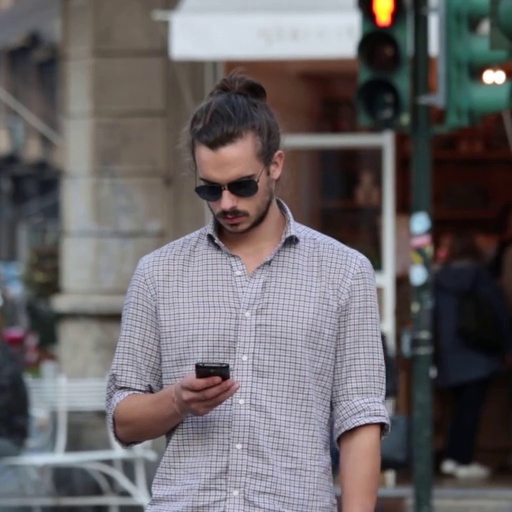} &
\includegraphics[width=0.137\linewidth,height=0.137\linewidth]{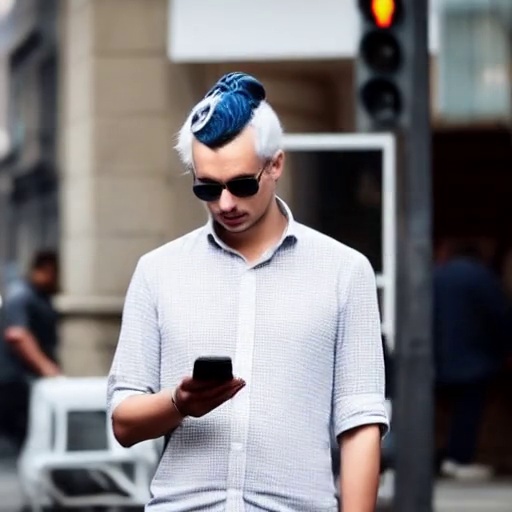} &
\includegraphics[width=0.137\linewidth,height=0.137\linewidth]{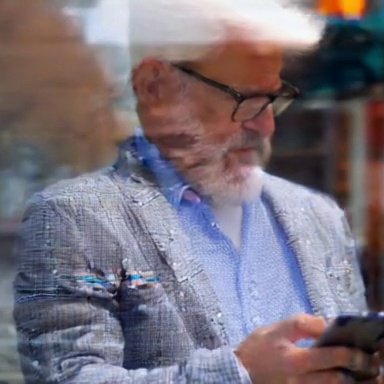} &
\includegraphics[width=0.137\linewidth,height=0.137\linewidth]{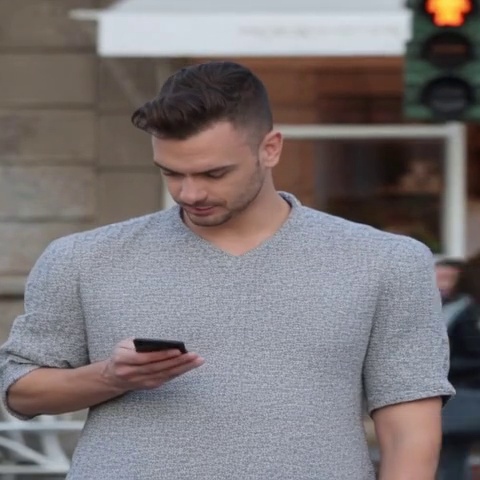} &
\includegraphics[width=0.137\linewidth,height=0.137\linewidth]{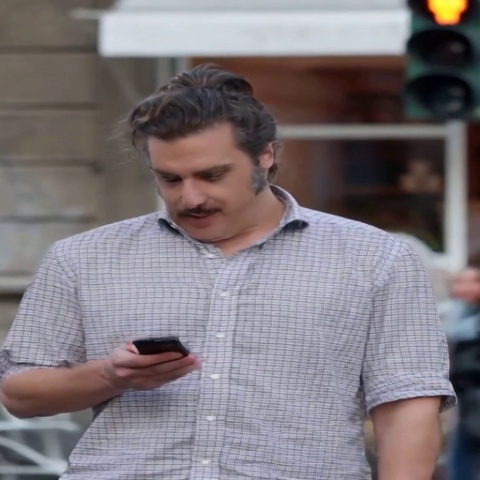} &
\includegraphics[width=0.137\linewidth,height=0.137\linewidth]{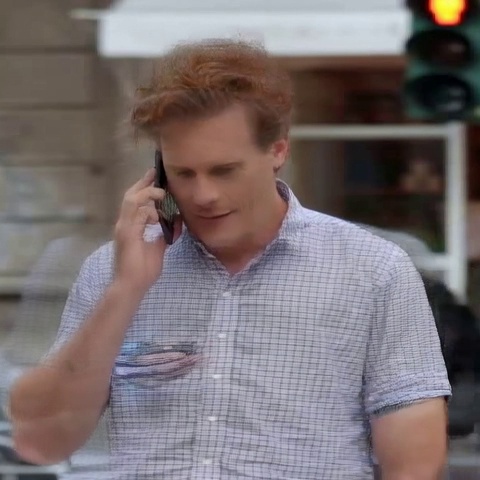} &
\includegraphics[width=0.137\linewidth,height=0.137\linewidth]{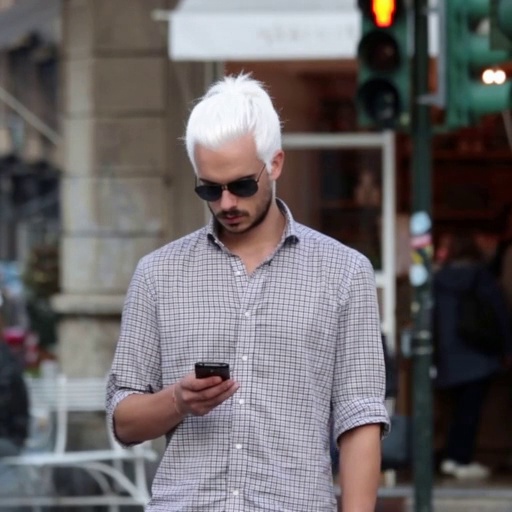} \\[-1.0pt]
\qualtime{$t_7$} &
\includegraphics[width=0.137\linewidth,height=0.137\linewidth]{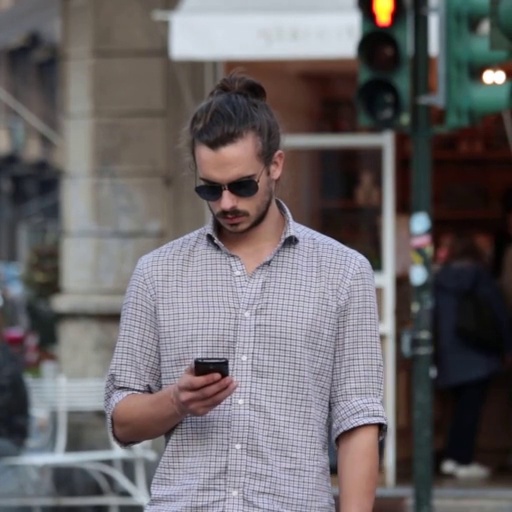} &
\includegraphics[width=0.137\linewidth,height=0.137\linewidth]{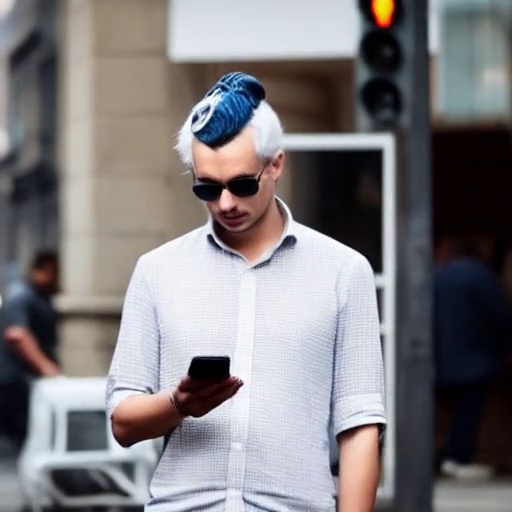} &
\includegraphics[width=0.137\linewidth,height=0.137\linewidth]{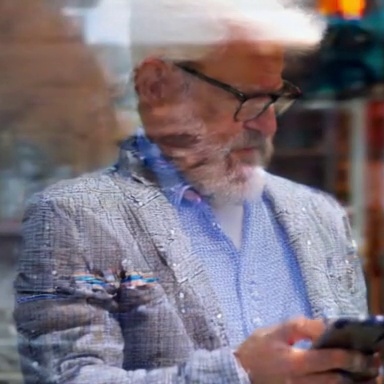} &
\includegraphics[width=0.137\linewidth,height=0.137\linewidth]{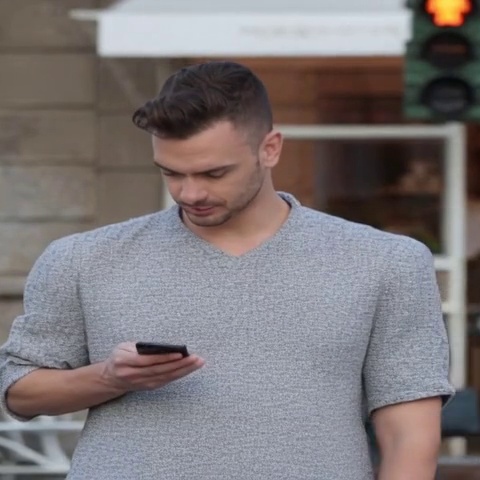} &
\includegraphics[width=0.137\linewidth,height=0.137\linewidth]{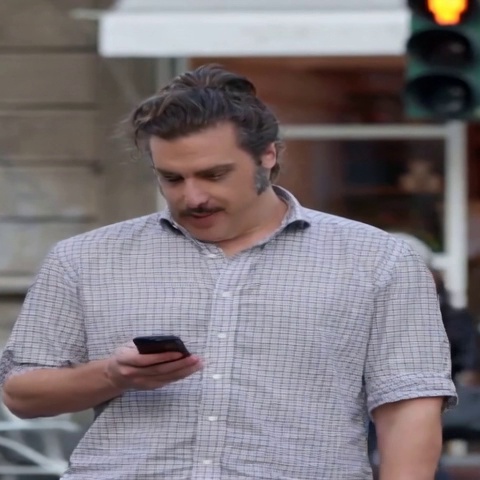} &
\includegraphics[width=0.137\linewidth,height=0.137\linewidth]{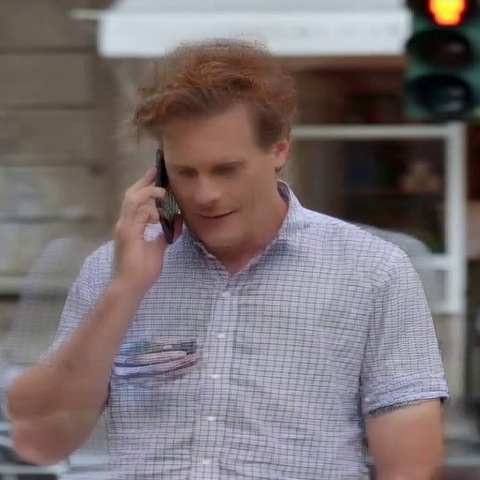} &
\includegraphics[width=0.137\linewidth,height=0.137\linewidth]{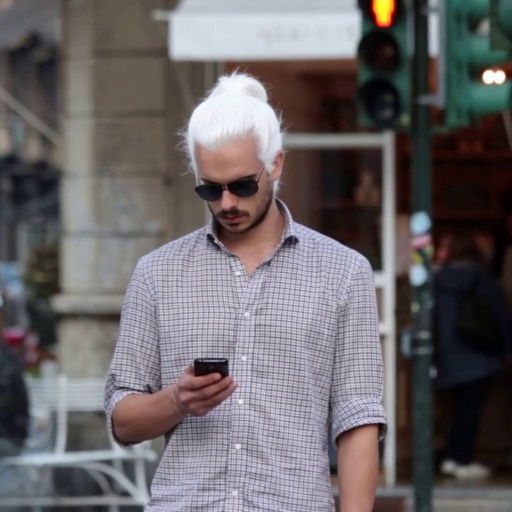} \\[-1.0pt]
\end{tabular}%
}
\caption{\textbf{Localized attribute editing.}
Given the instruction to change the person's hair color to white while keeping the hairstyle unchanged, \modelnamebf{} consistently changes the hair color while preserving the hairstyle, clothing, pose, and surrounding scene. Several competing methods either make only weak changes or alter other aspects of the subject. Frames are sampled at uniformly spaced temporal positions throughout each video.}
\label{fig:qual_man_white_hair}
\end{figure*}

\begin{figure*}[t!]
\centering
\setlength{\tabcolsep}{0.45pt}
\renewcommand{\arraystretch}{0.62}
\scriptsize
\def\qualtime#1{\raisebox{0.050\linewidth}{\makebox[0.018\linewidth][r]{#1}}}
\resizebox{\textwidth}{!}{%
\begin{tabular}{@{}r@{\hspace{1pt}}ccccccccc@{}}
\multicolumn{10}{c}{\small\textbf{Prompt:} Change the dog in the image to the lion shown in the reference image.} \\[3pt]
 & \textbf{Ref.} & \textbf{Source} & \textbf{AnyV2V~\cite{ku2024anyv2v}} & \textbf{VidToMe~\cite{li2024vidtome}} & \textbf{Pyramid-Edit~\cite{li2025five}} & \textbf{StreamEdit~\cite{jiao2026streamedit}} & \textbf{Wan-Edit~\cite{li2025five}} & \textbf{FlowDirector~\cite{li2026flowdirector}} & \makecell{\textbf{\modelname{}}\\\textbf{(Ours)}} \\[2pt]
\qualtime{$t_1$} &
\includegraphics[width=0.106\linewidth,height=0.106\linewidth]{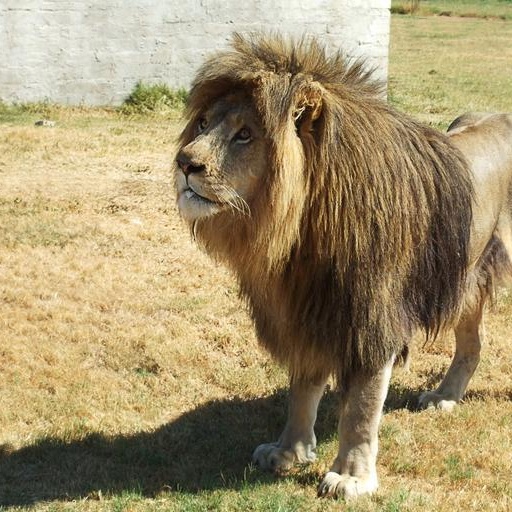} &
\includegraphics[width=0.106\linewidth,height=0.106\linewidth]{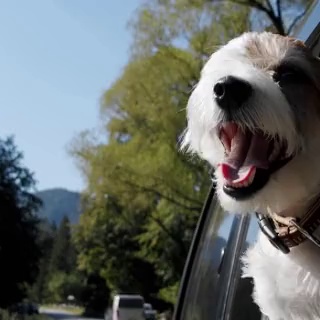} &
\includegraphics[width=0.106\linewidth,height=0.106\linewidth]{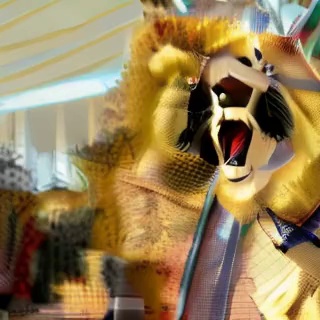} &
\includegraphics[width=0.106\linewidth,height=0.106\linewidth]{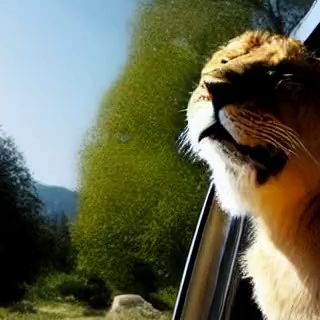} &
\includegraphics[width=0.106\linewidth,height=0.106\linewidth]{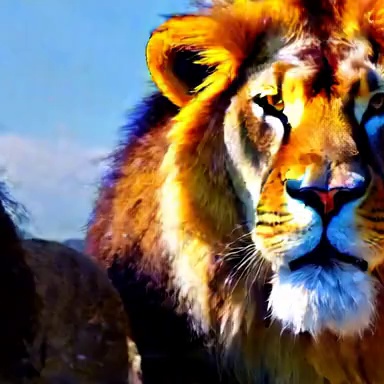} &
\includegraphics[width=0.106\linewidth,height=0.106\linewidth]{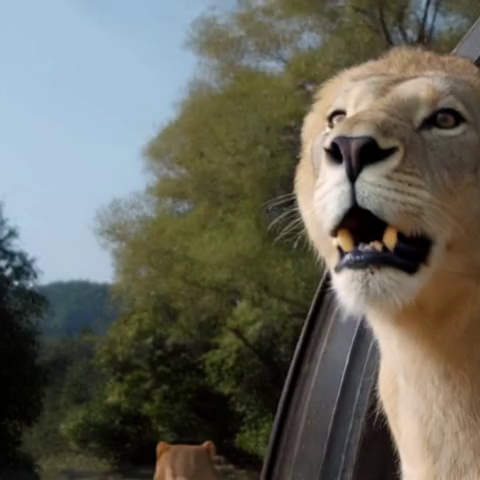} &
\includegraphics[width=0.106\linewidth,height=0.106\linewidth]{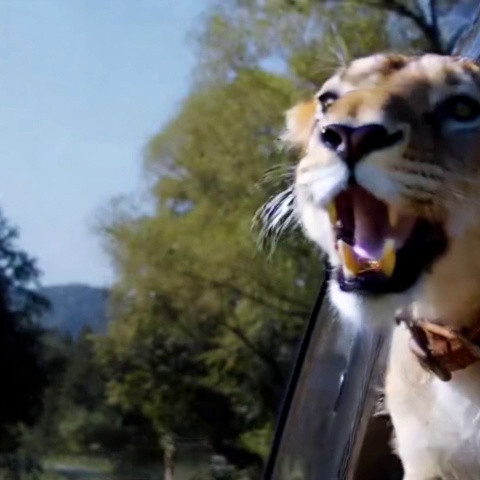} &
\includegraphics[width=0.106\linewidth,height=0.106\linewidth]{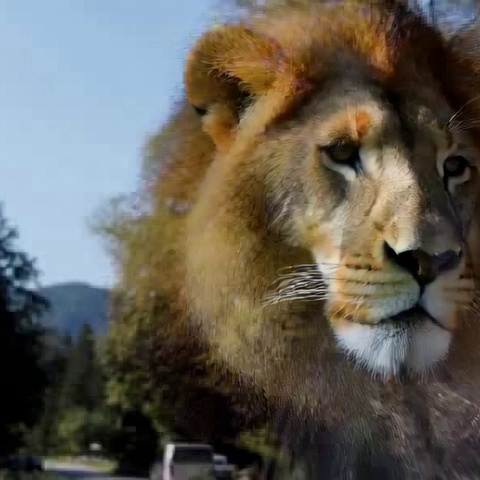} &
\includegraphics[width=0.106\linewidth,height=0.106\linewidth]{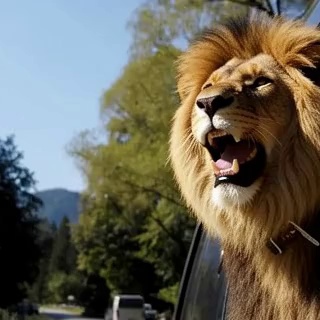} \\[-1.0pt]
\qualtime{$t_2$} &
\includegraphics[width=0.106\linewidth,height=0.106\linewidth]{assets/misc/figures/dog_lion_frames/selection_12/ref_01.jpg} &
\includegraphics[width=0.106\linewidth,height=0.106\linewidth]{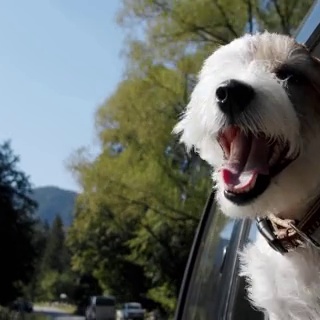} &
\includegraphics[width=0.106\linewidth,height=0.106\linewidth]{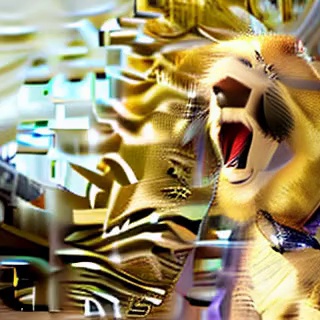} &
\includegraphics[width=0.106\linewidth,height=0.106\linewidth]{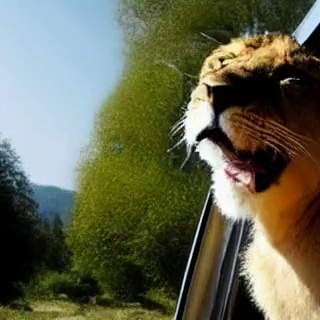} &
\includegraphics[width=0.106\linewidth,height=0.106\linewidth]{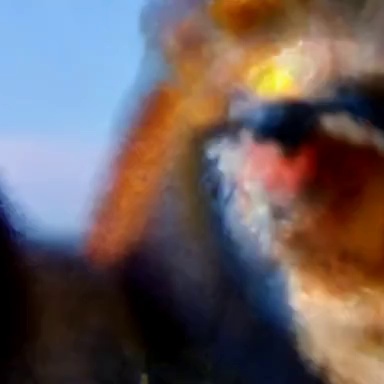} &
\includegraphics[width=0.106\linewidth,height=0.106\linewidth]{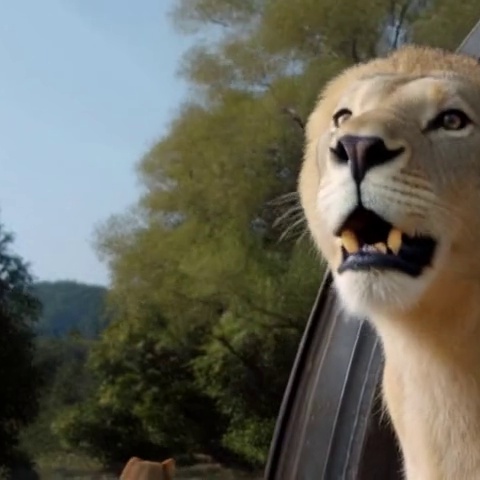} &
\includegraphics[width=0.106\linewidth,height=0.106\linewidth]{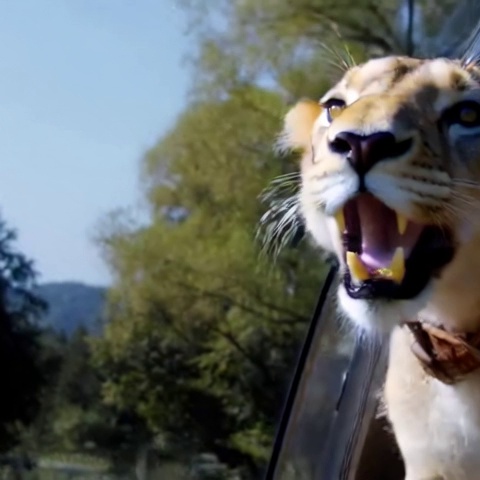} &
\includegraphics[width=0.106\linewidth,height=0.106\linewidth]{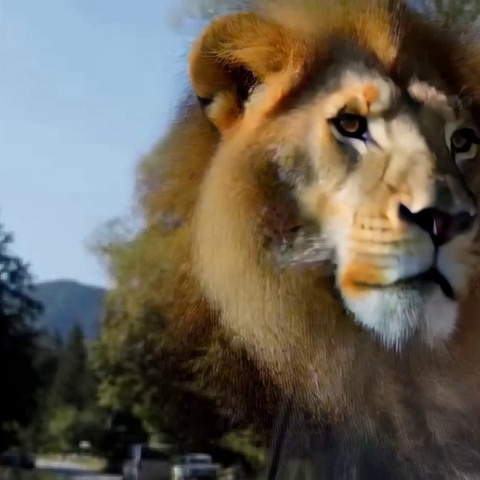} &
\includegraphics[width=0.106\linewidth,height=0.106\linewidth]{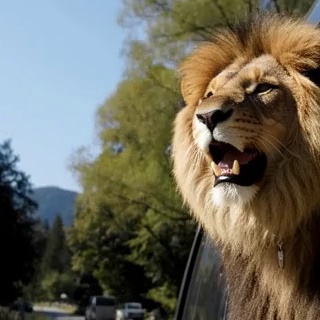} \\[-1.0pt]
\qualtime{$t_3$} &
\includegraphics[width=0.106\linewidth,height=0.106\linewidth]{assets/misc/figures/dog_lion_frames/selection_12/ref_01.jpg} &
\includegraphics[width=0.106\linewidth,height=0.106\linewidth]{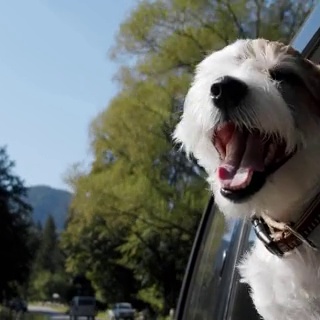} &
\includegraphics[width=0.106\linewidth,height=0.106\linewidth]{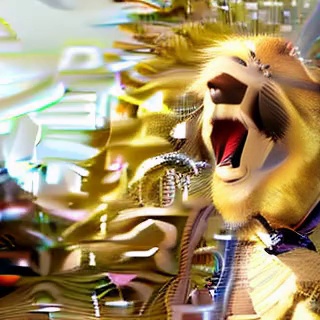} &
\includegraphics[width=0.106\linewidth,height=0.106\linewidth]{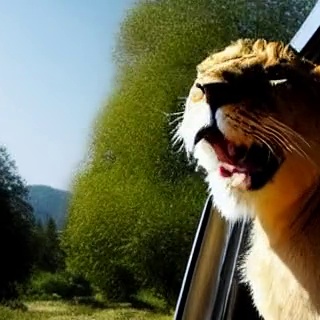} &
\includegraphics[width=0.106\linewidth,height=0.106\linewidth]{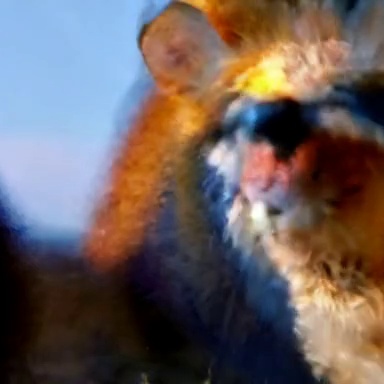} &
\includegraphics[width=0.106\linewidth,height=0.106\linewidth]{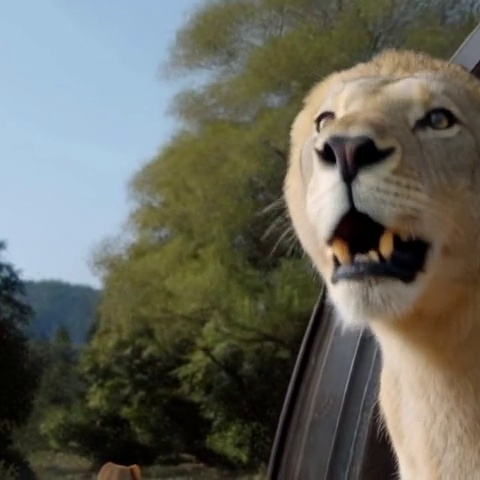} &
\includegraphics[width=0.106\linewidth,height=0.106\linewidth]{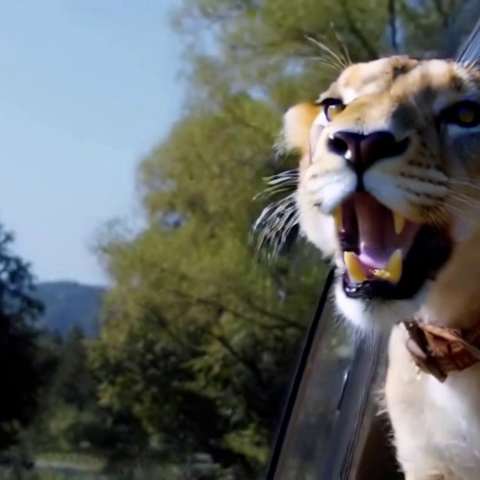} &
\includegraphics[width=0.106\linewidth,height=0.106\linewidth]{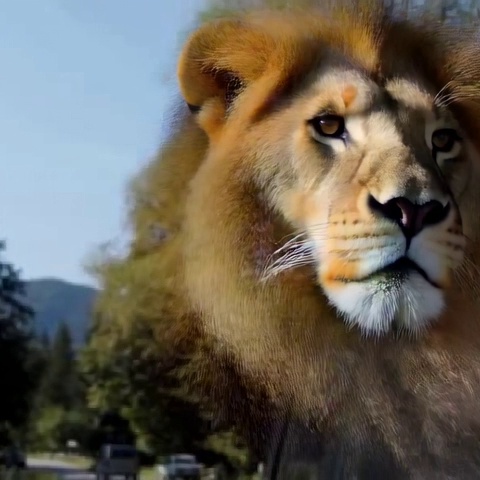} &
\includegraphics[width=0.106\linewidth,height=0.106\linewidth]{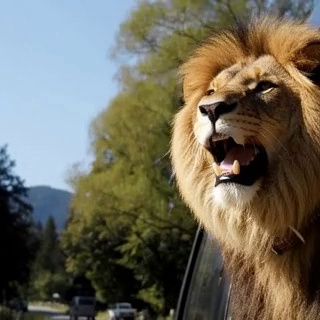} \\[-1.0pt]
\qualtime{$t_4$} &
\includegraphics[width=0.106\linewidth,height=0.106\linewidth]{assets/misc/figures/dog_lion_frames/selection_12/ref_01.jpg} &
\includegraphics[width=0.106\linewidth,height=0.106\linewidth]{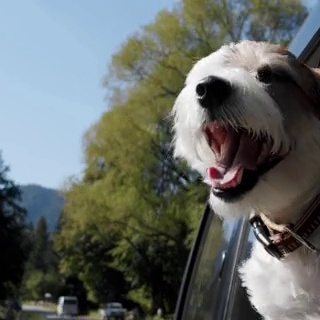} &
\includegraphics[width=0.106\linewidth,height=0.106\linewidth]{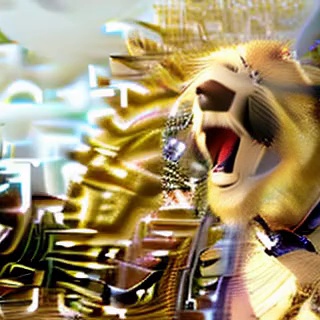} &
\includegraphics[width=0.106\linewidth,height=0.106\linewidth]{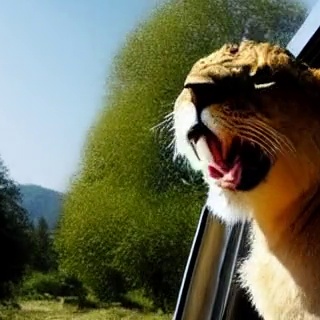} &
\includegraphics[width=0.106\linewidth,height=0.106\linewidth]{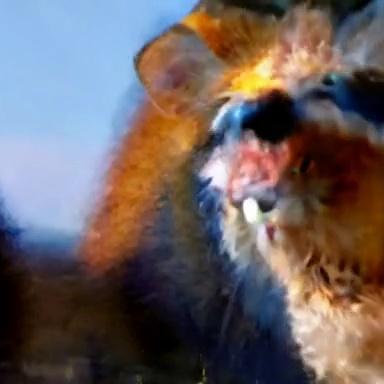} &
\includegraphics[width=0.106\linewidth,height=0.106\linewidth]{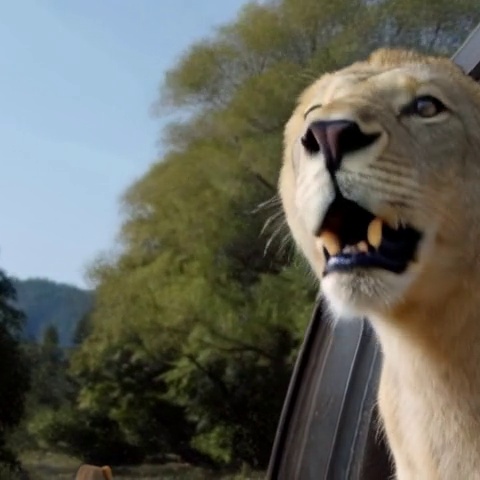} &
\includegraphics[width=0.106\linewidth,height=0.106\linewidth]{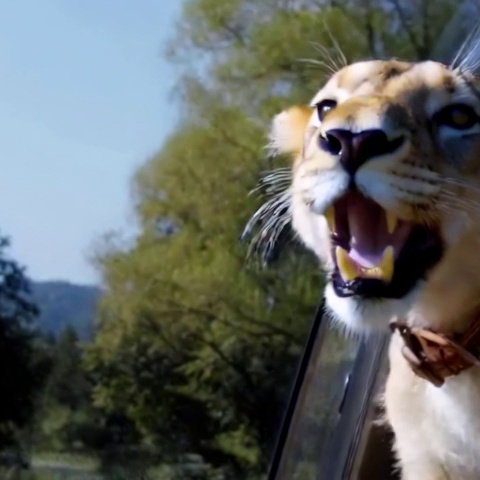} &
\includegraphics[width=0.106\linewidth,height=0.106\linewidth]{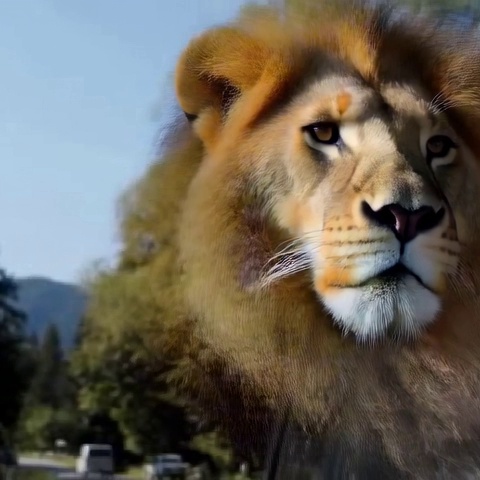} &
\includegraphics[width=0.106\linewidth,height=0.106\linewidth]{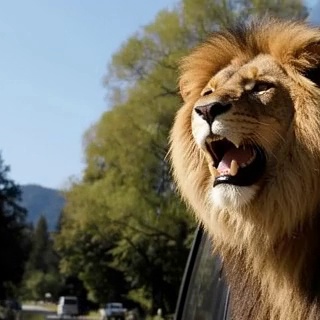} \\[-1.0pt]
\qualtime{$t_5$} &
\includegraphics[width=0.106\linewidth,height=0.106\linewidth]{assets/misc/figures/dog_lion_frames/selection_12/ref_01.jpg} &
\includegraphics[width=0.106\linewidth,height=0.106\linewidth]{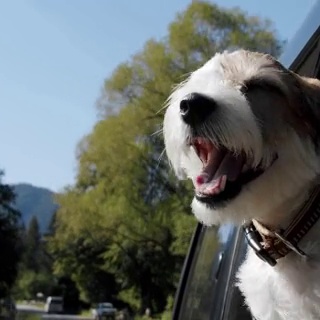} &
\includegraphics[width=0.106\linewidth,height=0.106\linewidth]{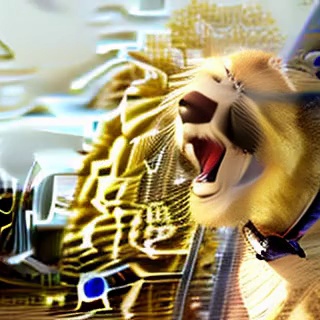} &
\includegraphics[width=0.106\linewidth,height=0.106\linewidth]{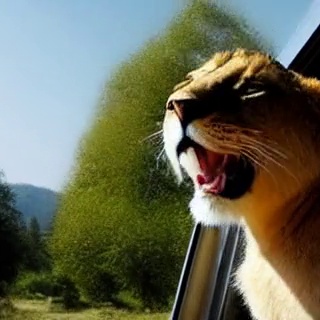} &
\includegraphics[width=0.106\linewidth,height=0.106\linewidth]{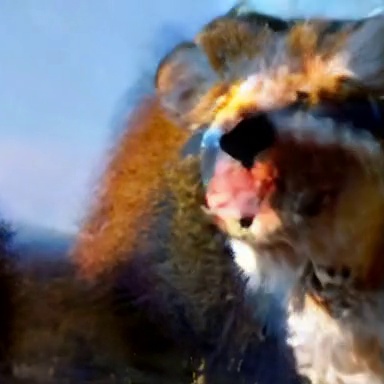} &
\includegraphics[width=0.106\linewidth,height=0.106\linewidth]{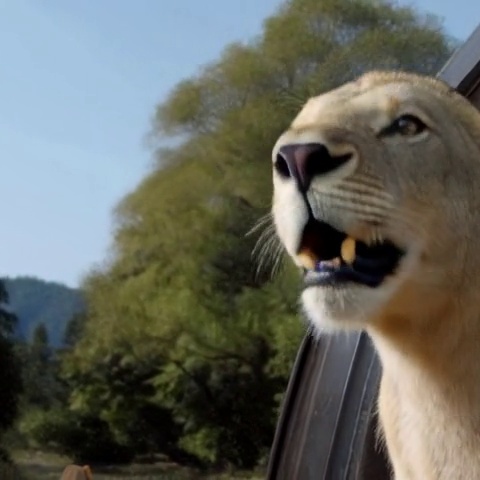} &
\includegraphics[width=0.106\linewidth,height=0.106\linewidth]{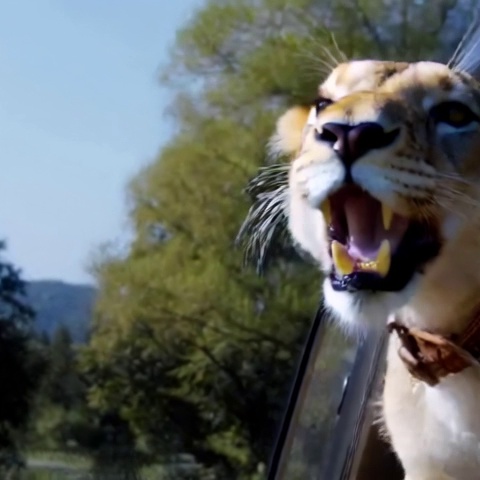} &
\includegraphics[width=0.106\linewidth,height=0.106\linewidth]{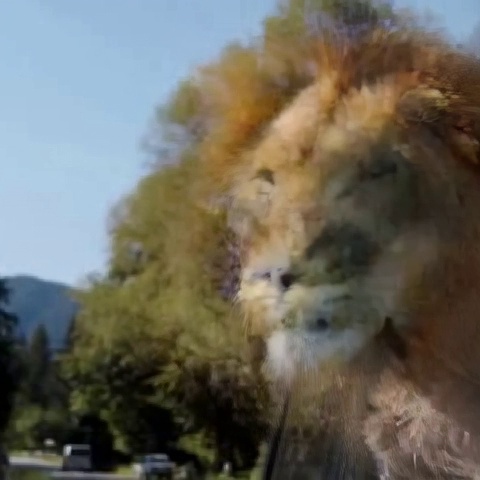} &
\includegraphics[width=0.106\linewidth,height=0.106\linewidth]{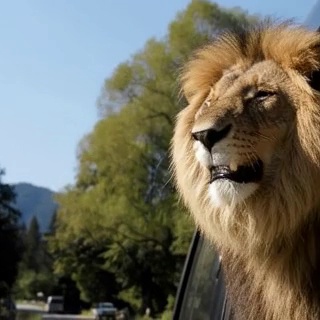} \\[-1.0pt]
\qualtime{$t_6$} &
\includegraphics[width=0.106\linewidth,height=0.106\linewidth]{assets/misc/figures/dog_lion_frames/selection_12/ref_01.jpg} &
\includegraphics[width=0.106\linewidth,height=0.106\linewidth]{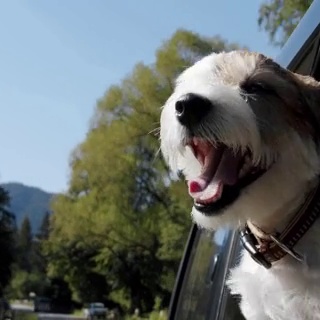} &
\includegraphics[width=0.106\linewidth,height=0.106\linewidth]{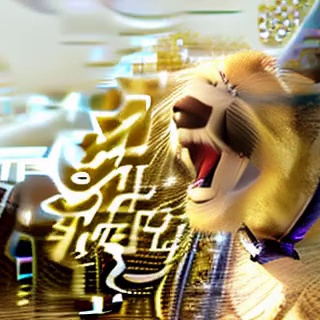} &
\includegraphics[width=0.106\linewidth,height=0.106\linewidth]{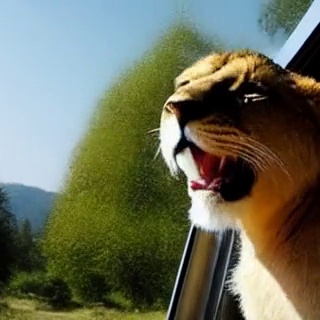} &
\includegraphics[width=0.106\linewidth,height=0.106\linewidth]{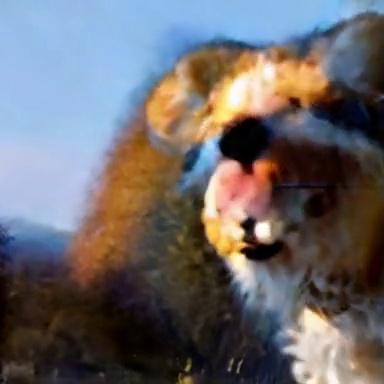} &
\includegraphics[width=0.106\linewidth,height=0.106\linewidth]{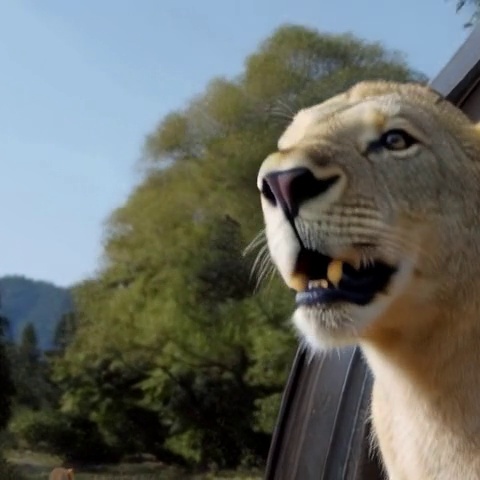} &
\includegraphics[width=0.106\linewidth,height=0.106\linewidth]{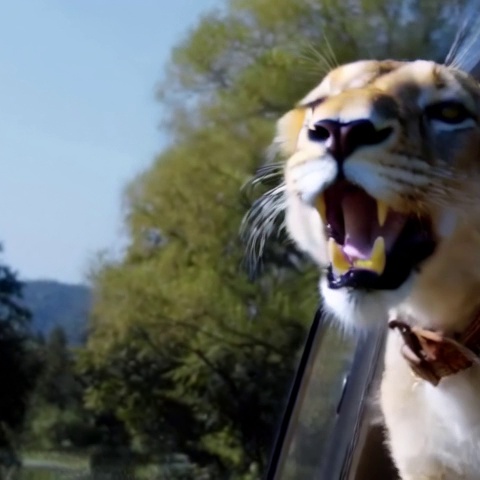} &
\includegraphics[width=0.106\linewidth,height=0.106\linewidth]{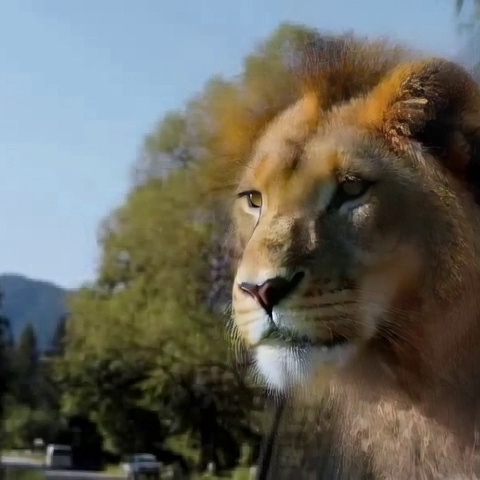} &
\includegraphics[width=0.106\linewidth,height=0.106\linewidth]{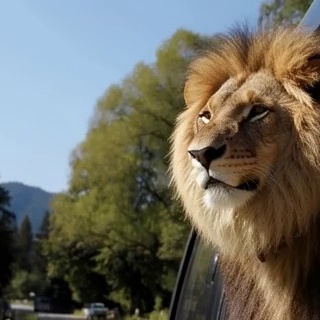} \\[-1.0pt]
\qualtime{$t_7$} &
\includegraphics[width=0.106\linewidth,height=0.106\linewidth]{assets/misc/figures/dog_lion_frames/selection_12/ref_01.jpg} &
\includegraphics[width=0.106\linewidth,height=0.106\linewidth]{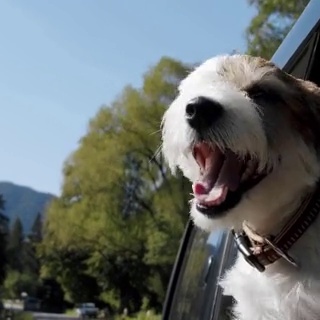} &
\includegraphics[width=0.106\linewidth,height=0.106\linewidth]{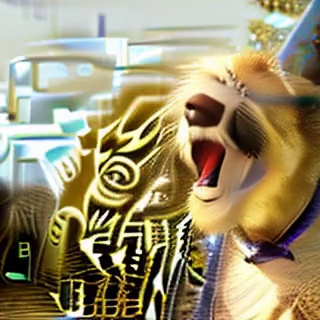} &
\includegraphics[width=0.106\linewidth,height=0.106\linewidth]{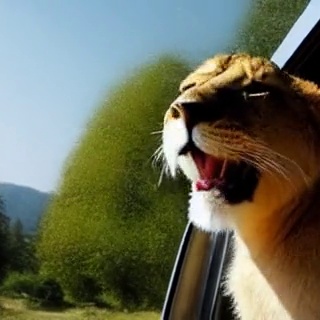} &
\includegraphics[width=0.106\linewidth,height=0.106\linewidth]{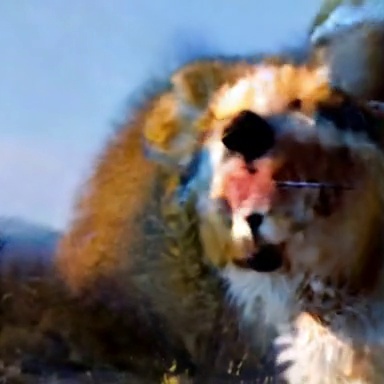} &
\includegraphics[width=0.106\linewidth,height=0.106\linewidth]{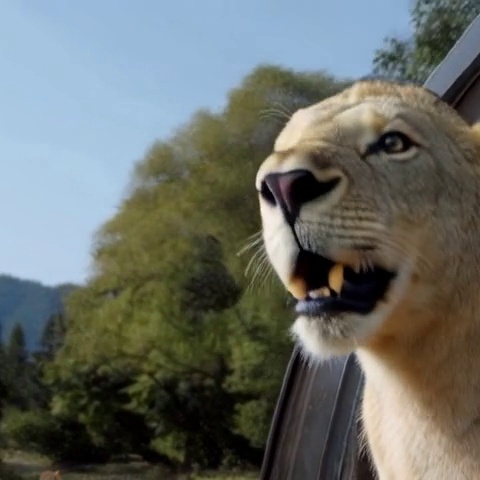} &
\includegraphics[width=0.106\linewidth,height=0.106\linewidth]{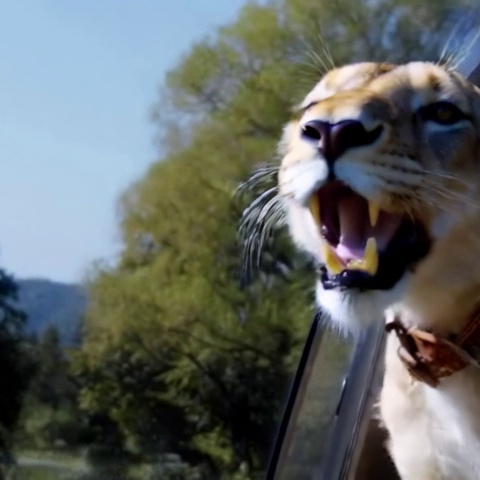} &
\includegraphics[width=0.106\linewidth,height=0.106\linewidth]{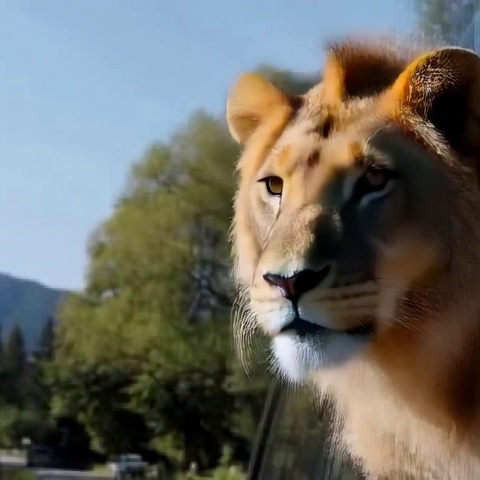} &
\includegraphics[width=0.106\linewidth,height=0.106\linewidth]{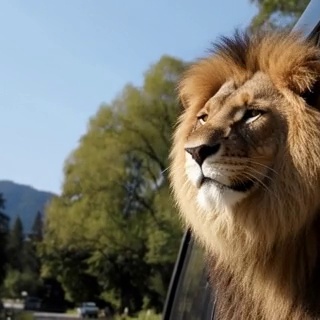} \\[-1.0pt]
\qualtime{$t_8$} &
\includegraphics[width=0.106\linewidth,height=0.106\linewidth]{assets/misc/figures/dog_lion_frames/selection_12/ref_01.jpg} &
\includegraphics[width=0.106\linewidth,height=0.106\linewidth]{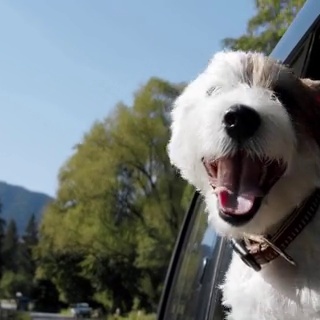} &
\includegraphics[width=0.106\linewidth,height=0.106\linewidth]{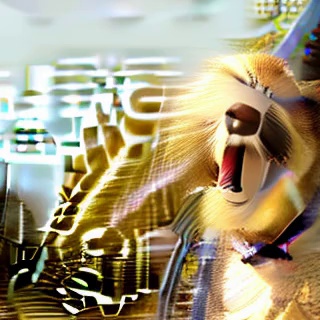} &
\includegraphics[width=0.106\linewidth,height=0.106\linewidth]{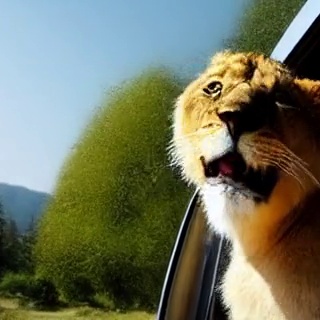} &
\includegraphics[width=0.106\linewidth,height=0.106\linewidth]{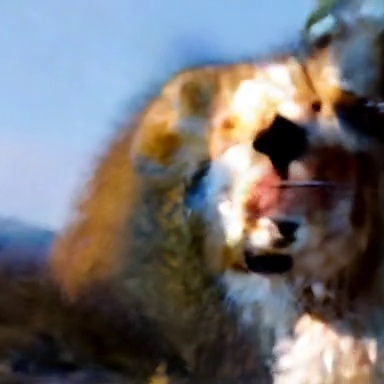} &
\includegraphics[width=0.106\linewidth,height=0.106\linewidth]{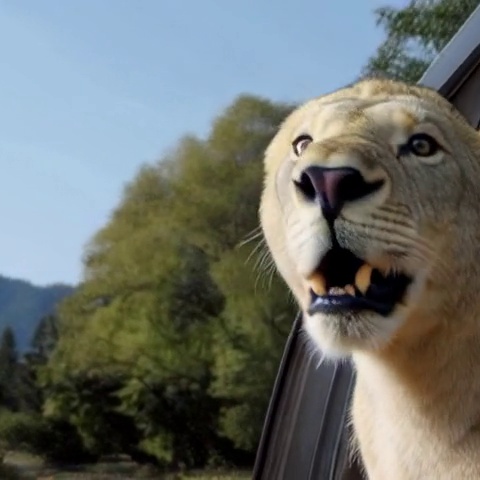} &
\includegraphics[width=0.106\linewidth,height=0.106\linewidth]{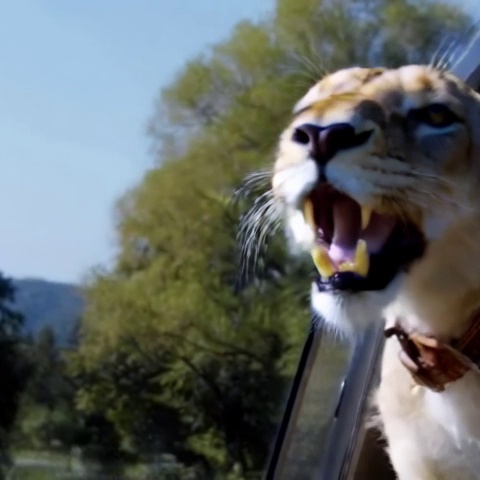} &
\includegraphics[width=0.106\linewidth,height=0.106\linewidth]{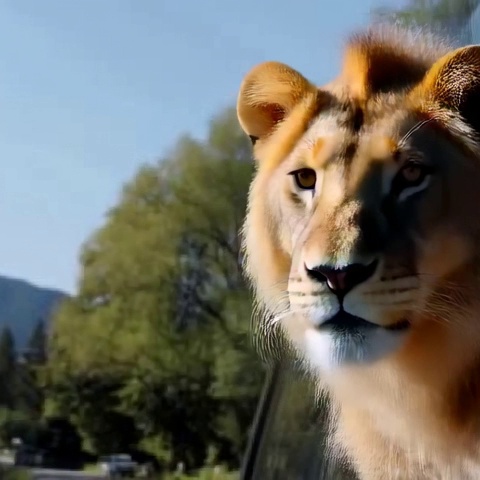} &
\includegraphics[width=0.106\linewidth,height=0.106\linewidth]{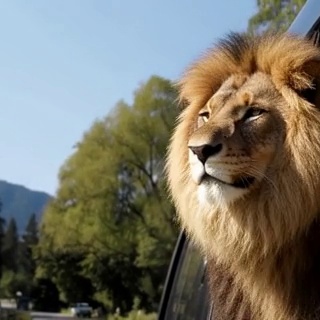} \\[-1.0pt]
\qualtime{$t_9$} &
\includegraphics[width=0.106\linewidth,height=0.106\linewidth]{assets/misc/figures/dog_lion_frames/selection_12/ref_01.jpg} &
\includegraphics[width=0.106\linewidth,height=0.106\linewidth]{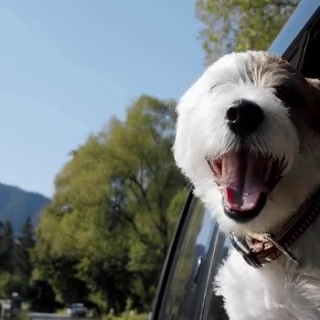} &
\includegraphics[width=0.106\linewidth,height=0.106\linewidth]{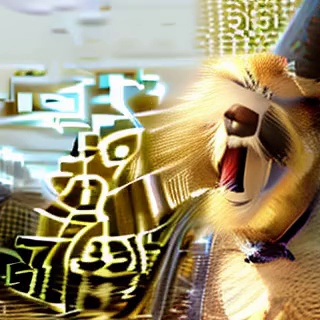} &
\includegraphics[width=0.106\linewidth,height=0.106\linewidth]{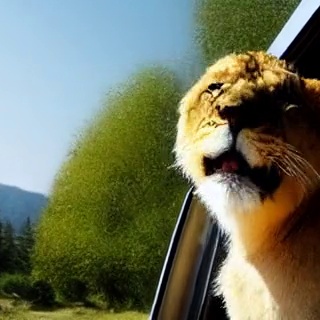} &
\includegraphics[width=0.106\linewidth,height=0.106\linewidth]{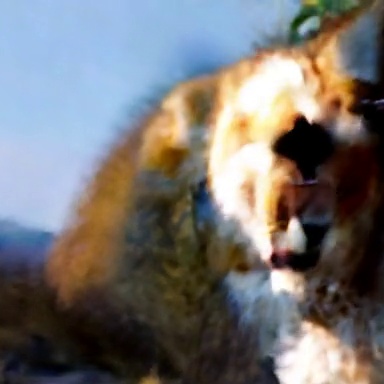} &
\includegraphics[width=0.106\linewidth,height=0.106\linewidth]{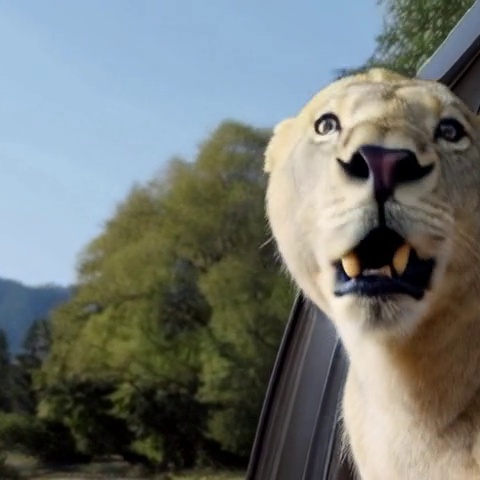} &
\includegraphics[width=0.106\linewidth,height=0.106\linewidth]{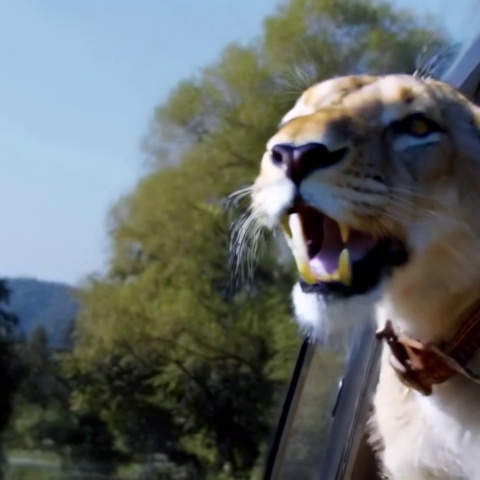} &
\includegraphics[width=0.106\linewidth,height=0.106\linewidth]{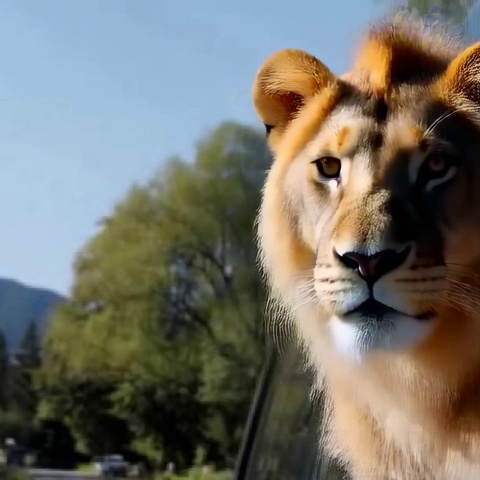} &
\includegraphics[width=0.106\linewidth,height=0.106\linewidth]{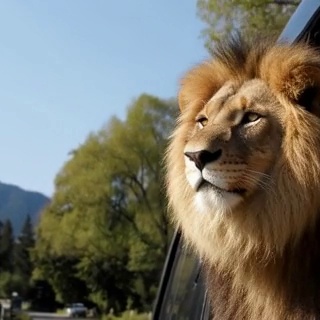} \\[-1.0pt]
\qualtime{$t_{10}$} &
\includegraphics[width=0.106\linewidth,height=0.106\linewidth]{assets/misc/figures/dog_lion_frames/selection_12/ref_01.jpg} &
\includegraphics[width=0.106\linewidth,height=0.106\linewidth]{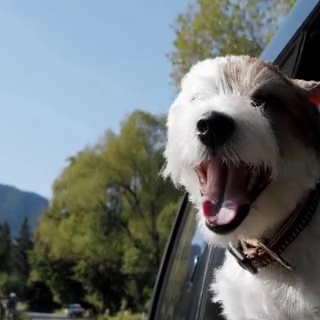} &
\includegraphics[width=0.106\linewidth,height=0.106\linewidth]{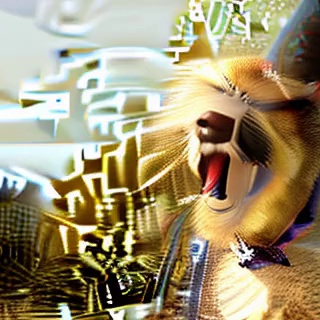} &
\includegraphics[width=0.106\linewidth,height=0.106\linewidth]{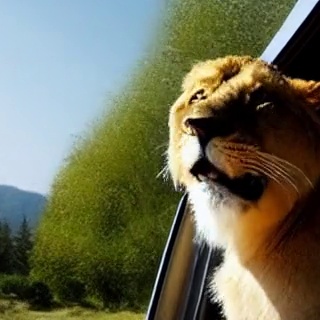} &
\includegraphics[width=0.106\linewidth,height=0.106\linewidth]{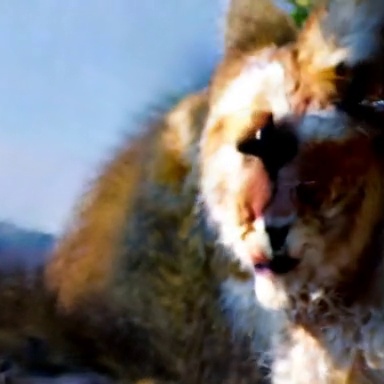} &
\includegraphics[width=0.106\linewidth,height=0.106\linewidth]{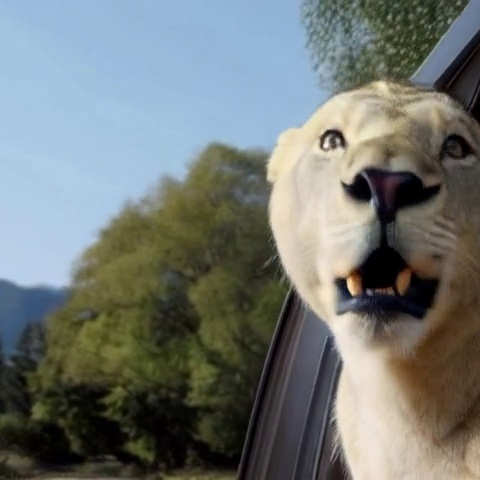} &
\includegraphics[width=0.106\linewidth,height=0.106\linewidth]{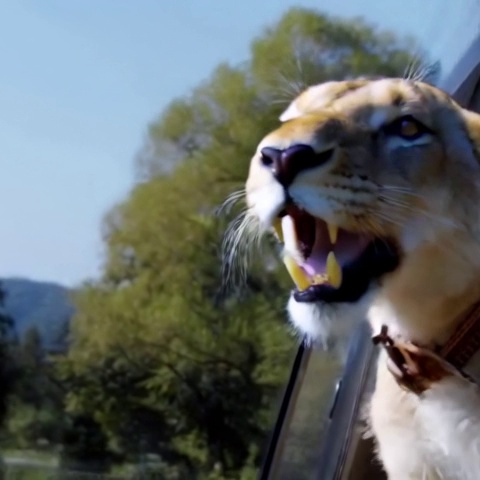} &
\includegraphics[width=0.106\linewidth,height=0.106\linewidth]{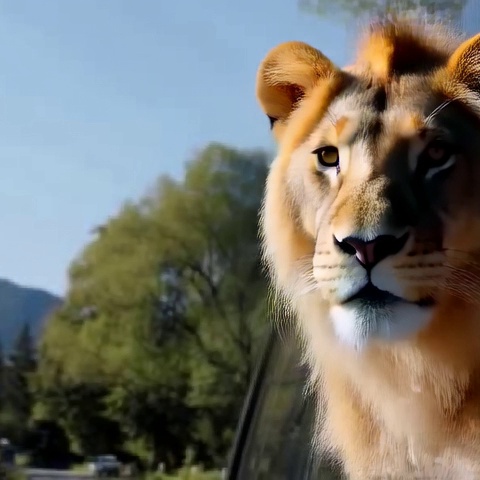} &
\includegraphics[width=0.106\linewidth,height=0.106\linewidth]{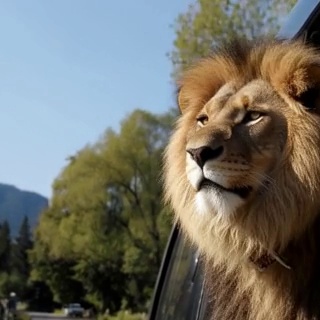} \\[-1.0pt]
\end{tabular}%
}
\caption{\textbf{Subject-guided identity replacement.}
Given the instruction to change the dog into the lion shown in the reference image, \modelnamebf{} transfers the lion's characteristic appearance while preserving the source pose, scale, motion, and scene structure. Several competing methods either retain substantial source-dog appearance, show weak correspondence to the reference, or introduce geometric and background changes. Frames are sampled at uniformly spaced temporal positions throughout each video.}
\label{fig:qual_subject_dog_lion}
\end{figure*}

\begin{figure*}[t!]
\centering
\setlength{\tabcolsep}{0.45pt}
\renewcommand{\arraystretch}{0.62}
\scriptsize
\def\qualtime#1{\raisebox{0.050\linewidth}{\makebox[0.018\linewidth][r]{#1}}}
\resizebox{\textwidth}{!}{%
\begin{tabular}{@{}r@{\hspace{1pt}}ccccccccc@{}}
\multicolumn{10}{c}{\small\textbf{Prompt:} Replace only the swan with the polar bear from the reference image.} \\[3pt]
& \textbf{Ref.} & \textbf{Source} & \textbf{AnyV2V~\cite{ku2024anyv2v}} & \textbf{VidToMe~\cite{li2024vidtome}} & \textbf{Pyramid-Edit~\cite{li2025five}} & \textbf{StreamEdit~\cite{jiao2026streamedit}} & \textbf{Wan-Edit~\cite{li2025five}} & \textbf{FlowDirector~\cite{li2026flowdirector}} & \makecell{\textbf{\modelname{}}\\\textbf{(Ours)}} \\[2pt]
\qualtime{$t_1$} &
\includegraphics[width=0.106\linewidth,height=0.106\linewidth]{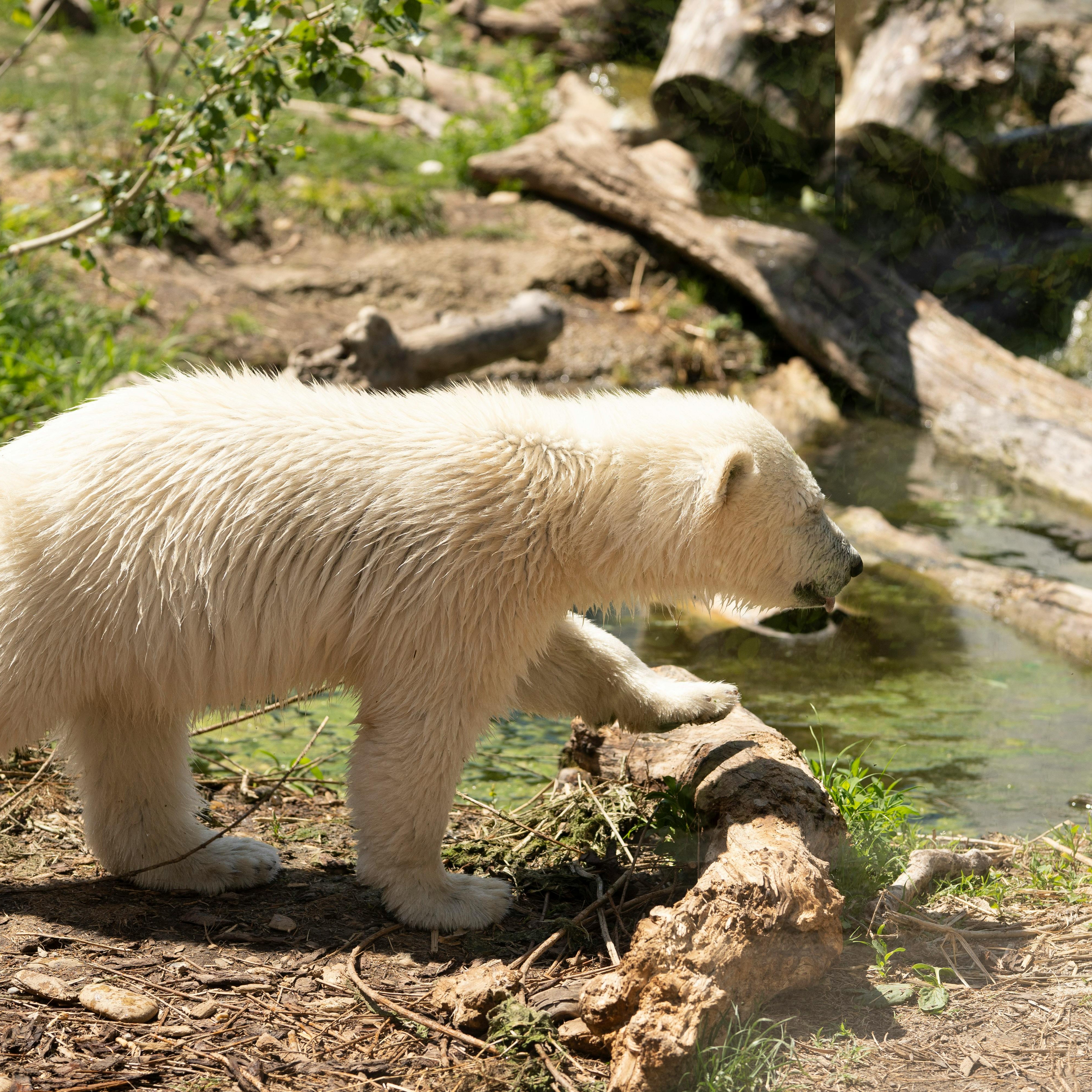} &
\includegraphics[width=0.106\linewidth,height=0.106\linewidth]{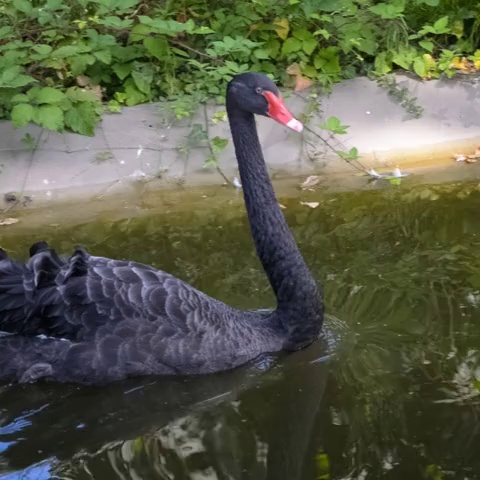} &
\includegraphics[width=0.106\linewidth,height=0.106\linewidth]{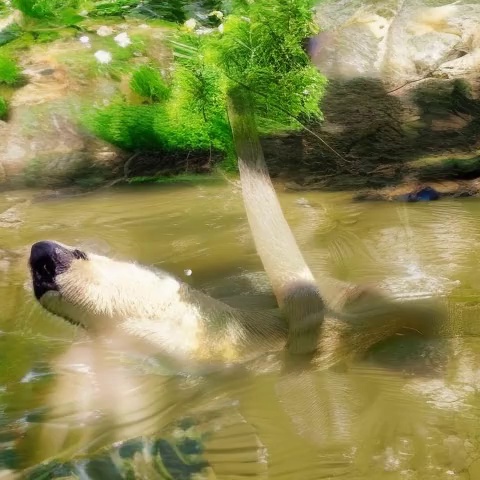} &
\includegraphics[width=0.106\linewidth,height=0.106\linewidth]{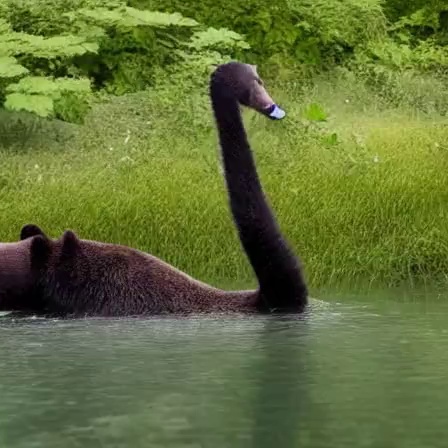} &
\includegraphics[width=0.106\linewidth,height=0.106\linewidth]{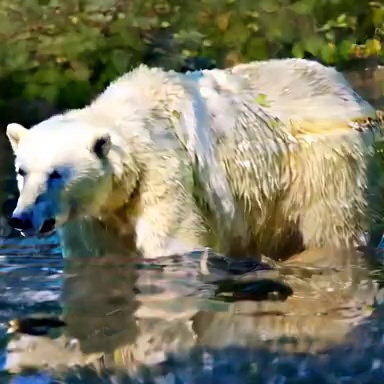} &
\includegraphics[width=0.106\linewidth,height=0.106\linewidth]{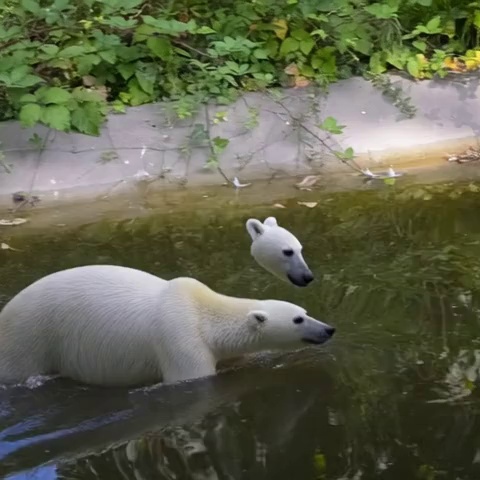} &
\includegraphics[width=0.106\linewidth,height=0.106\linewidth]{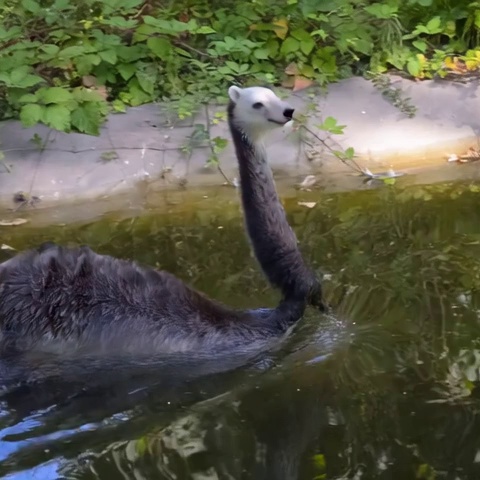} &
\includegraphics[width=0.106\linewidth,height=0.106\linewidth]{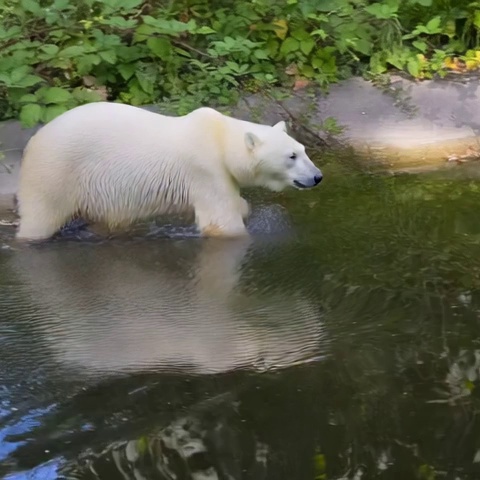} &
\includegraphics[width=0.106\linewidth,height=0.106\linewidth]{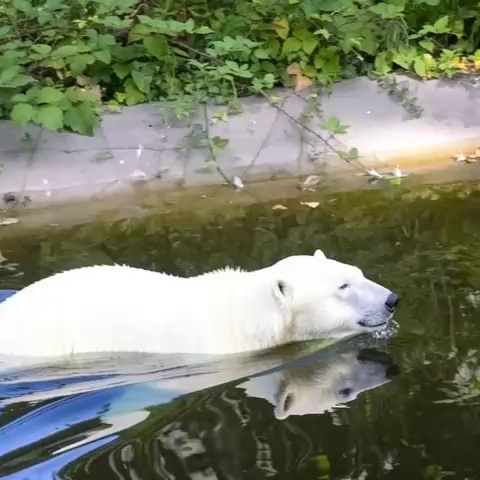} \\[-1.0pt]
\qualtime{$t_2$} &
\includegraphics[width=0.106\linewidth,height=0.106\linewidth]{assets/misc/figures/swan_polar_bear_frames/selection_12/ref_01.jpg} &
\includegraphics[width=0.106\linewidth,height=0.106\linewidth]{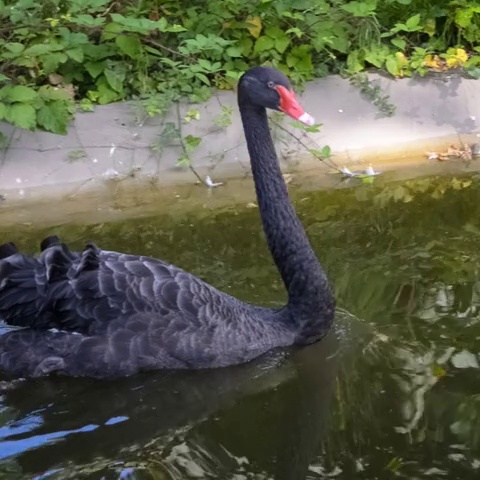} &
\includegraphics[width=0.106\linewidth,height=0.106\linewidth]{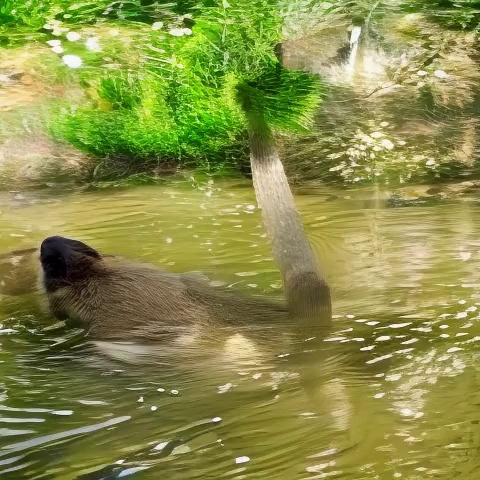} &
\includegraphics[width=0.106\linewidth,height=0.106\linewidth]{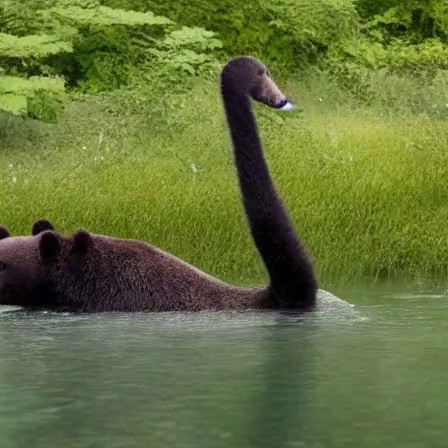} &
\includegraphics[width=0.106\linewidth,height=0.106\linewidth]{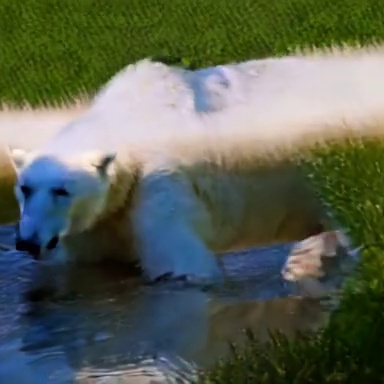} &
\includegraphics[width=0.106\linewidth,height=0.106\linewidth]{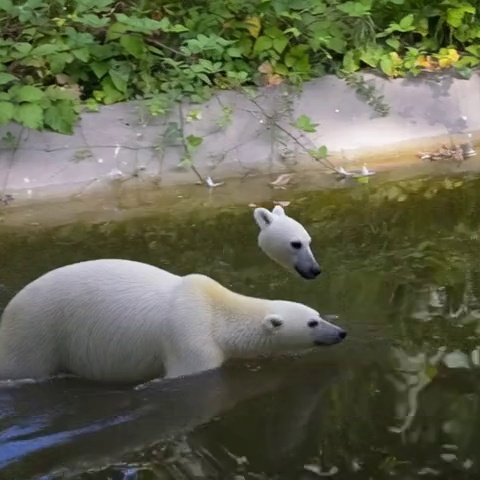} &
\includegraphics[width=0.106\linewidth,height=0.106\linewidth]{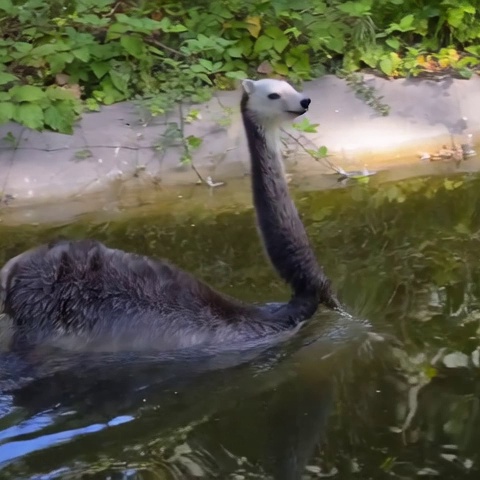} &
\includegraphics[width=0.106\linewidth,height=0.106\linewidth]{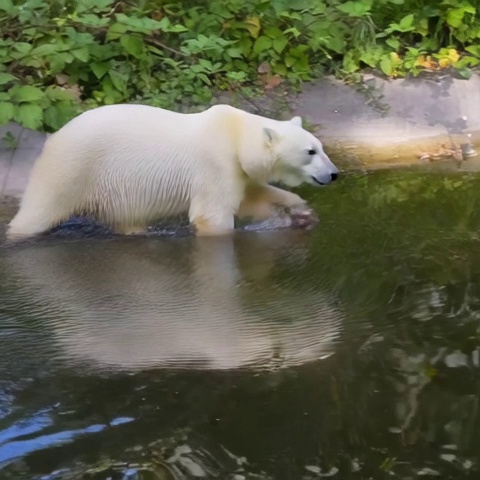} &
\includegraphics[width=0.106\linewidth,height=0.106\linewidth]{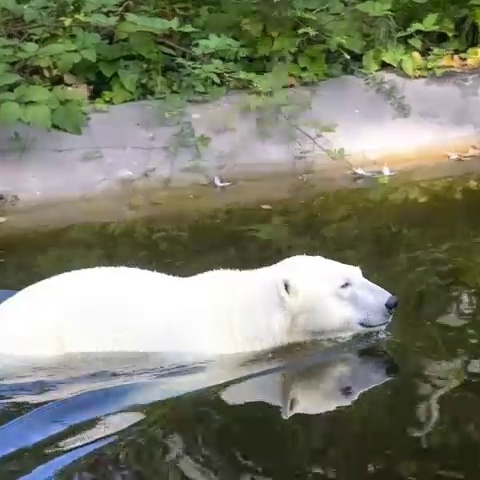} \\[-1.0pt]
\qualtime{$t_3$} &
\includegraphics[width=0.106\linewidth,height=0.106\linewidth]{assets/misc/figures/swan_polar_bear_frames/selection_12/ref_01.jpg} &
\includegraphics[width=0.106\linewidth,height=0.106\linewidth]{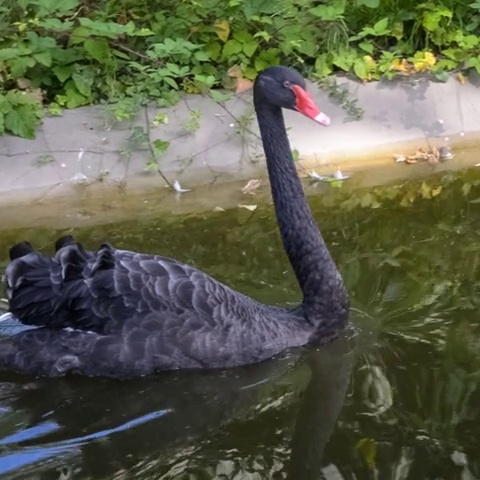} &
\includegraphics[width=0.106\linewidth,height=0.106\linewidth]{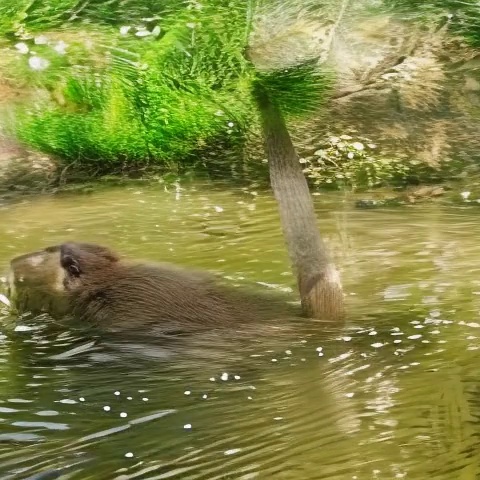} &
\includegraphics[width=0.106\linewidth,height=0.106\linewidth]{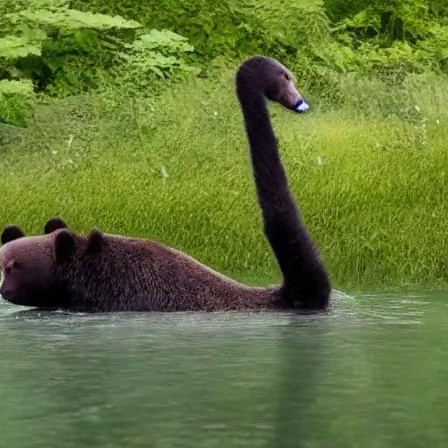} &
\includegraphics[width=0.106\linewidth,height=0.106\linewidth]{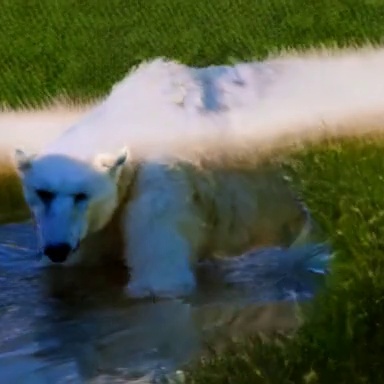} &
\includegraphics[width=0.106\linewidth,height=0.106\linewidth]{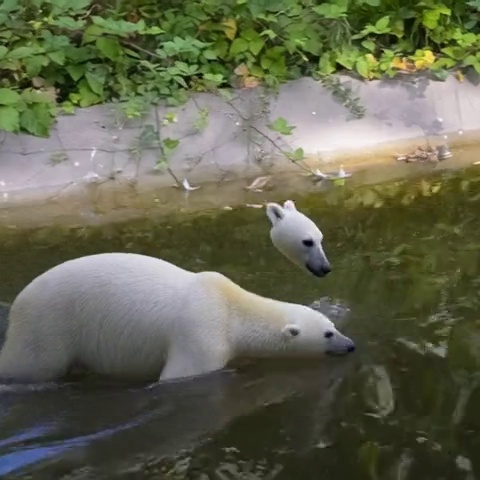} &
\includegraphics[width=0.106\linewidth,height=0.106\linewidth]{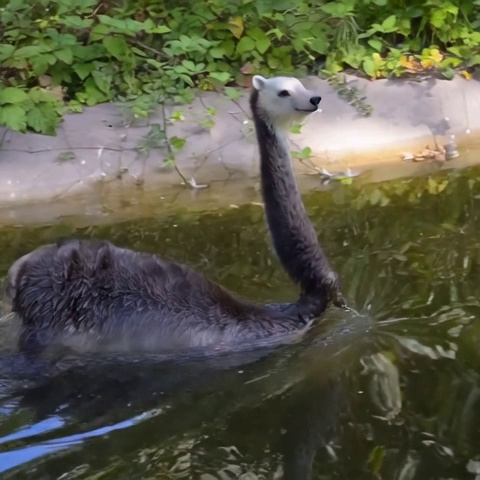} &
\includegraphics[width=0.106\linewidth,height=0.106\linewidth]{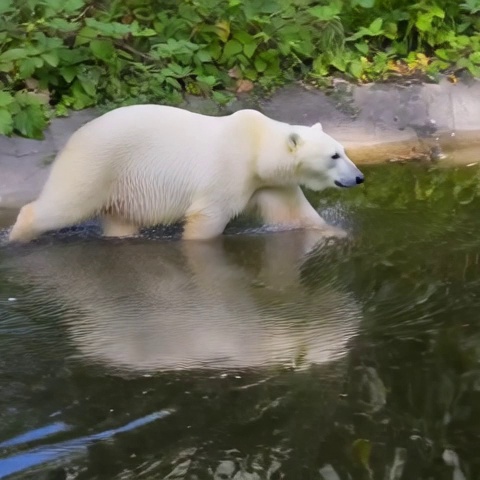} &
\includegraphics[width=0.106\linewidth,height=0.106\linewidth]{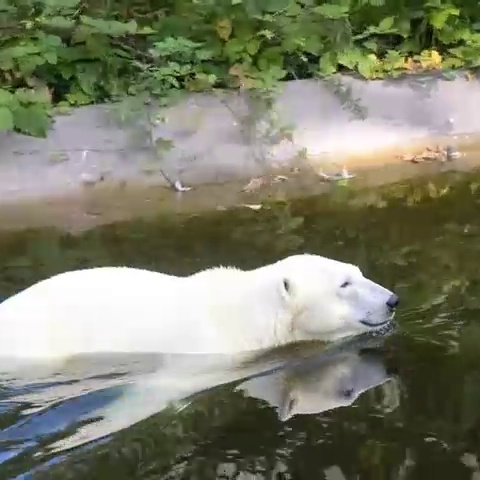} \\[-1.0pt]
\qualtime{$t_4$} &
\includegraphics[width=0.106\linewidth,height=0.106\linewidth]{assets/misc/figures/swan_polar_bear_frames/selection_12/ref_01.jpg} &
\includegraphics[width=0.106\linewidth,height=0.106\linewidth]{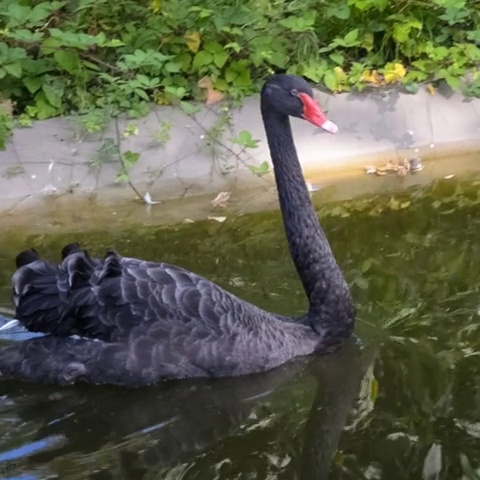} &
\includegraphics[width=0.106\linewidth,height=0.106\linewidth]{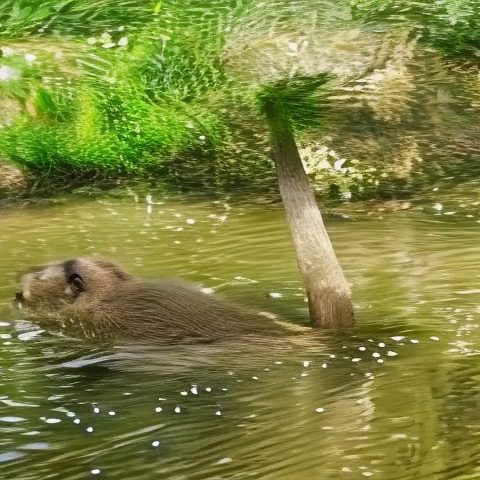} &
\includegraphics[width=0.106\linewidth,height=0.106\linewidth]{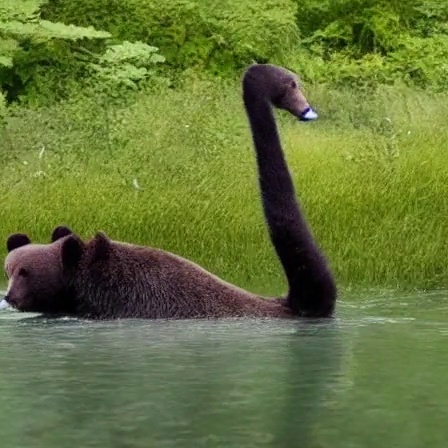} &
\includegraphics[width=0.106\linewidth,height=0.106\linewidth]{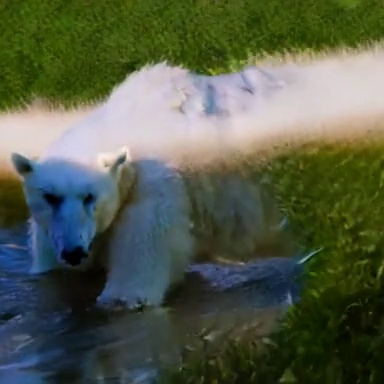} &
\includegraphics[width=0.106\linewidth,height=0.106\linewidth]{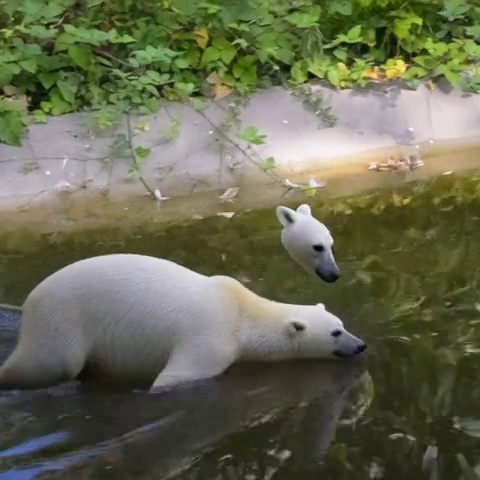} &
\includegraphics[width=0.106\linewidth,height=0.106\linewidth]{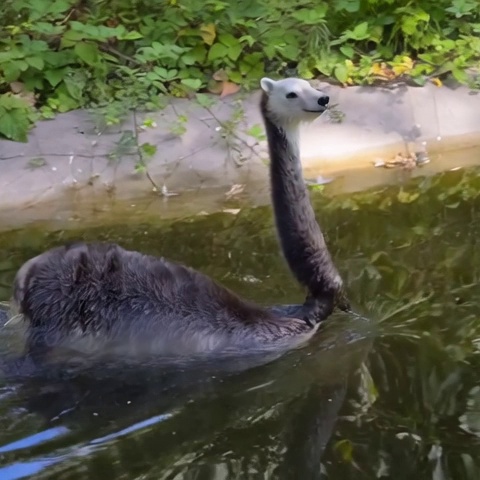} &
\includegraphics[width=0.106\linewidth,height=0.106\linewidth]{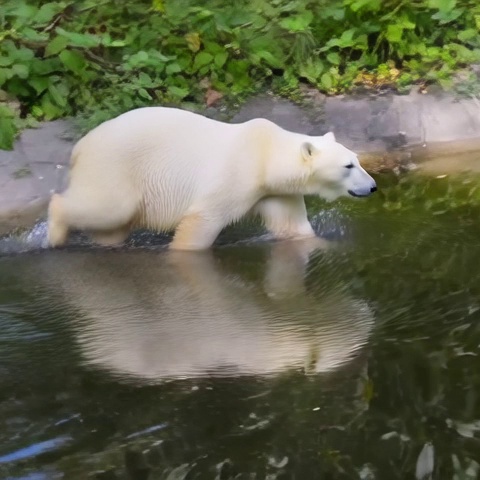} &
\includegraphics[width=0.106\linewidth,height=0.106\linewidth]{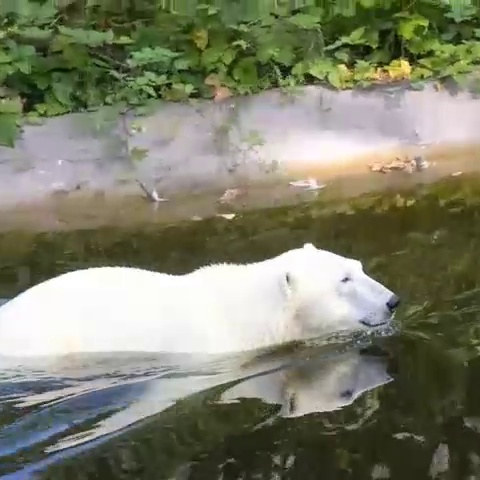} \\[-1.0pt]
\qualtime{$t_5$} &
\includegraphics[width=0.106\linewidth,height=0.106\linewidth]{assets/misc/figures/swan_polar_bear_frames/selection_12/ref_01.jpg} &
\includegraphics[width=0.106\linewidth,height=0.106\linewidth]{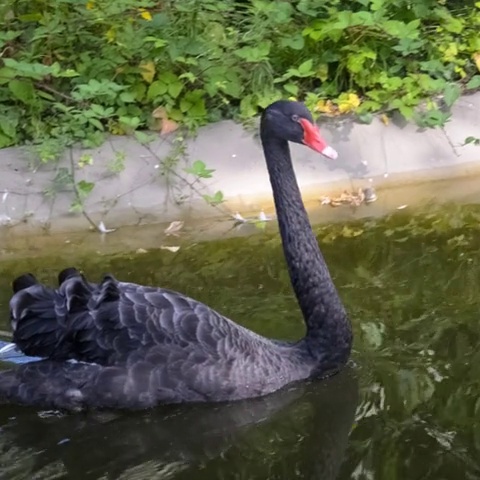} &
\includegraphics[width=0.106\linewidth,height=0.106\linewidth]{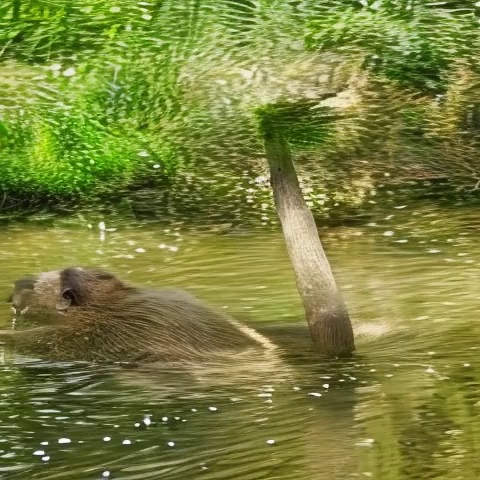} &
\includegraphics[width=0.106\linewidth,height=0.106\linewidth]{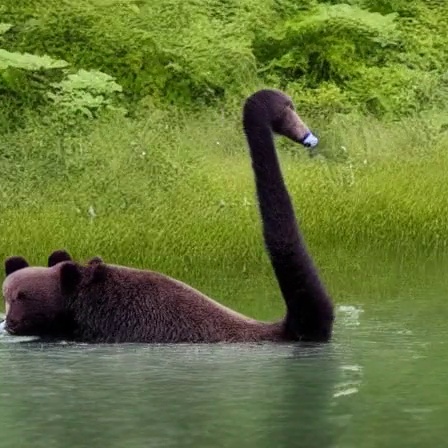} &
\includegraphics[width=0.106\linewidth,height=0.106\linewidth]{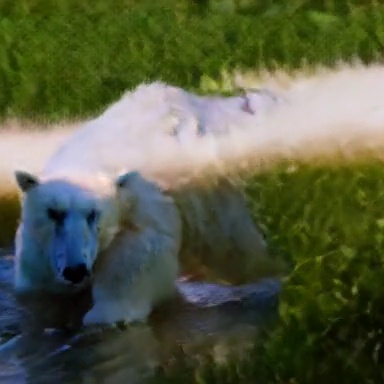} &
\includegraphics[width=0.106\linewidth,height=0.106\linewidth]{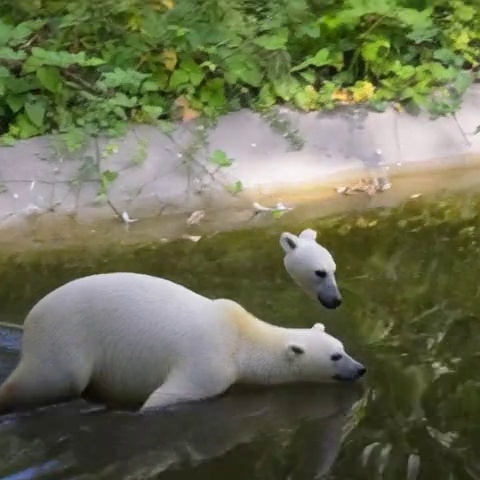} &
\includegraphics[width=0.106\linewidth,height=0.106\linewidth]{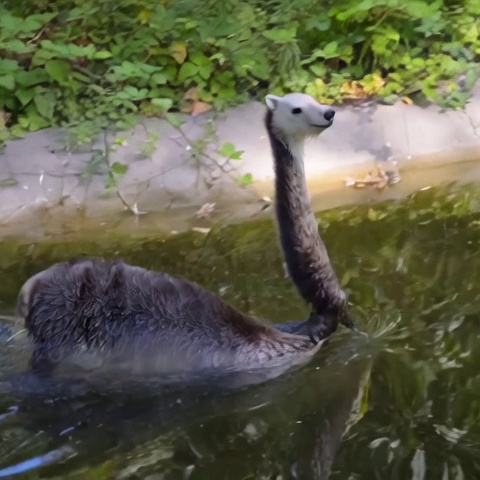} &
\includegraphics[width=0.106\linewidth,height=0.106\linewidth]{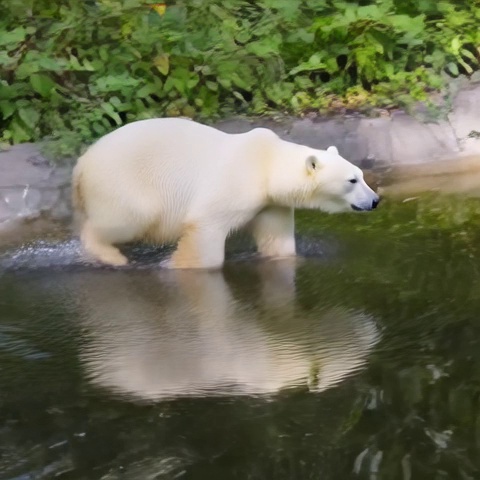} &
\includegraphics[width=0.106\linewidth,height=0.106\linewidth]{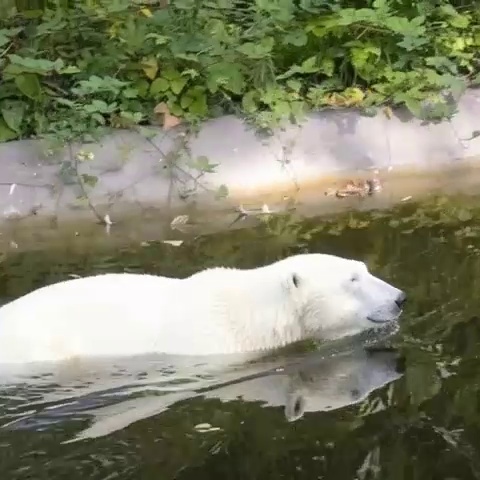} \\[-1.0pt]
\qualtime{$t_6$} &
\includegraphics[width=0.106\linewidth,height=0.106\linewidth]{assets/misc/figures/swan_polar_bear_frames/selection_12/ref_01.jpg} &
\includegraphics[width=0.106\linewidth,height=0.106\linewidth]{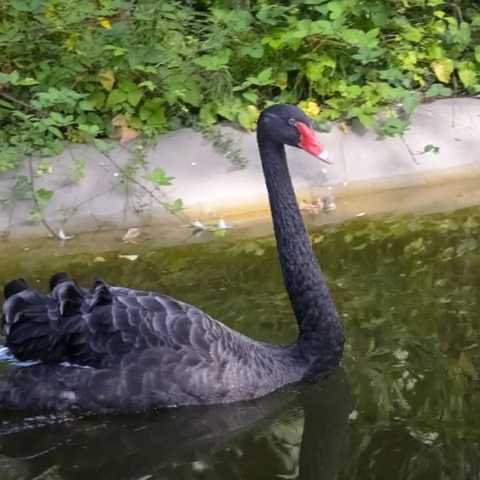} &
\includegraphics[width=0.106\linewidth,height=0.106\linewidth]{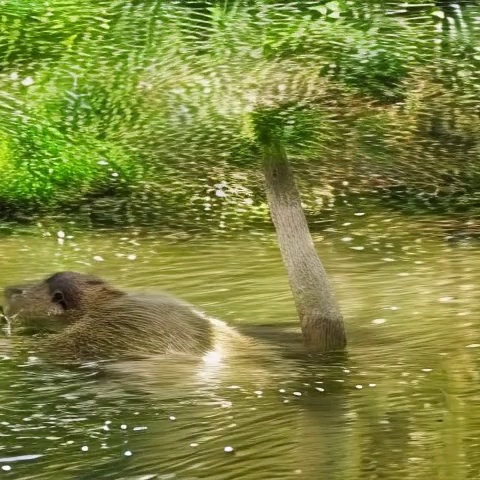} &
\includegraphics[width=0.106\linewidth,height=0.106\linewidth]{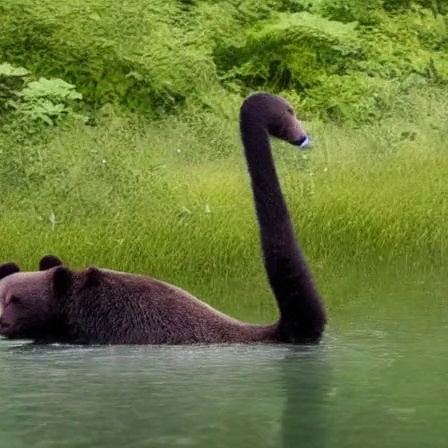} &
\includegraphics[width=0.106\linewidth,height=0.106\linewidth]{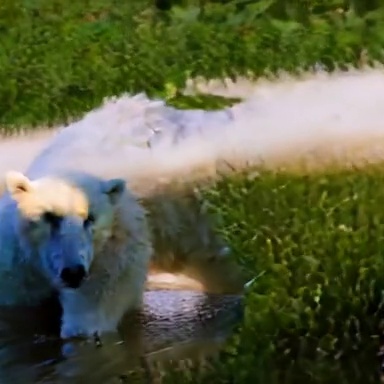} &
\includegraphics[width=0.106\linewidth,height=0.106\linewidth]{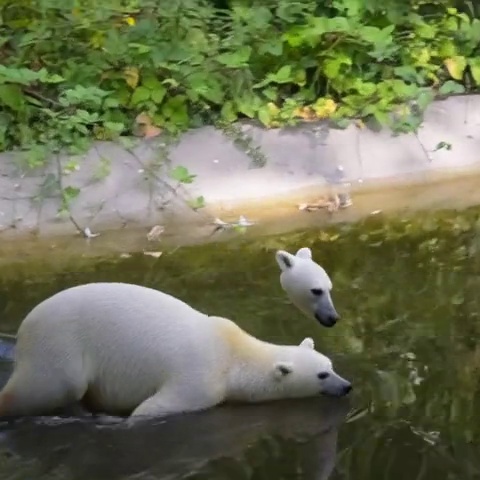} &
\includegraphics[width=0.106\linewidth,height=0.106\linewidth]{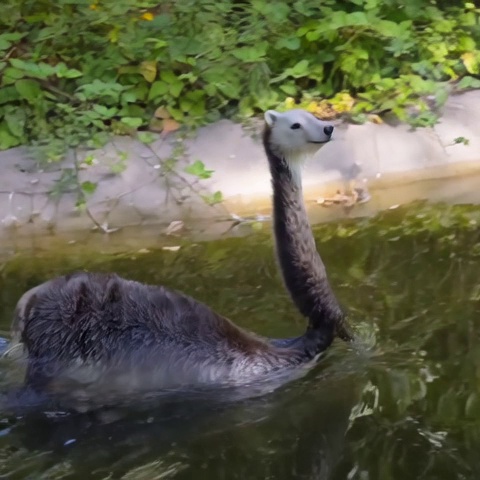} &
\includegraphics[width=0.106\linewidth,height=0.106\linewidth]{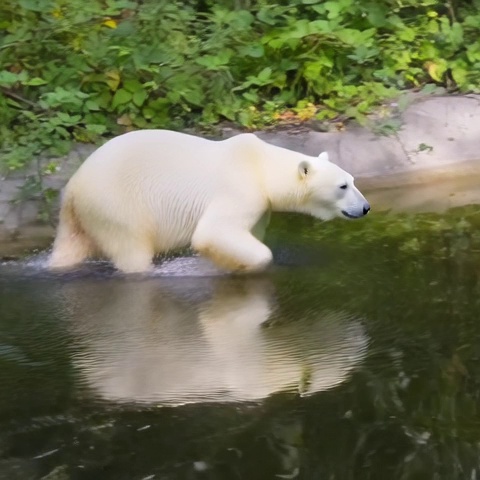} &
\includegraphics[width=0.106\linewidth,height=0.106\linewidth]{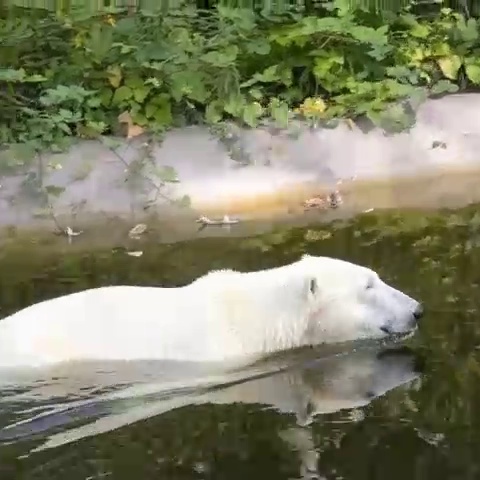} \\[-1.0pt]
\qualtime{$t_7$} &
\includegraphics[width=0.106\linewidth,height=0.106\linewidth]{assets/misc/figures/swan_polar_bear_frames/selection_12/ref_01.jpg} &
\includegraphics[width=0.106\linewidth,height=0.106\linewidth]{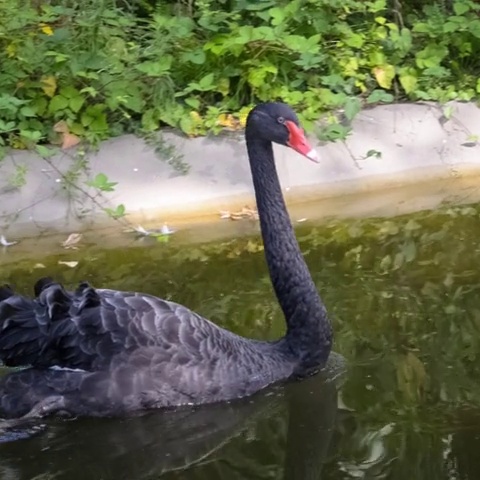} &
\includegraphics[width=0.106\linewidth,height=0.106\linewidth]{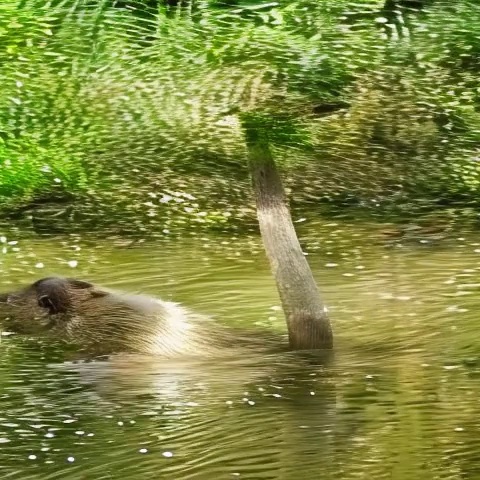} &
\includegraphics[width=0.106\linewidth,height=0.106\linewidth]{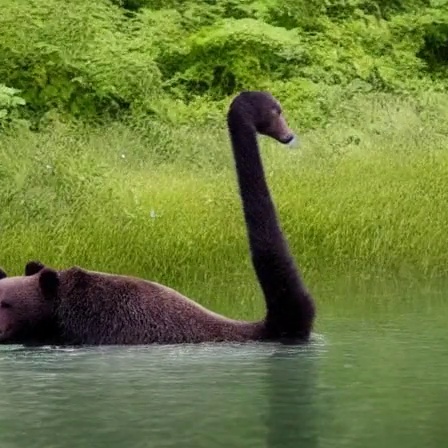} &
\includegraphics[width=0.106\linewidth,height=0.106\linewidth]{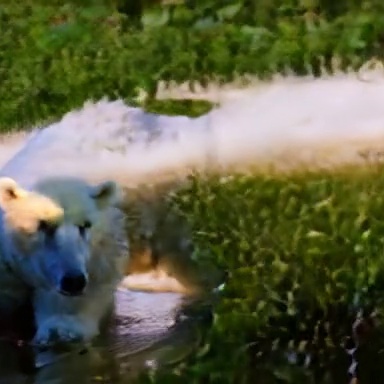} &
\includegraphics[width=0.106\linewidth,height=0.106\linewidth]{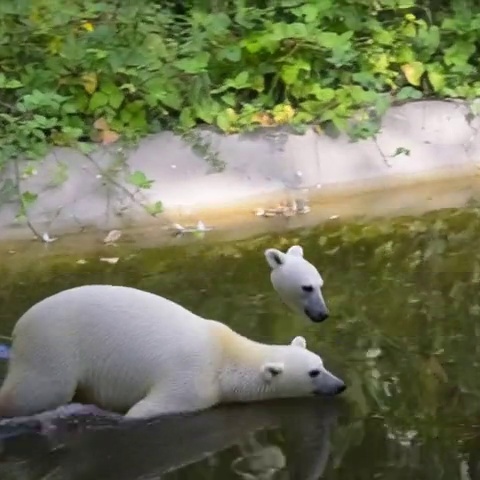} &
\includegraphics[width=0.106\linewidth,height=0.106\linewidth]{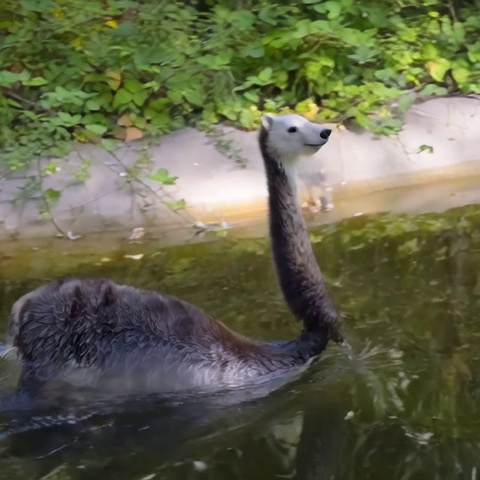} &
\includegraphics[width=0.106\linewidth,height=0.106\linewidth]{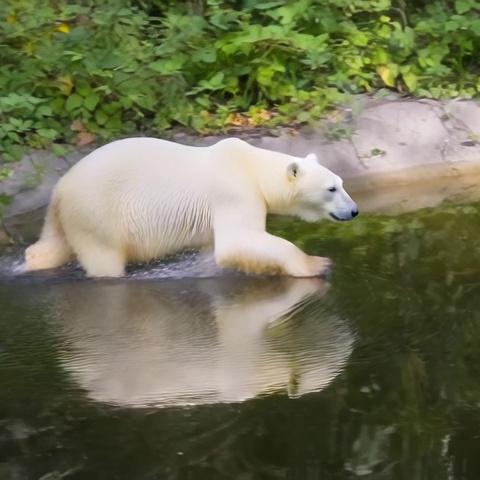} &
\includegraphics[width=0.106\linewidth,height=0.106\linewidth]{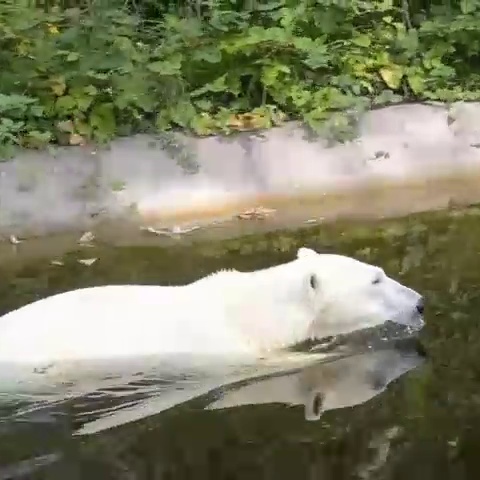} \\[-1.0pt]
\qualtime{$t_8$} &
\includegraphics[width=0.106\linewidth,height=0.106\linewidth]{assets/misc/figures/swan_polar_bear_frames/selection_12/ref_01.jpg} &
\includegraphics[width=0.106\linewidth,height=0.106\linewidth]{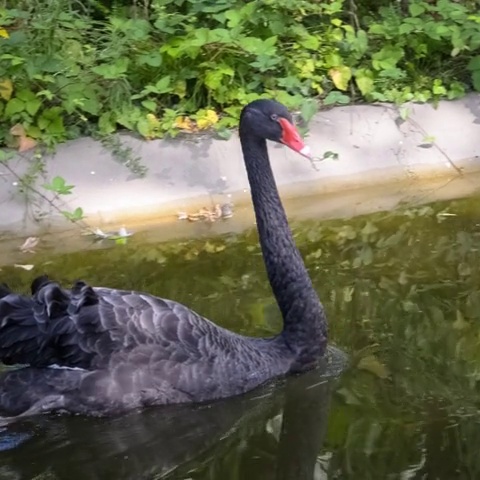} &
\includegraphics[width=0.106\linewidth,height=0.106\linewidth]{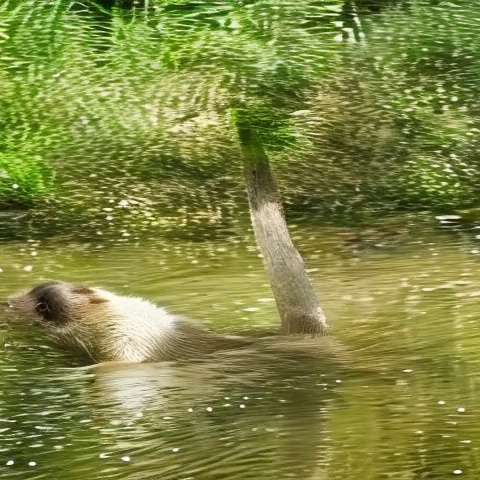} &
\includegraphics[width=0.106\linewidth,height=0.106\linewidth]{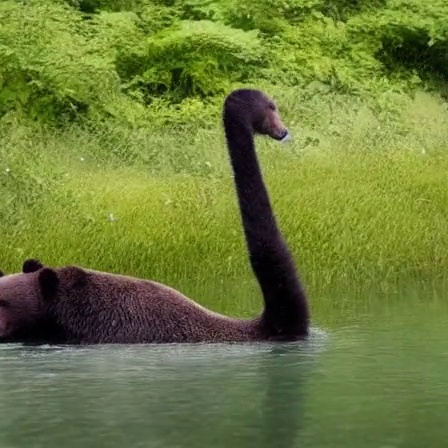} &
\includegraphics[width=0.106\linewidth,height=0.106\linewidth]{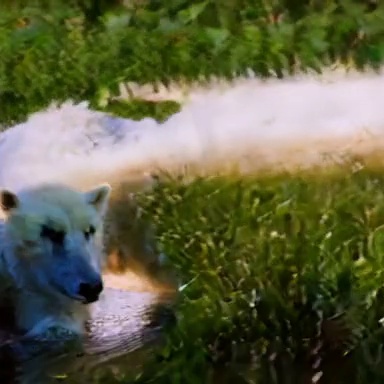} &
\includegraphics[width=0.106\linewidth,height=0.106\linewidth]{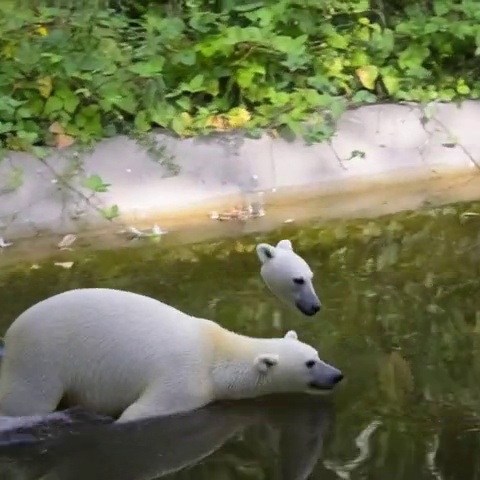} &
\includegraphics[width=0.106\linewidth,height=0.106\linewidth]{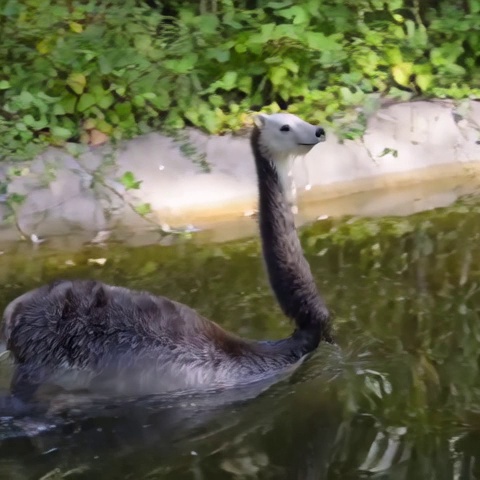} &
\includegraphics[width=0.106\linewidth,height=0.106\linewidth]{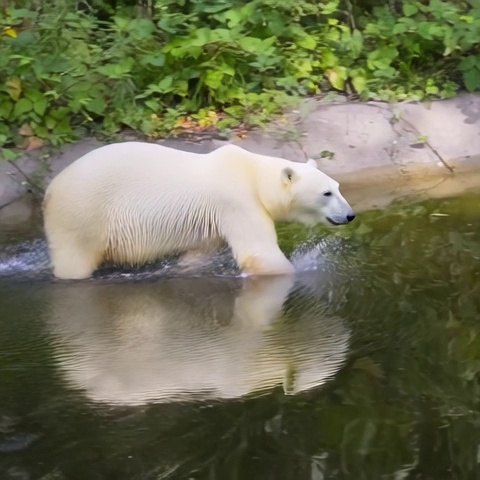} &
\includegraphics[width=0.106\linewidth,height=0.106\linewidth]{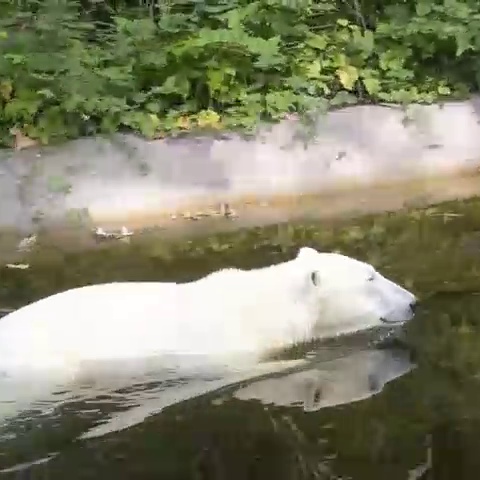} \\[-1.0pt]
\qualtime{$t_9$} &
\includegraphics[width=0.106\linewidth,height=0.106\linewidth]{assets/misc/figures/swan_polar_bear_frames/selection_12/ref_01.jpg} &
\includegraphics[width=0.106\linewidth,height=0.106\linewidth]{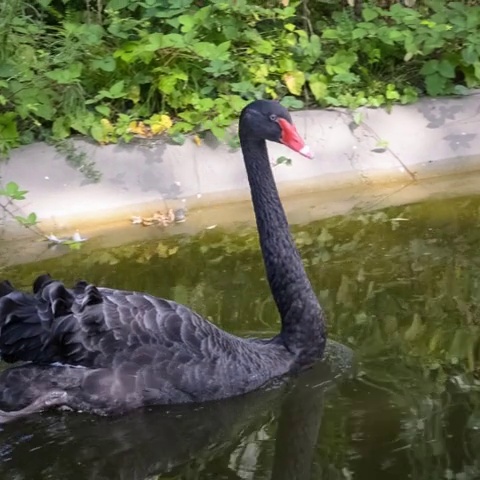} &
\includegraphics[width=0.106\linewidth,height=0.106\linewidth]{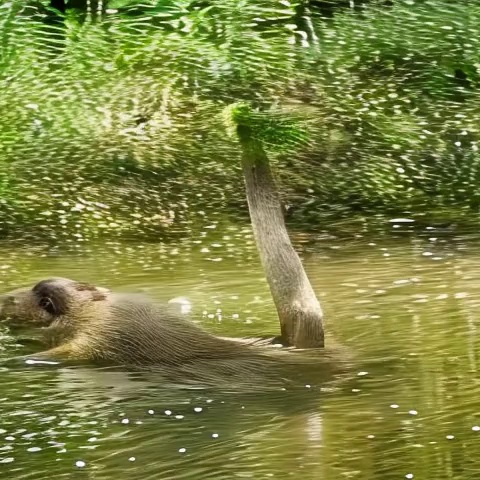} &
\includegraphics[width=0.106\linewidth,height=0.106\linewidth]{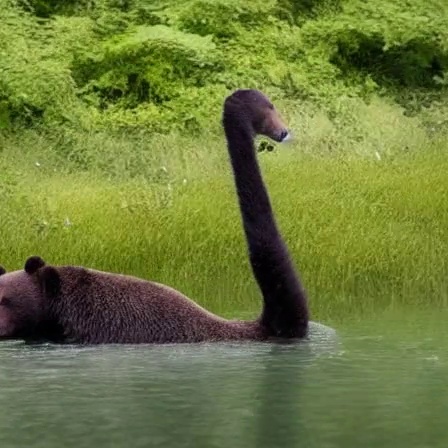} &
\includegraphics[width=0.106\linewidth,height=0.106\linewidth]{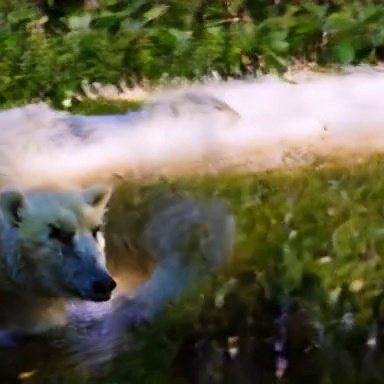} &
\includegraphics[width=0.106\linewidth,height=0.106\linewidth]{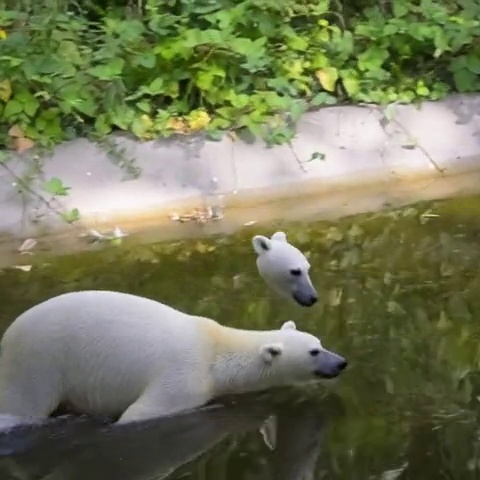} &
\includegraphics[width=0.106\linewidth,height=0.106\linewidth]{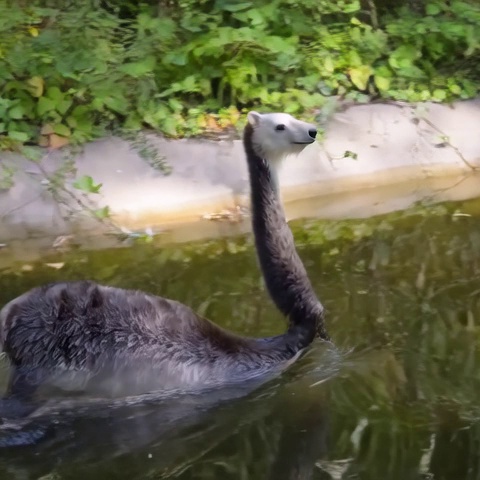} &
\includegraphics[width=0.106\linewidth,height=0.106\linewidth]{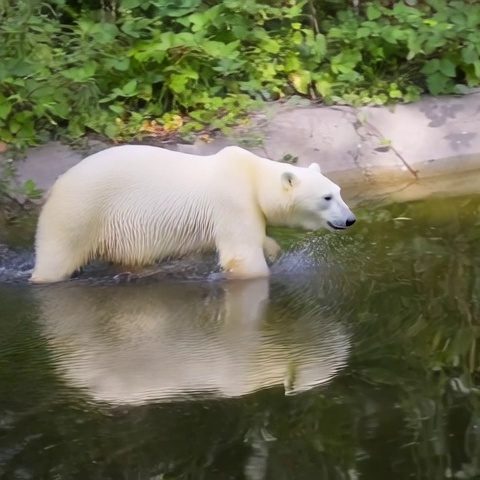} &
\includegraphics[width=0.106\linewidth,height=0.106\linewidth]{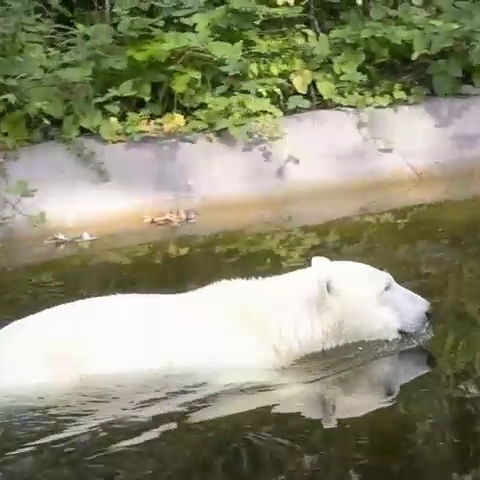} \\[-1.0pt]
\qualtime{$t_{10}$} &
\includegraphics[width=0.106\linewidth,height=0.106\linewidth]{assets/misc/figures/swan_polar_bear_frames/selection_12/ref_01.jpg} &
\includegraphics[width=0.106\linewidth,height=0.106\linewidth]{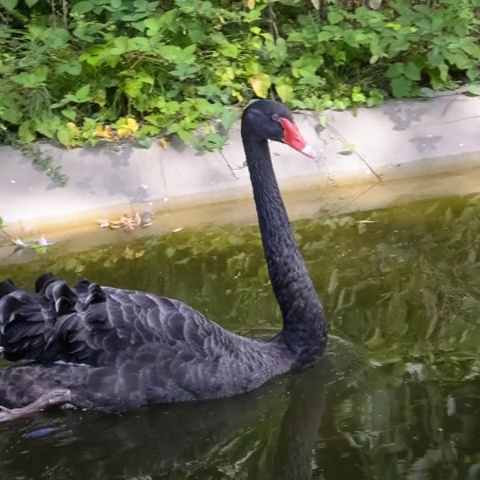} &
\includegraphics[width=0.106\linewidth,height=0.106\linewidth]{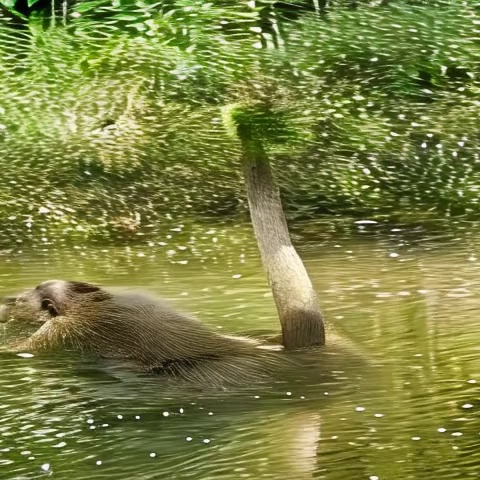} &
\includegraphics[width=0.106\linewidth,height=0.106\linewidth]{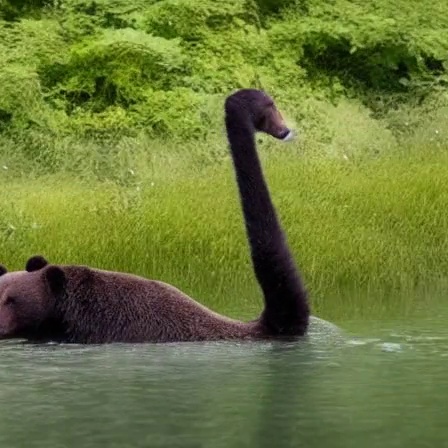} &
\includegraphics[width=0.106\linewidth,height=0.106\linewidth]{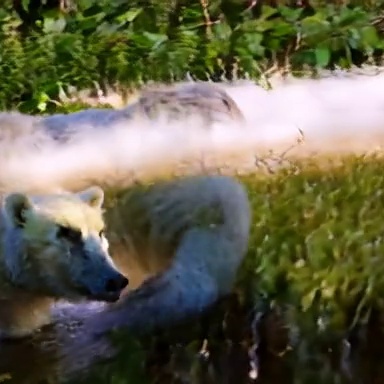} &
\includegraphics[width=0.106\linewidth,height=0.106\linewidth]{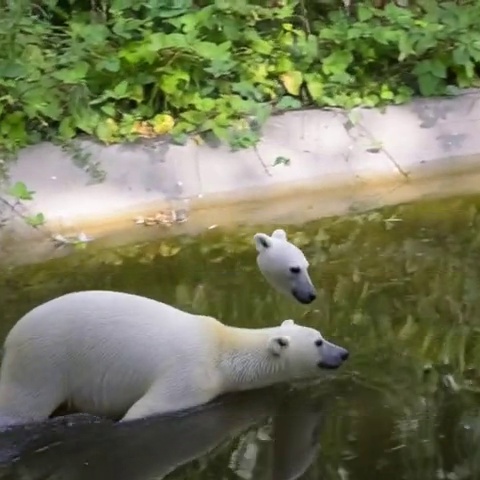} &
\includegraphics[width=0.106\linewidth,height=0.106\linewidth]{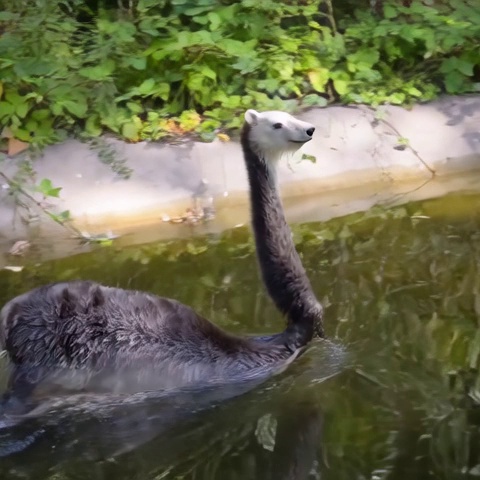} &
\includegraphics[width=0.106\linewidth,height=0.106\linewidth]{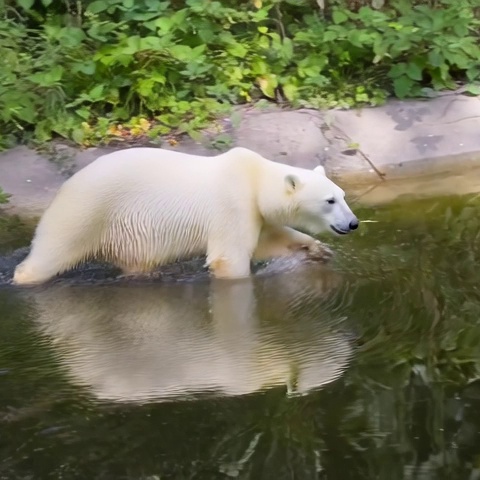} &
\includegraphics[width=0.106\linewidth,height=0.106\linewidth]{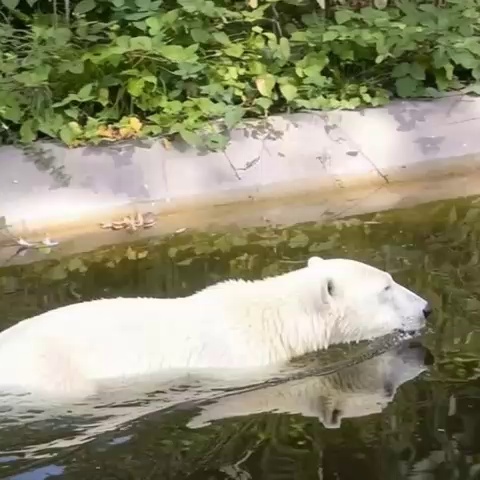} \\[-1.0pt]
\end{tabular}%
}
\caption{\textbf{Subject-guided editing under large semantic change.}
Given the instruction to replace the source subject with the polar bear shown in the reference image, \modelnamebf{} maintains a coherent polar-bear appearance while preserving the source trajectory, motion, and interaction with the surrounding water. Competing methods more often exhibit incomplete identity transfer, geometric deformation, or temporal inconsistency. Frames are sampled at uniformly spaced temporal positions throughout each video.}
\label{fig:qual_subject_swan_polar_bear}
\end{figure*}

We provide additional qualitative comparisons covering global style transfer, object-level material and appearance edits, localized part-level modifications, and subject-guided identity replacement. \cref{fig:qual_ghibli_duck} demonstrates global style transfer, where the challenge is to apply a strong appearance change while preserving the source geometry and motion. \cref{fig:qual_butterfly_glass,fig:qual_holographic_breakdancer} evaluate stronger object-level transformations that substantially change material or appearance while the edited subject undergoes non-rigid motion and large pose variation. \cref{fig:qual_wolf_blue_eyes,fig:qual_man_white_hair} examine localized edits in which only a small semantic attribute should change while the remaining subject appearance is preserved. Finally, \cref{fig:qual_subject_dog_lion,fig:qual_subject_swan_polar_bear} evaluate subject-guided replacement, requiring reference-identity transfer while retaining the source trajectory, pose evolution, and scene interaction.\looseness-1

Across these settings, the qualitative results reveal a consistent trade-off among existing methods. Several preservation-oriented baselines such as FlowDirector, StreamEdit, and Wan-Edit retain the source video well but under-apply the requested transformation, particularly in \cref{fig:qual_butterfly_glass,fig:qual_holographic_breakdancer,fig:qual_wolf_blue_eyes,fig:qual_man_white_hair}. Conversely, methods that produce stronger edits can introduce substantial deviations in object geometry, pose, or surrounding content; this behavior is especially apparent for Pyramid-Edit across multiple examples. The global-style result in \cref{fig:qual_ghibli_duck} further shows that strong, faithful stylization remains challenging for existing baselines. The subject-guided examples in \cref{fig:qual_subject_dog_lion,fig:qual_subject_swan_polar_bear} further highlight a key limitation of existing baselines, which generally lack effective reference-identity-guided video editing capability.
In contrast, \modelname{} consistently performs the requested edit while retaining the source motion, pose, and scene structure across both text-guided and subject-guided settings.\looseness-1

\section{Limitations}
\label{sec:limitations}
\modelname{} is designed for appearance-focused spatial video editing with improved temporal consistency, rather than explicit motion manipulation. Consequently, edits that require changing object dynamics, such as speed, trajectory, or motion direction, are outside the primary scope of the current framework. In addition, our method relies on previous-frame KV context and anchor-to-frame correspondences, which are most effective when temporal associations remain sufficiently reliable. Performance may therefore become less stable under severe occlusion or very large appearance changes that substantially weaken correspondence quality. Addressing explicit motion control and stronger long-range temporal reasoning is an important direction for future work.

\section{Broader Impacts}
\label{sec:broader_impact}
Our training-free framework can broaden access to high-quality video editing by reducing the need for task-specific training, large compute budgets, and specialized pipelines, benefiting research, education, accessibility, and creative media workflows. As a practical safeguard against misuse, we will release code under the MIT License for reproducibility and open research, accompanied by responsible-use guidelines that explicitly discourage deceptive, non-consensual, or harmful applications. These measures should be complemented by continued research on provenance, watermarking, and detection methods beyond temporal inconsistency.

\FloatBarrier

\end{document}